\documentclass[10pt,journal,compsoc]{IEEEtran}
\usepackage[pdftex]{graphicx} 

\usepackage{amsmath}
\usepackage{amsfonts}
\usepackage{amsthm}
\usepackage[utf8]{inputenc}
\usepackage[english]{babel}
\usepackage[usenames, dvipsnames]{color}

\usepackage{subfig}
\usepackage{array}
\usepackage{multirow}

\theoremstyle{definition}
\newtheorem{definition}{Definition}[section]

\ifCLASSOPTIONcompsoc
  \usepackage[nocompress]{cite}
\else
  \usepackage{cite}
\fi
\ifCLASSINFOpdf
\else
\fi
\begin{document}
%
\title{Mesh Denoising based on Normal Voting Tensor and Binary Optimization}
%
%
%
%

\author{Sunil~Kumar~Yadav,
	Ulrich~Reitebuch,
	and~Konrad~Polthier
	\IEEEcompsocitemizethanks{\IEEEcompsocthanksitem Sunil~Kumar~Yadav, Ulrich~Reitebuch and Konrad~Polthier are with the Department
		of Mathematics and Computer Science, Freie Universit{\"a}t Berlin.\protect\\
		E-mail: sunil.yadav@fu-berlin.de}
}

\IEEEtitleabstractindextext{%
\begin{abstract}
This paper presents a two-stage mesh denoising algorithm. Unlike other traditional averaging approaches, our approach uses an element-based normal voting tensor to compute smooth surfaces. By introducing a binary optimization on the proposed tensor together with a local binary neighborhood concept, our algorithm better retains sharp features and produces smoother umbilical regions than previous approaches. On top of that, we provide a stochastic analysis on the different kinds of noise based on the average edge length. The quantitative results demonstrate that the performance of our method is better compared to state of the art smoothing approaches.
\end{abstract}

}

\maketitle

\IEEEdisplaynontitleabstractindextext

%
\IEEEpeerreviewmaketitle

\IEEEraisesectionheading{\section{Introduction}\label{sec:introduction}}

%
%
%
%
\IEEEPARstart{M}{esh} denoising is a central preprocessing tool in discrete geometry processing with many applications in computer graphics such as CAD, reverse engineering, virtual reality and medical diagnosis. The acquisition of 3D surface data takes place using 3D measurement technologies such as 3D cameras and laser scanners. During the surface measurement, noise is inevitable due to various internal and external factors; this degrades surface data quality and its usability. The main goal of any mesh denoising algorithm is to remove spurious noise and compute a high quality smooth function on the triangle mesh while preserving sharp features. 

In general, noise and sharp features both have high frequency components, so decoupling the sharp features from noise is still a challenging problem in mesh denoising algorithms. Traditionally, noise is removed by using a low pass filtering approach, but this operation leads to feature blurring. A variety of Laplacian-based surface smoothing algorithms are available to overcome the problem of feature blurring. Our smoothing approach uses eigenanalysis and a binary optimization of the proposed element based normal voting tensor to decouple noise and sharp features. We design an iterative denoising method that removes low and high frequency noise while preserving sharp features in smooth surfaces. Our algorithm does not produce piecewise flat areas (false features) in the denoised triangular mesh. 
\subsection{Contributions}
We introduce a simple and effective mesh denoising algorithm which does not follow the classic Laplacian approach of surface smoothing. Our algorithm follows a two stage denoising process. In the first stage, we process noisy face normals. In the second stage, we update the vertex positions accordingly. Our main contributions are as follows:
\begin{itemize}
	\item We propose a tensor-based smoothing technique with stable and fast convergence property to remove the undesired noise from noisy surfaces. 
	\item We apply a binary optimization technique on the eigenvalues of the proposed element-based normal voting tensor (ENVT) that helps us to retain sharp features in the concerned geometry and improves the convergence rate of the algorithm.
	\item We give a stochastic analysis of the effect of noise on the triangular mesh based on the minimum edge length of the elements in the geometry. It gives an upper bound to the noise standard deviation to have minimum probability for flipped element normals.   
\end{itemize}

 \begin{figure*}[h]\centering
 	\includegraphics[width=0.96\linewidth]{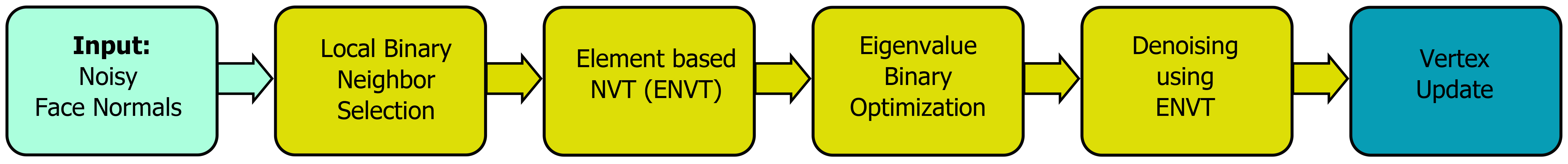}
 	\caption{The pipeline for the proposed smoothing algorithm. The yellow blocks show the face normal processing and the blue block represents the last stage (the vertex update) of the algorithm.
 		}
 	\label{fig:pipeline}
 \end{figure*}
\section{Related Work}
In the last two decades, a wide variety of smoothing algorithms have been introduced to remove undesired noise while preserving sharp features in the geometry. The most common technique for noise reduction is mainly based on the Laplacian on surfaces.
For a comprehensive review on mesh denoising, we refer to \cite{Botsch} and \cite{polyMeshAna}. We give a short overview of major related works in this section.

Isotropic smoothing methods are the earliest smoothing algorithms. These algorithms have low complexity but suffer from severe shrinkage and further feature blurring\cite{lapDel}. Desbrun et al.\cite{Desbrunlambda/u} introduced an implicit smoothing algorithm that produces stable results against irregular meshes and avoids the shrinkage by volume preservation. Later, the concept of the differential coordinates was introduced by Alexa\cite{Alexa} as a local shape descriptor of a geometry. Su et al. exploited the differential coordinates concept for mesh denoising by computing the mean of the differential coordinates at each vertex and then computes a smooth surface according to the mean differential coordinates\cite{diffCoordinate}. This method produces less shrinkage but is unable to preserve shallow features. The differential coordinates framework has been extended for a variety of mesh processing algorithms by Sorkine \cite{CGF:CGF999}. In general, isotropic smoothing methods are prone to shrink volumes and blur features, but effective in noise removal. 

Anisotropic diffusion is a PDE-based de-noising algorithm introduced by Perona and Malik \cite{peronaMalik}. The same concept was extended for surface denoising using a diffusion tensor\cite{Clarenz:2000:AGD:375213.375276} \cite{Bajaj:2003:ADS:588272.588276}. Similarly, the anisotropic diffusion of surface normals was introduced for surface smoothing by Ohtake et al.\cite{Ohtake}\cite{Ohtake2}. These methods compute smooth normals by a weighted sum of the neighborhood element normals. The sharp feature identification, using the surface normals, has been introduced by computing the angle between the neighbor normals \cite{KobbeltFaeture}. Tasziden et al.\cite{AnisoNormals} 
exploited the level set surface model along the anisotropic diffusion of surface normals to produce desired results. Later, the prescribed mean curvature-based surface evolution algorithm was introduced by Hildebrandt et al.\cite{aniso}. It avoids the volume shrinkage and preserves features effectively during the denoising process. Several other algorithms related to anisotropic diffusion are based on the bilateral smoothing\cite{bilAniso}, which was initially introduced by Tomasi et al. \cite{Tomasi:1998:BFG:938978.939190} for image smoothing. Later, the Gaussian kd-tree was introduced to accelerate the bilateral and local mean filtering\cite{AdamsKd}. Researchers have proposed a general framework for bilateral and mean shift filtering in any arbitrary domain\cite{Solomon}. These algorithms are simple and effective against noise and feature blurring. In general, anisotropic denoising methods are more robust against volume shrinkage and are better in terms of feature preservation, but the algorithm complexity is higher compared to isotropic algorithms.

The two step denoising methods are simple and quite robust against noise. These algorithms consist of face normal smoothing and vertex position updates\cite{Ohtake}. Face normals are treated as signals on the dual graph of the mesh with values in the unit sphere. The Laplacian smoothing of the face normals on the sphere was introduced by Taubin\cite{Taubin01linearanisotropic} where displacement of the concerned face normal is computed along a geodesic on the unit sphere. The face normal smoothing is done by rotating the face normal on the unit sphere according to the weighted average of the neighbor face normals. Different linear and non-linear weighting functions have been introduced by different algorithms for face normal smoothing. For example, Yogou et al.\cite{meanface} computed the mean and the median filtering of face normals to remove noise. Later, a modified Gaussian weighting function was applied to face normal smoothing in an adaptive manner to reduce feature blurring\cite{automaticSmoothing}. In continuation, the alpha trimming method introduced a non-linear weighting factor which approximates both, the mean and the median filtering\cite{alphaTrimming}. Bilateral normal is one of the most effective and simple algorithms among the two step methods\cite{BilNorm}, where the weighting function is computed based on the normal differences (similarity measurement) and spatial distances between neighboring faces. Recently, a total variational method has been also introduced for mesh normal filtering\cite{peicewiseLin}. After the preprocessing of the face normals, vertex position updates are done by using the orthogonality between the corresponding edge vector and the face normal\cite{vertexUpdate}. The two step denoising methods are simple in implementation and produce effective results. However, on noisy surfaces, it is difficult to compute the similarity function because of the ambiguity between noise and sharp features; this may lead to unsatisfactory results.

In recent mesh denoising methods, compressed sensing techniques are involved to preserve sharp features precisely and remove noise effectively\cite{optSurvey}. For example, the $L^0$ mesh denoising method assumes that features are sparse on general surfaces and introduces an area based differential operator. This method utilizes the $L^0$-optimization to maximize flat regions on noisy surfaces to remove noise\cite{L0Mesh}. The $L^0$ method is effective against high noise but also produces piecewise flat area on smooth surfaces. Later, the weighted $L^1$-analysis compressed sensing optimization was utilized to recover sharp features from the residual data after global Laplacian smoothing \cite{L1Mesh}. Recently, the ROF (Rudin, Osher and Fatemi) based algorithm has been introduced in \cite{ROFL1}. This method applies $L^1$-optimization on both data fidelity and regularization to remove noise without volume shrinkage. In general, the compressed sensing based denoising algorithms are robust against high intensity noise and recover not only the sharp but also the shallow features, but at the same time these algorithms produce false features (piecewise flat areas) on smooth geometries.

A multistage denoising framework was applied in recent methods \cite{anisobil} and \cite{NVTsmooth} where feature identification is done by the eigenanalysis of the NVT (normal voting tensor)\cite{NVT},\cite{NVTb}. Then, the whole geometry is divided into different clusters based on the features and then smoothing is applied on different clusters independently. Later, the guided mesh normals were computed based on the consistent normal orientation and bilateral filtering was applied \cite{Guidedmesh}. Recently, Wei et. al.\cite{binormal} exploited both vertex and face normal information for feature classification and surface smoothing. In continuation, researchers detected features on the noisy surface using quadratic optimization and then remove noise using $L^1$-optimization while preserving features\cite{robust16}. Multistage denoising algorithms produce effective results against different levels of noise but have higher algorithm complexity because of the different stages.   
 
 In our method, the face normal smoothing is motivated by the NVT-based algorithms. Noise and features are decoupled using the eigenanalysis of the ENVT and noise is removed by the multiplication of the ENVT to the corresponding face normal.  

\section{Method}
Figure~\ref{fig:pipeline} shows the whole pipeline of our algorithm. The face normal smoothing (the yellow blocks in Figure \ref{fig:pipeline}) consists of four steps: (1) We compute the geometric neighborhood for the concerned face using a local binary scheme. (2) We define the element based normal voting tensor within its geometric neighborhood. (3) To remove noise effectively, we apply a binary optimization on the eigenvalues of the computed tensor. (4) We multiply the modified ENVT to the corresponding face normal to suppress the noise. In the last stage (the blue block in Figure \ref{fig:pipeline}), we update the vertex positions using the orthogonality between the edge vectors and the face normals. In this section, we explain each stage of the proposed algorithm briefly.
\subsection{Local Binary Neighbor Selection} 
\label{locneigh}
The first step of our denoising scheme is the preprocessing of the face normals using the neighboring face normals. To select the neighborhood area $\Omega$, there are three possibilities: Combinatorial, geodesic and geometrical neighborhood. Each of these terms are explained in Appendix \ref{app:Neigh}. The geometrical neighborhood is applied in the proposed algorithm because it depends only on the radius of the disk irrespective of mesh resolution unlike the topological neighborhood. 

\begin{figure}[bth]%
	\centering
	\subfloat[Noisy input]{{\includegraphics[width=2.4cm]{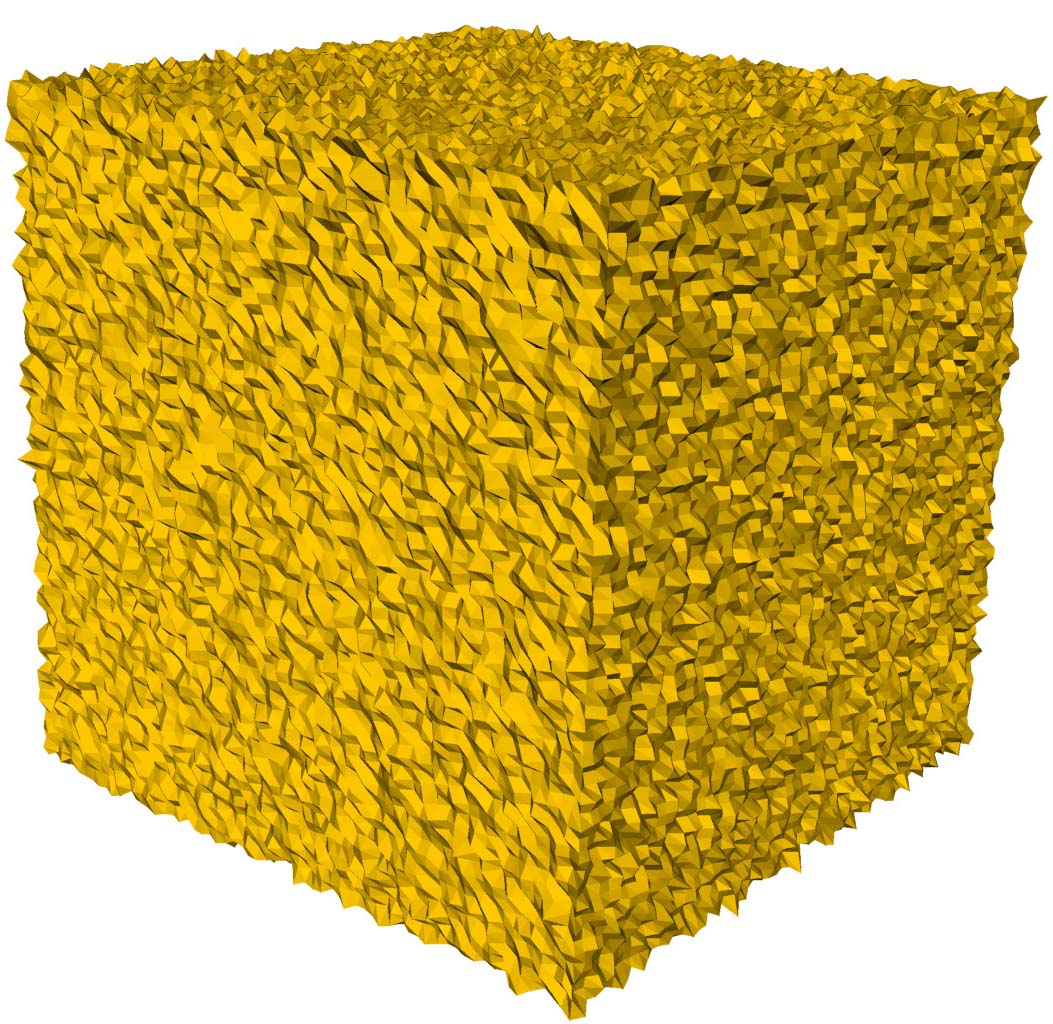} }}%
	\subfloat[$w_{ij}\in\{0,1\}$]{{\includegraphics[width=2.4cm]{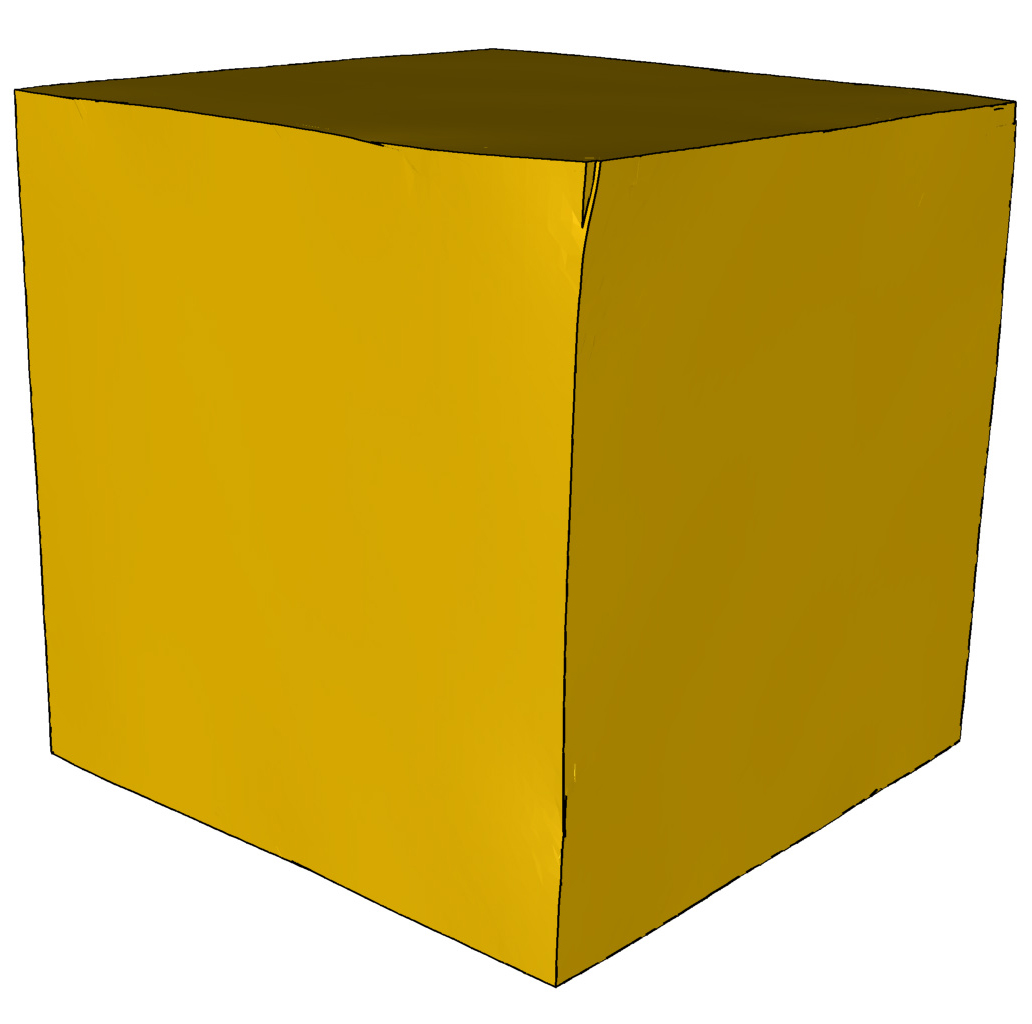} }}%
	\subfloat[$w_{ij}\in\{0.1,1\}$]{{\includegraphics[width=2.4cm]{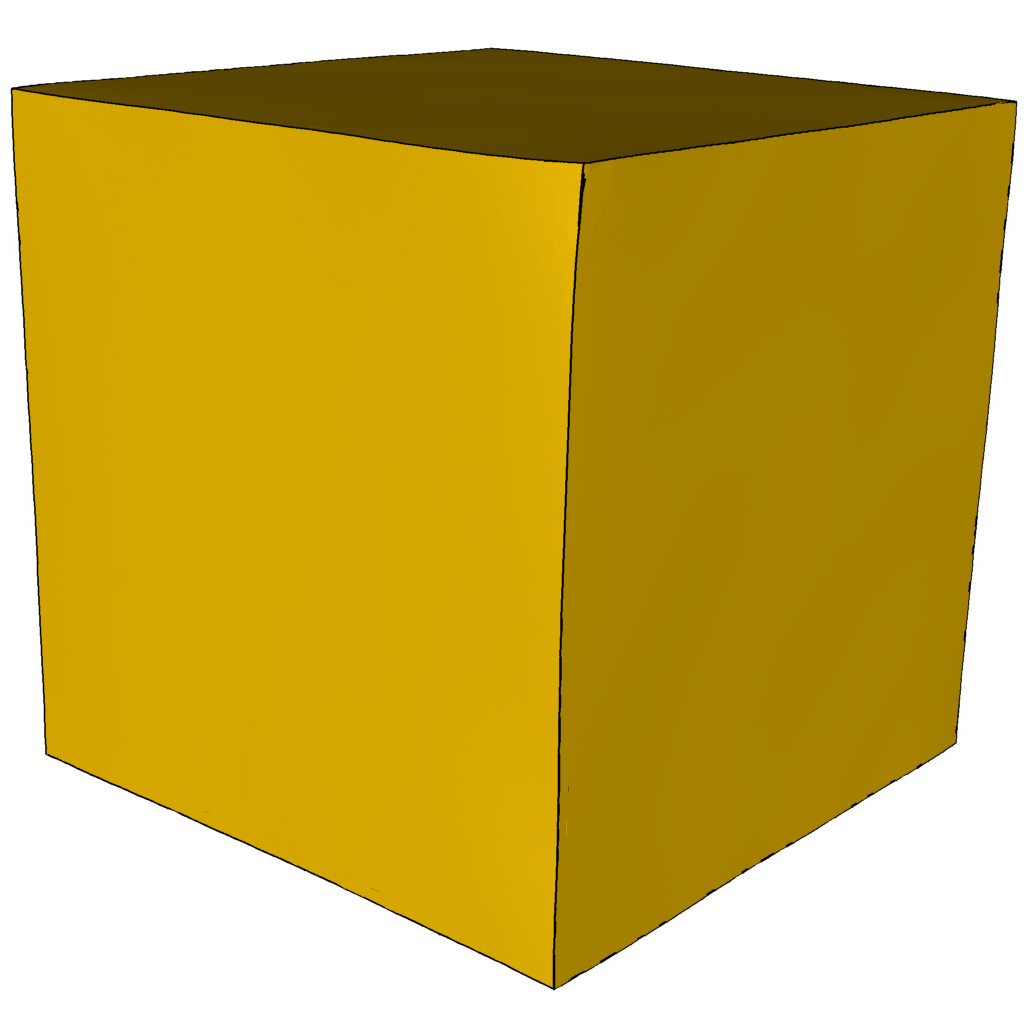} }}%
	\subfloat[zoom]{{\includegraphics[width=1.2cm]{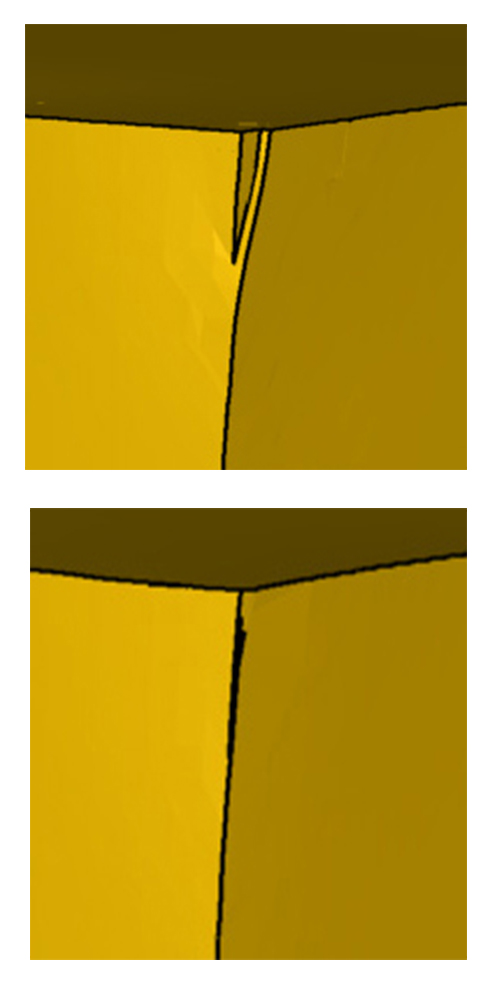} }}%
	\caption{ The results with the two different binary weighting functions in the proposed method. (a) Noisy cube model. (b) The weight function ($w_{ij}\in\{0,1\}$). (c) The weight function mentioned in Equation~\ref{equ:LBP}. (d) The magnified view shows that the weight function mentioned in Equation~\ref{equ:LBP} is more effective compared to the exact binary weighting ($w_{ij}\in\{0,1\}$). }%
	\label{fig:neighW}%
\end{figure}

\begin{figure}[h]%
	\centering
	\subfloat[Noisy input]{{\includegraphics[width=2cm]{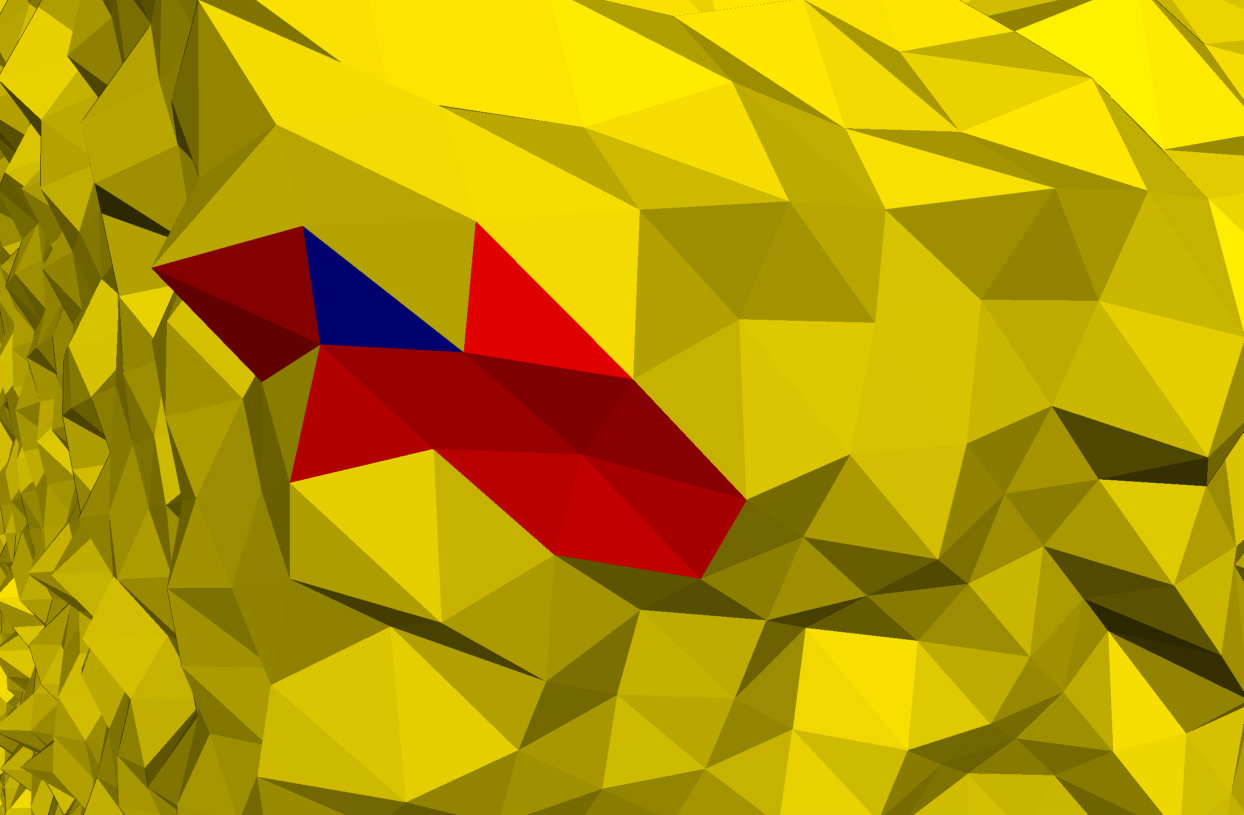} }}%
	\subfloat[10 iterations]{{\includegraphics[width=2cm]{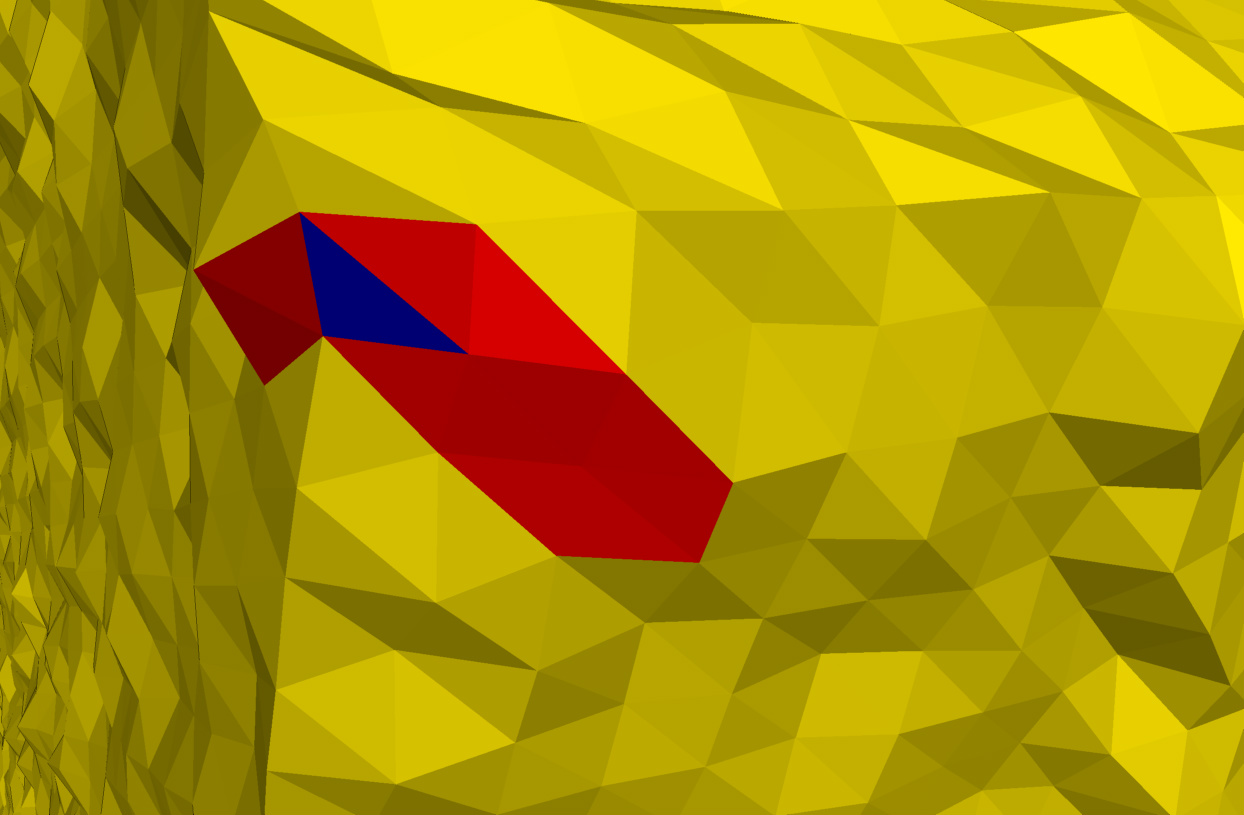} }}%
	\subfloat[20 iterations]{{\includegraphics[width=2cm]{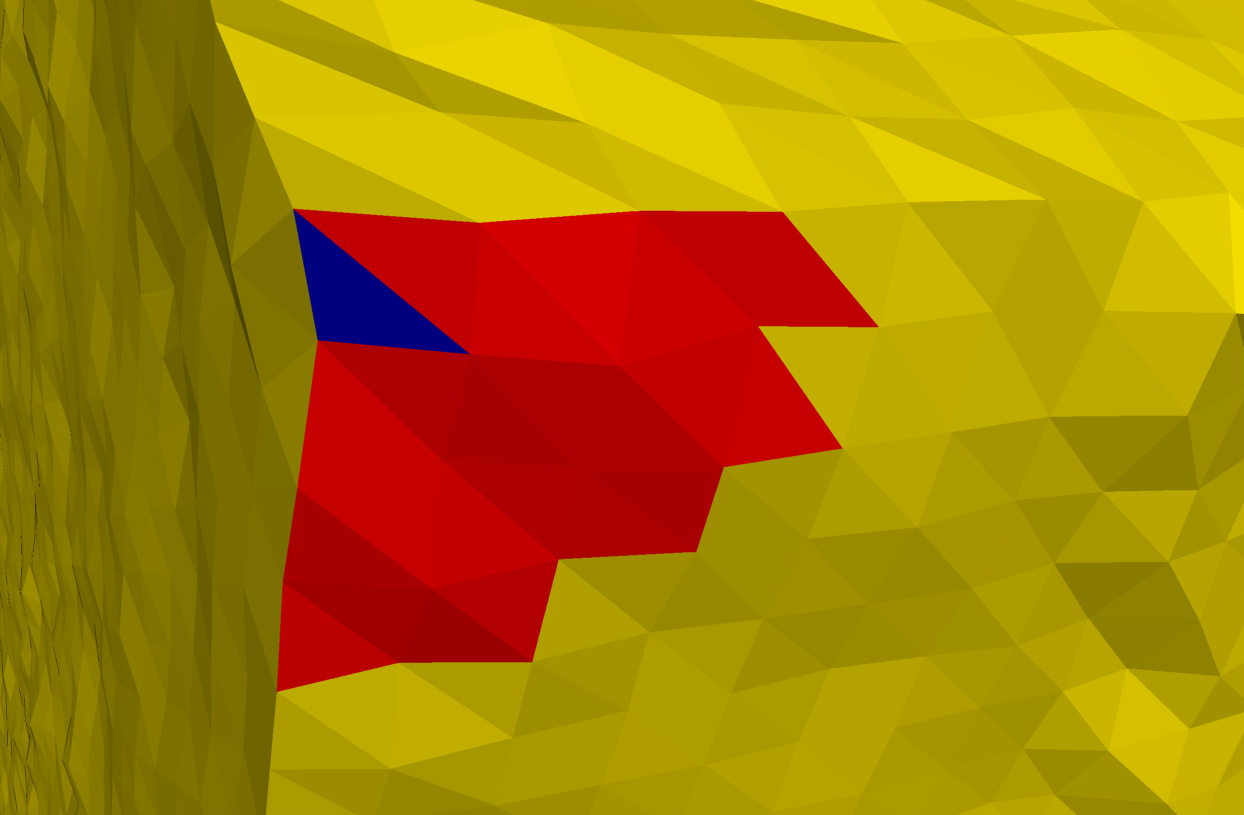} }}%
	\subfloat[40 iterations]{{\includegraphics[width=2cm]{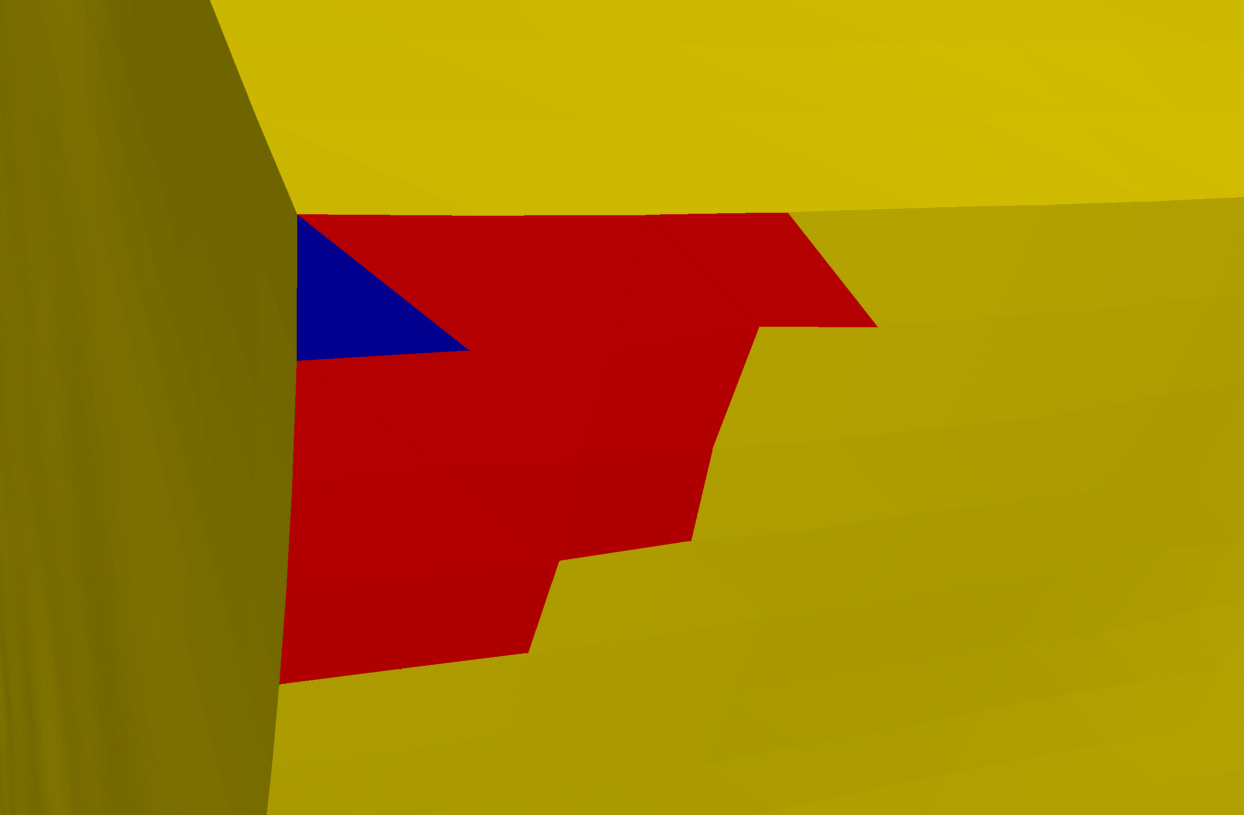} }}
	\caption{ The local binary neighbor selection in the proposed algorithm. Intially, it selects very few neighbor elements $f_j$ (red color) around the central element $f_i$(blue) because of the dihedral angle threshold. As iterations increase, it selects more elements with similar element normals.   }%
	\label{fig:neighC}%
\end{figure}

 The geometric neighborhood elements are weighted based on an angle threshold value $\rho$ \cite{LBP}. Based on $\rho$, we assign a binary value to the neighborhood elements $f_j$ w.r.t. the central element $f_i$ using the following function:
\begin{equation}
w_{ij} = \begin{cases} 1 &\mbox{if } \angle (\mathbf{n}_i,\mathbf{n}_j)\leq \rho \\
0.1 & \mbox{if } \angle (\mathbf{n}_i,\mathbf{n}_j) > \rho, \end{cases}
\label{equ:LBP} 
\end{equation} 
where $\mathbf{n}_i$ and $\mathbf{n}_j$ are the face normals of the central element and the neighbor elements. By using the value of 0.1, close to a feature, we still allow the area on the other side of the feature to contribute (so the edge direction can be detected from the computed tensor), but the area on the "same" side of the feature will be dominant. Figure~\ref{fig:neighW} shows that the contribution of the other side of the feature helps to enhance the sharp corner (Figure \ref{fig:neighW}(d)). Equation \ref{equ:LBP} shows a discontinuous box filter which takes similar faces into consideration and avoids blurring  features within the user defined geometric neighborhood.  Figure~\ref{fig:neighC} shows that the weighting function depends on the dihedral angle, which can be unstable intially but stabilizes after a few iterations. In further discussion, local binary neighbor refers to \mbox{Equation \ref{equ:LBP}}. 

\subsection{Element-based Normal Voting Tensor (ENVT)}
\label{ENVT}
We define an ENVT on every element of a properly oriented triangulated mesh similar to the vertex based voting tensor proposed in \cite{NVT}. The ENVT $C_i$ is a covariance matrix, defined on the face $f_i$:
\begin{equation}
C_i=\frac{1}{\sum_{j \in \Omega_i }^{} w_{ij}} \sum_{j\in\Omega}^{} w_{ij} A_j  \mathbf{n}_j \cdot \mathbf{n}_j^T,
\end{equation}
where $A_j$ is the area of the corresponding neighbor element $f_j$ and $w_{ij}$ is the weighting function as mentioned in Equation~\ref{equ:LBP}. Weighting by corresponding element areas makes the ENVT more robust against irregular sampling. 
 The eigenanalyis of the given tensor identifies features on triangulated surfaces similar to the methods \cite{NVT} and \cite{NVTsmooth}. In our algorithm, the ENVT is represented as a mesh denoising operator which is able to suppress the noise contents from noisy surfaces while preserving sharp features. The similarity between the ENVT and the shape operator is discussed in Appendix \ref{app:B}. The ENVT 
 is a symmetric and positive semi definite matrix so we can represent $C_i$ using an orthonormal basis of the eigenvectors $\mathbf{e}_k$ and real eigenvalues $\lambda_k$:
\begin{equation}
C_i = \sum_{k=0}^{2}\lambda_k\mathbf{e}_k\mathbf{e}_k^T.
\end{equation}  
 
\textbf{Geometrical Interpretation:} On a noise free triangulated mesh, a planar area has only one dominant eigenvalue in surface normal direction. Two dominant eigenvalues indicate edge features where the weakest eigenvector will be along the edge direction. At a corner all three eigenvalues are dominant. Consider a cube model where the eigenvalues of the ENVT are sorted in decreasing order ($\lambda_1\geq\lambda_2\geq\lambda_3\geq0$) and normalized, then for orthogonal features we can write: $\{\lambda_1=1, \lambda_2=\lambda_3=0\}$ (face), $\{\lambda_1=\lambda_2=\frac{\sqrt{2}}{2}, \lambda_3=0\}$ (edge) and $\{\lambda_1= \lambda_2=\lambda_3=\frac{\sqrt{3}}{3}\}$ (corner).

\subsection{Eigenvalues Binary Optimization }
\label{sec:binaryOpti}
Let us consider a noisy mesh, corrupted by a random noise with standard deviation $\sigma_n$ bounded by minimum edge length. On a planar area (face) of the geometry: $\lambda_1 \gg \sigma_n$, and the other two eigenvalues will be proportional to noise intensity, $\lambda_2,\lambda_3 \propto \sigma_n$. Similarly, on an edge of the geometry: $\lambda_1, \lambda_2 \gg \sigma_n$ and $\lambda_3 \propto \sigma_n$. On a corner of the geometry: $\lambda_1,\lambda_2, \lambda_3 \gg \sigma_n$.

We apply binary optimization to remove noise effectively by setting the less dominant eigenvalues to zero and the dominant eigenvalues to one. Our optimization technique removes noise not only from the planar area but also along the edge direction of sharp features during the denoising process.
We implemented the binary optimization by introducing a scalar threshold value $\tau$ which is proportional to the noise intensity $\tau \propto \sigma_n$ and smaller than the dominant eigenvalues. The $\tilde{\lambda}$ are modified eigenvalues of the ENVT using the following optimization technique. There are three eigenvalues for feature classification so our optimization method checks the following three cases:   
\begin{itemize}
	\item At corners of noisy surfaces (smooth or sharp), the smallest eigenvalues should be bigger than the threshold value i.e. $\lambda_3\geq \tau$. Hence:
	\begin{equation*}
	\begin{aligned} \tilde{\lambda}_i= 1 , \quad i \in \{1,2,3\} \quad &\mbox{if} \quad \lambda_3\geq \tau.  \end{aligned} 
	\end{equation*}
	
	\item At edges of noisy geometry (smooth or sharp), the less dominant eigenvalue should be smaller than the threshold value i.e. $\lambda_3 < \tau $ and $\lambda_2 \geq \tau $. Hence:
	\begin{equation*}
	\begin{aligned} \tilde{\lambda}_2=\tilde{\lambda}_1=1 \text{,} \quad{\tilde{\lambda}_3=0} \quad &\mbox  {if } \quad \lambda_2\geq \tau \text{,} \quad{\lambda_3 < \tau}.   
	\end{aligned} 
	\end{equation*}
	
	\item In the last case, we check for planar area of the geometry. Having $\lambda_2< \tau$ and $\lambda_3< \tau$ show that the only dominant eigenvalue is $\lambda_1$. Hence:
	\begin{equation*}
	\begin{aligned} \tilde{\lambda}_1=1 \text{,} \quad{\tilde{\lambda}_2=\tilde{\lambda}_3=0} \quad &\mbox{if } \quad \lambda_1\geq \tau \text{,} \quad{\lambda_3\text{,}\lambda_2 < \tau.}   \end{aligned} 
	\end{equation*}  
\end{itemize} 
There are the three possible combinations during the eigenvalue binary optimization. The threshold $\tau$ has to be set by the user according to the noise intensity.
\begin{figure}[h]%
	\centering
	\subfloat[Noisy]{{\includegraphics[width=2.1cm]{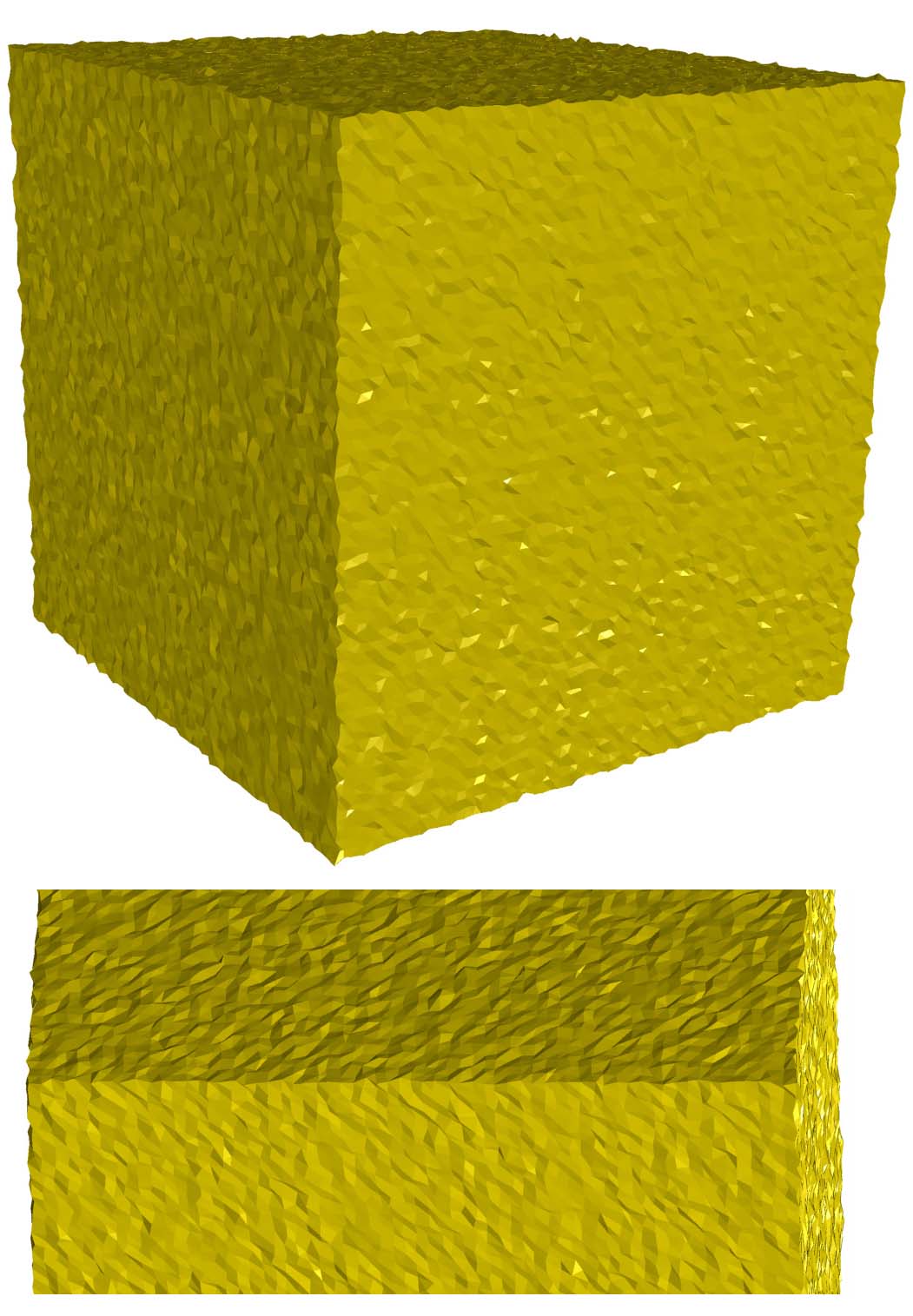}}}%
	\subfloat[]{{\includegraphics[width=2.1cm]{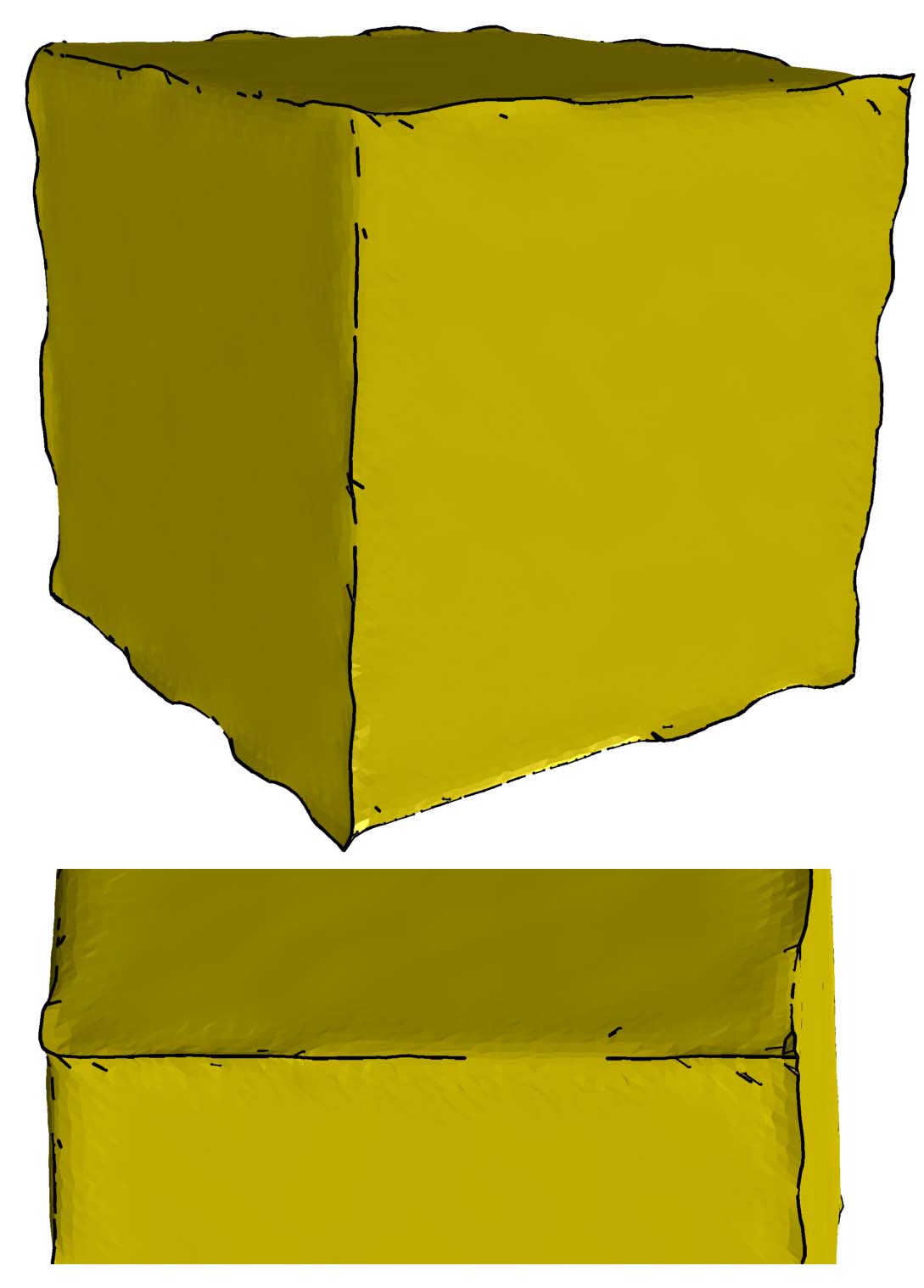}}}%
	\subfloat[]{{\includegraphics[width=2.1cm]{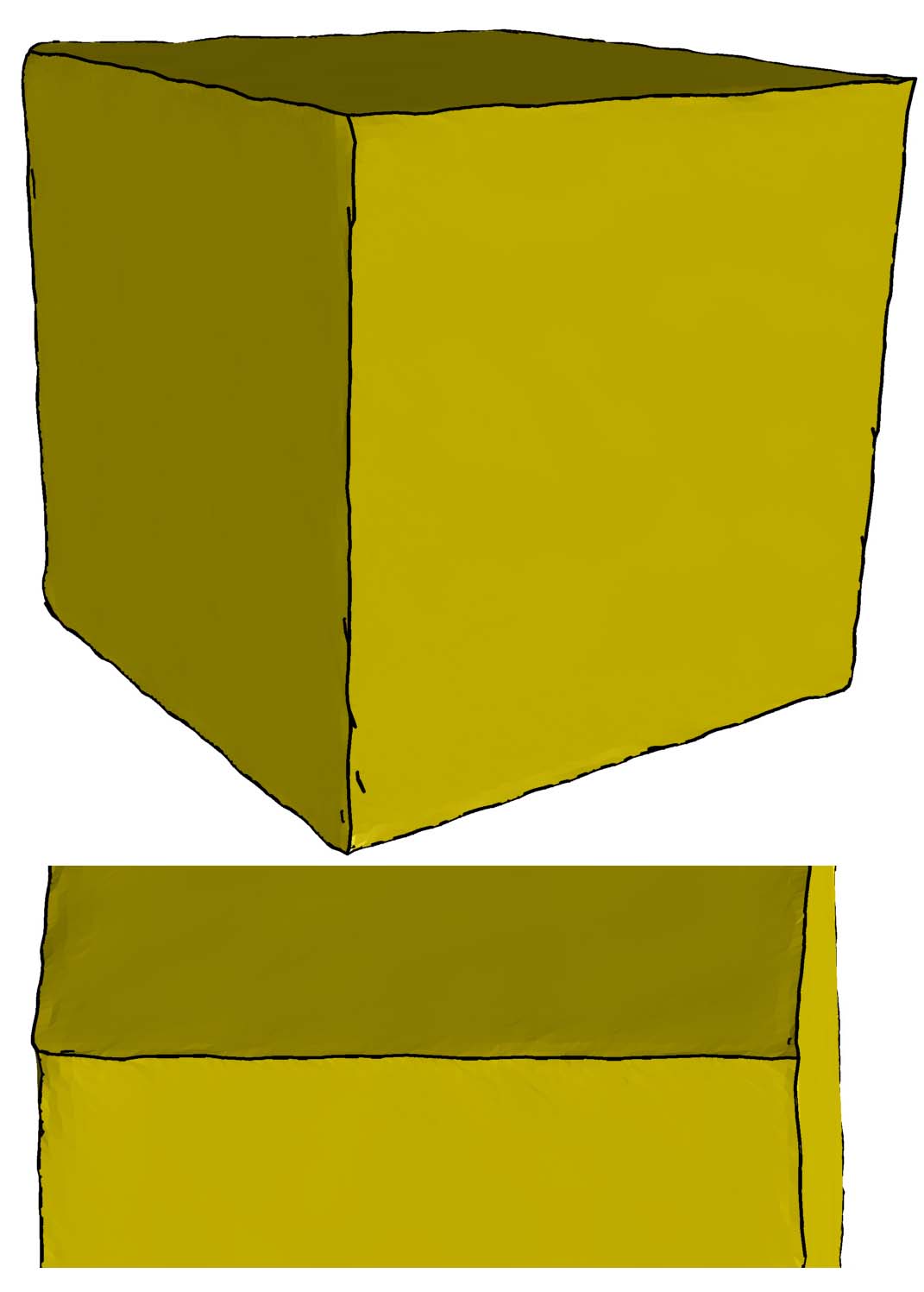}}}%
	\subfloat[]{{\includegraphics[width=2.1cm]{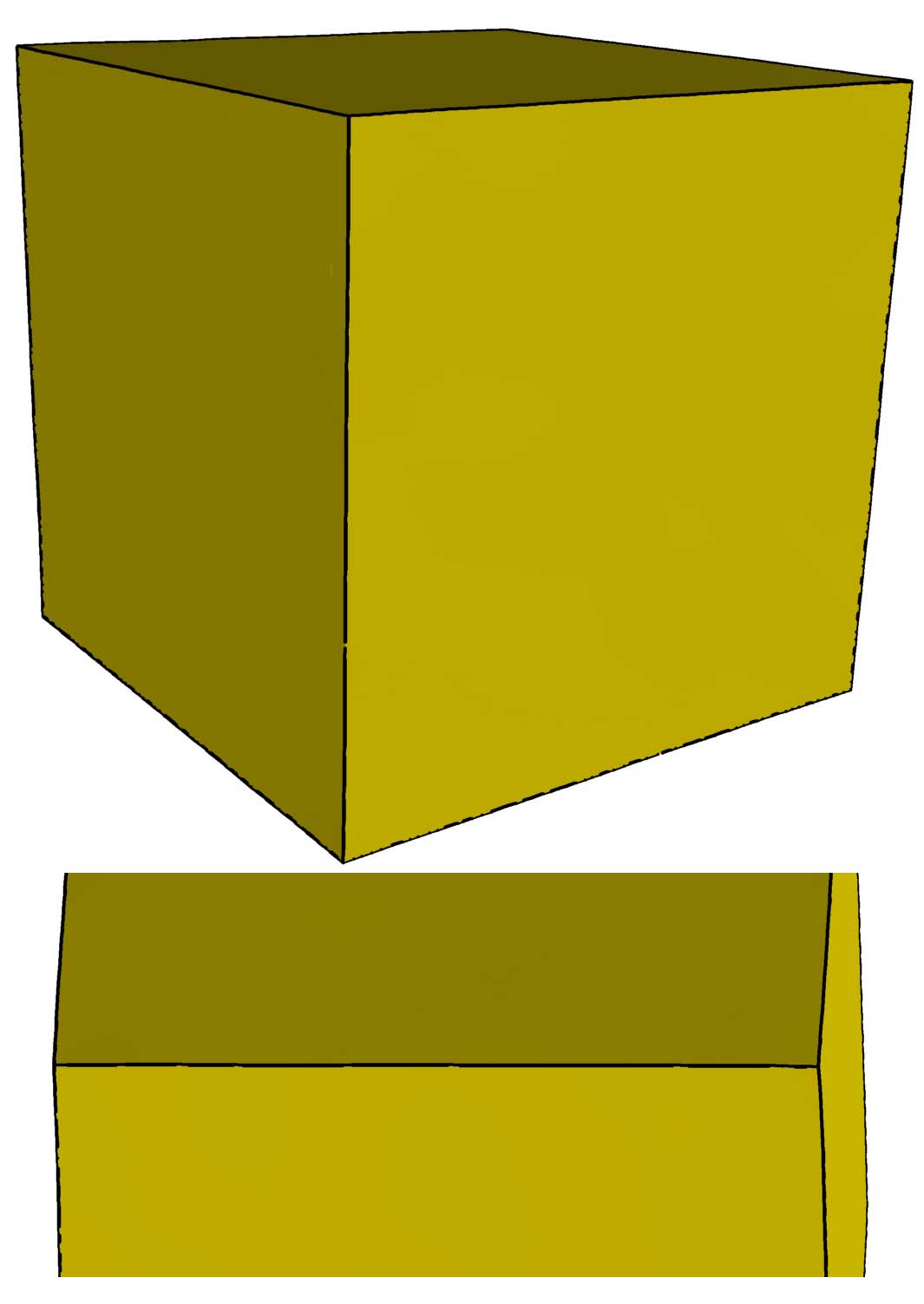}}}%
	\caption{ The results with the different combinations of steps of the proposed algorithm. (a) Noisy cube model, (b) Without using the eigenvalues binary optimization and local binary neighborhood selection. (c) Without the local binary neighbor selection. (d) Using both eigenvalues binary optimization and local binary neighborhood selection. The black curve shows the edge information in the smooth geometry and it is detected using the dihedral angle.}%
	\label{fig:stages}%
\end{figure}

\subsection{De-Noising using ENVT}
 
Our denoising method is inspired by the feature classification characteristic of the eigenvalues of the ENVT. The smallest eigendirections (one for edges of the geometry, two in planar areas) represent noise. Multiplication of the ENVT with the corresponding element normal will suppress noise in the weak eigendirection. This operation also strengthens the element normal in the strongest eigendirection. The visual representation of this operation is shown and explained in Appendix \ref{denoise}.  
\subsubsection{Anisotropic Face Normal Denoising}
\label{sec:AFNP}
\begin{figure}[h]%
	\centering
	\subfloat[Noisy input]{{\includegraphics[width=2.1cm]{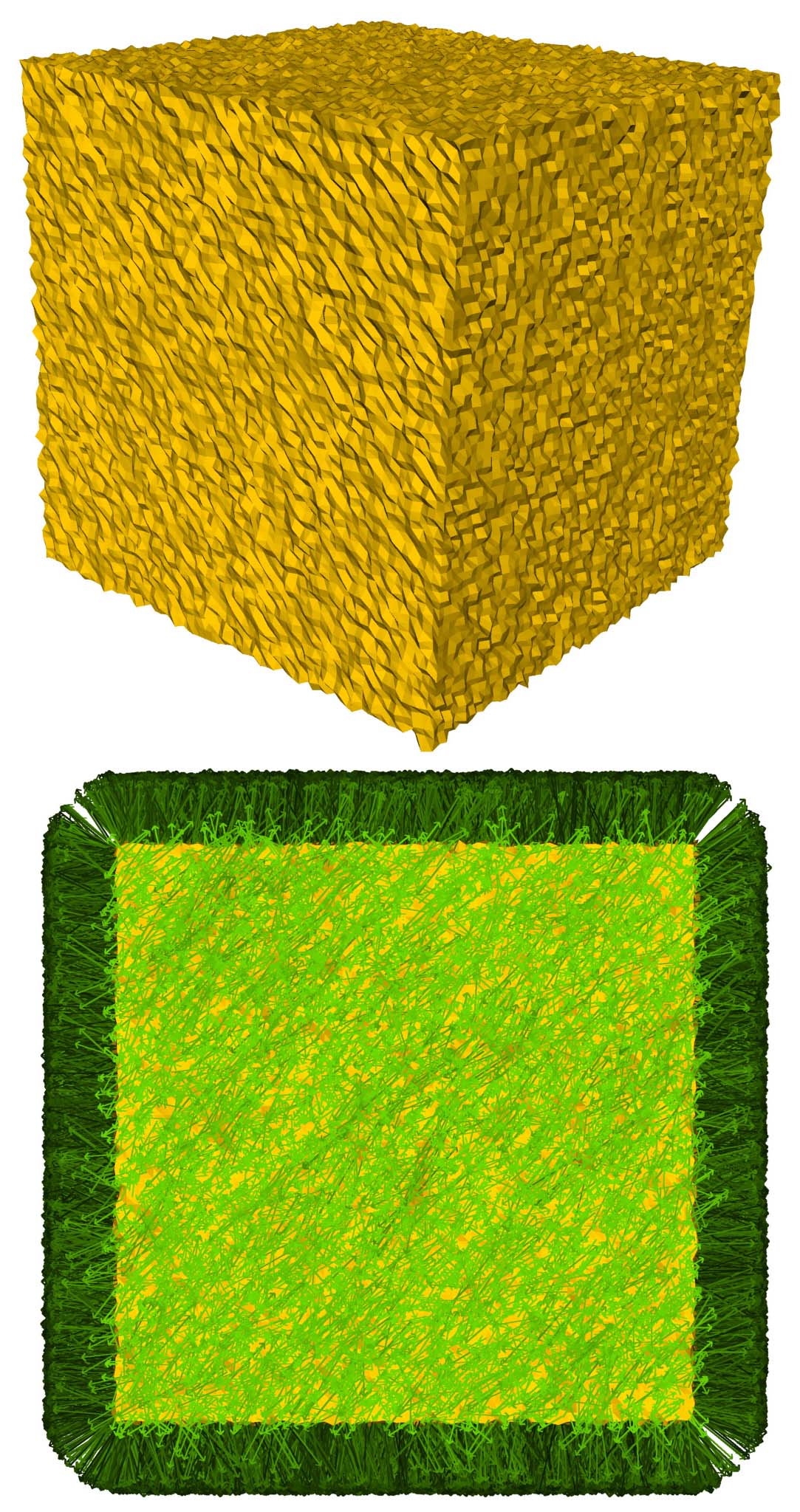}}}%
	\subfloat[30 iterations]{{\includegraphics[width=2.1cm]{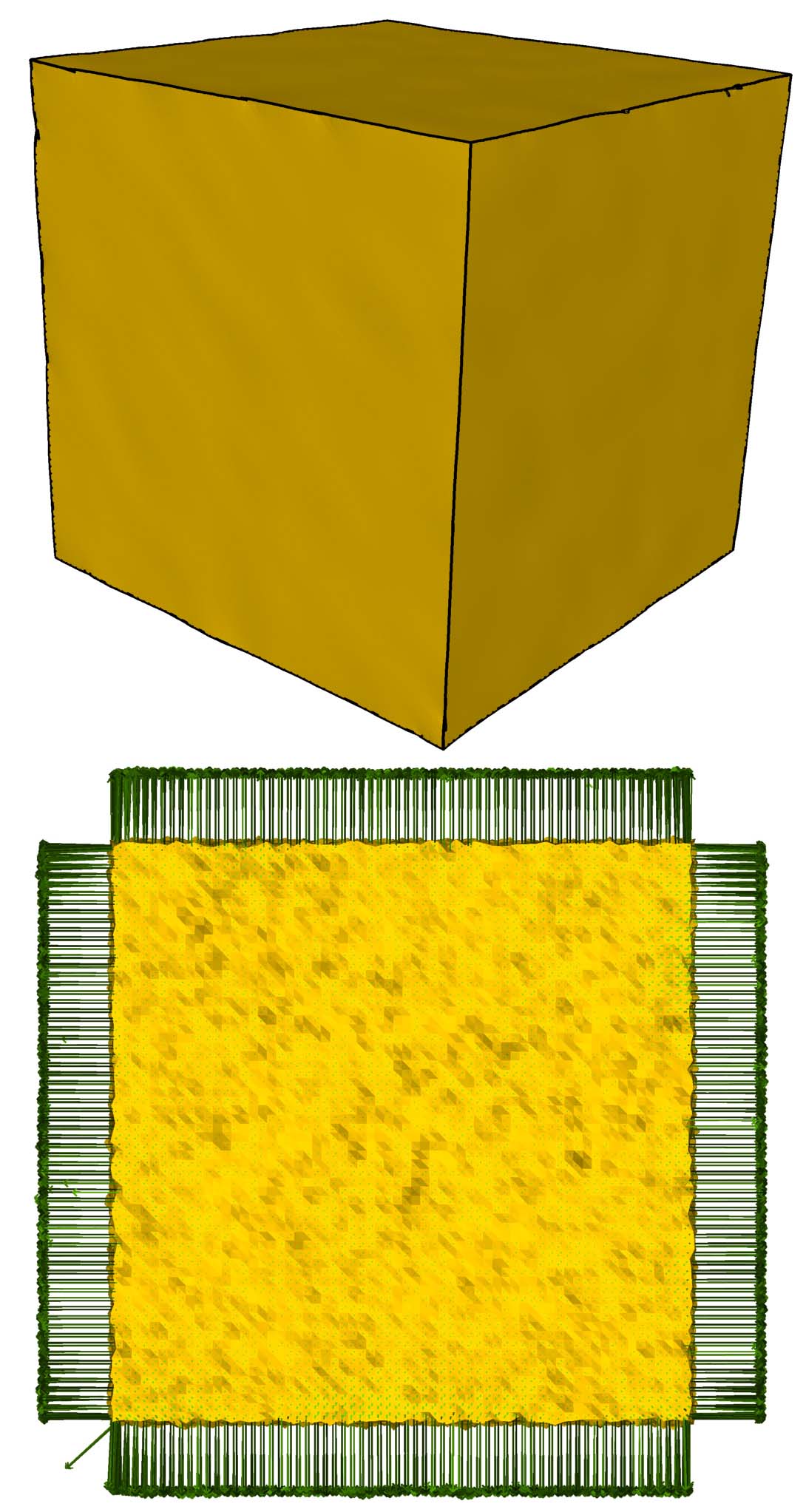}}}%
	\subfloat[60 iterations]{{\includegraphics[width=2.1cm]{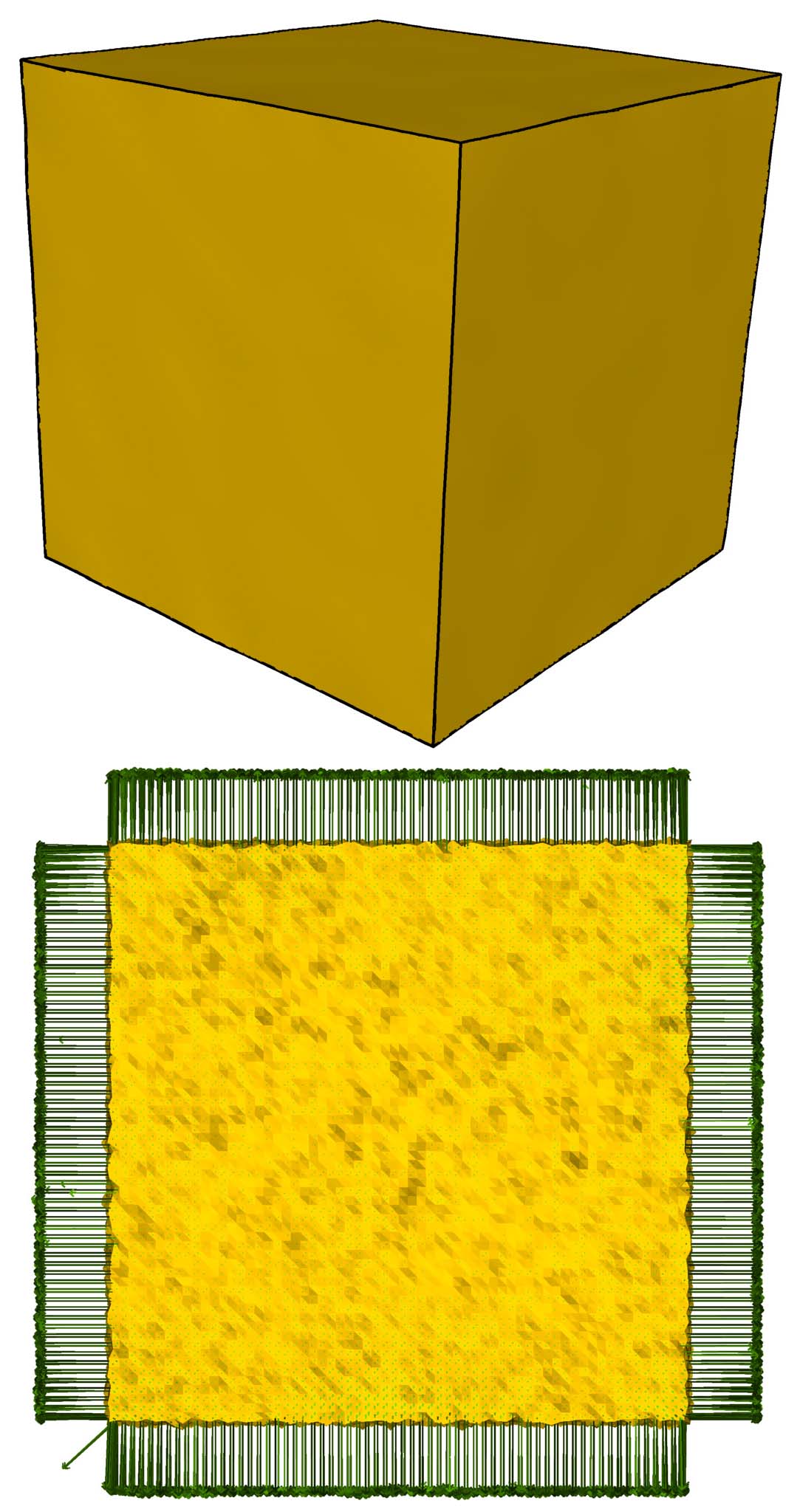}}}%
	\subfloat[200 iterations]{{\includegraphics[width=2.1cm]{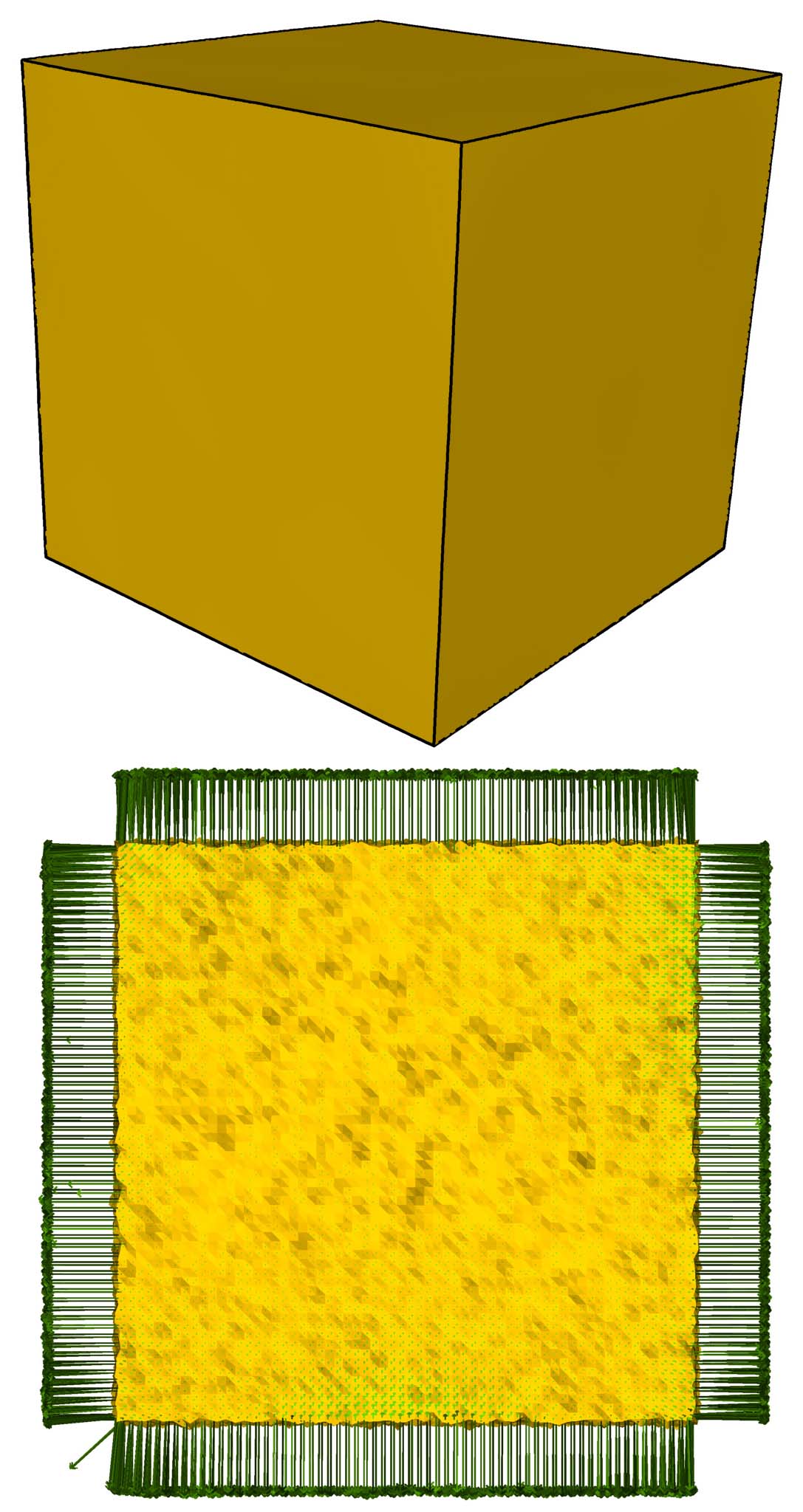}}}%
	\caption{Stable convergence with different number of iterations and the corresponding processed face normals (XY-plane view). (a) Noisy cube model, (b) The result after 30 iterations, low frequency noise can be seen on the face of the Cube model. (c) After 60 iterations, smoother compared to Figure (b). (d) After 200 iterations. There is no significant difference between Figure (c) and (d). }%
	\label{fig:iteration}%
\end{figure}

We recompute the ENVT by using the same eigenvectors with modified eigenvalues:
\begin{equation}
\label{equ:aniso}
\tilde{C}_f=\sum_{k=0}^{2} \tilde{\lambda}_k \mathbf{e}_k \mathbf{e}_k^T.
\end{equation}
Now, $\tilde{C_f}$ will have the quantized eigenvalues according to the different features on the surface. To remove noise, we multiply the corresponding element normal with the newly computed tensor $\tilde{C_f}$. The multiplication  will lead to noise removal while retaining sharp features. 
\begin{equation}
\label{equ:faceP}
\tilde{\mathbf{n}}_i=d\mathbf{n}_i+ \tilde{C}_f \mathbf{n}_i = d\mathbf{n}_i+ \sum_{k=0}^{2} \tilde{\lambda}_k \langle \mathbf{e}_k,\mathbf{n}_i\rangle \mathbf{e}_k,
\end{equation}  
where 
$d$ is the damping factor to control the speed of preprocessing of the face normals. We use $d=3$ for all experiments. The second row of Figure \ref{fig:iteration} shows the face normal denoising using the tensor multiplication.

\subsubsection{Vertex Update} 
\label{sec:vertUpdate}
In the last stage of the denoising algorithm, we synchronize the vertex position with corresponding newly computed face normals. To compute the proper vertex position, the orthogonality between edge vectors and face normals is used \cite{vertexUpdate}. An energy function is then defined as follows:
\begin{equation}
\begin{aligned}
& \underset{v_i}{\text{min}}
& & \sum_{j = 0 }^{N_v(i)-1}\sum_{(i,j)\in \partial F_k}^{} \lVert \tilde{\mathbf{n}}_k\cdot (v_i-v_j)\rVert^2,
\end{aligned}
\label{equ:vertUp}
\end{equation}
where $v_i$ is the vertex position and $N_v$ represents the number of vertices of the vertex star of $v_i$, $\partial F_k$ is the boundary edge of the vertex star of $v_i$ shared with face $f_k$. $\tilde{\mathbf{n}_k}$ is the smooth face normal at $f_k$. Taubin \cite{Taubin01linearanisotropic} explained that the face normal vector can be decomposed into a normal and a tangential component and the main problem here is to find the vertex positions which minimize the tangential error. The possible solution of Equation~\ref{equ:vertUp} may be a mesh with degenerate triangular faces. Like Taubin \cite{Taubin01linearanisotropic}, to avoid the degenerate solution, we are using gradient descent that will lead to the optimal vertex positions.
\begin{equation}
\tilde{v}_i= v_i + \frac{1}{F(v_i)}\sum_{j = 0 }^{N_v(i)-1}\sum_{(i,j)\in \partial F_k}^{} \tilde{\mathbf{n}}_k\cdot (v_i-v_j), 
\label{equ:dampingFactor}
\end{equation}
 where $F(v_i)$ is the number of faces connected to the vertex $v_i$. We iterate the whole procedure several times and the number of iterations depends on the noise intensity. Figure \ref{fig:stages} shows the effect of each stage of the face normal processing in the proposed algorithm.

\subsection{Effect of Noise on the Proposed Method}
\label{sec:noise}
Noise is inevitable during digital data acquisition of real life objects. The high intensity of noise flips edges in the geometry and that leads to the inconsistent face normals on a geometry. As we mentioned in section 3.2, the ENVT is defined on properly oriented surfaces with consistence face normals because the spectral decomposition of the ENVT is invariant to the face normal orientation. 
In this section, we give a stochastic approximation about the relation between noise and geometry resolution to prevent edge flips in the geometry. Let us consider a smooth triangular mesh $\mathcal{M}_s$ which is corrupted by noise $\mathcal{N}$, $\mathcal{M}=\mathcal{M}_s+\mathcal{N}$. 
The noise $\mathcal{N}$ can be approximated by a random vector $X_n$ consisting of three independent random variables. 
We assume, that the random vector $X_n$ follows the Gaussian distribution; this is a realistic model for noise from 3D scanning \cite{noiseScanner}.
Let $\sigma_n$ be the standard deviation of noise in each independent direction, then:
\begin{align*} 
P\{\lvert X_n \rvert \leq \sigma_n \} &=0.682, \\ 
P\{\lvert X_n \rvert\leq 2\sigma_n\}&=0.954, \\
P\{\lvert X_n \rvert\leq 3\sigma_n\}&=0.997.
\end{align*}

To explain the probability of normal flips, we switch to the 2D case of a polygon in $\mathbb{R}^2$.
Let us consider an edge vector $l$ between two vertices $v_0$ and $v_1$ in $\mathbb{R}^2$: $l =(\vec{v_0}-\vec{v_1})$. We give a probabilistic estimation of the effect of noise on the edge $l$ w.r.t. the noise intensity (standard deviation) $\sigma_n$. Our analysis is mainly focused on the proper orientation of the edge normal. Wrong orientation of the edge normal $\mathbf{n}_l$ leads to an edge flip in the smooth geometry. We denote by $\Omega_1$ and $\Omega_2$ the sets of correctly oriented edge normals and wrong oriented edge normals respectively. The probabilistic estimation of the orientation of the edge normals based on noise intensity and edge length is given as follows:
\begin{itemize}
	\item Probability of an edge to have a correctly oriented edge normal:
	\begin{equation*}
	P\{\vec{\mathbf{n}_l}\in \Omega_1 \} \geq \begin{cases} 0.682 &\mbox{if } \sigma_n \leq \frac{\lvert l \rvert}{2} \\
	0.954  & \mbox{if } \sigma_n \leq \frac{\lvert l \rvert}{4}. \end{cases}
	\end{equation*}
	\item Similarly, the probability of an edge to have a wrong oriented edge normal:
	\begin{equation*}
	P\{\vec{\mathbf{n}_l}\in \Omega_2 \} \leq \begin{cases} 0.318 &\mbox{if } \sigma_n \leq \frac{\lvert l \rvert}{2} \\
	0.046  & \mbox{if } \sigma_n \leq \frac{\lvert l \rvert}{4}. \end{cases}
	\end{equation*}
\end{itemize}
Due to the presence of noise, edge flipping may occur when the vector sum of the vertex dislocations at the edge is bigger than the edge length. 
This is similar to the sampling theorem, where a signal can be reconstructed properly if and only if the data is sampled with a frequency bigger than twice the highest frequency of the data signal.


Using the given analysis, for a given probability density function and an upper bound to the standard deviation, we can estimate the expected number of edge flips in the geometry.
If a surface is affected by noise only in normal direction, then there is no edge flip, irrespective of the probability density function of the noise.
 
\begin{figure}[h]%
	\centering
	\subfloat[Original]{{\includegraphics[width=1.6cm]{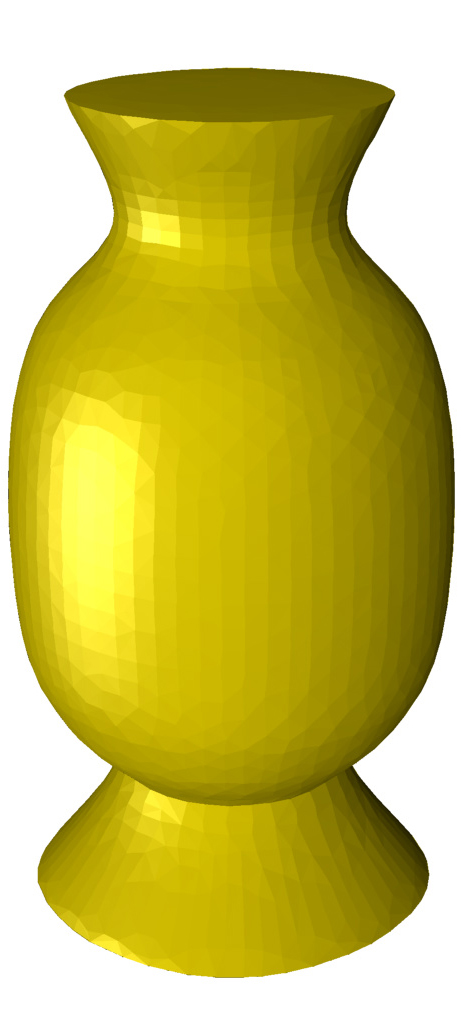} }}%
	\subfloat[Impulse]{{\includegraphics[width=1.6cm]{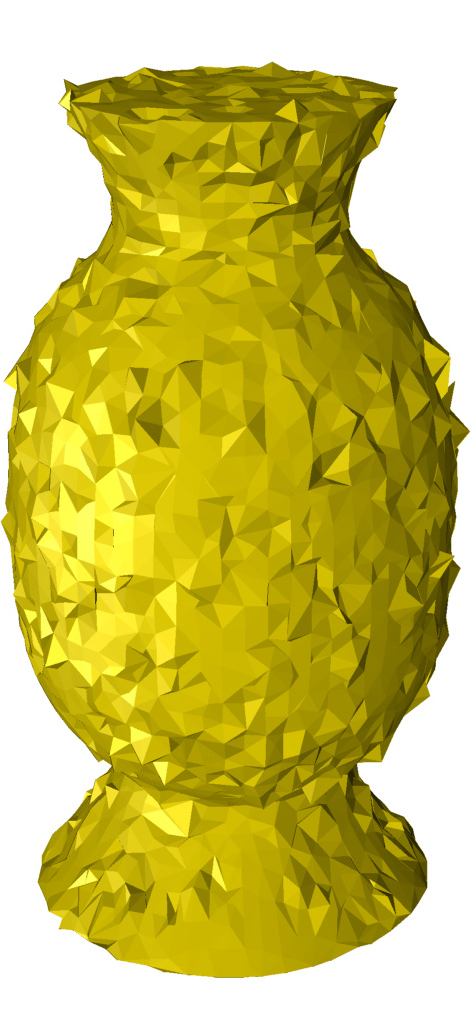} }}%
	\subfloat[Ours]{{\includegraphics[width=1.6cm]{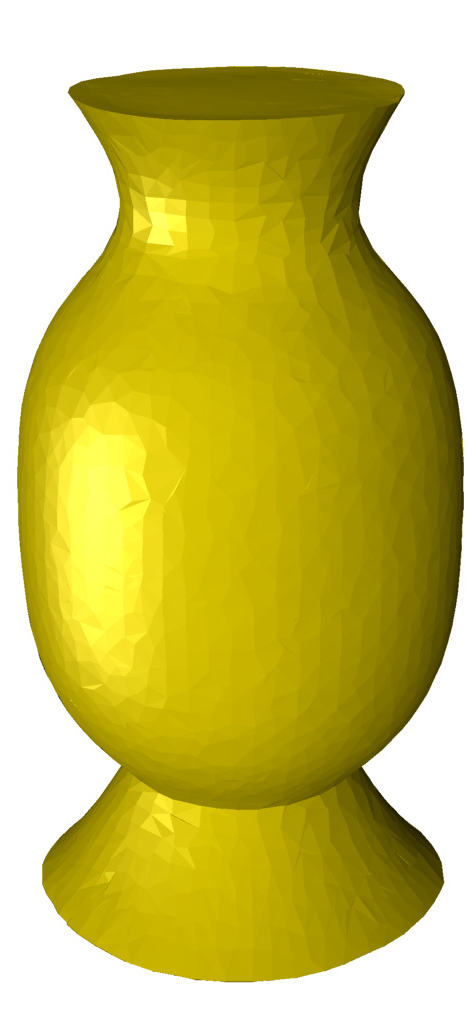} }}%
	\subfloat[Uniform]{{\includegraphics[width=1.6cm]{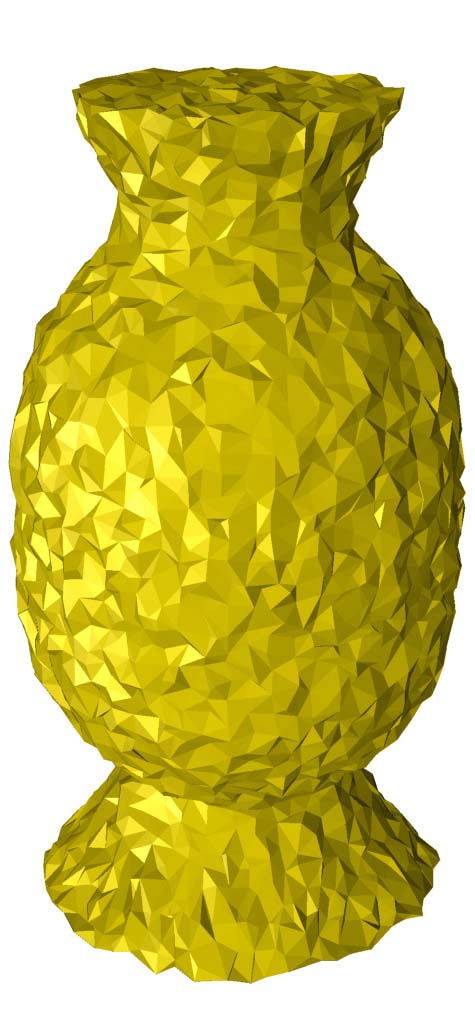} }}%
	\subfloat[Ours]{{\includegraphics[width=1.6cm]{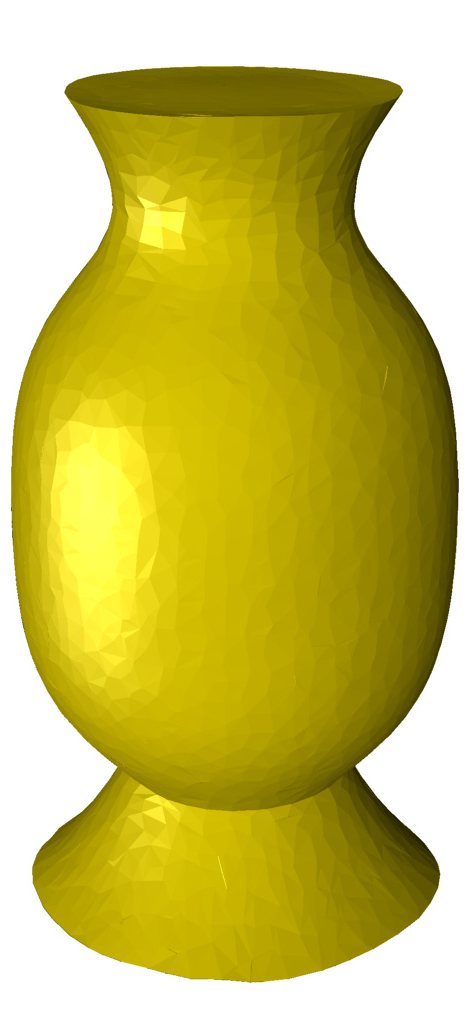} }}%
	\caption{The results obtained by our method against different kinds of noise. (a) Original vase model. (b) $1/3$ of the vertices of the vase model are corrupted by impulsive random noise. (c) Corresponding result with our method. (d) The vase model is corrupted by uniformly distributed noise and (e) corresponding result.}
	\label{fig:diffNoise}
\end{figure}

We also experimented with uniformly distributed noise where the random variable $X_n$ follows the uniform distribution so we can write:  $\	P\{\lvert X_n \rvert \leq \sigma_n \}= 1$. If the noise intensity is less than half of the minimum edge length in the geometry then there will be no edge flip as shown in \mbox{Figure ~\ref{fig:diffNoise}}.

\begin{figure*}[h]%
	\centering
	\subfloat[Noisy]{{\includegraphics[width=2.25cm]{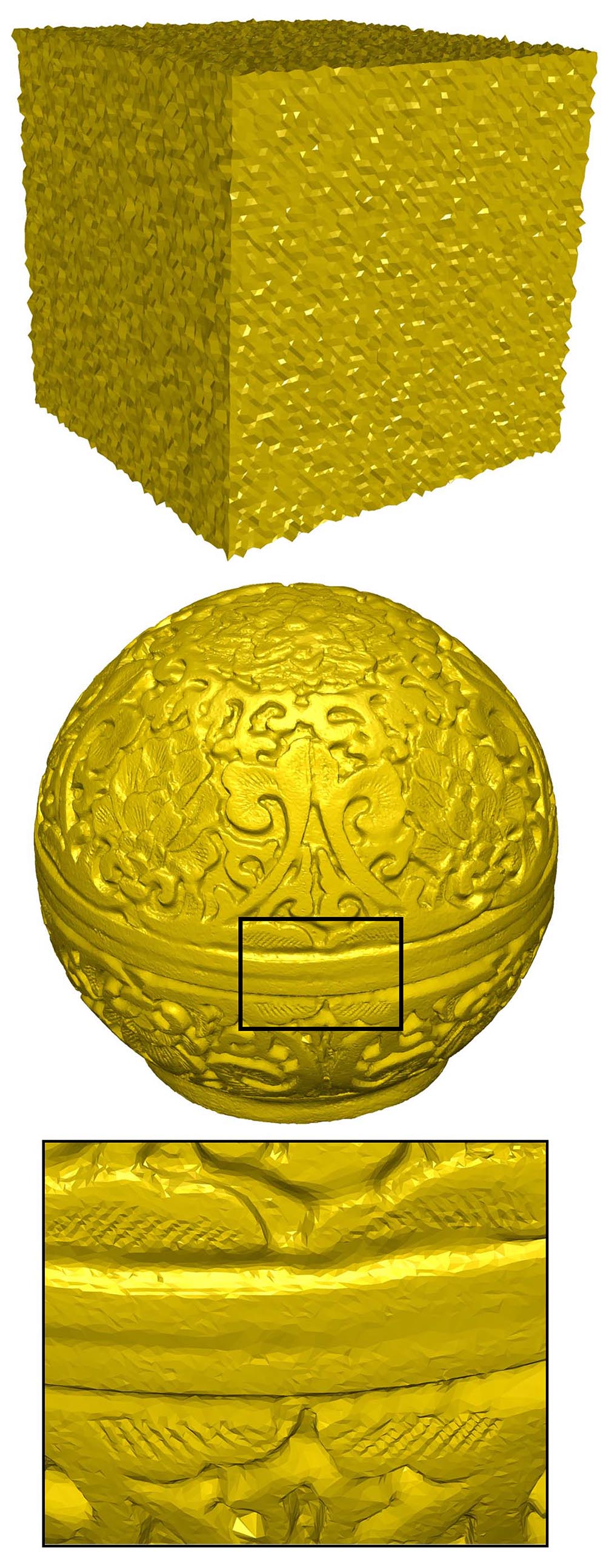} }}%
	\subfloat[{\boldmath$\tau =0.01$}]{{\includegraphics[width=2.25cm]{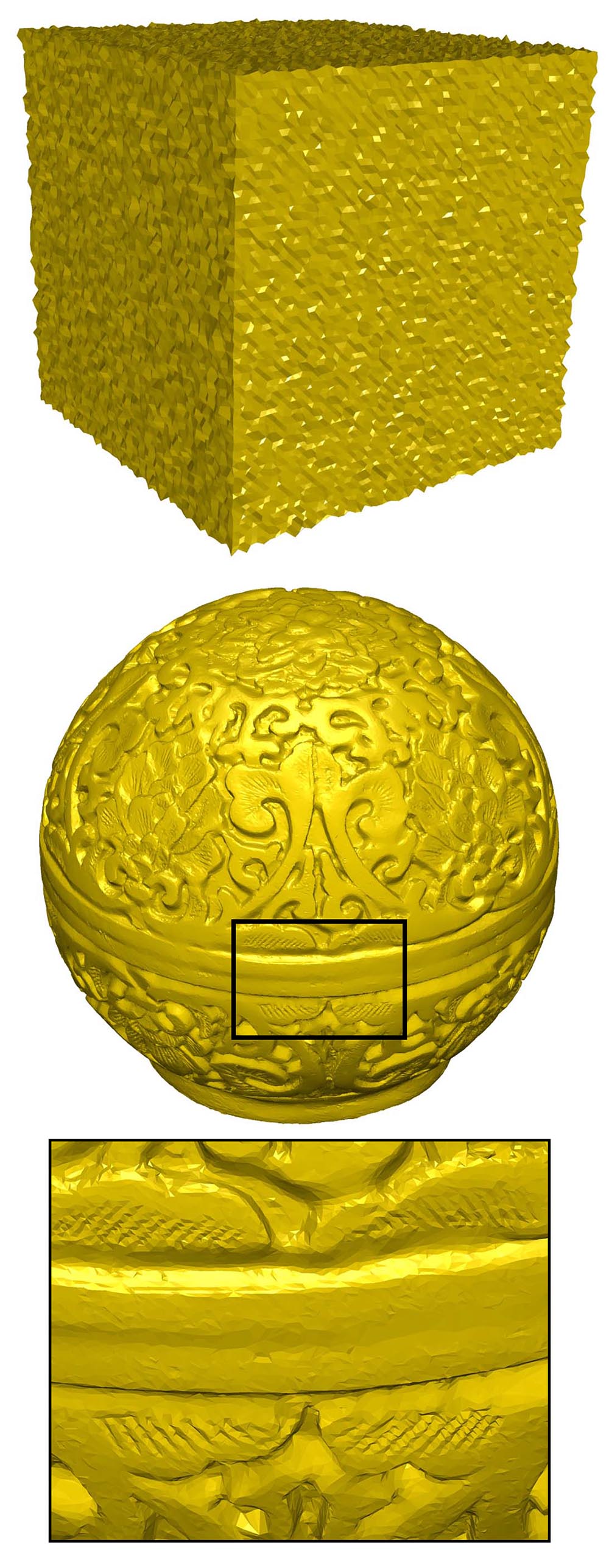} }}%
	\subfloat[{\boldmath$\tau = 0.02$}]{{\includegraphics[width=2.25cm]{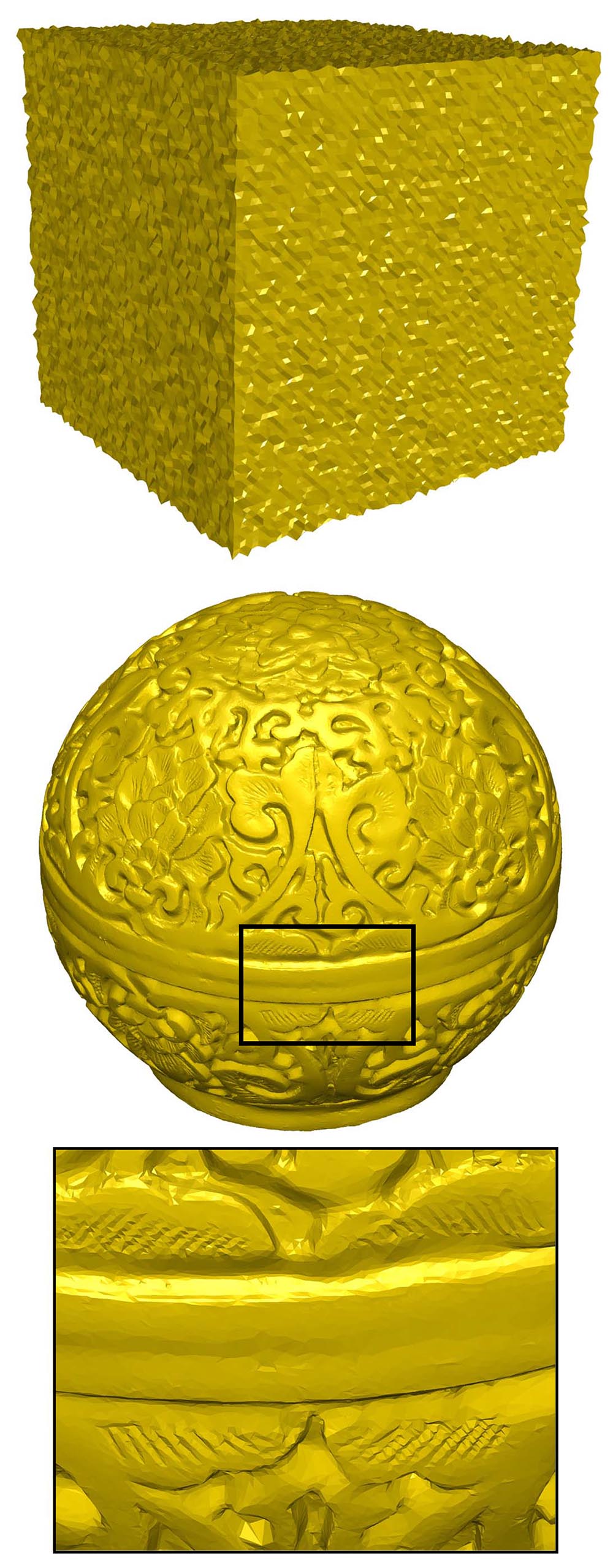} }}%
	\subfloat[{\boldmath$\tau =0.03$}]{{\includegraphics[width=2.2cm]{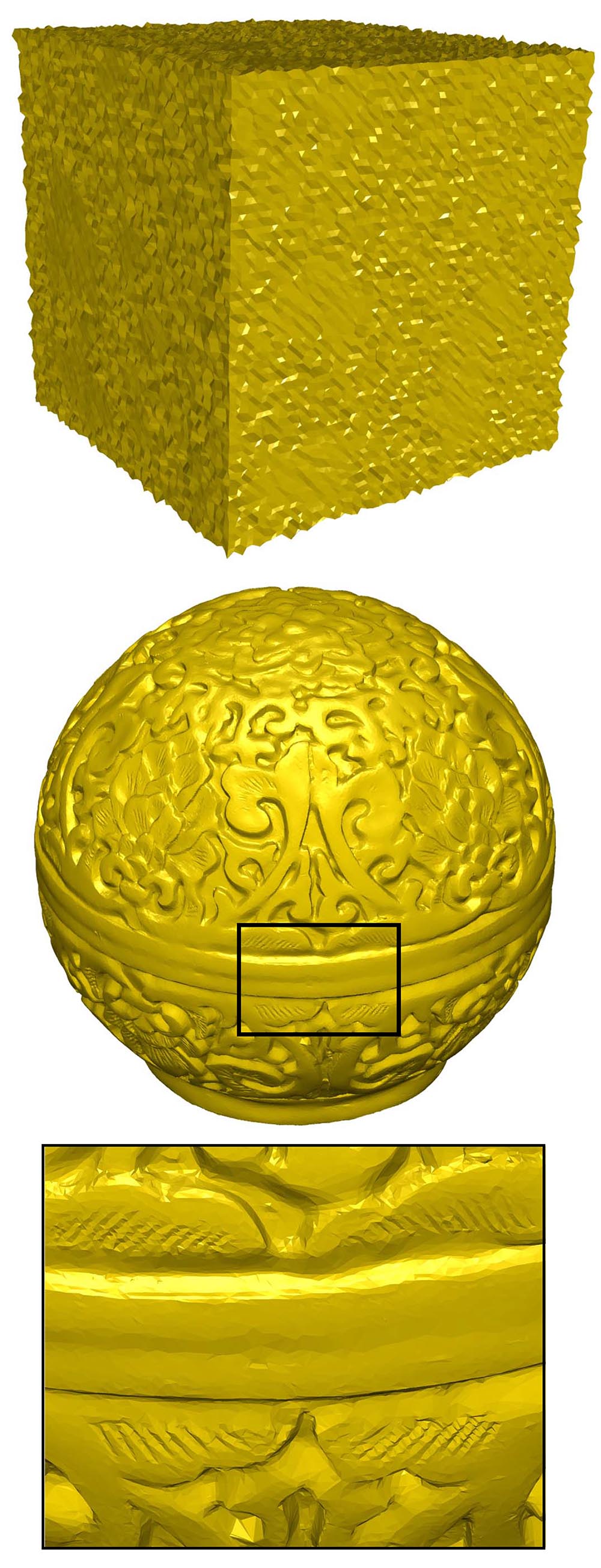} }}%
	\subfloat[{\boldmath$\tau =0.04$}]{{\includegraphics[width=2.2cm]{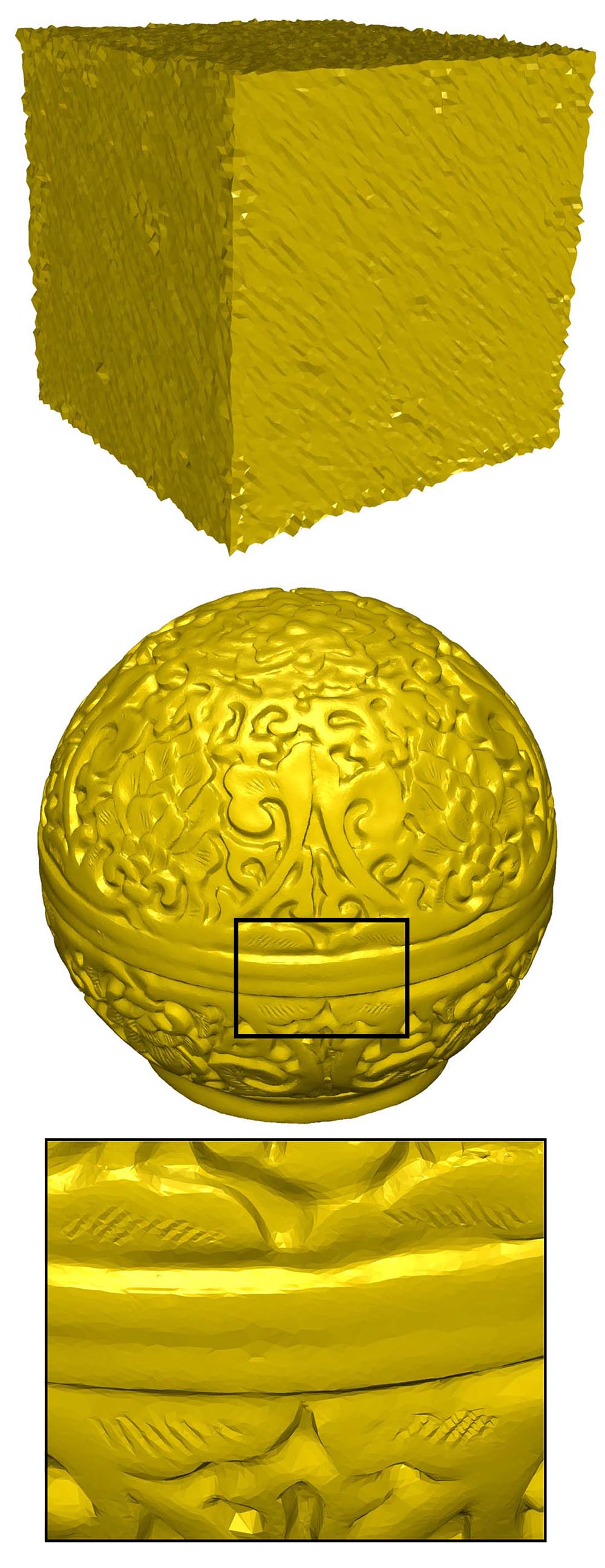} }}%
	\subfloat[{\boldmath$\tau =0.05$}]{{\includegraphics[width=2.25cm]{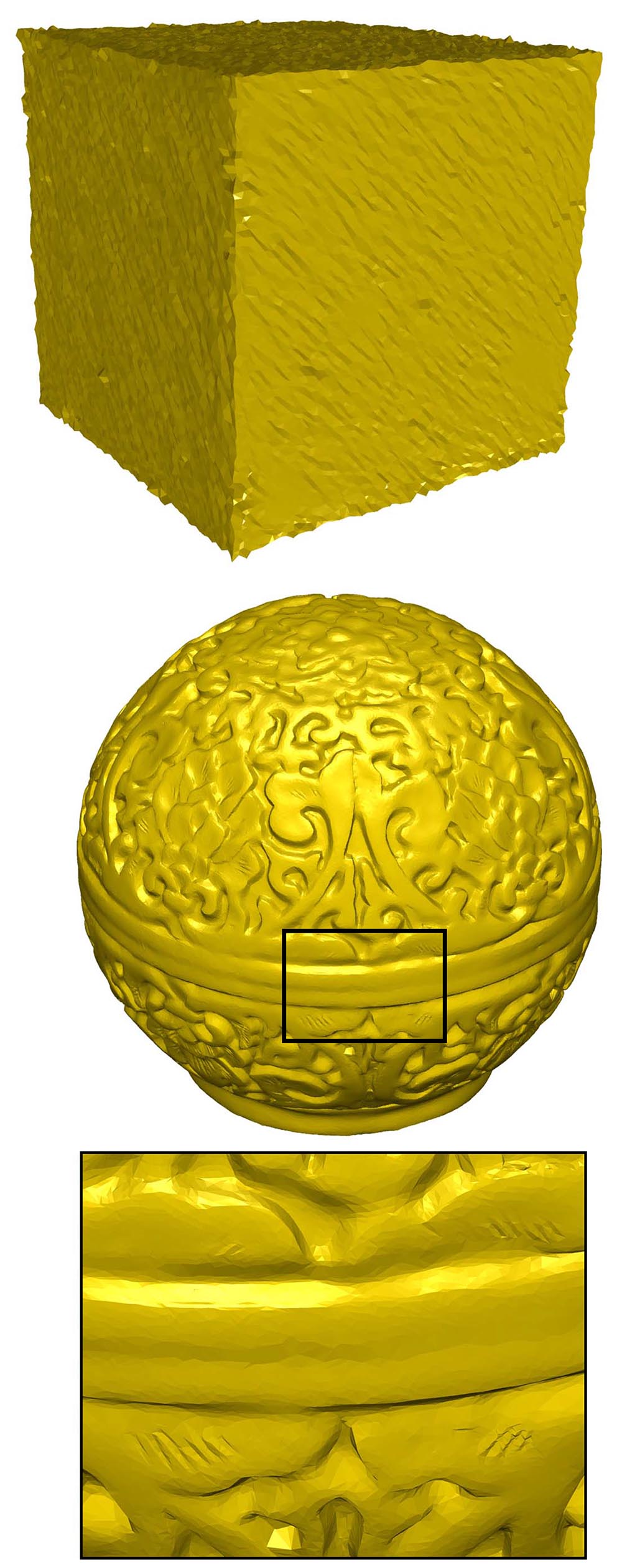} }}%
	\subfloat[{\boldmath$\tau =0.1$}]{{\includegraphics[width=2.2cm]{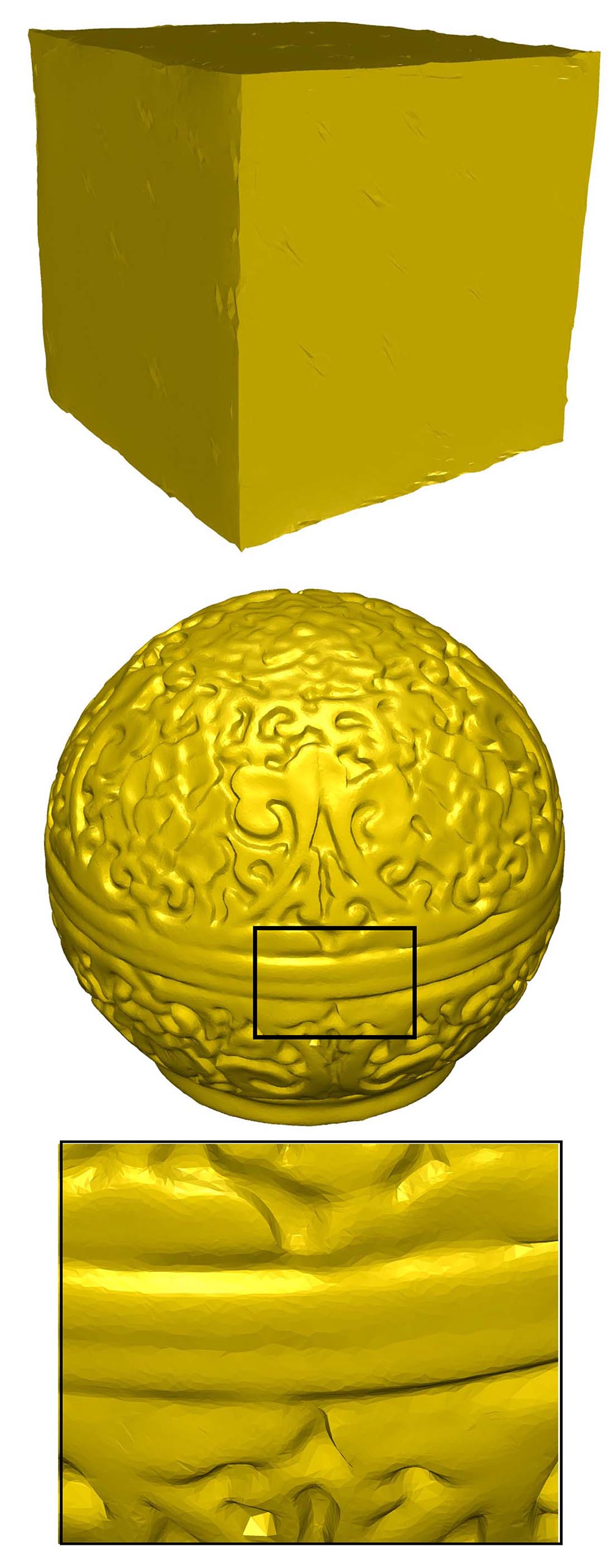} }}%
	\subfloat[{\boldmath$\tau =0.5$}]{{\includegraphics[width=2.15cm]{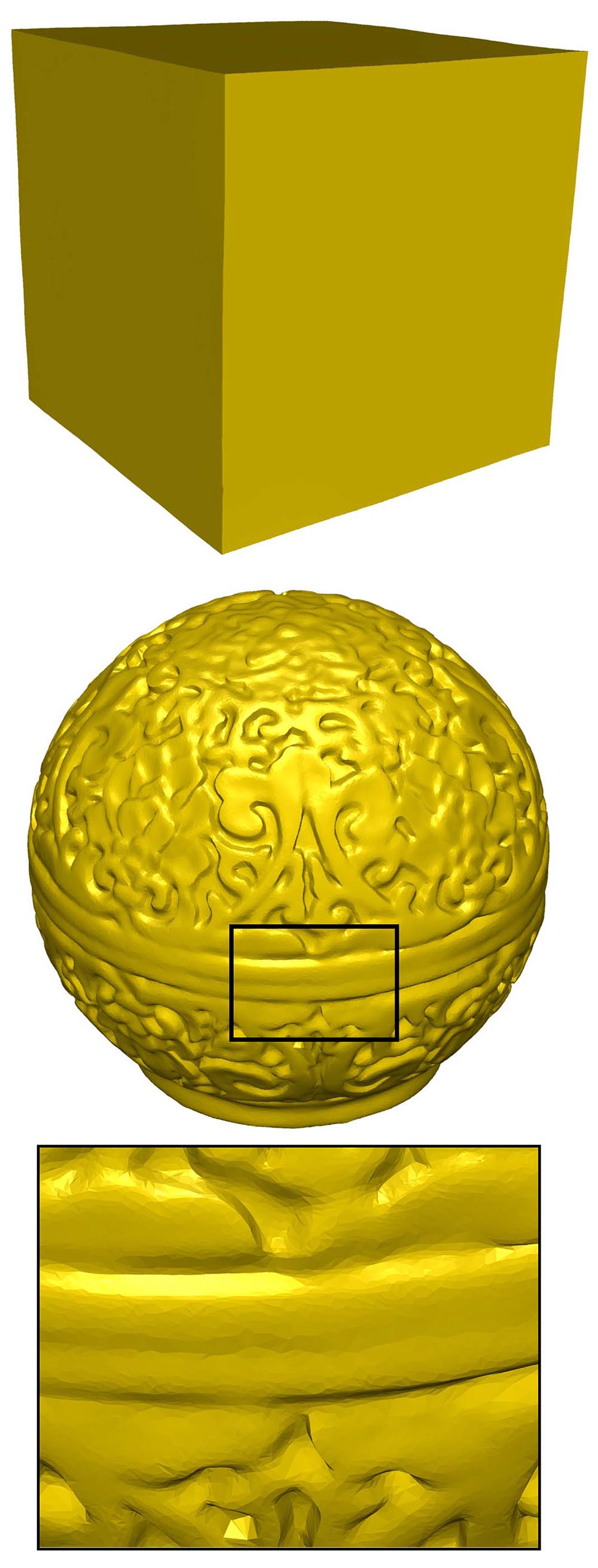} }}%
	\caption{The effect of the eigenvalue binary optimization threshold value {\boldmath$\tau$} on results of the proposed algorithm. The first row shows the cube model ($\lvert V \rvert=24578$, $\lvert F \rvert = 49152$) corrupted by synthetic Gaussian noise ($\sigma_n=0.4l_e$)  where $l_e$ is the average edge length and the second row shows the scanned box model ({real data} with $\lvert V \rvert = 149992$, $\lvert F \rvert = 299980$) and corresponding results regarding different values of {\boldmath$\tau$}. The third row shows the magnified area of the box model. For the box model, the algorithm produced the optimal result with smaller value of {\boldmath$\tau = 0.03$} because of low noise whereas the cube model needed the higher value of {\boldmath$\tau = 0.5$} because of high intensity noise. The bigger value of {\boldmath$\tau$} can lead to the feature blurring as shown in Figure (h) for the box model.}%
	\label{fig:tauinc}%
\end{figure*}

\begin{figure}[h]%
	\centering
	\subfloat[{\boldmath$\sigma_n = 0.4l_e$}]{{\includegraphics[width=2.1cm]{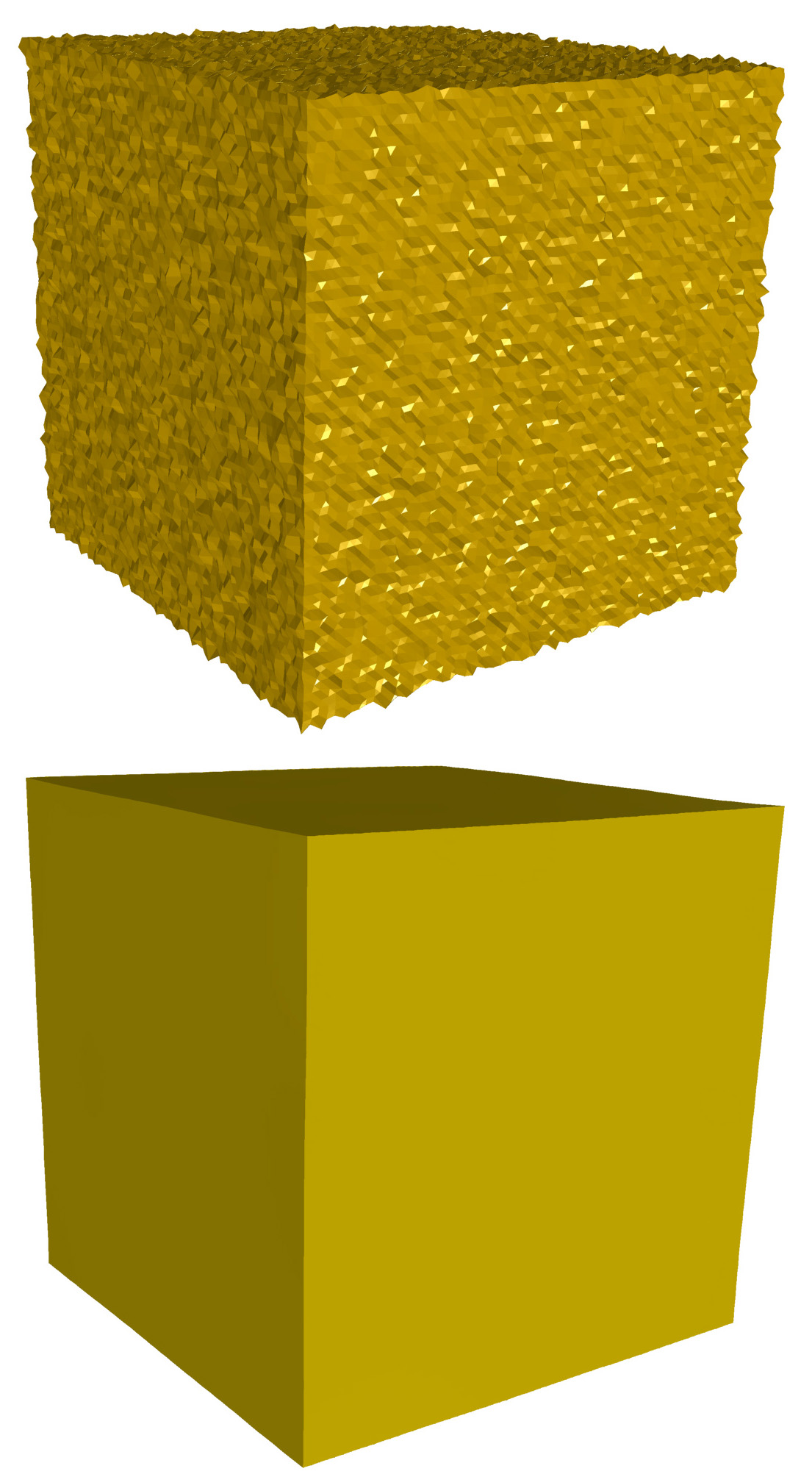}}}%
	\subfloat[{\boldmath$\sigma_n = 0.6l_e$}]{{\includegraphics[width=2.0cm]{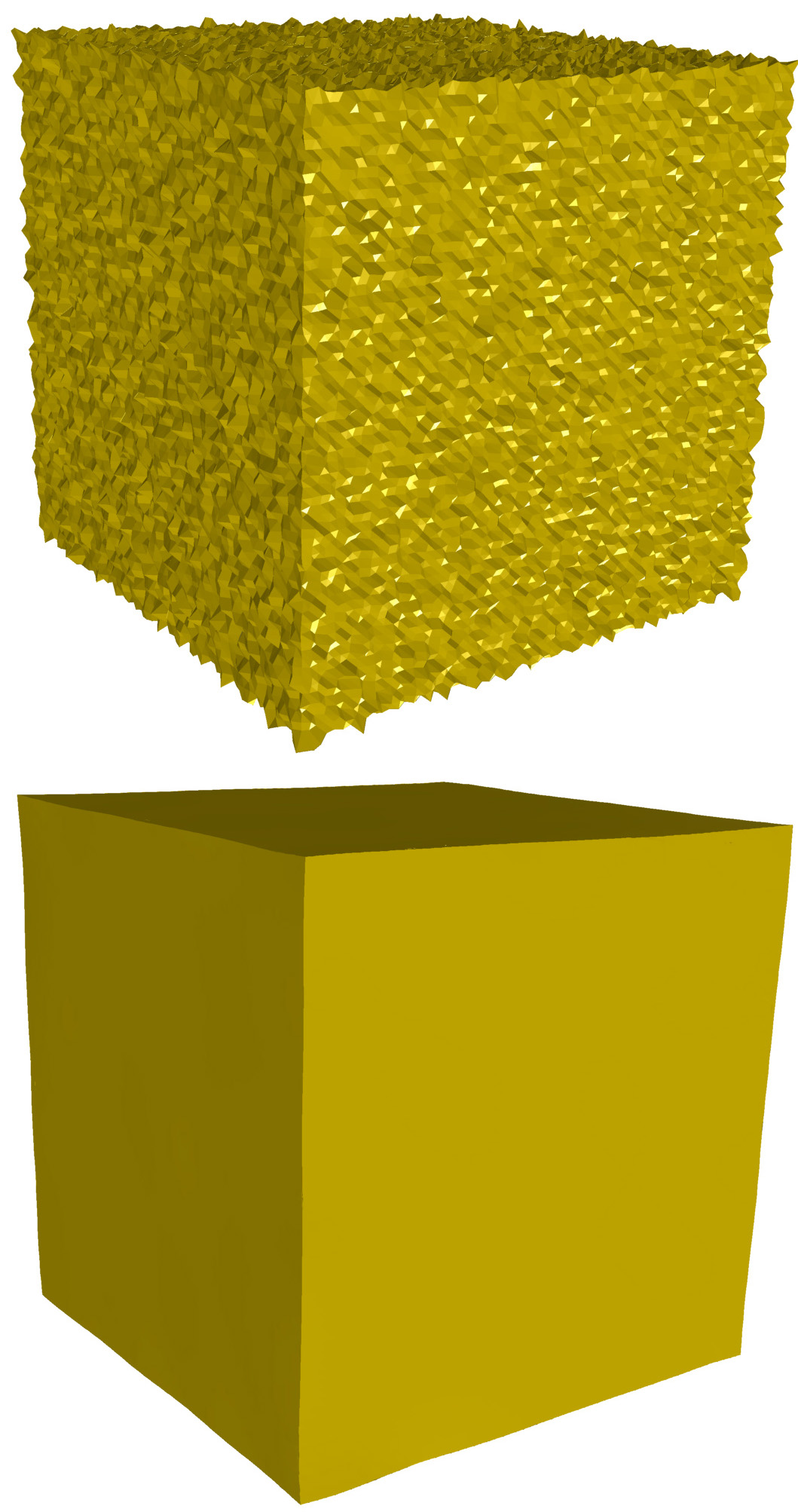}}}%
	\subfloat[{\boldmath$\sigma_n = 0.8l_e$}]{{\includegraphics[width=2.05cm]{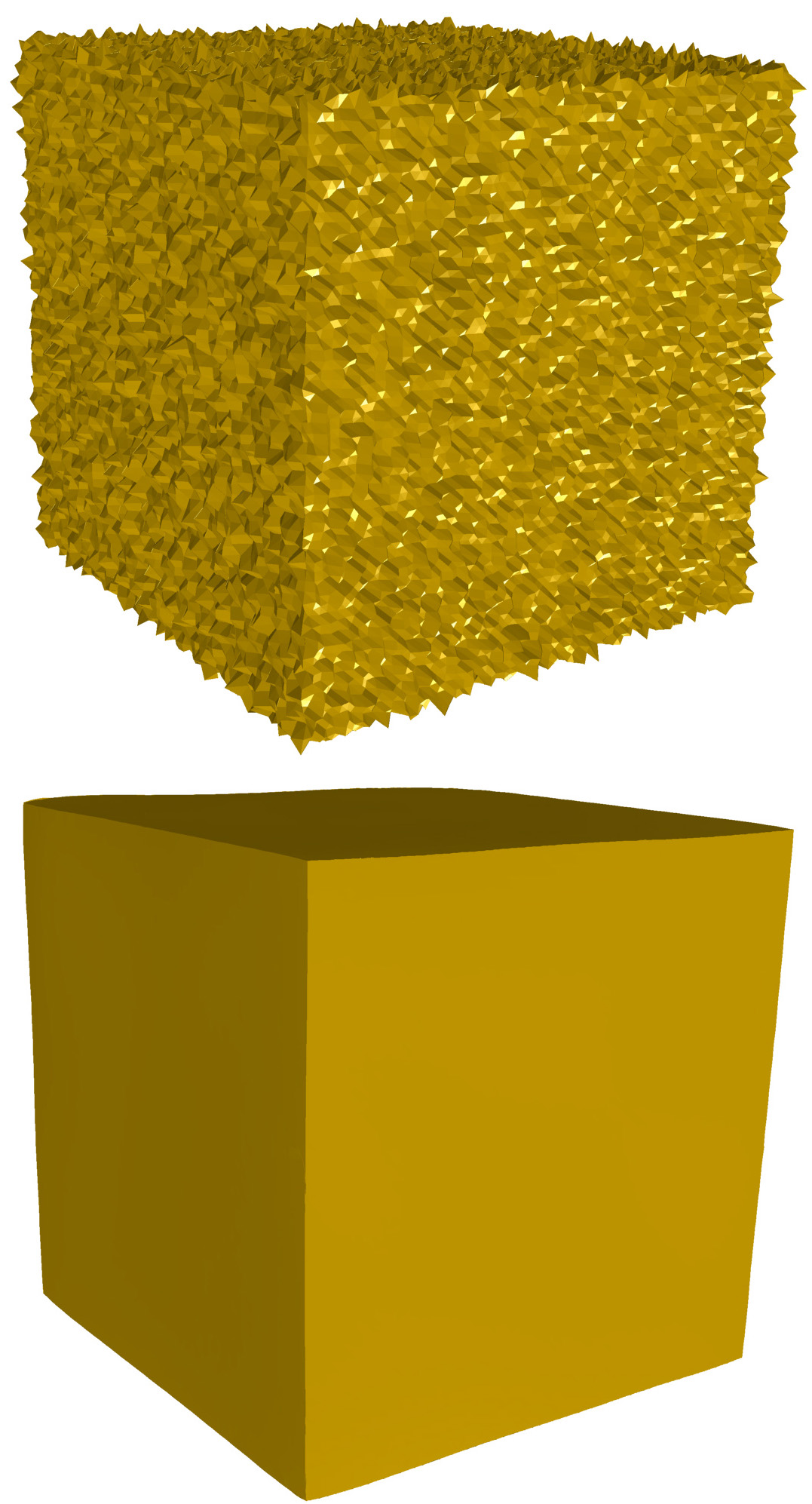}}}%
	\subfloat[{\boldmath$\sigma_n = l_e$}]{{\includegraphics[width=2.0cm]{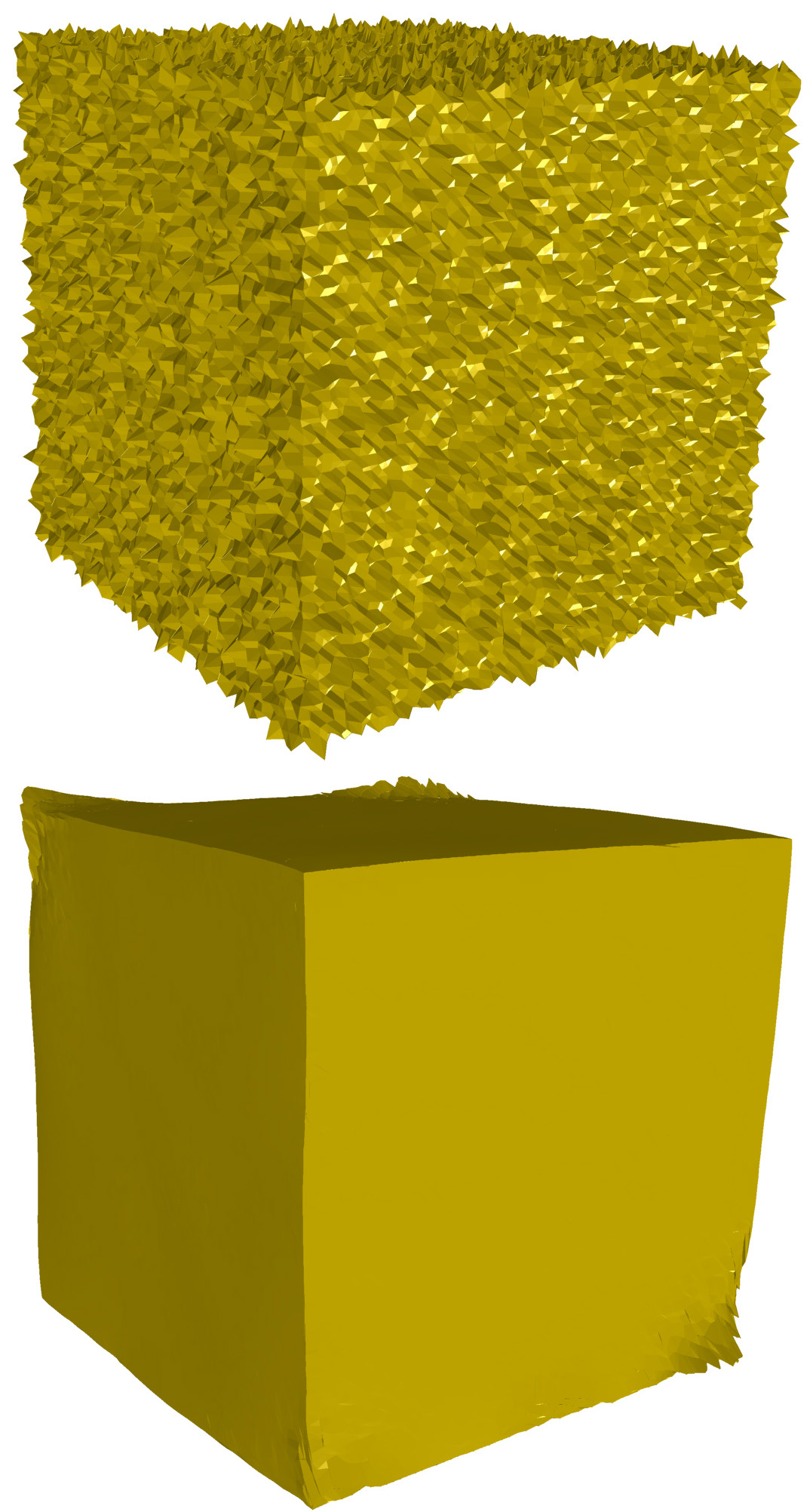}}}%
	\caption{Robustness against different levels of noise: The first row shows the cube model corrupted with different levels of noise. The second row shows the corresponding results obtained by the proposed method. In Figure (d), the noise level is bigger than the feature size and it is impossible to decouple features from noise. As a consequence, we are not able to recover the perfect cube. }%
	\label{fig:noiseLevel}%
\end{figure}

\begin{figure}[h]%
	\centering
	\subfloat[Original]{{\includegraphics[width=2.1cm]{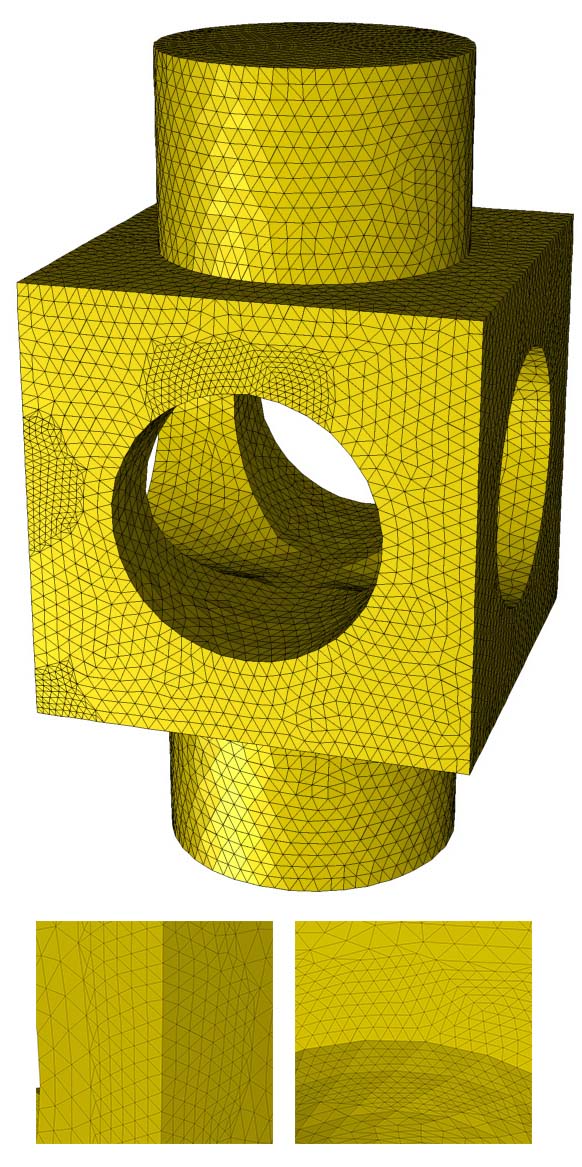} }}%
	\subfloat[Noisy]{{\includegraphics[width=2.1cm]{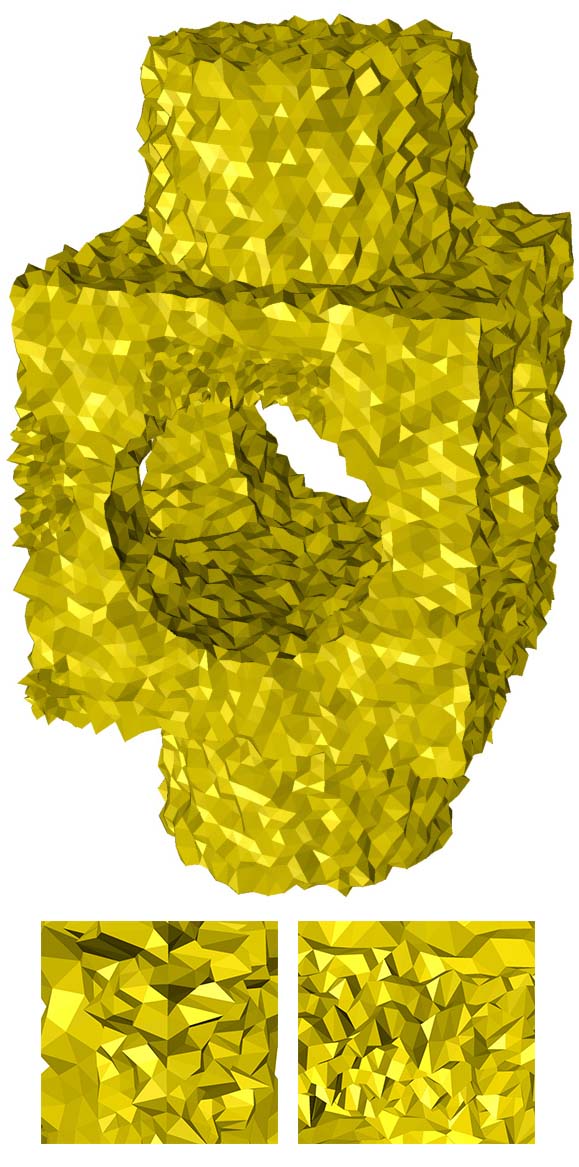} }}%
	\subfloat[Combinatorial]{{\includegraphics[width=2.1cm]{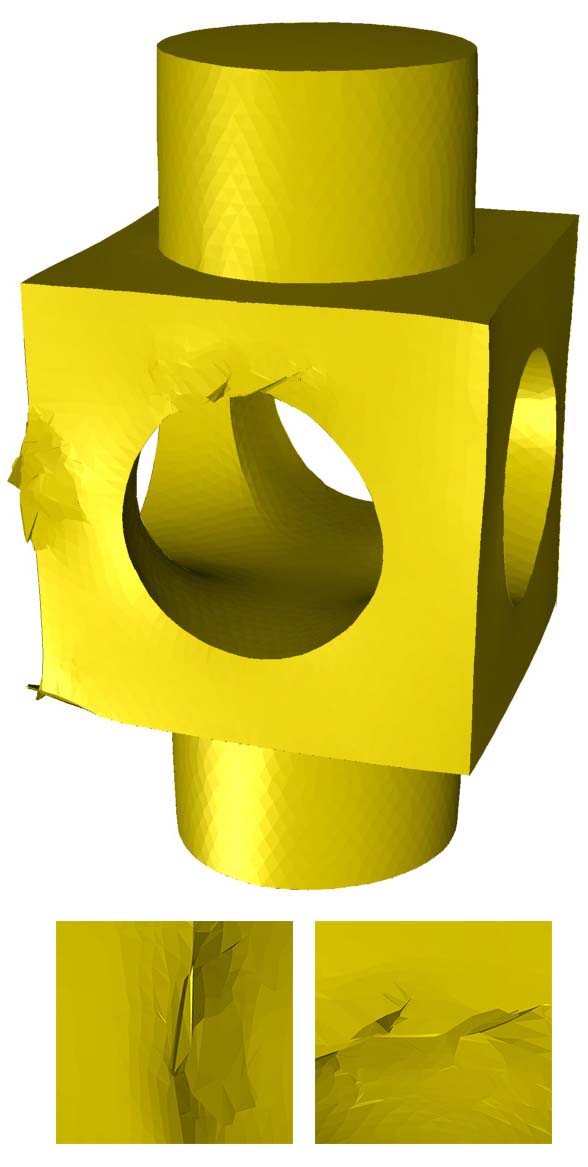} }}%
	\subfloat[Geometric]{{\includegraphics[width=2.1cm]{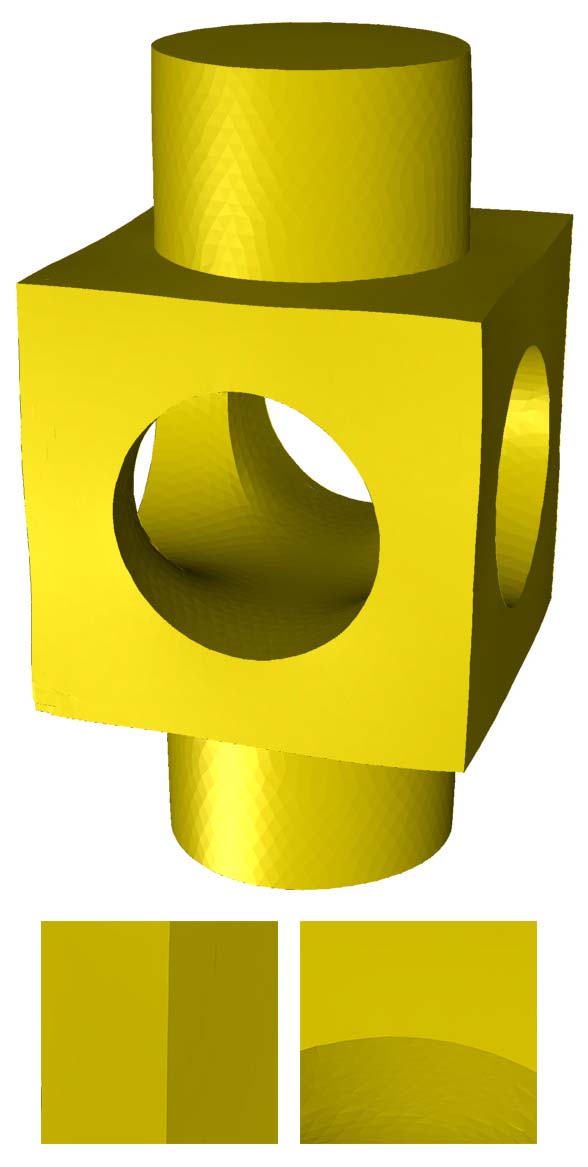} }}%
	\caption{Comparison between geometric and combinatorial neighborhood on a non-uniform mesh block model. Figure (c) and (d) show the results obtained by our method using the topological and the geometric neighborhood.}%
	\label{fig:neighComp}%
\end{figure}

\begin{figure*}[h]%
	\centering
	\subfloat[Noisy]{{\includegraphics[width=2.5cm]{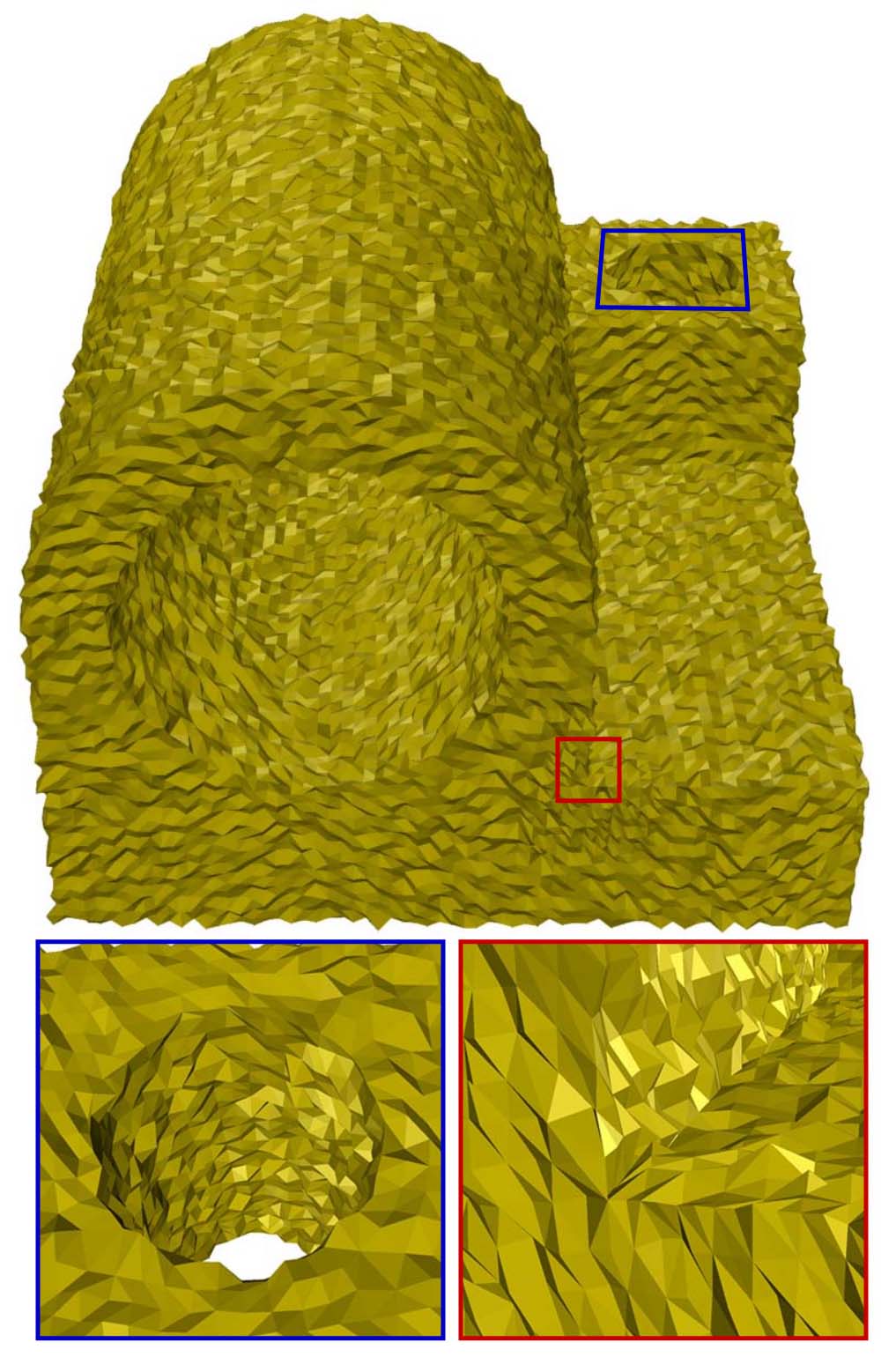} }}%
	\subfloat[\cite{aniso}]{{\includegraphics[width=2.5cm]{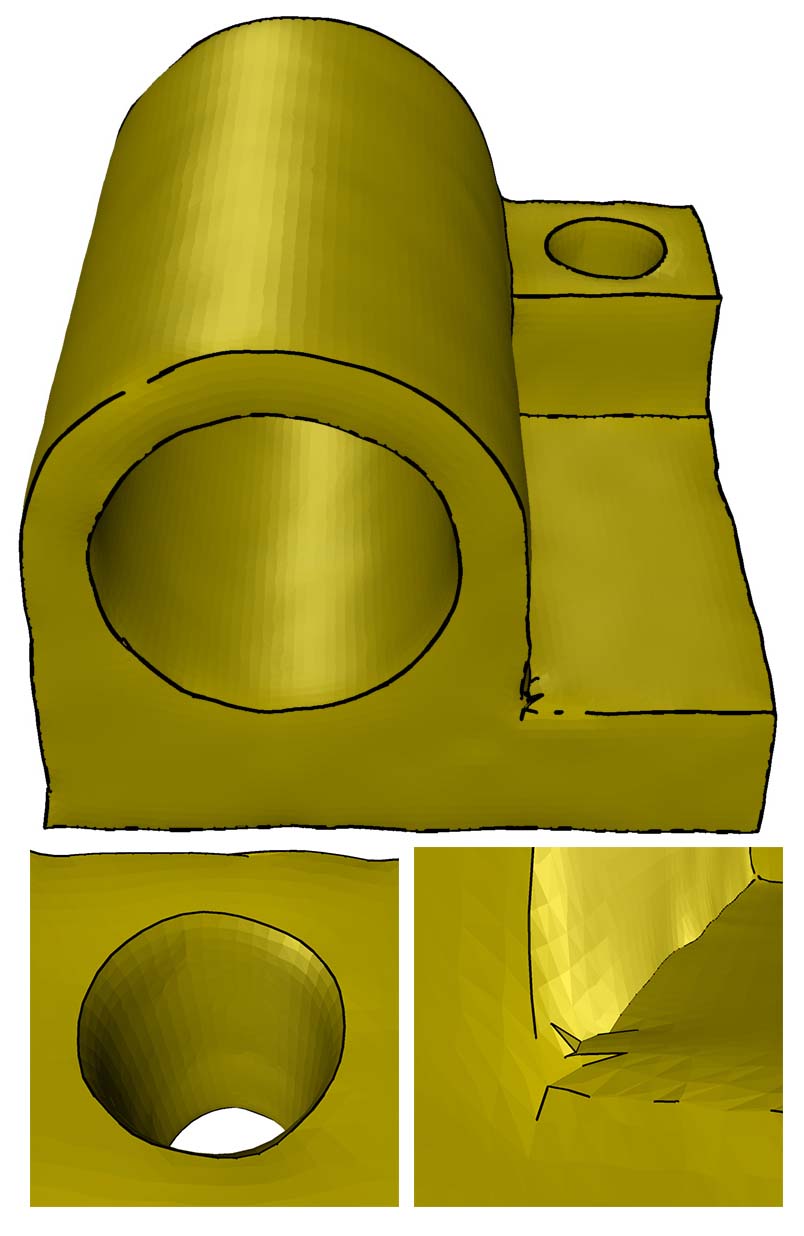} }}%
	\subfloat[\cite{BilNorm}]{{\includegraphics[width=2.5cm]{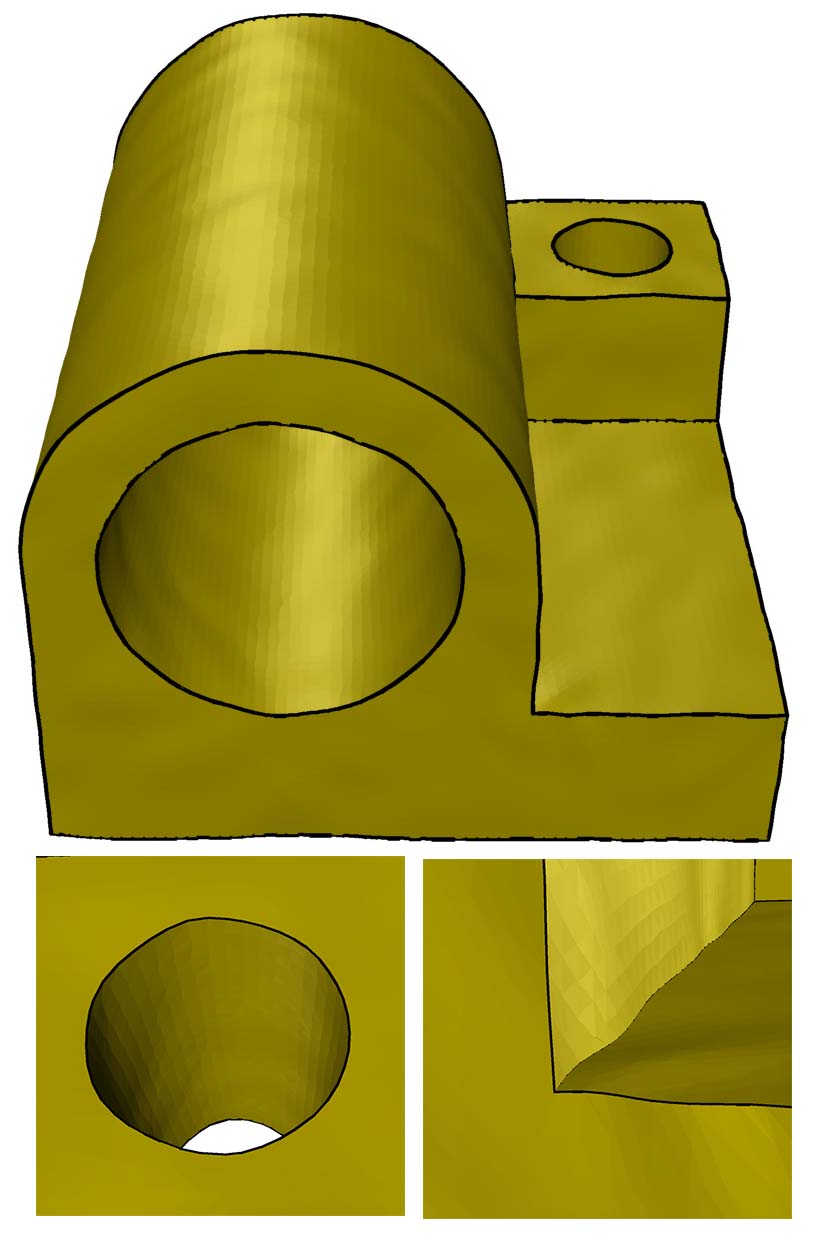} }}%
	\subfloat[\cite{L0Mesh}]{{\includegraphics[width=2.5cm]{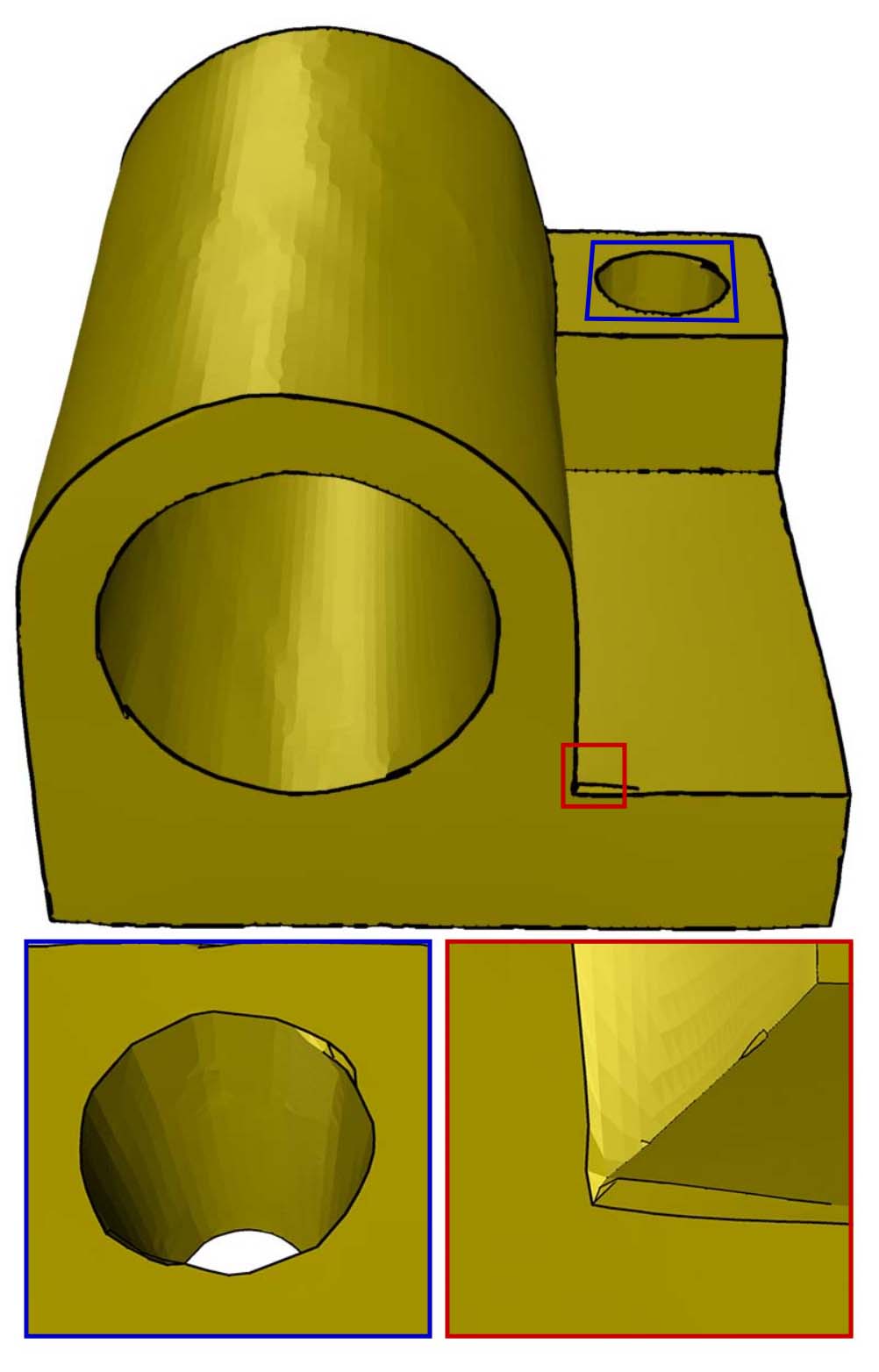} }}%
	\subfloat[\cite{Guidedmesh}]{{\includegraphics[width=2.5cm]{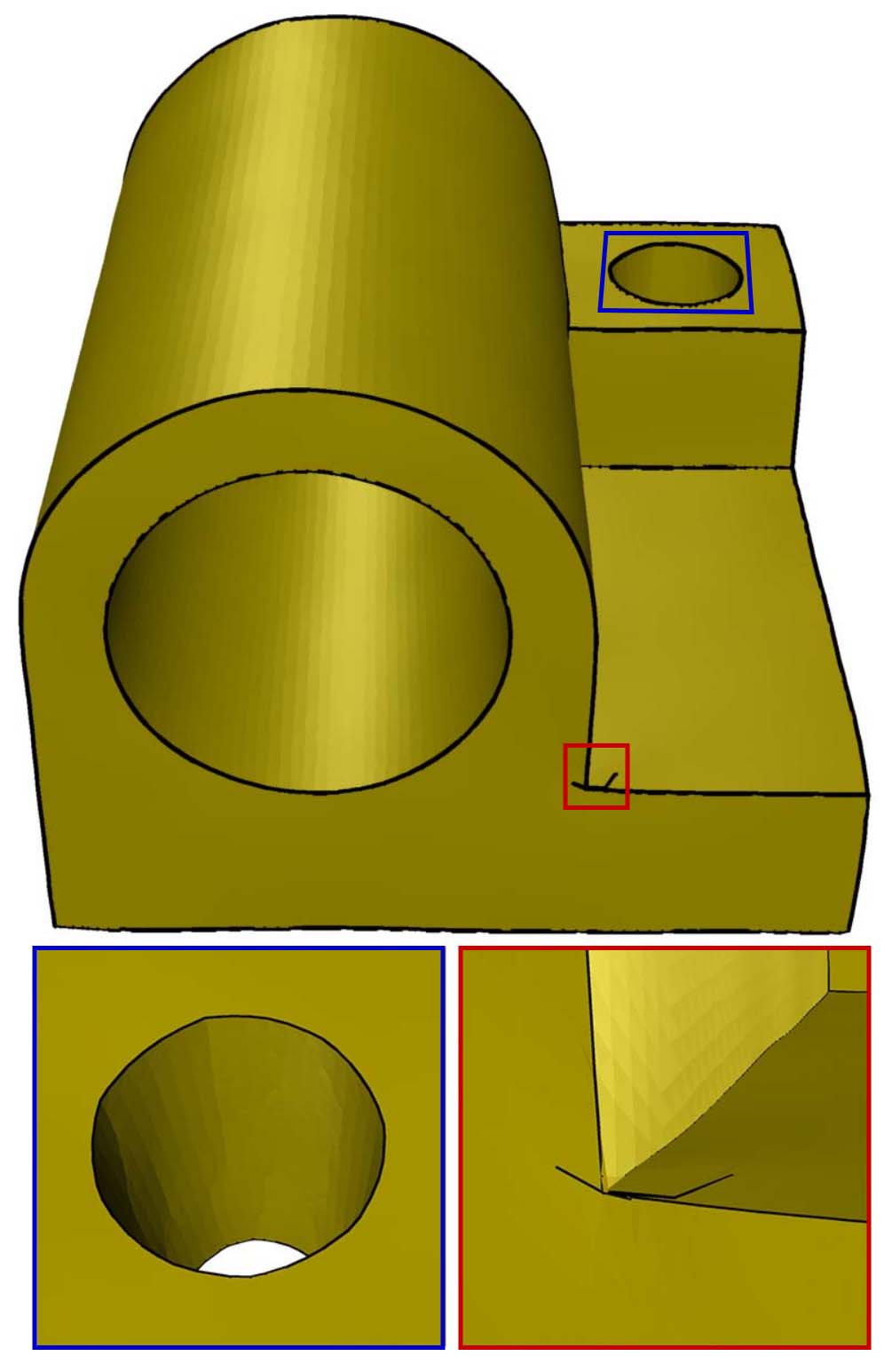} }}%
	\subfloat[\cite{binormal}]{{\includegraphics[width=2.5cm]{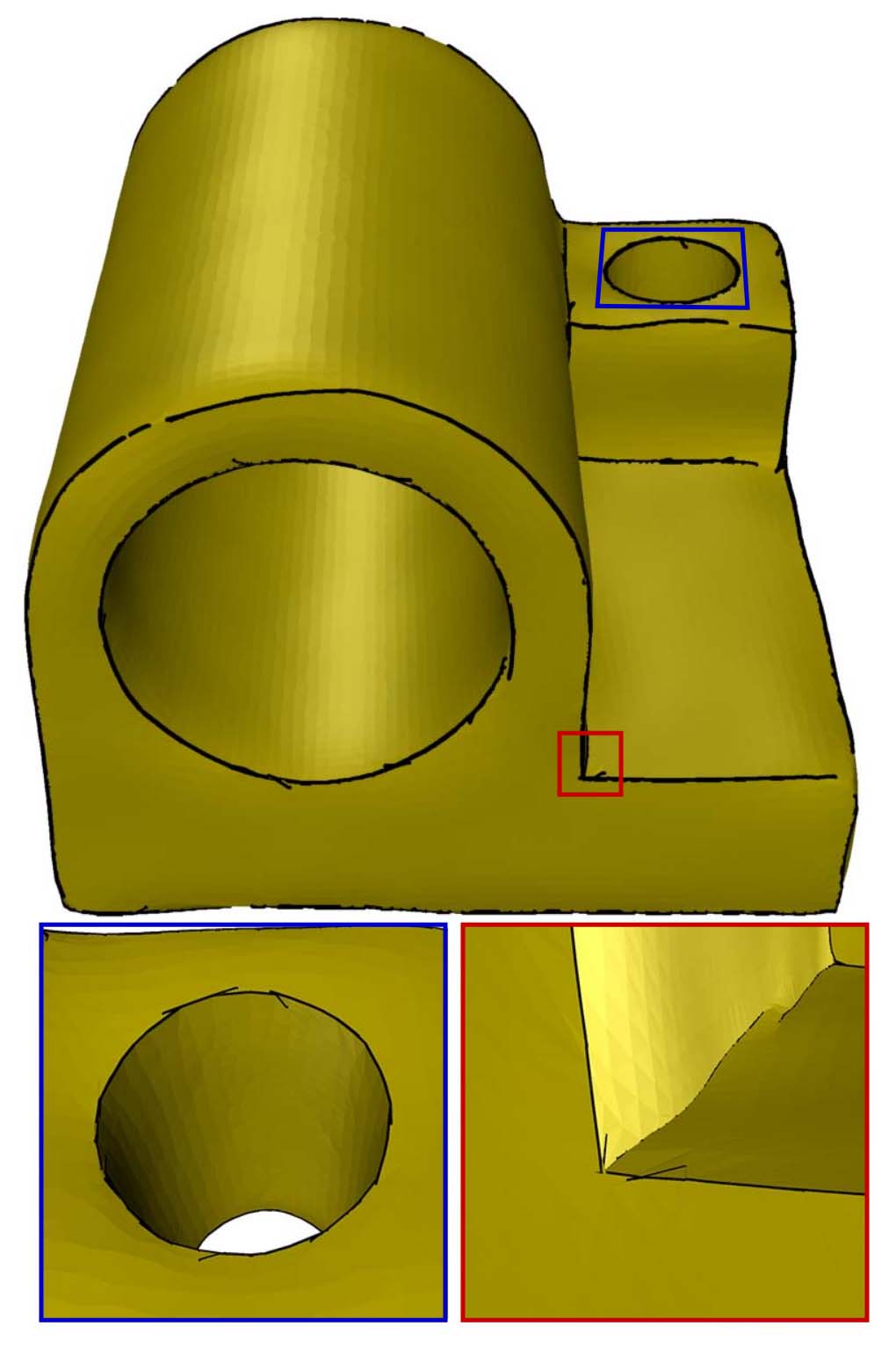} }}%
	\subfloat[Ours]{{\includegraphics[width=2.5cm]{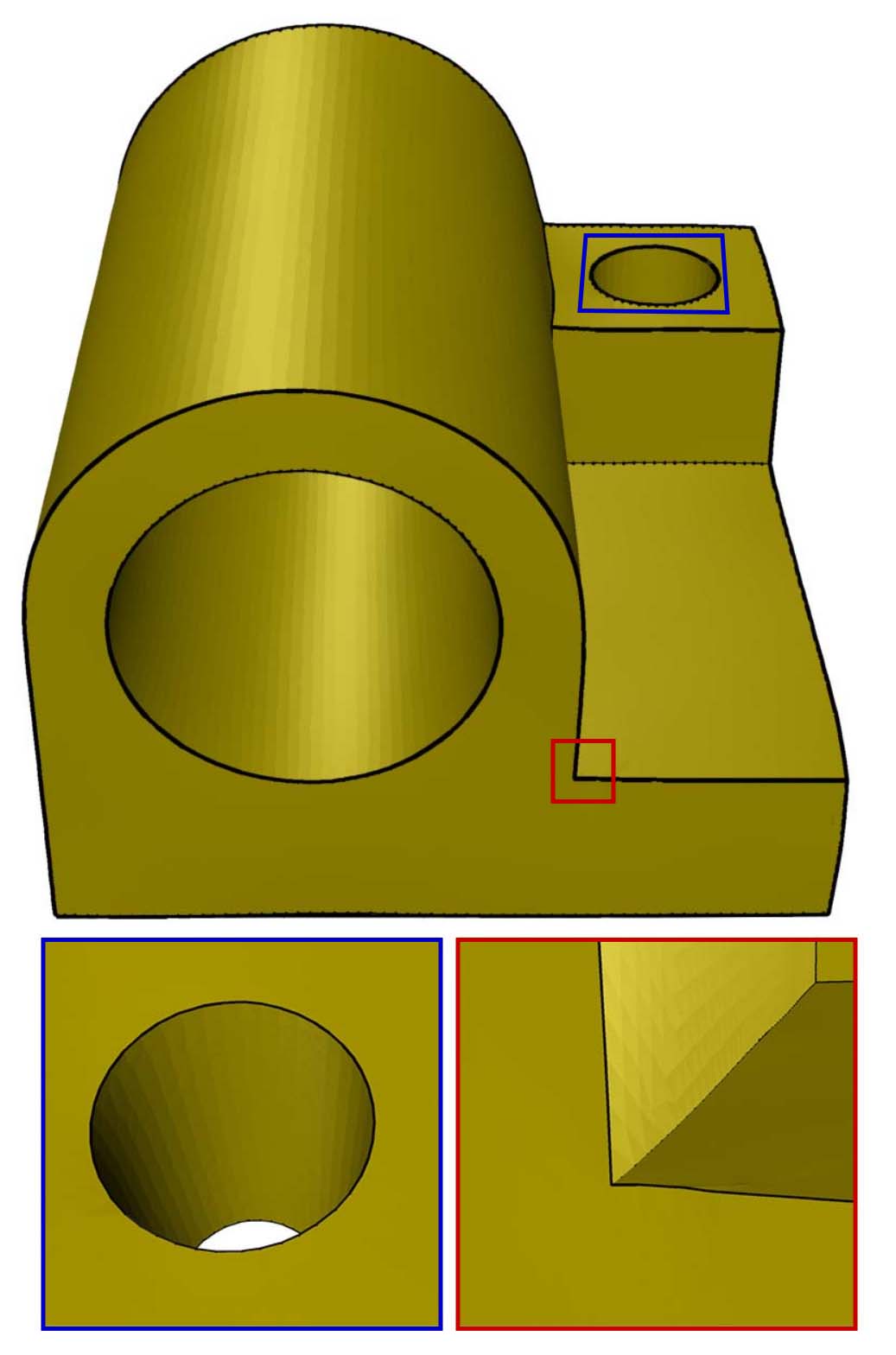} }}%
	\centering
	\caption{Non-uniform triangulated mesh surface corrupted by Gaussian noise ($\sigma_n = 0.35l_e$) in normal direction where $l_e$ is the average edge length. The first row shows the results obtained by state-of-the-art methods and the proposed method. The second row shows the magnified view of the corner and the cylindrical hole of the corresponding geometry.  }
	\label{fig:jointSharp}
\end{figure*}

\begin{figure*}%
	\centering
	\subfloat[Original]{{\includegraphics[width=2.1cm]{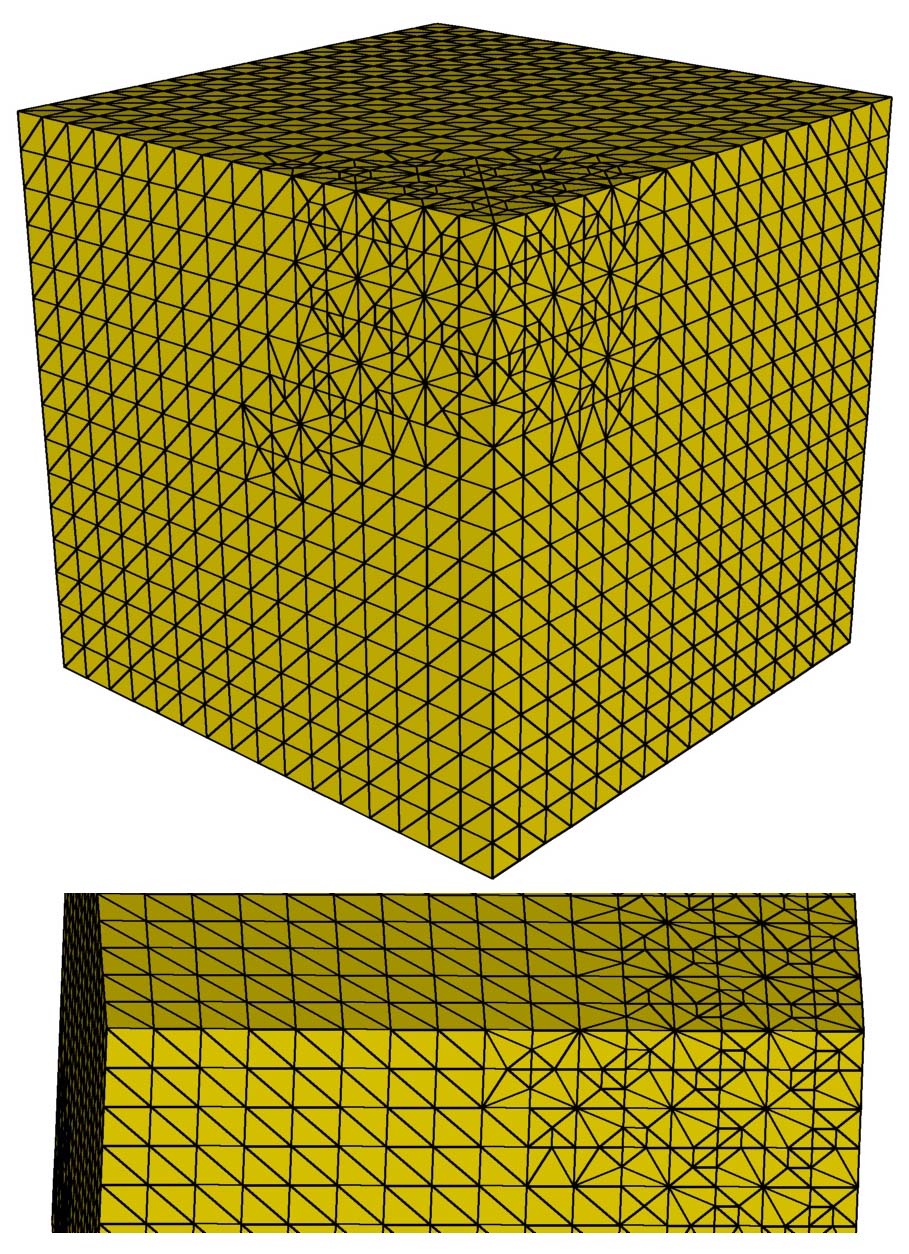} }}%
	\subfloat[Noisy]{{\includegraphics[width=2.1cm]{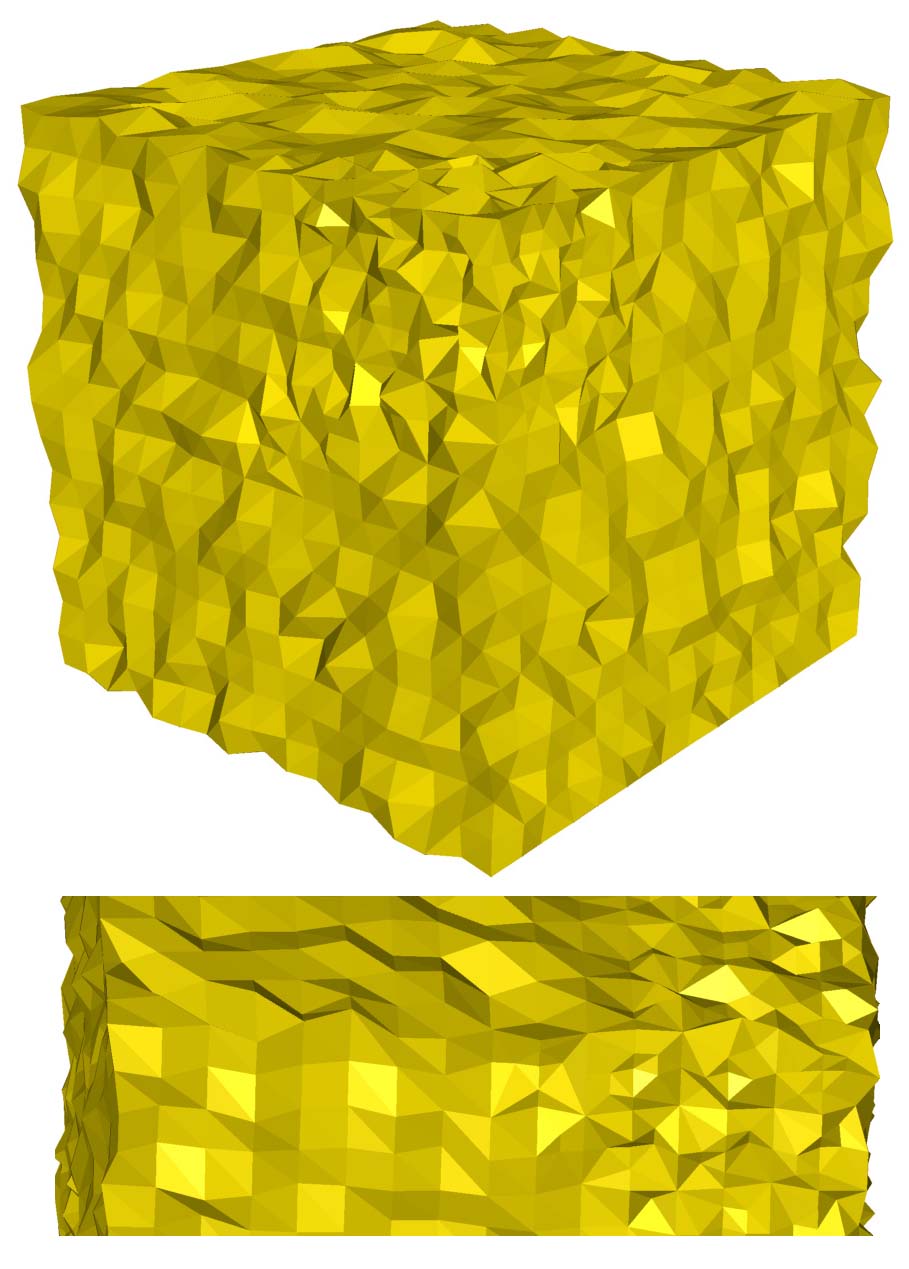} }}%
	\subfloat[\cite{aniso}]{{\includegraphics[width=2.1cm]{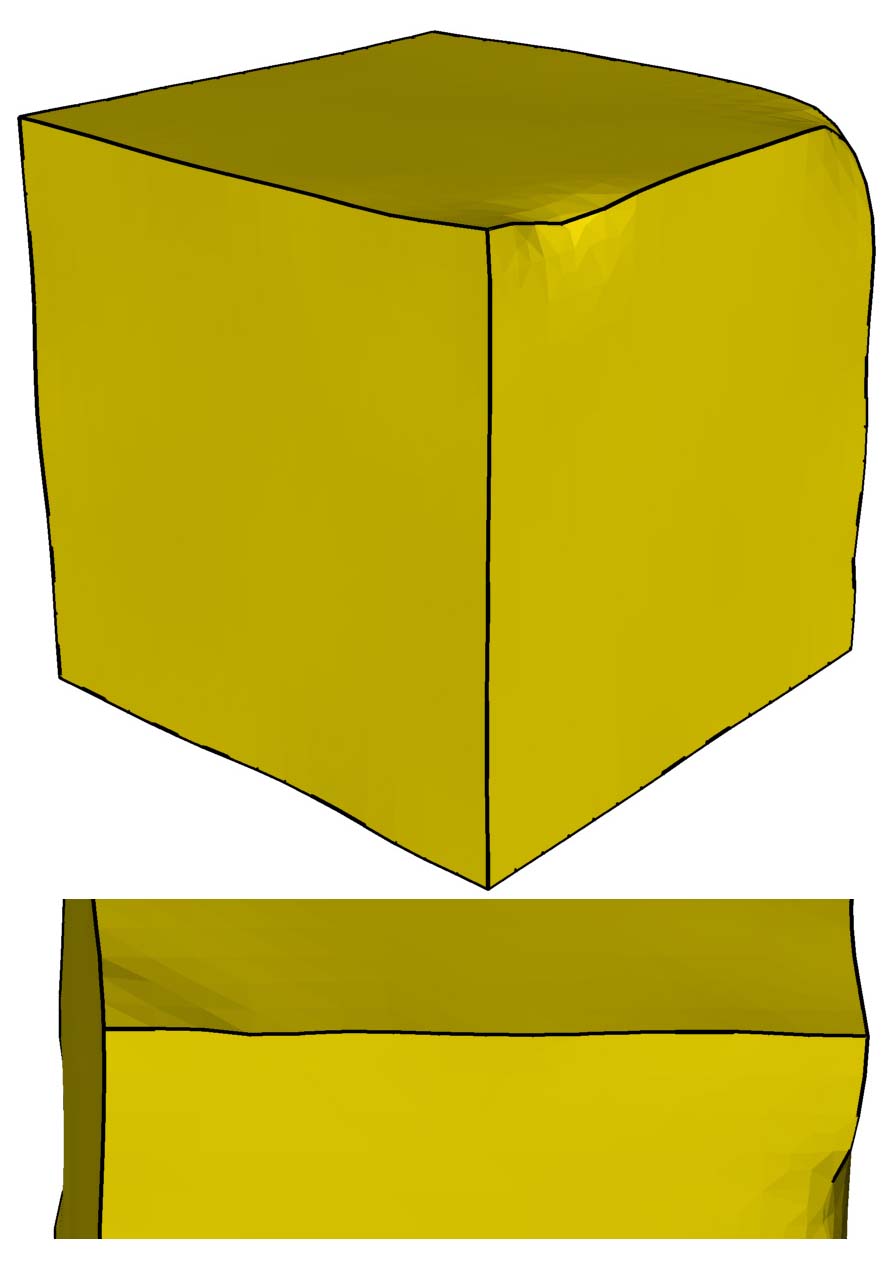} }}%
	\subfloat[\cite{BilNorm}]{{\includegraphics[width=2.1cm]{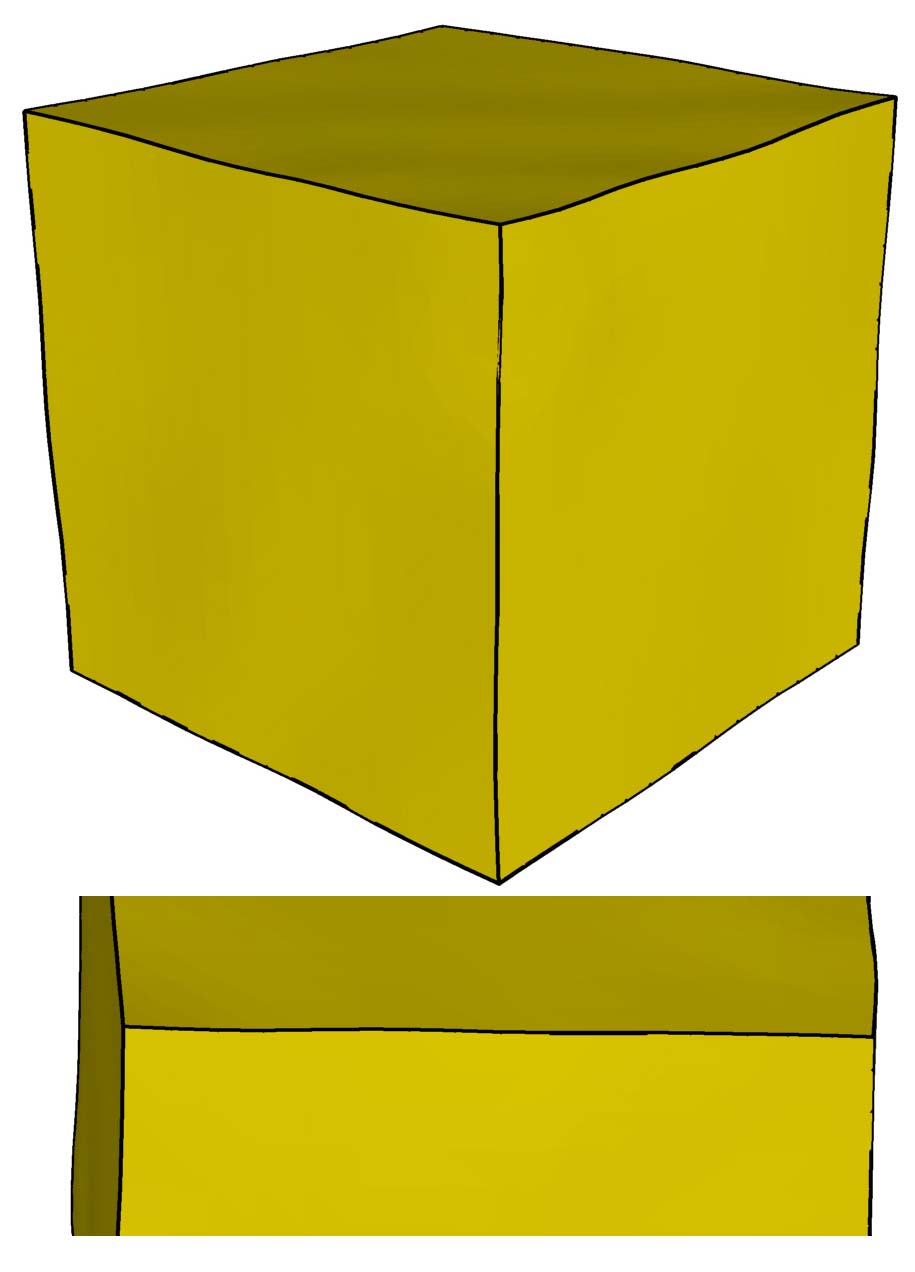} }}%
	\subfloat[\cite{L0Mesh}]{{\includegraphics[width=2.1cm]{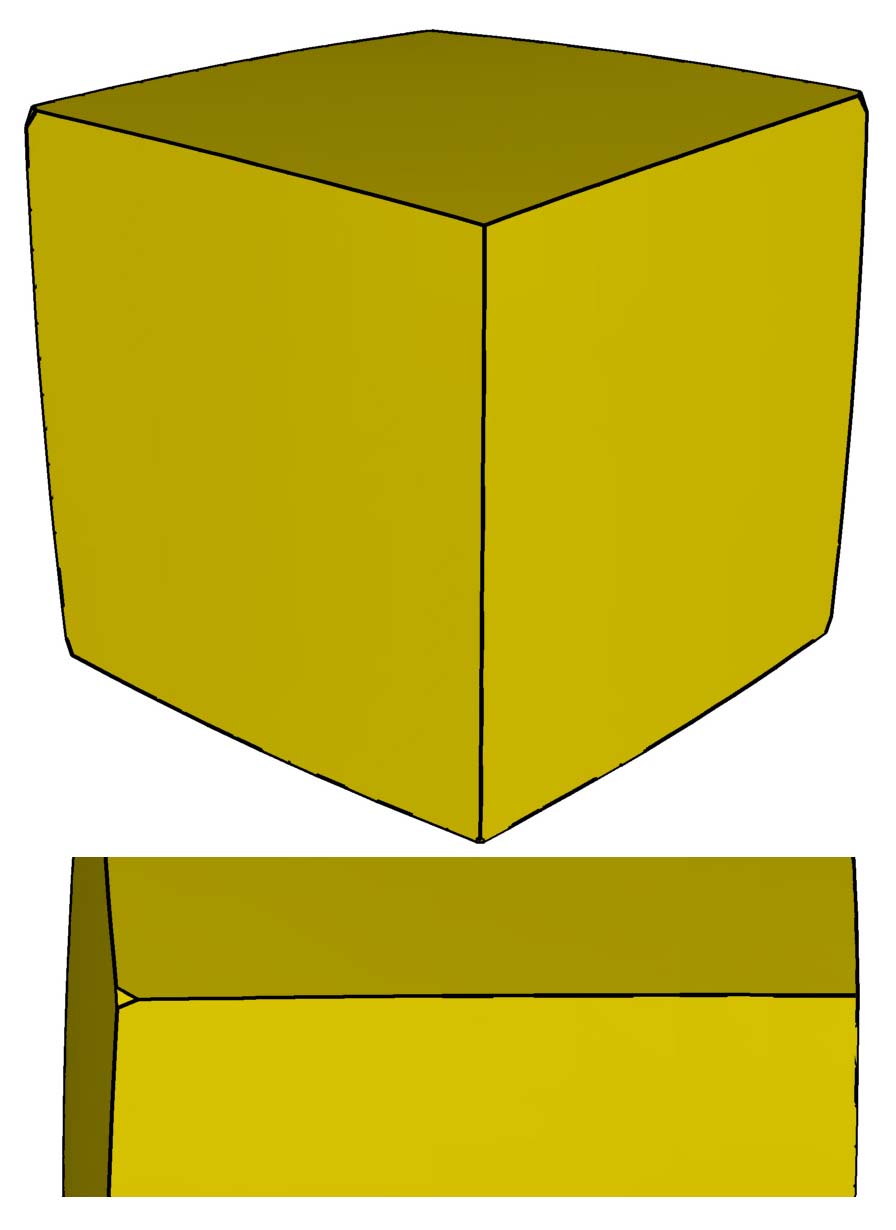} }}%
	\subfloat[\cite{Guidedmesh}]{{\includegraphics[width=2.1cm]{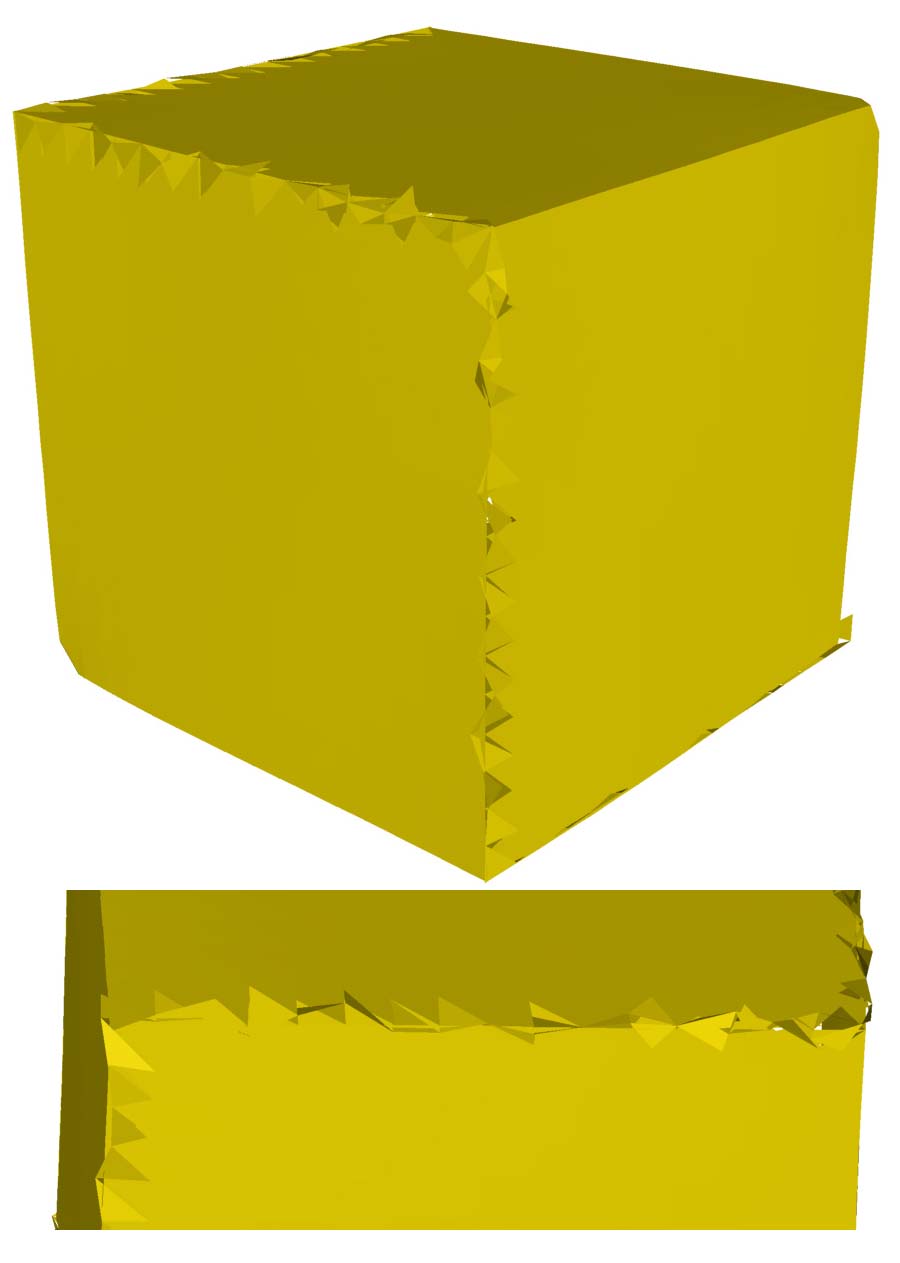} }}%
	\subfloat[\cite{binormal}]{{\includegraphics[width=2.1cm]{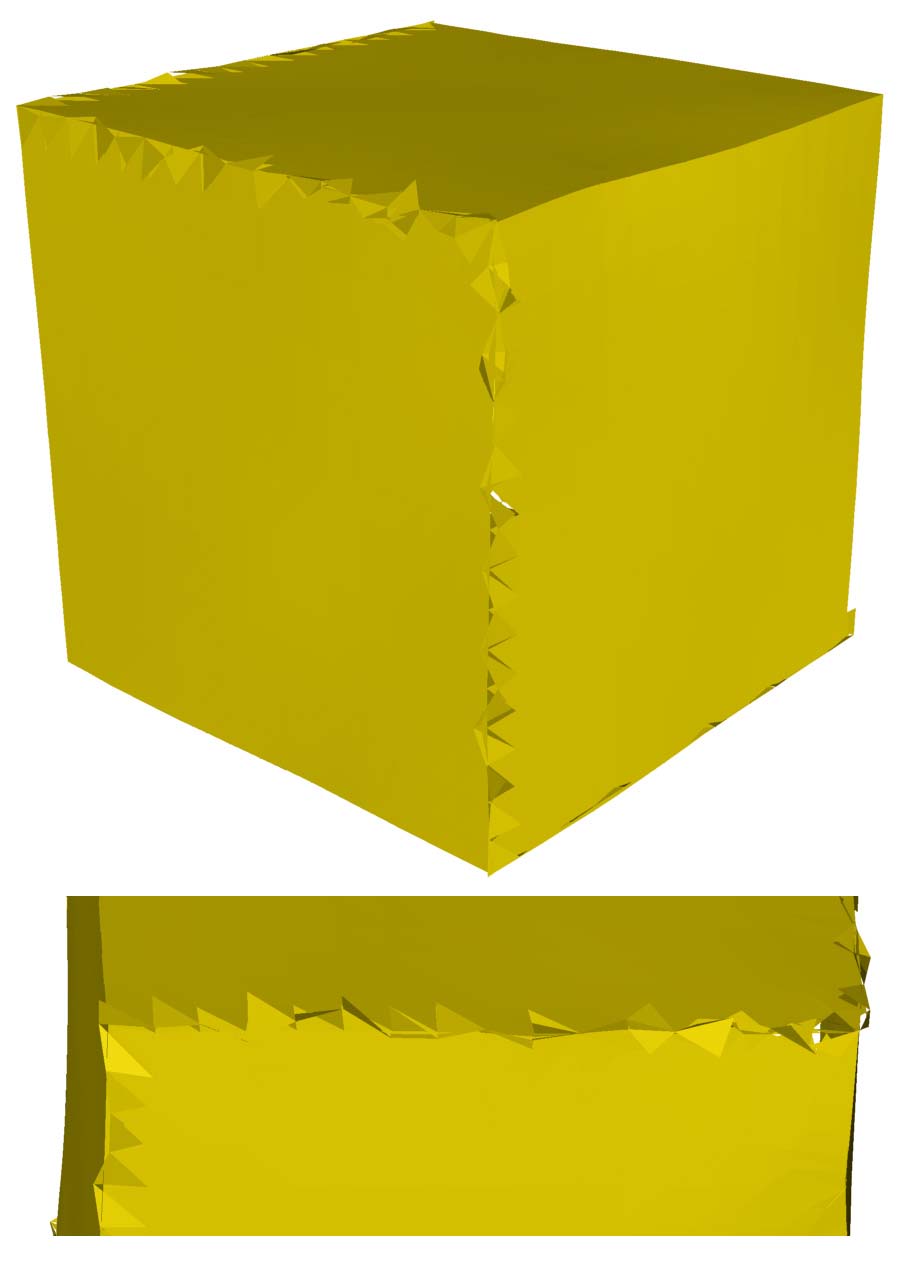} }}%
	\subfloat[Ours]{{\includegraphics[width=2.1cm]{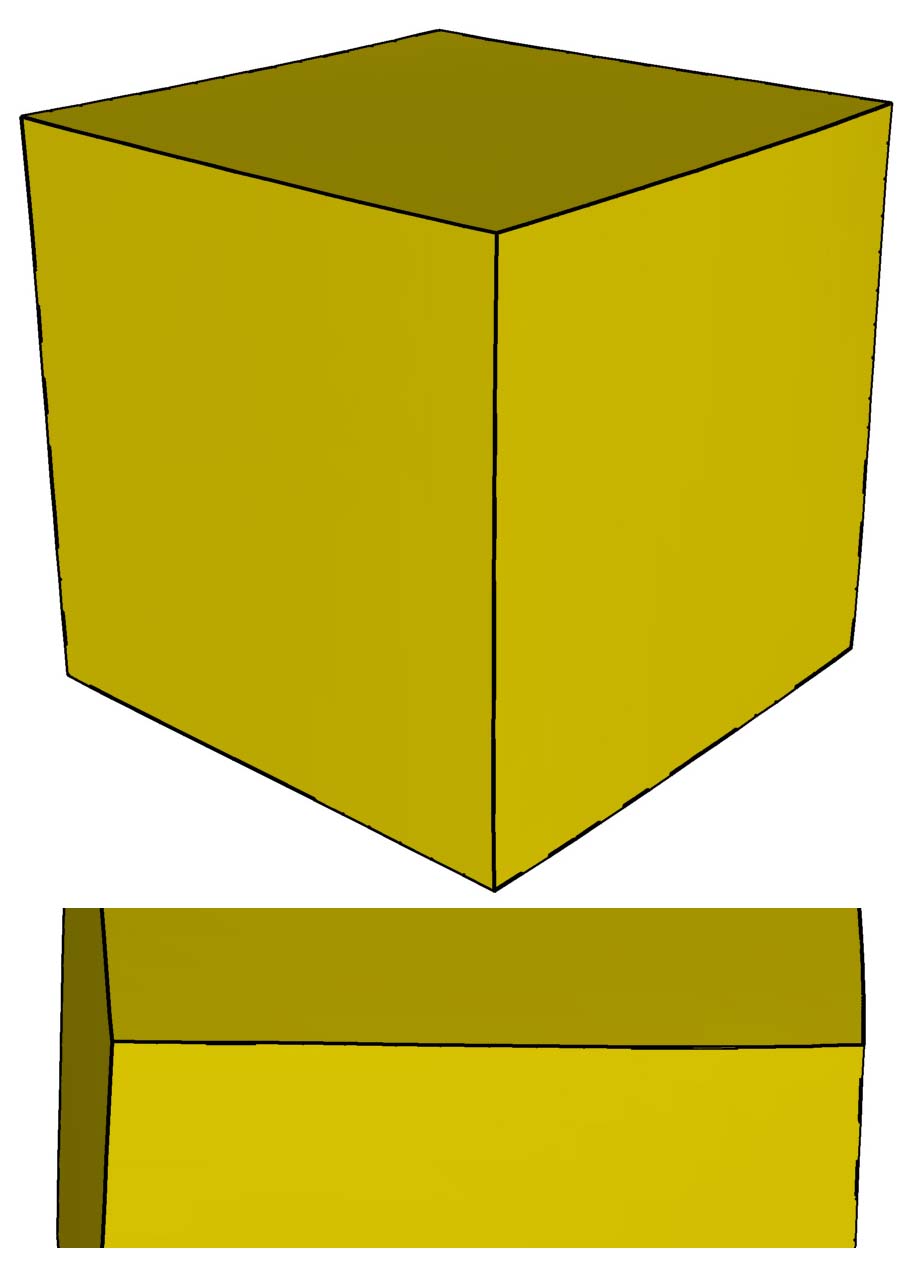} }}%
	\caption{ The Cube model consists of non-uniform triangles corrupted by Gaussian noise ($\sigma_n= 0.3l_e$) in normal direction. The first row shows the results produced by state-of-the-art methods and our proposed method. The second row shows magnified view of one of the sharp edges in the cube model. The results show that the proposed method has sharper and straighter edges compared to state-of-the-art methods. }%
	\label{fig:cube}%
\end{figure*}
{\footnotesize
	\begin{center}
		\captionof{table}{Quantitative Comparison}
		\label{tab:quant}
		\scalebox{0.8}{
			\begin{tabular}{ |c|c|c|c|l| }
				\hline
				Models & Methods & MSAE (degrees) & $E_v(\times 10^{-3})$ & Parameters \\ \hline
				& \cite{BilFleish} & 5.267 & 3.142 & (0.3, 0.3, 30)  \\
				& \cite{aniso} & 1.935 & 1.432 & (0.05, 0.05, 50) \\
				Cube & \cite{BilNorm} & 1.159 & 1.207& (0.3, 60)\\
				\small{${\lvert F \rvert = 3808 }$} & \cite{L0Mesh} & 1.343 & 4.476 & (1.4)\\ 
				\small{${\lvert V \rvert = 1906 }$}& \cite{Guidedmesh} & 40.41 & 1.166 & (Default)\\
				& \cite{binormal} & 41.63 & 1.446 & (Default)\\
				& \textbf{Ours} & \textbf{0.700} & \textbf{1.034} & (0.3, 0.15, 50)\\ \hline
				& \cite{BilFleish} & 4.983 & 58.86  & (0.3, 0.3, 20)  \\
				& \cite{aniso} & 3.444 & 32.53 & (0.2, 0.05, 30 ) \\
				Devil & \cite{BilNorm} & 2.641 & 8.95  & (0.3, 50)\\
				\small{${\lvert F \rvert = 25906 }$}& \cite{L0Mesh} & 6.699 & 26.28 & (4.0)\\ 
				\small{${\lvert V \rvert =  12986}$}& \cite{Guidedmesh} & 2.870 & 9.44 & (Default)\\
				& \cite{binormal} & 5.396 & 13.52 & (Default)\\
				& \textbf{Ours} & \textbf{2.702} & \textbf{6.901} & (0.1, 1.0, 30)\\ \hline
				& \cite{BilFleish} & 3.777 & 0.530   & (0.3, 0.3, 40) \\
				& \cite{aniso} & 2.630 & 0.766  & (0.009, 0.05, 50)\\
				Joint & \cite{BilNorm} & 1.808 & 0.263 & (0.4, 100) \\
				\small{${\lvert F \rvert = 52226  }$} & \cite{L0Mesh} &1.768 & 0.500 & (1.4)\\ 
				\small{${\lvert V \rvert =  26111}$}& \cite{Guidedmesh} & 0.956 & 0.179 & (Default)\\
				& \cite{binormal} & 2.874 & 0.366 & (Default)\\
				& \cite{robust16} & {1.16} & {1.49}& (Default)\\
				& \textbf{Ours} & \textbf{0.829} & \textbf{0.171} & (0.3, 0.05, 60)\\ \hline
				& \cite{BilFleish} & 8.567 & 4.422 & (0.4, 0.4, 40)   \\
				Fandisk & \cite{aniso} & 5.856 & 4.910  & (0.07, 0.05, 30)\\
				\small{${\lvert F \rvert = 12946  }$}& \cite{BilNorm} & 2.727 & 1.877 & (0.4,70) \\
				\small{${\lvert V \rvert = 6475 }$}& \cite{L0Mesh} &4.788 & 5.415 & (1.4)\\ 
				& \cite{Guidedmesh} & \textbf{2.221} & \textbf{1.702}& (Default)\\ 
				& \cite{robust16} & {3.1} & {4.42}& (Default)\\ 
				& \textbf{Ours} & {2.692} & {1.964}& (0.3, 0.2, 50)\\ \hline
				& \cite{BilFleish} & 5.737 & 44.22 & (0.3, 0.3, 40)  \\
				Rockerarm& \cite{aniso} & 5.982 & 51.35& (0.2, 0.05, 50) \\
				\small{${\lvert F \rvert = 48212  }$}& \cite{BilNorm} & 5.713 & 23.47& (1.0, 100) \\
				\small{${\lvert V \rvert = 24106 }$}& \cite{L0Mesh} & 7.468 & 34.70 & (1.4)\\ 
				& \cite{Guidedmesh} & 6.846 & 30.45& (Default)\\
				& \textbf{Ours} & \textbf{5.410} & \textbf{20.05}& (0.3, 1.0, 60)\\ 
				\hline
				Vase& \cite{robust16} & \textbf{2.92} & 1.72& (Default) \\
				
				\small{${\lvert V \rvert = 3827 }$}& \cite{L0Mesh} & 4.26 & 0.83 & (1.6)\\
				\small{${\lvert F \rvert = 7650  }$}& \textbf{Ours} & 3.34 & \textbf{0.42} & (0.3, 0.03, 30) \\
				& & & &\\
				\hline
			\end{tabular}
			
		}
	\end{center}
}

\begin{figure*}%
	\centering
	\subfloat[Original]{{\includegraphics[width=2.1cm]{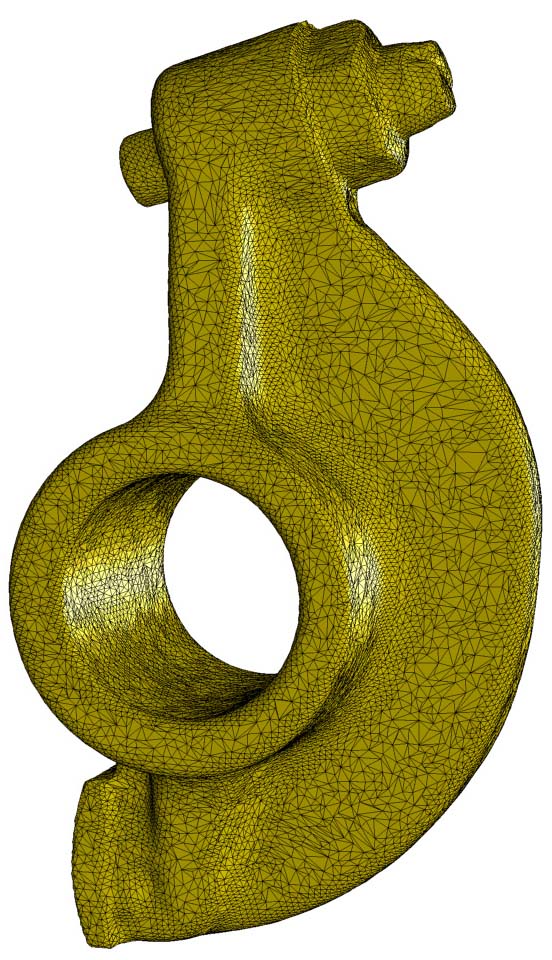} }}%
	\subfloat[Noisy]{{\includegraphics[width=2.1cm]{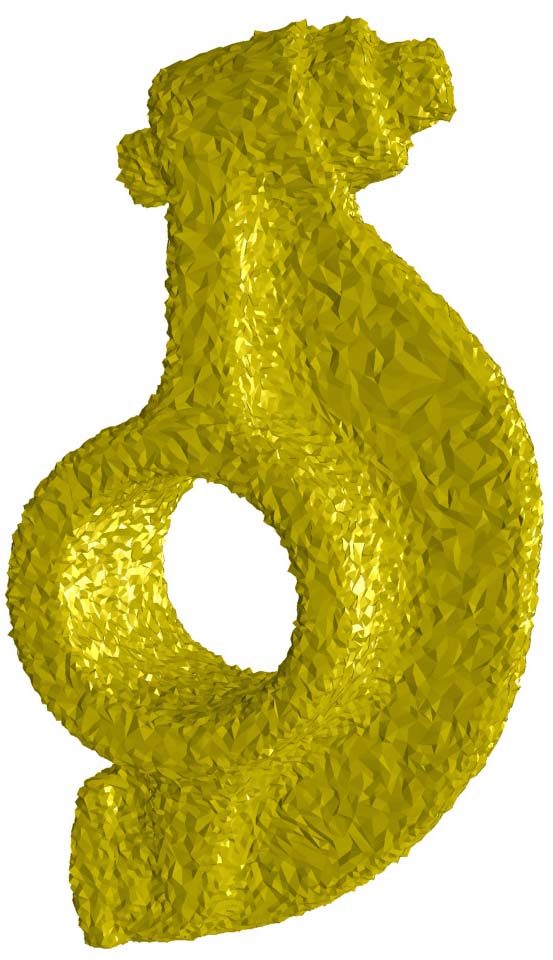} }}%
	\subfloat[\cite{BilFleish}]{{\includegraphics[width=2.1cm]{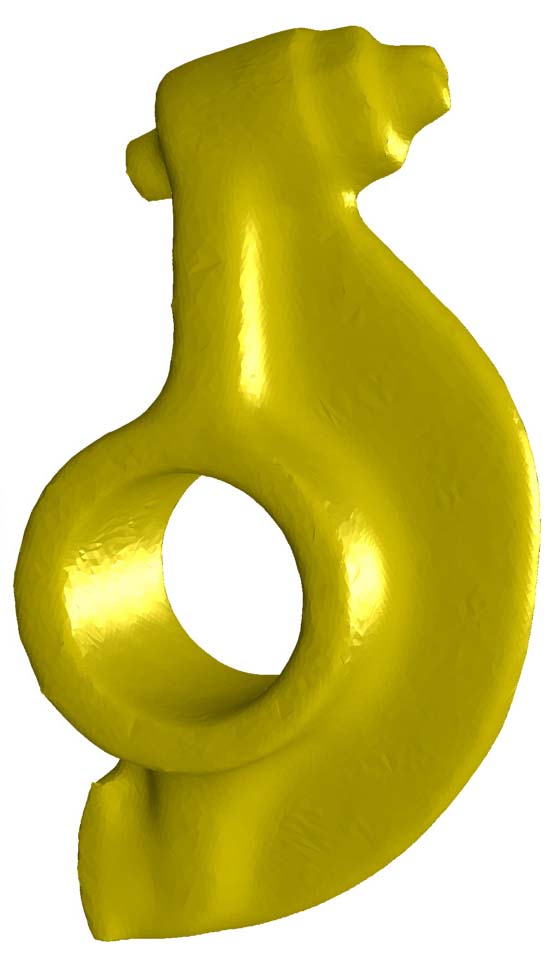} }}%
	\subfloat[\cite{aniso}]{{\includegraphics[width=2.1cm]{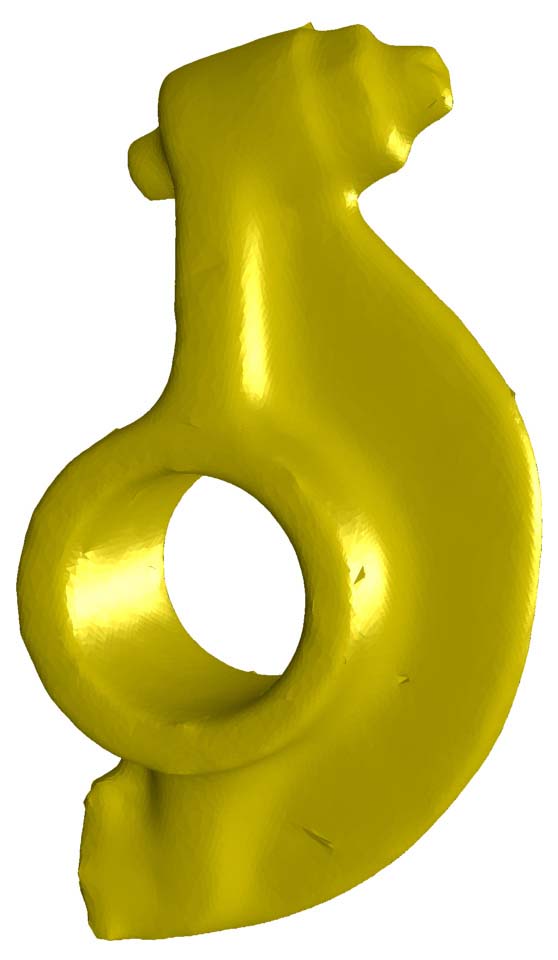} }}%
	\subfloat[\cite{BilNorm}]{{\includegraphics[width=2.1cm]{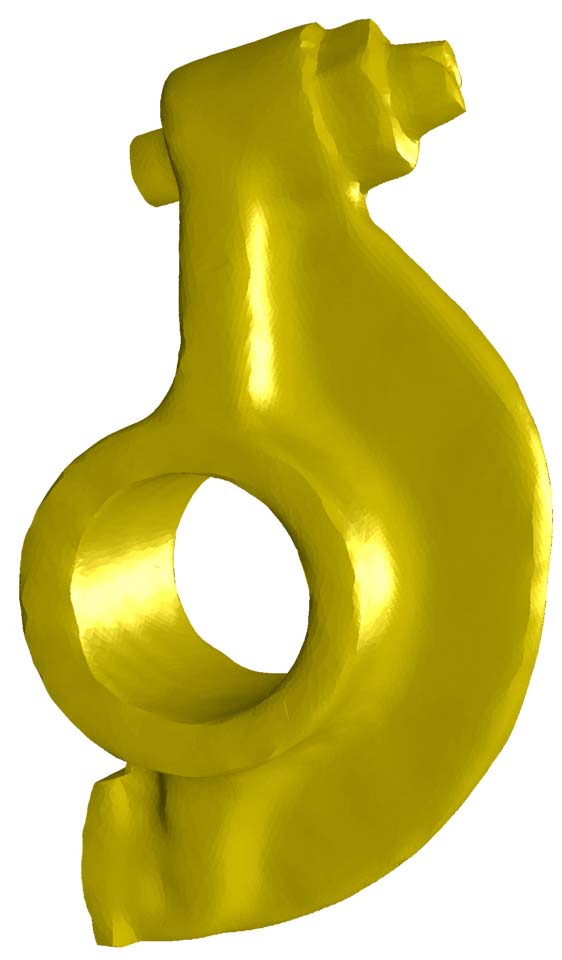} }}%
	\subfloat[\cite{L0Mesh}]{{\includegraphics[width=2.1cm]{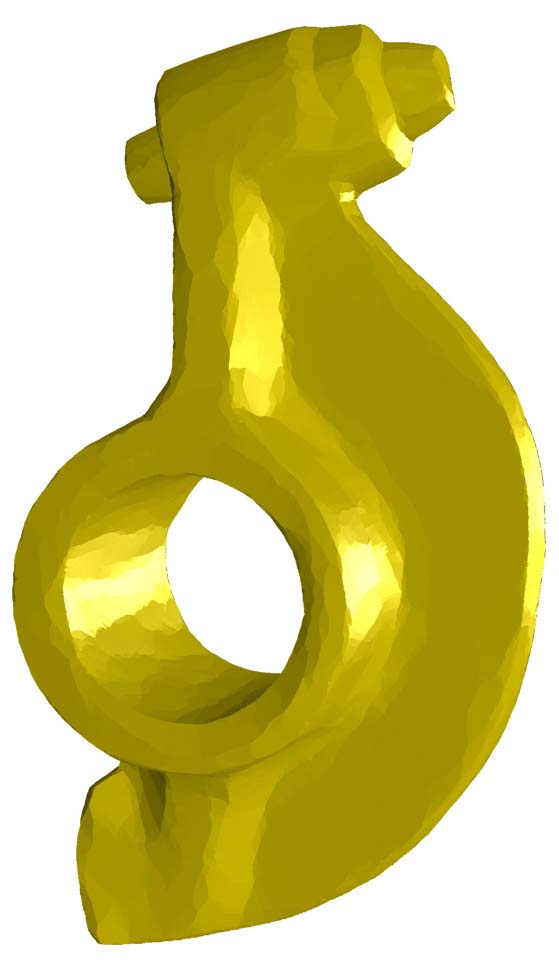} }}%
	\subfloat[\cite{Guidedmesh}]{{\includegraphics[width=2.1cm]{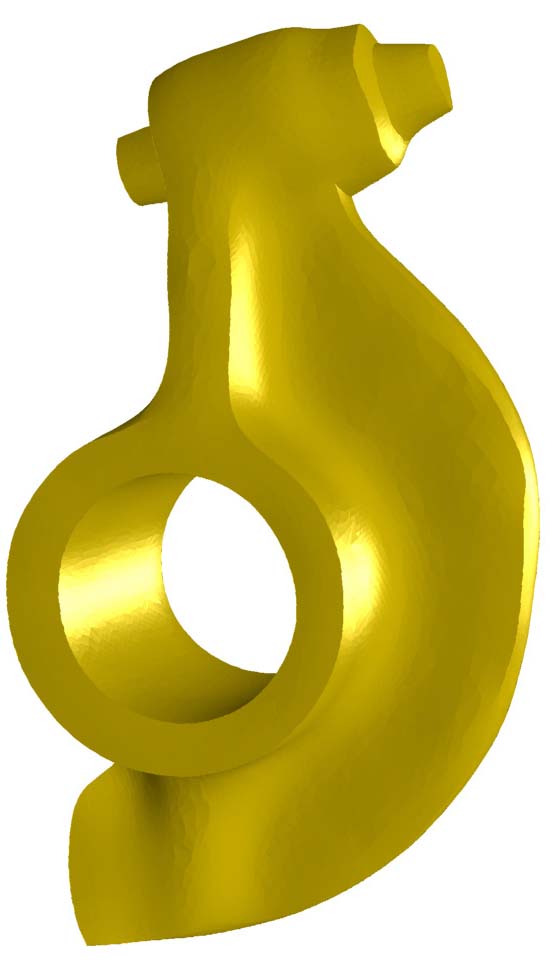} }}%
	\subfloat[Ours]{{\includegraphics[width=2.1cm]{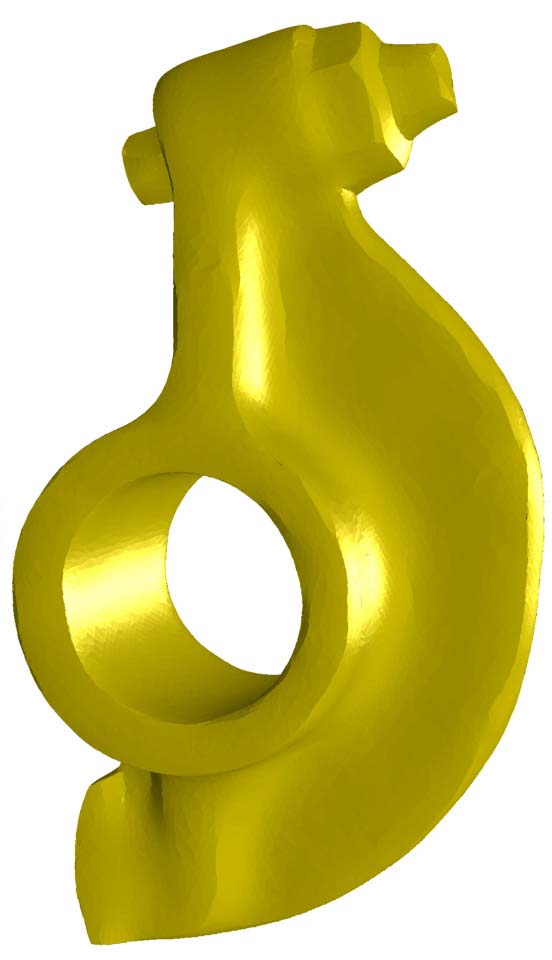} }}%
	\caption{Rockerarm model corrupted by Gaussian noise ($\sigma_n= 0.3l_e$) in normal direction. The results are produced by state-of-the-art methods and our proposed method. }%
	\label{fig:rocker}%
\end{figure*}

\begin{figure*}%
	\centering
	\subfloat[Original]{{\includegraphics[width=2.1cm]{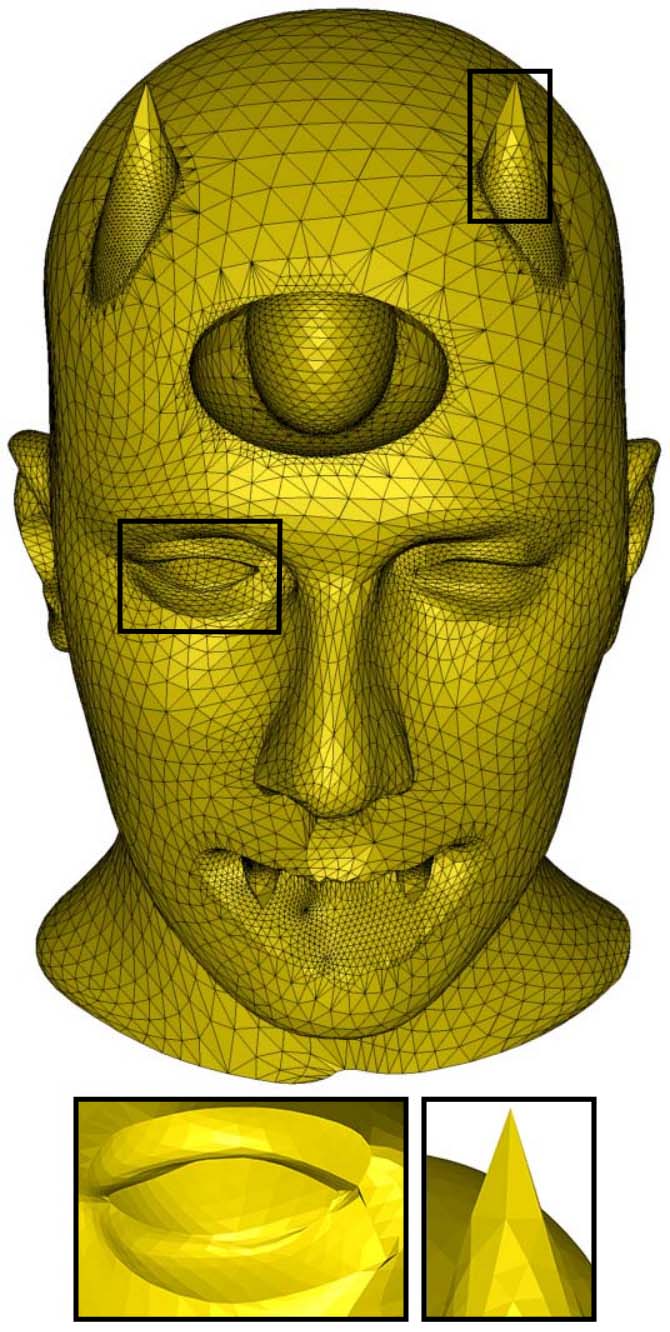} }}%
	\subfloat[Noisy]{{\includegraphics[width=2.1cm]{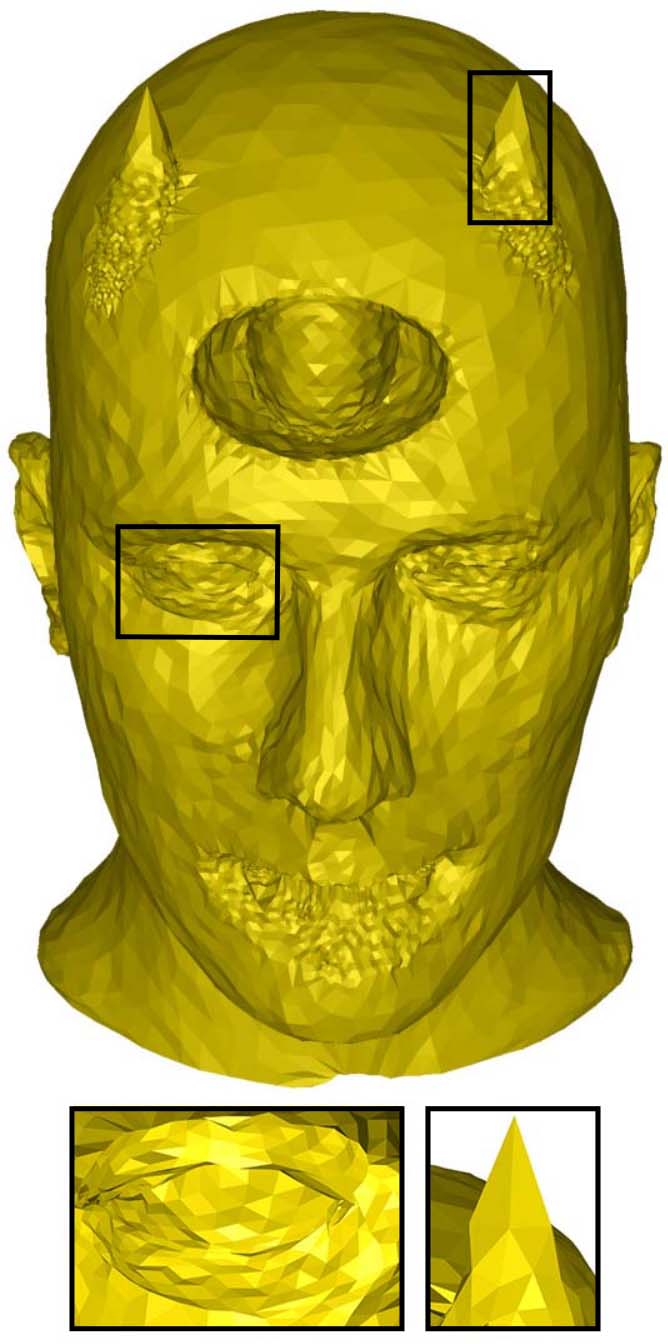} }}%
	\subfloat[\cite{aniso}]{{\includegraphics[width=2.1cm]{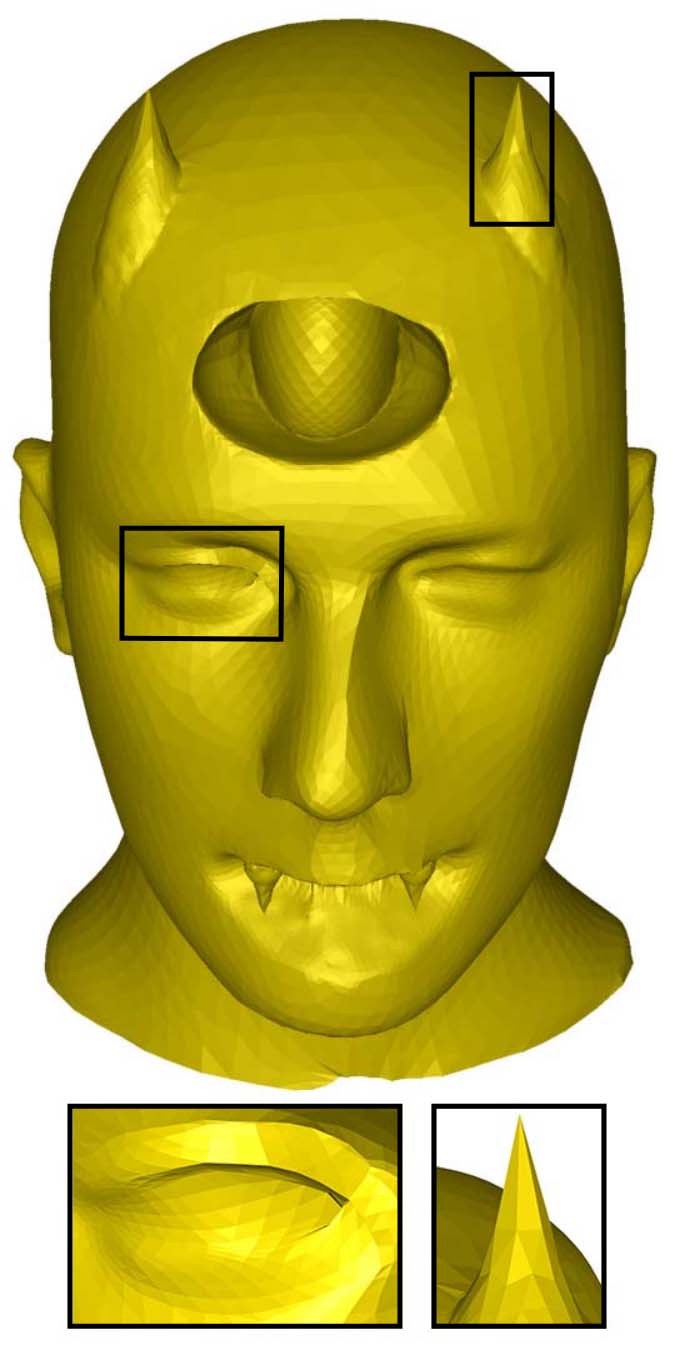} }}%
	\subfloat[\cite{BilNorm}]{{\includegraphics[width=2.1cm]{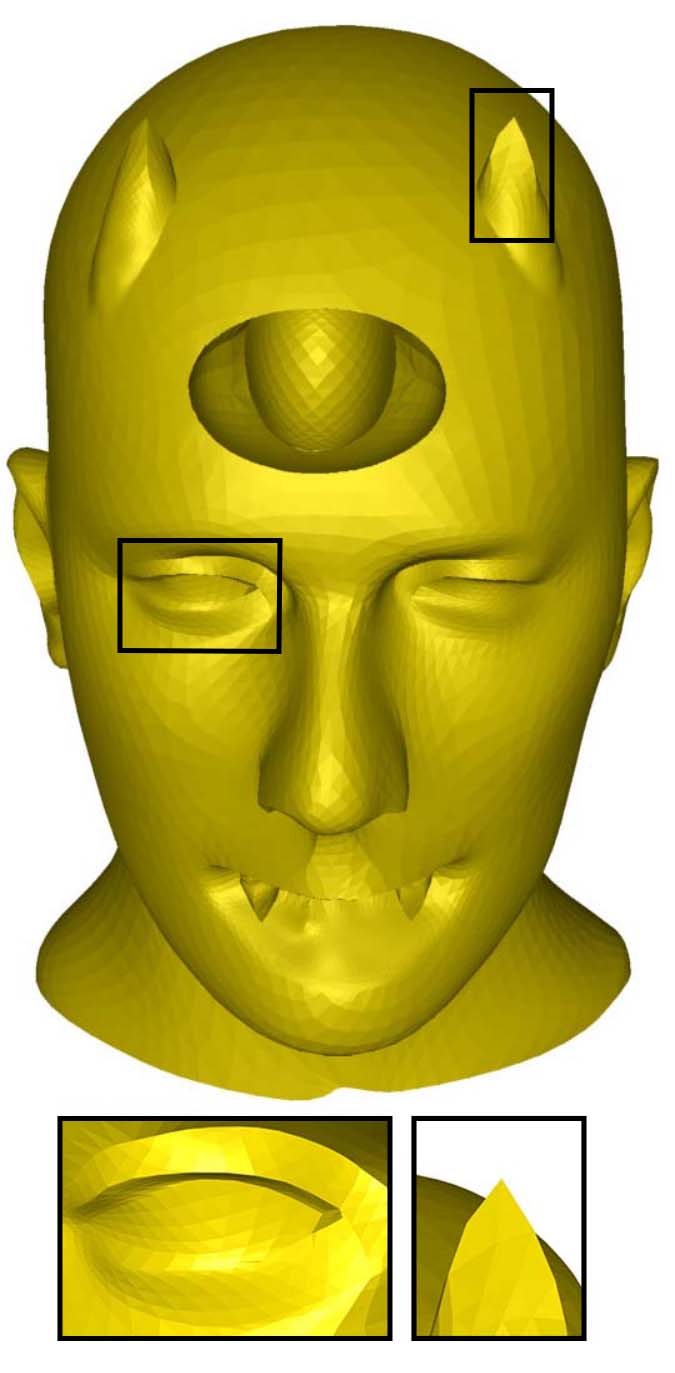} }}%
	\subfloat[\cite{L0Mesh}]{{\includegraphics[width=2.1cm]{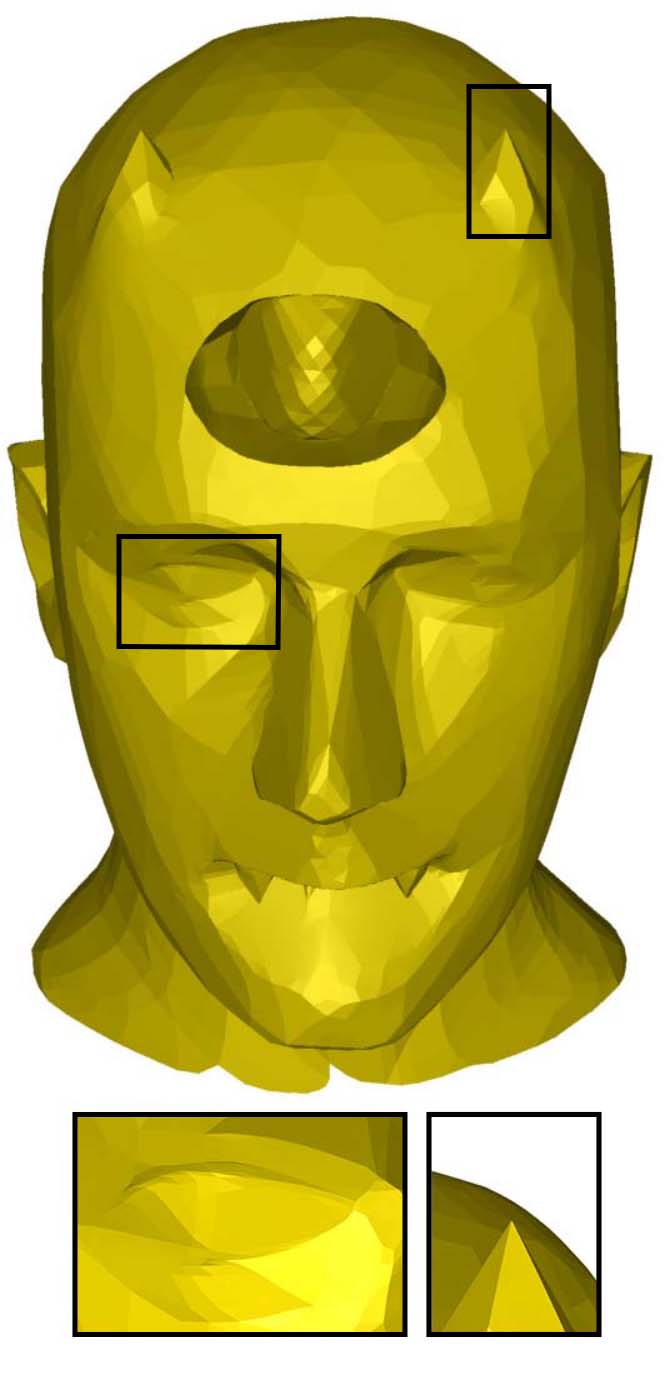} }}%
	\subfloat[\cite{Guidedmesh}]{{\includegraphics[width=2.1cm]{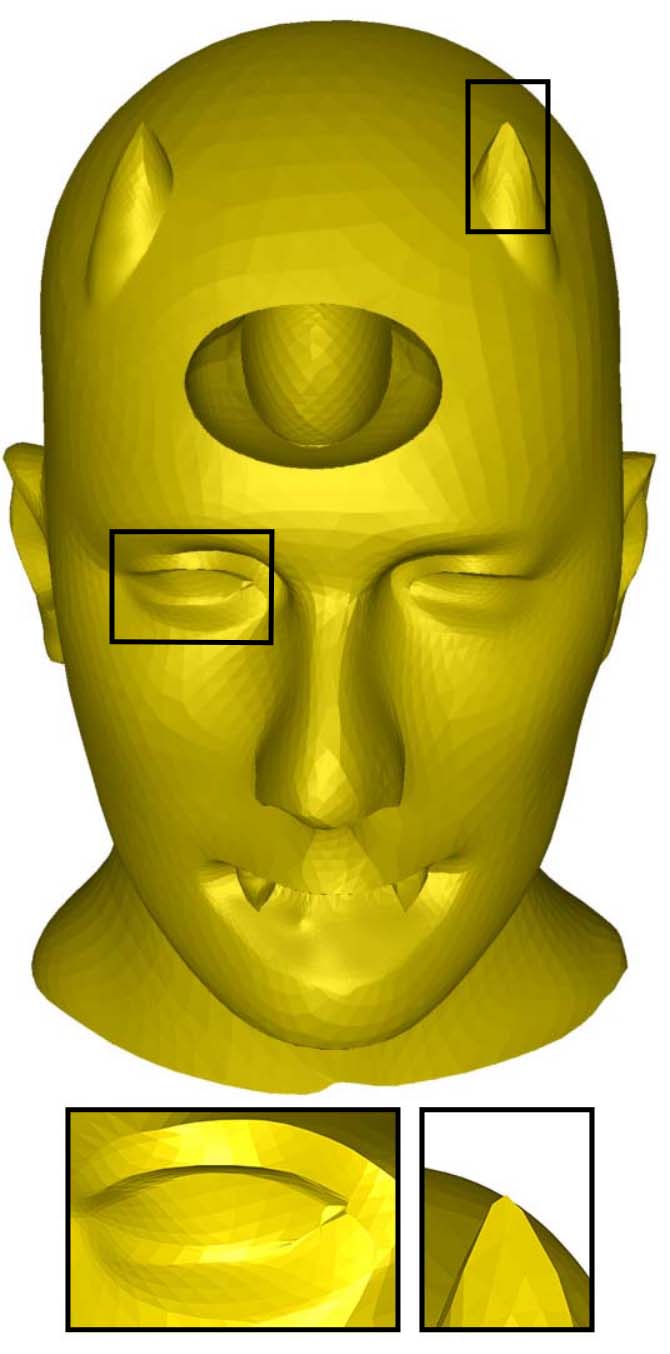} }}%
	\subfloat[\cite{binormal}]{{\includegraphics[width=2.1cm]{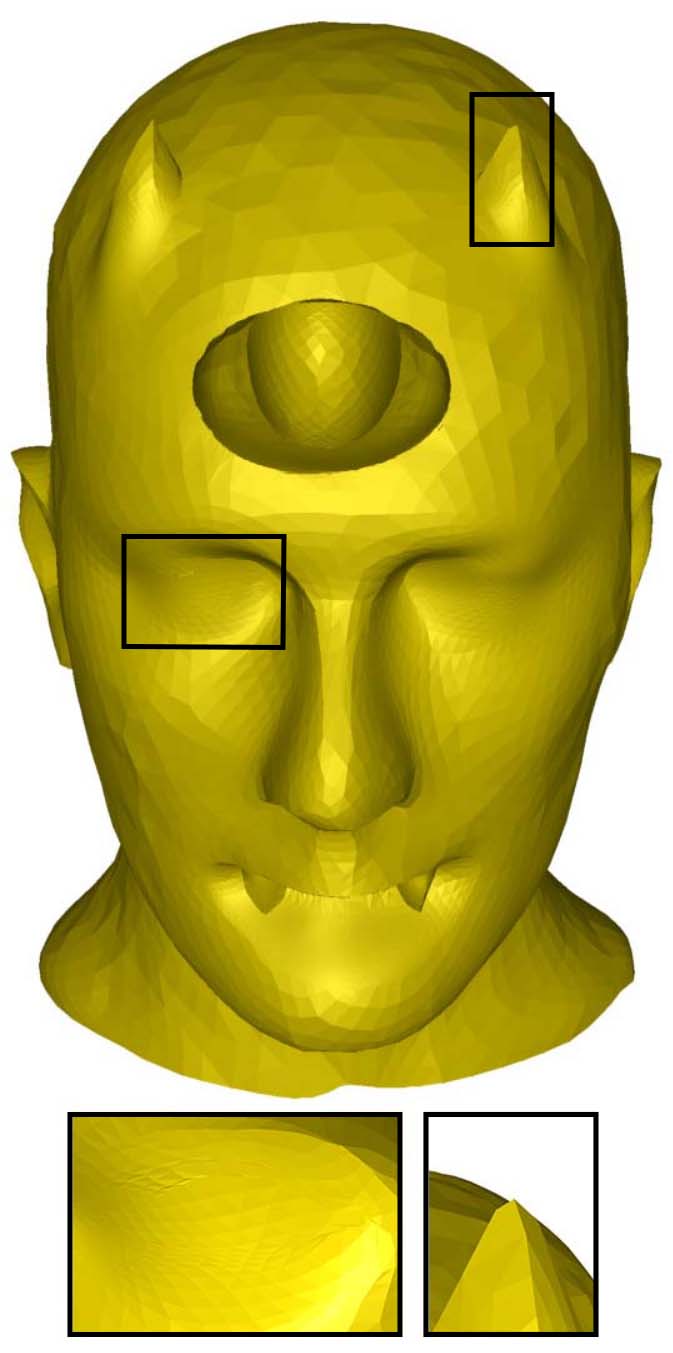} }}%
	\subfloat[Ours]{{\includegraphics[width=2.1cm]{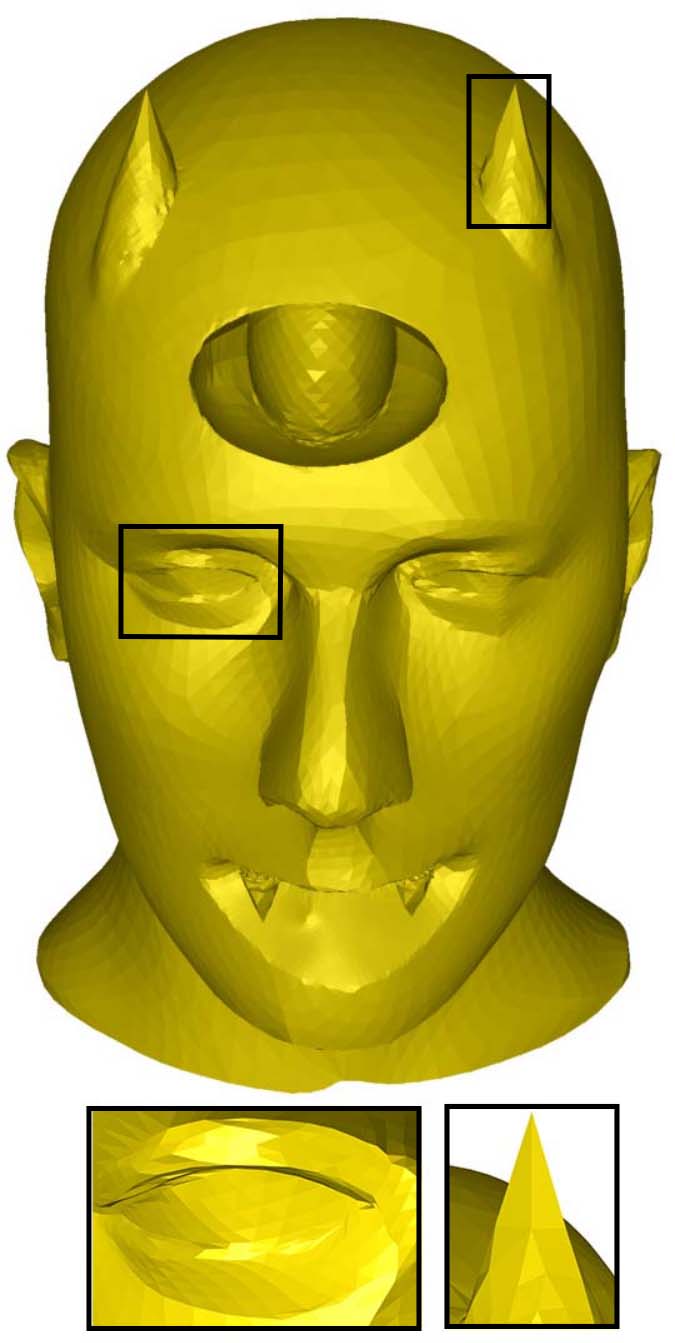} }}%
	\caption{ The Devil model consists of non-uniform triangles corrupted by Gaussian noise with standard deviation $\sigma_n= 0.15l_e$. The first row shows the results produced by state-of-the-art methods and our proposed method. The second row shows the magnified view of the left eye and the right horn of the devil. The results show that the proposed method has minimum shrinkage at the horn area.}%
	\label{fig:devRock}%
\end{figure*}

\begin{figure*}[bth]%
	\centering
	\subfloat[Noisy]{{\includegraphics[width=2.4cm]{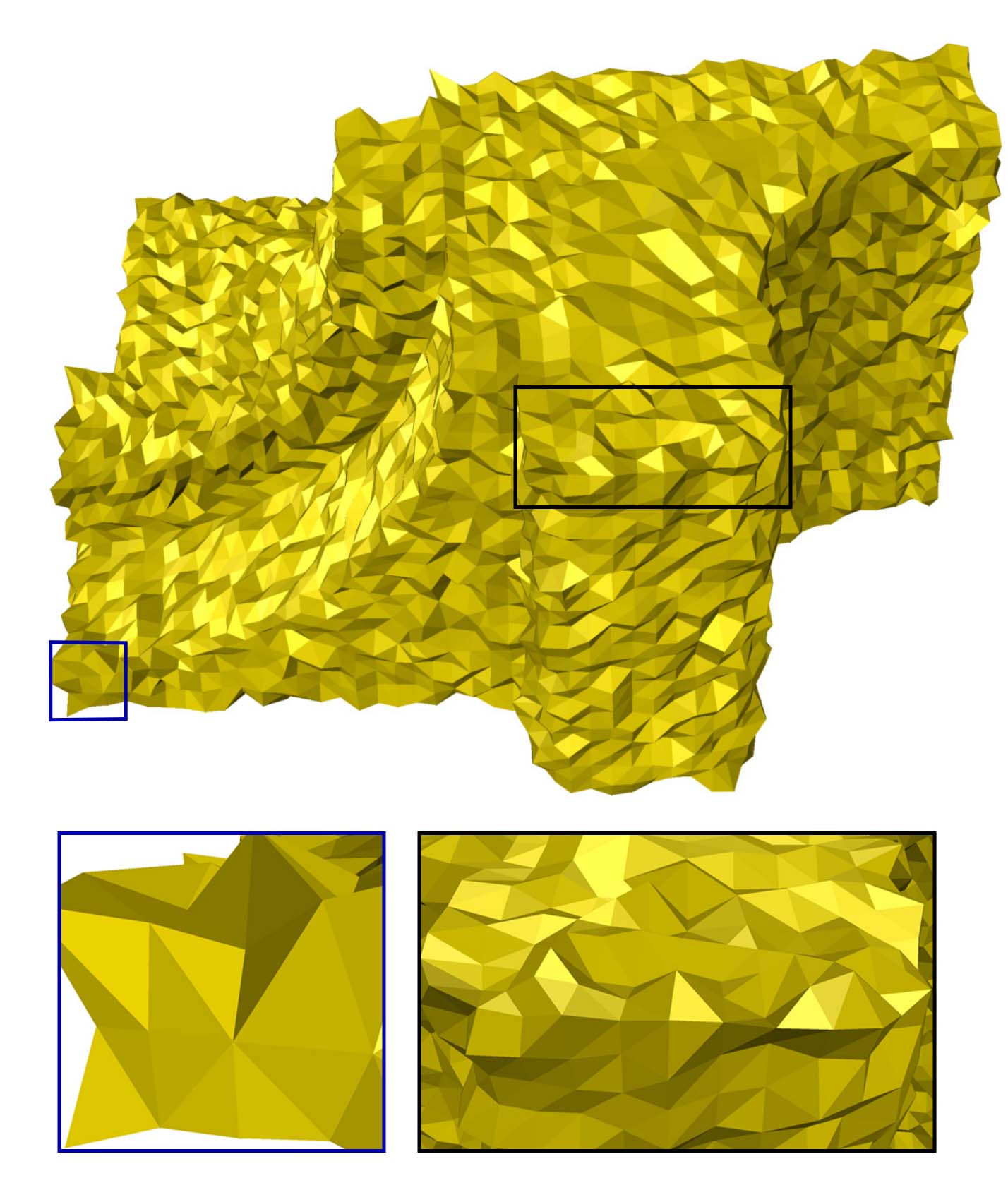} }}%
	\subfloat[\cite{BilFleish}]{{\includegraphics[width=2.4cm]{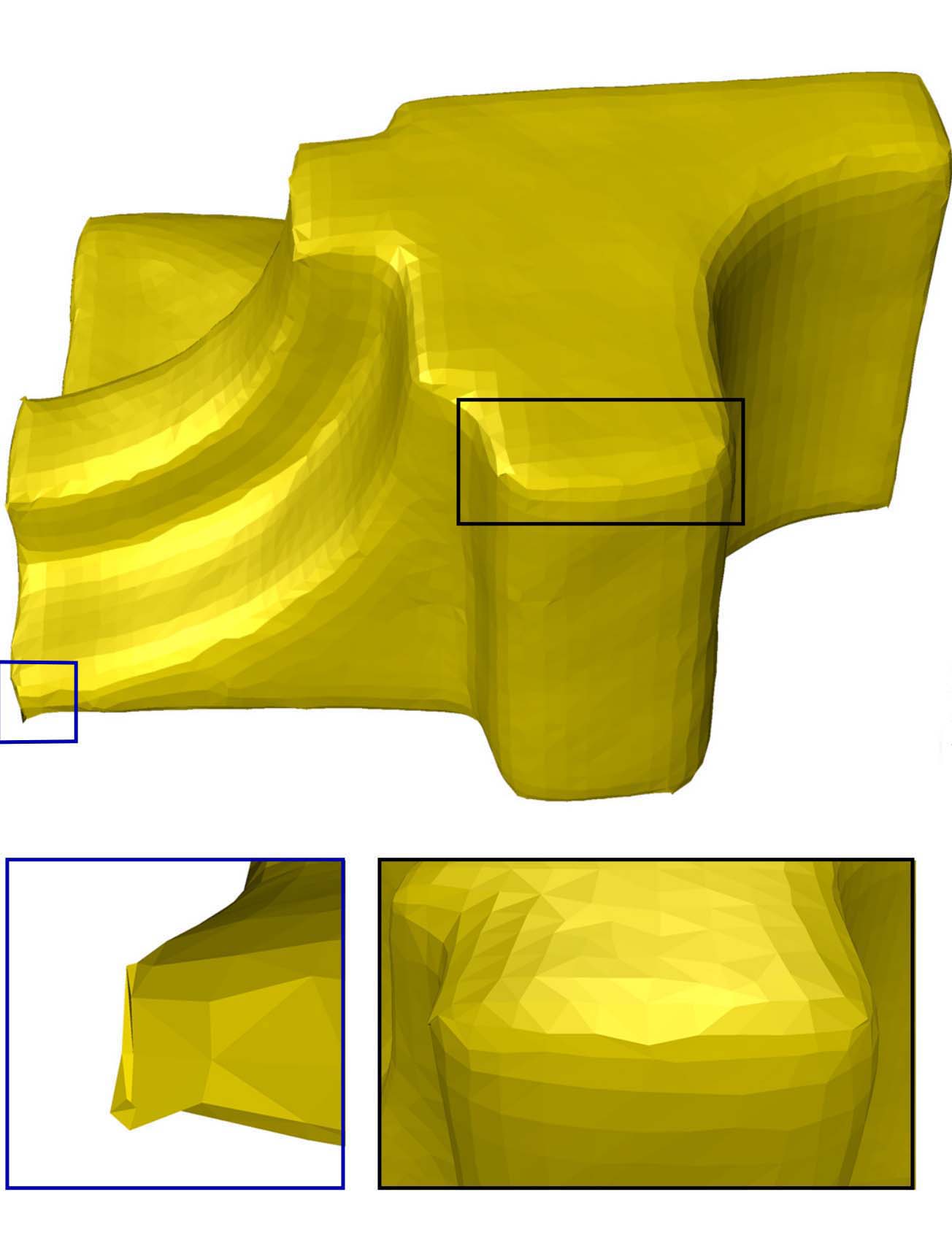} }}%
	\subfloat[\cite{aniso}]{{\includegraphics[width=2.4cm]{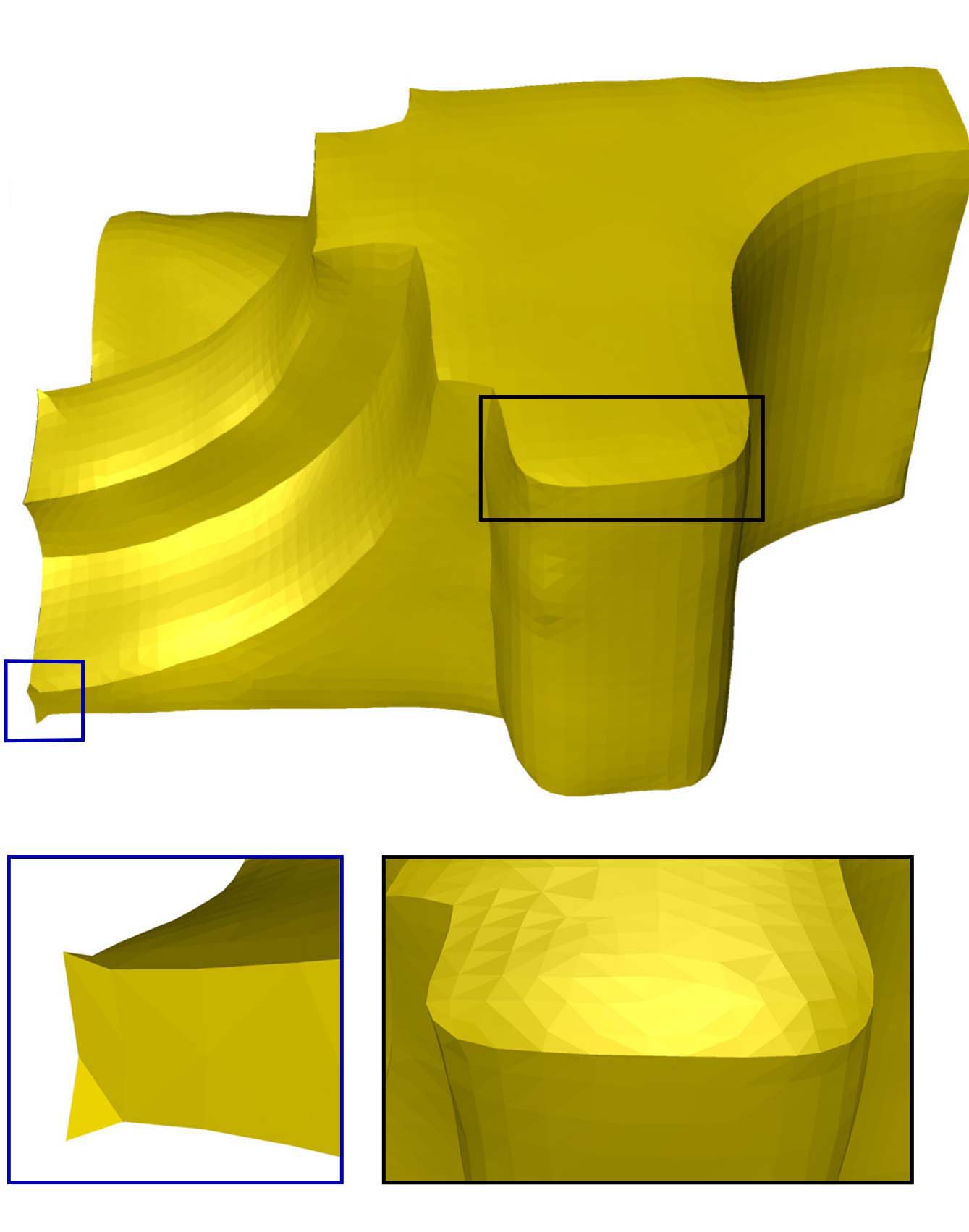} }}%
	\subfloat[\cite{BilNorm}]{{\includegraphics[width=2.4cm]{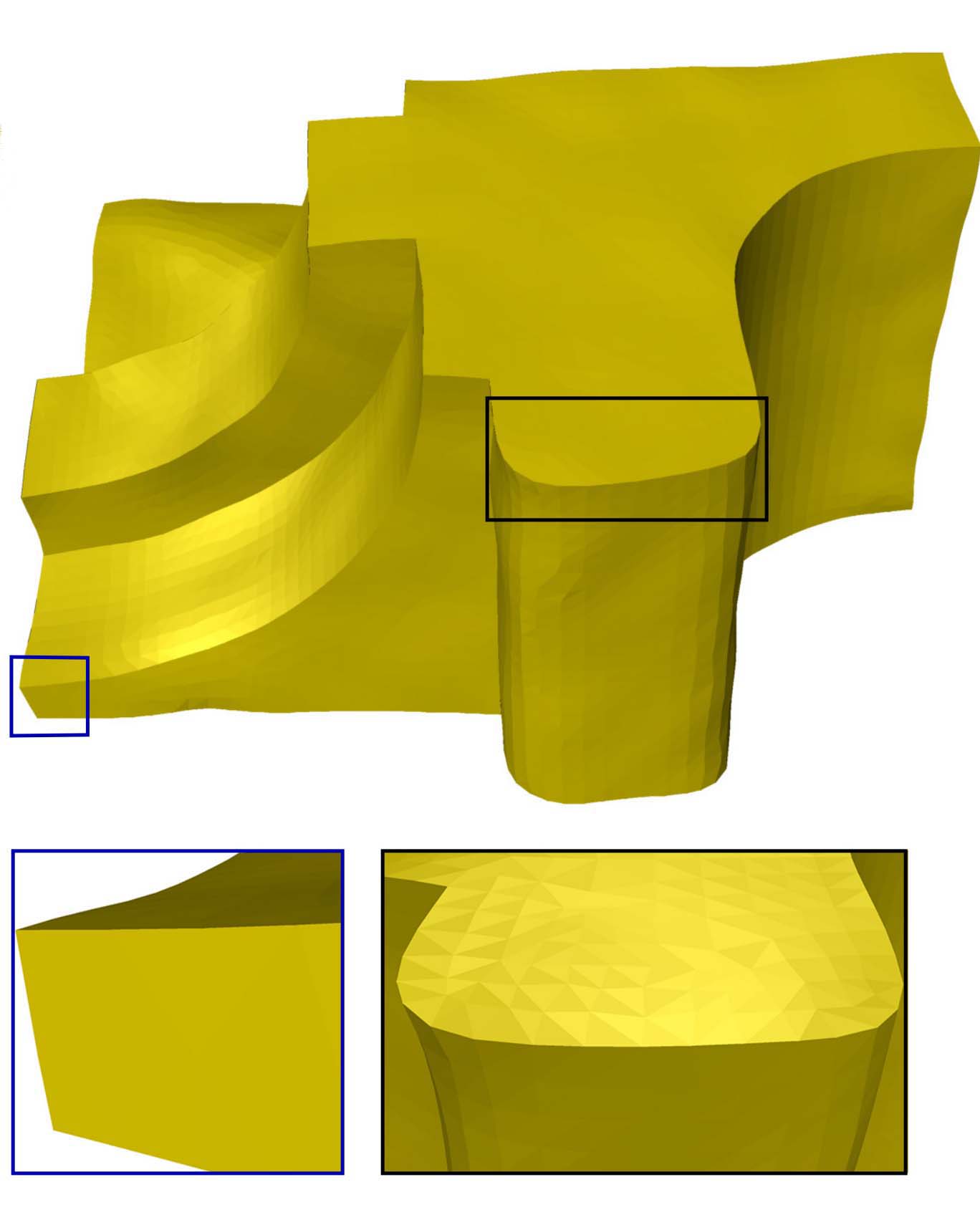} }}%
	\subfloat[\cite{L0Mesh}]{{\includegraphics[width=2.4cm]{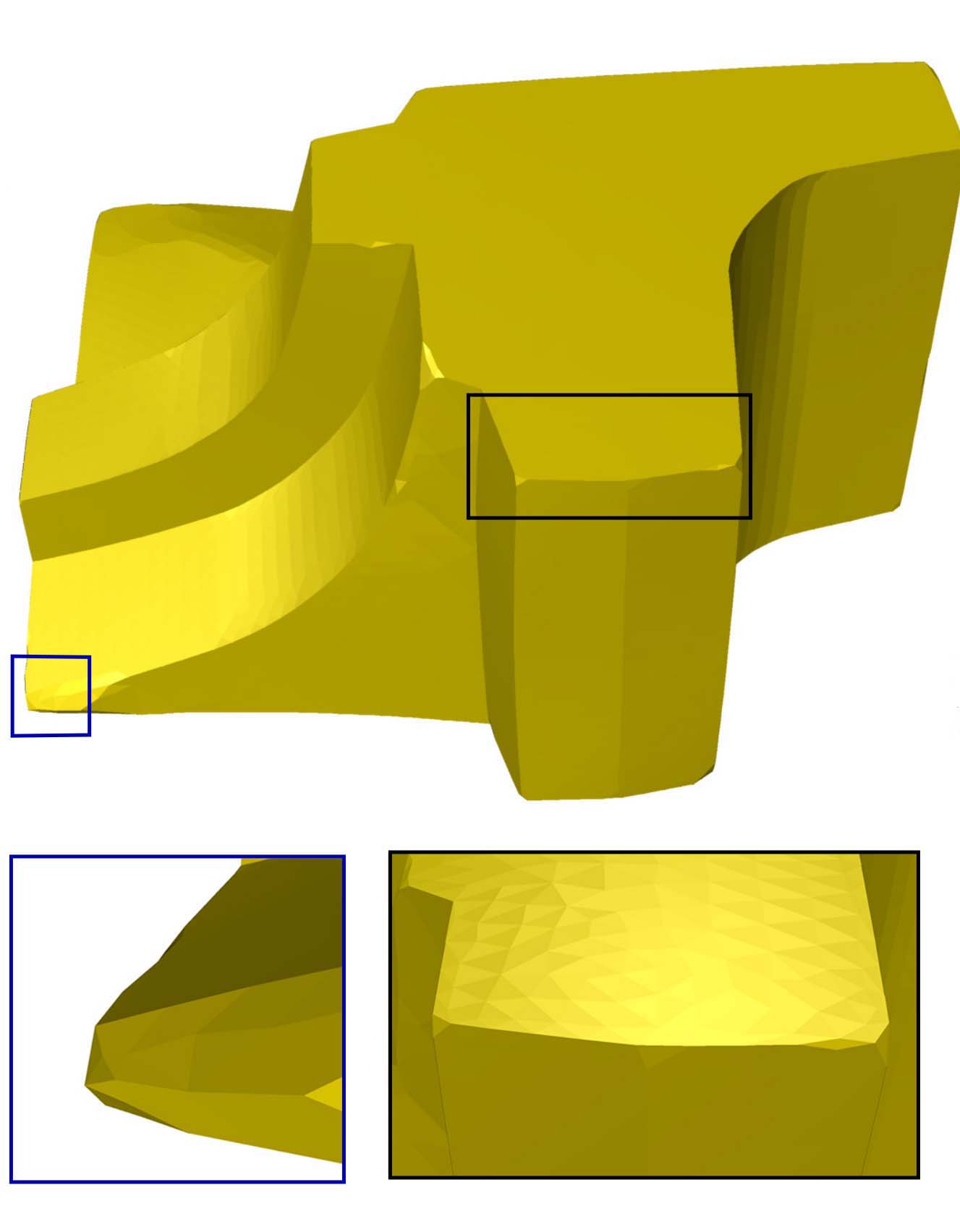} }}%
	\subfloat[\cite{Guidedmesh}]{{\includegraphics[width=2.4cm]{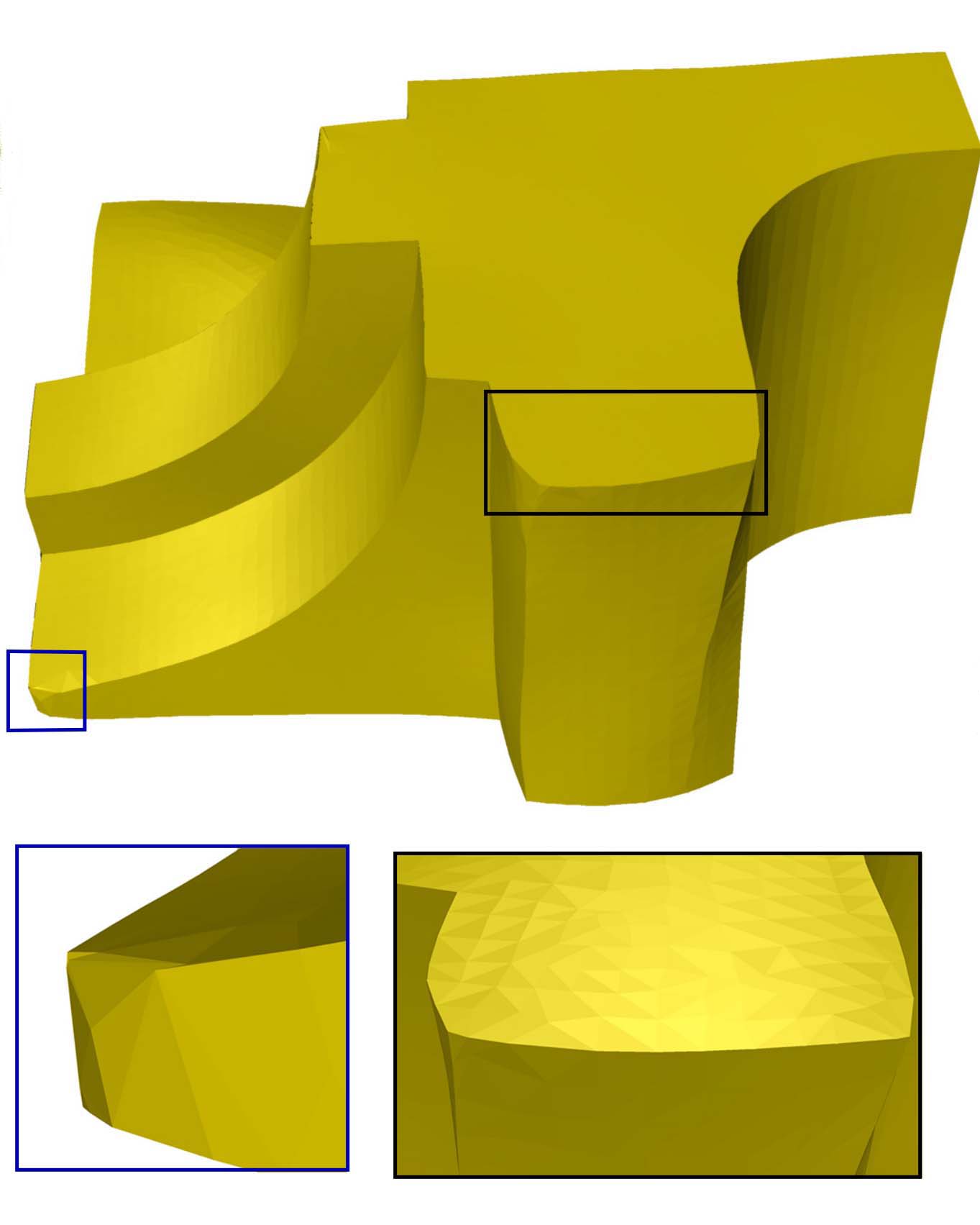} }}%
	\subfloat[Ours]{{\includegraphics[width=2.4cm]{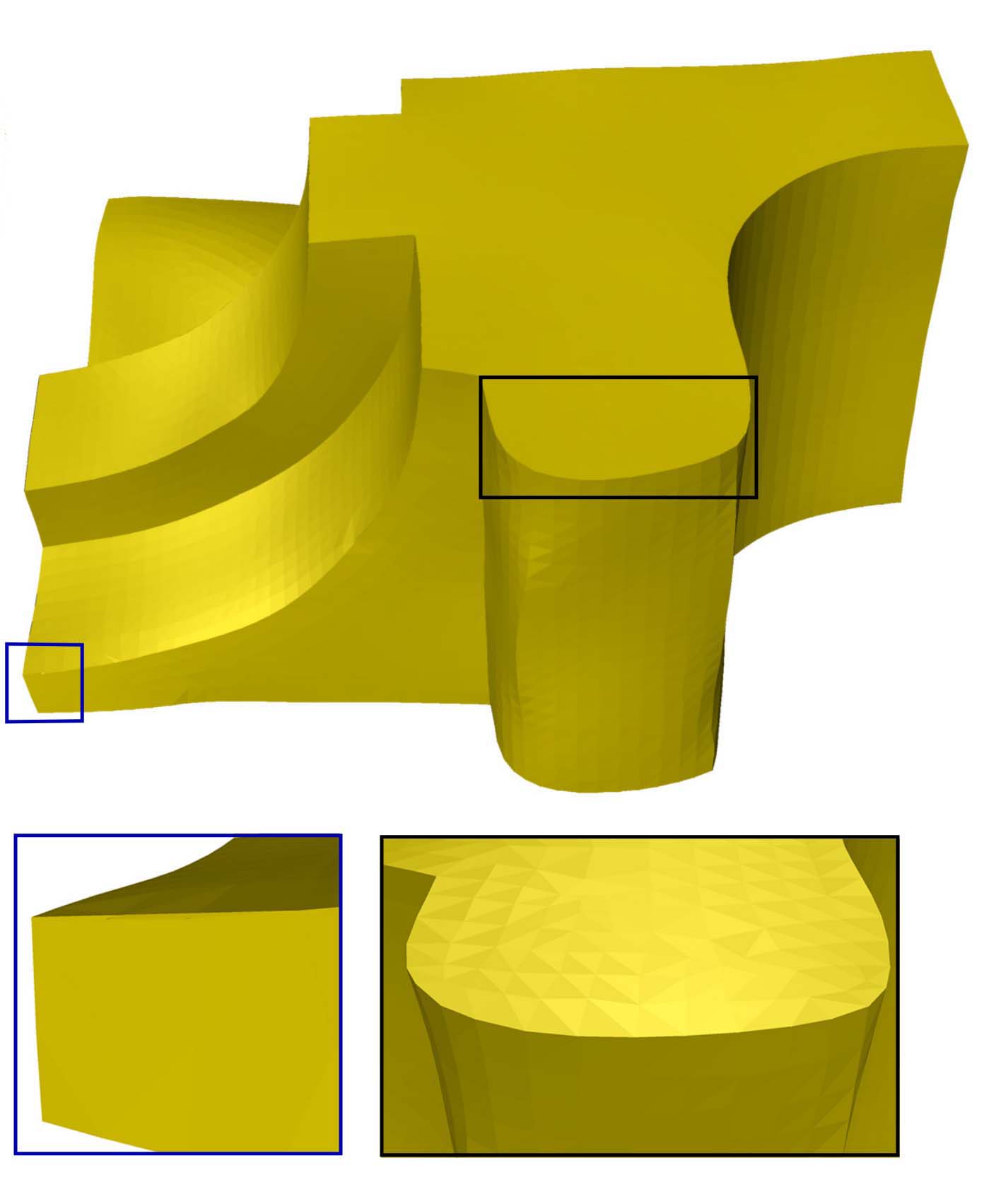} }}%
	\caption{Fandisk model corrupted with a Gaussian noise ($\sigma_e= 0.3l_e$) in random direction. The first row shows the results produced by state-of-the-art methods and our proposed method. The second row shows the corner and the cylindrical region of the concerned model. The proposed method does not produce any false features in the umbilical region while retains sharp features and corners.}%
	\label{fig:fanDisk}%
\end{figure*}
\begin{figure}%
	\centering
	\subfloat[Noisy]{{\includegraphics[width=2.8cm]{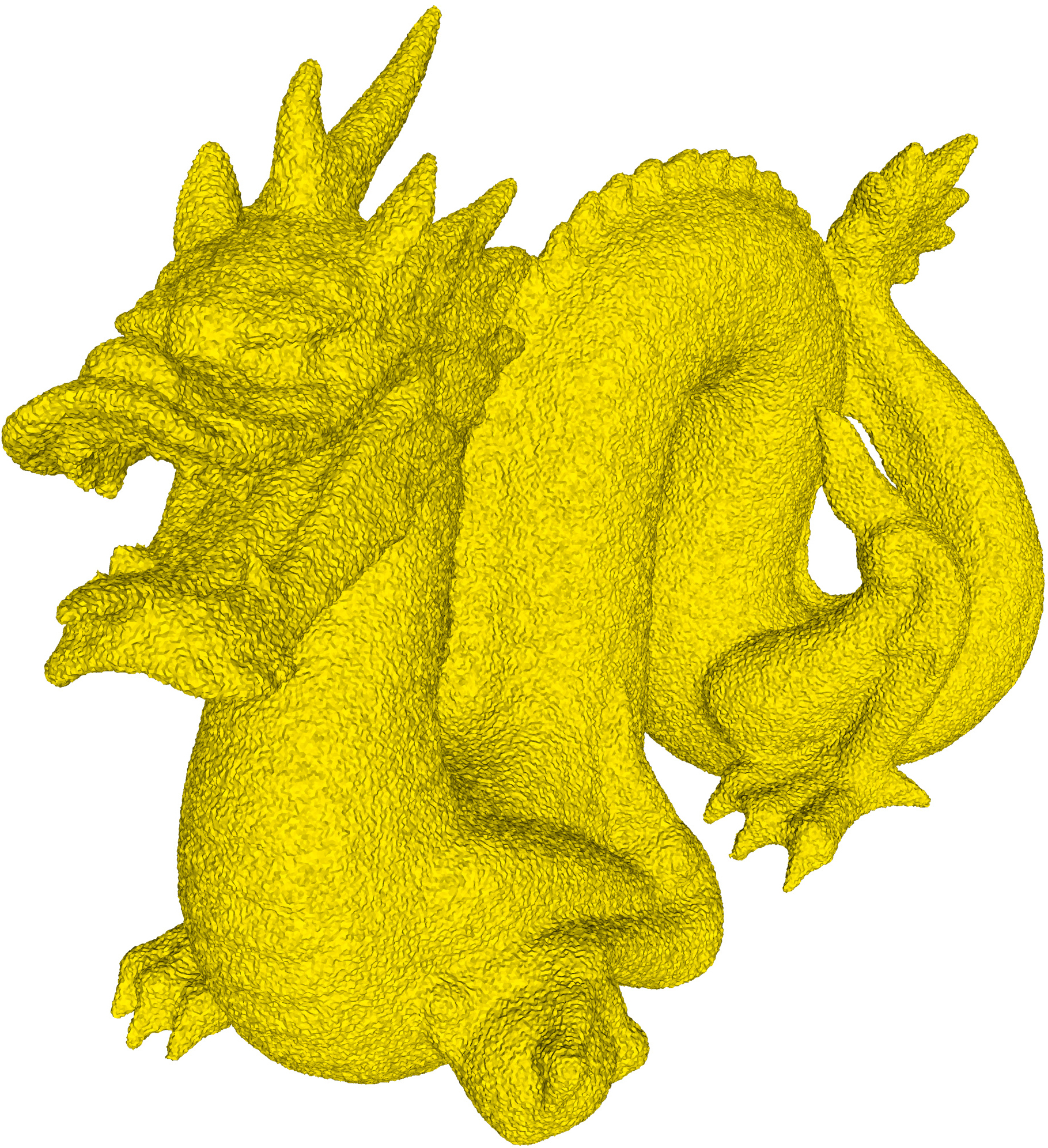} }}%
	\subfloat[\cite{AdamsKd}]{{\includegraphics[width=2.8cm]{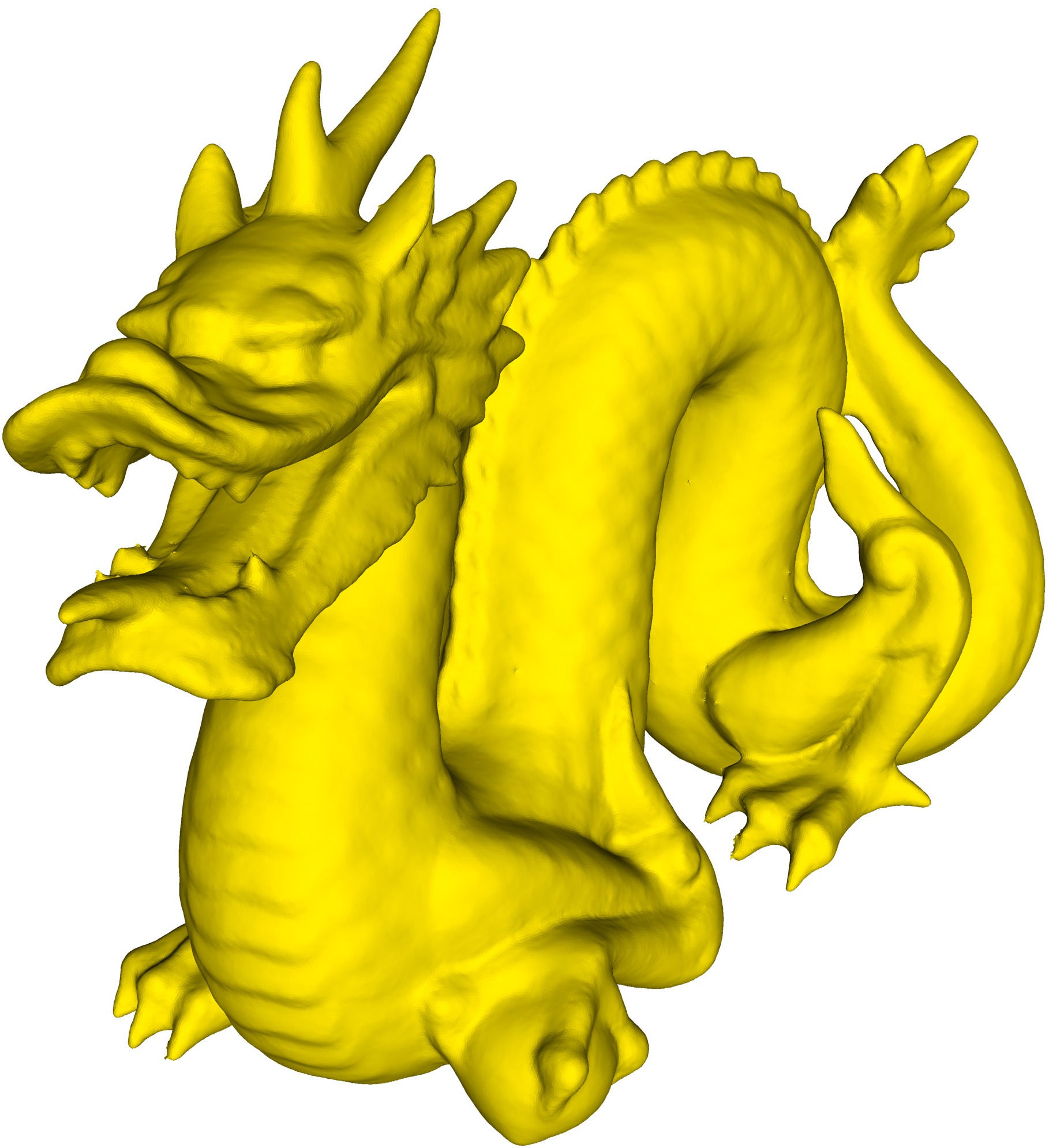} }}%
	\subfloat[Ours]{{\includegraphics[width=2.8cm]{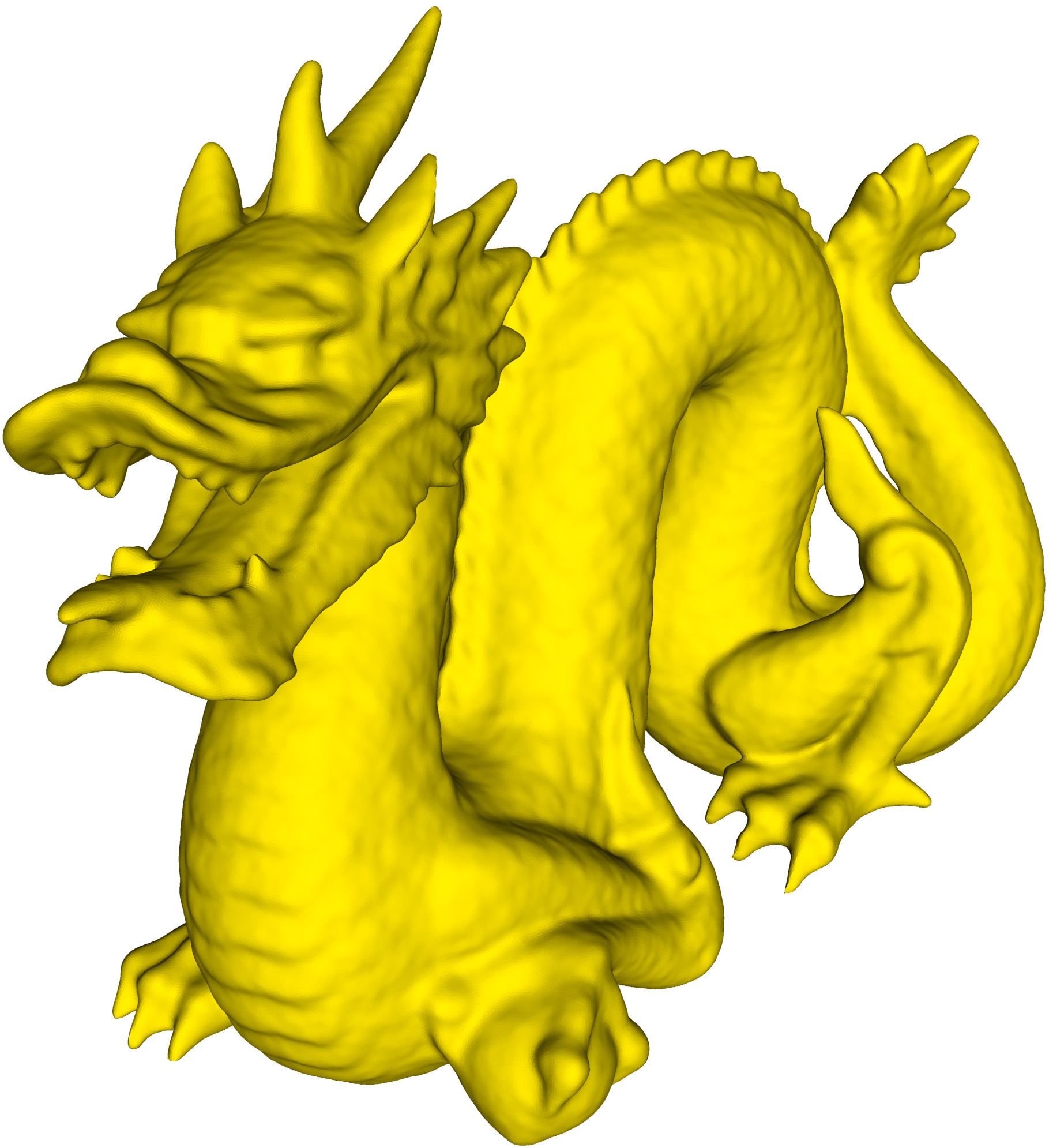} }}%
	
	\caption{Left to right: (a) Noisy Dragon model corrupted with synthetic noise. (b) Denoised model using \cite{AdamsKd}. (c) The result obtained using the proposed method .}%
	\label{fig:dragKd}%
\end{figure}

\begin{figure*}[h]%
	\centering
	\subfloat[Noisy]{{\includegraphics[width=2.4cm]{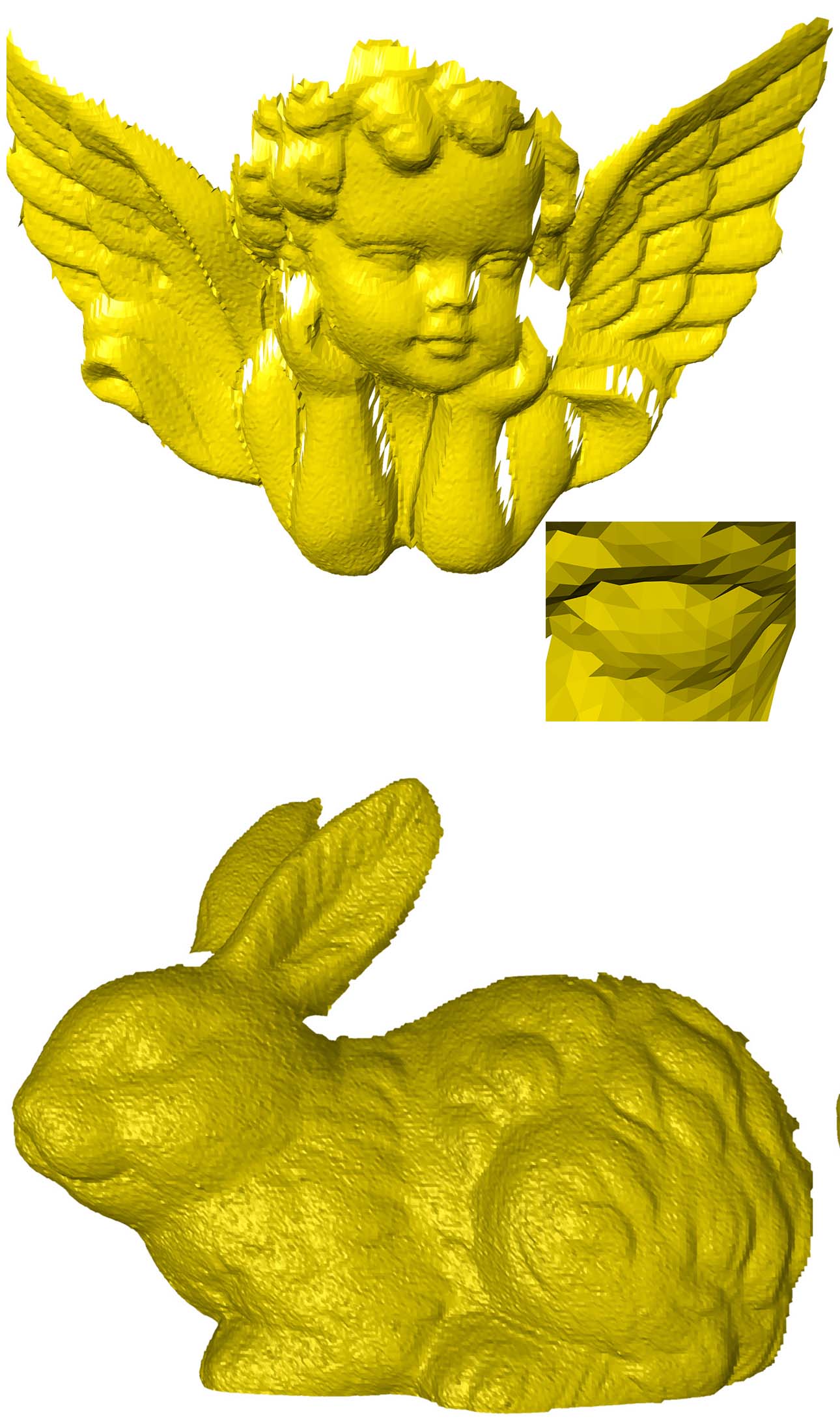} }}%
	\subfloat[\cite{BilFleish}]{{\includegraphics[width=2.4cm]{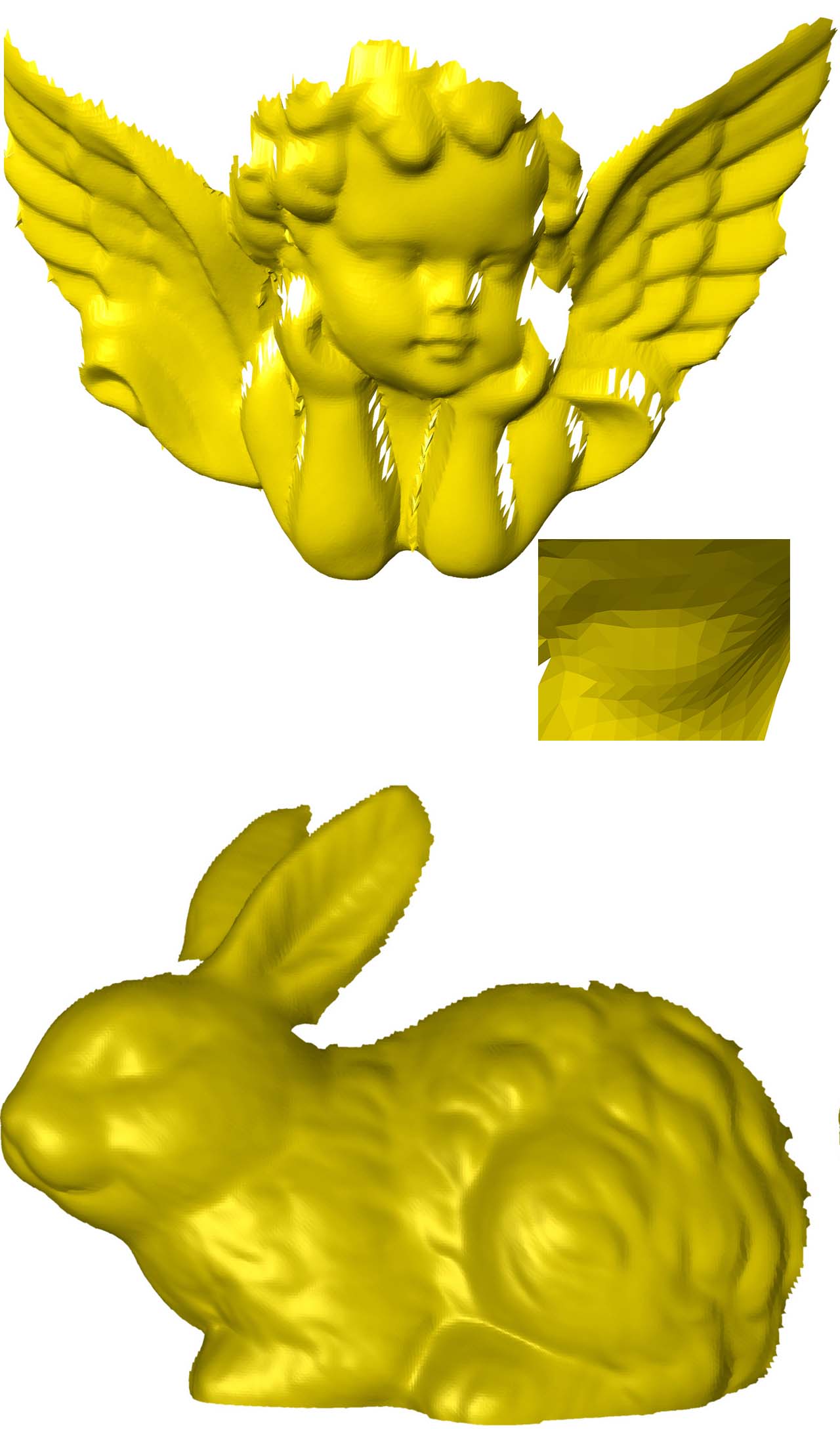} }}%
	\subfloat[\cite{aniso}]{{\includegraphics[width=2.4cm]{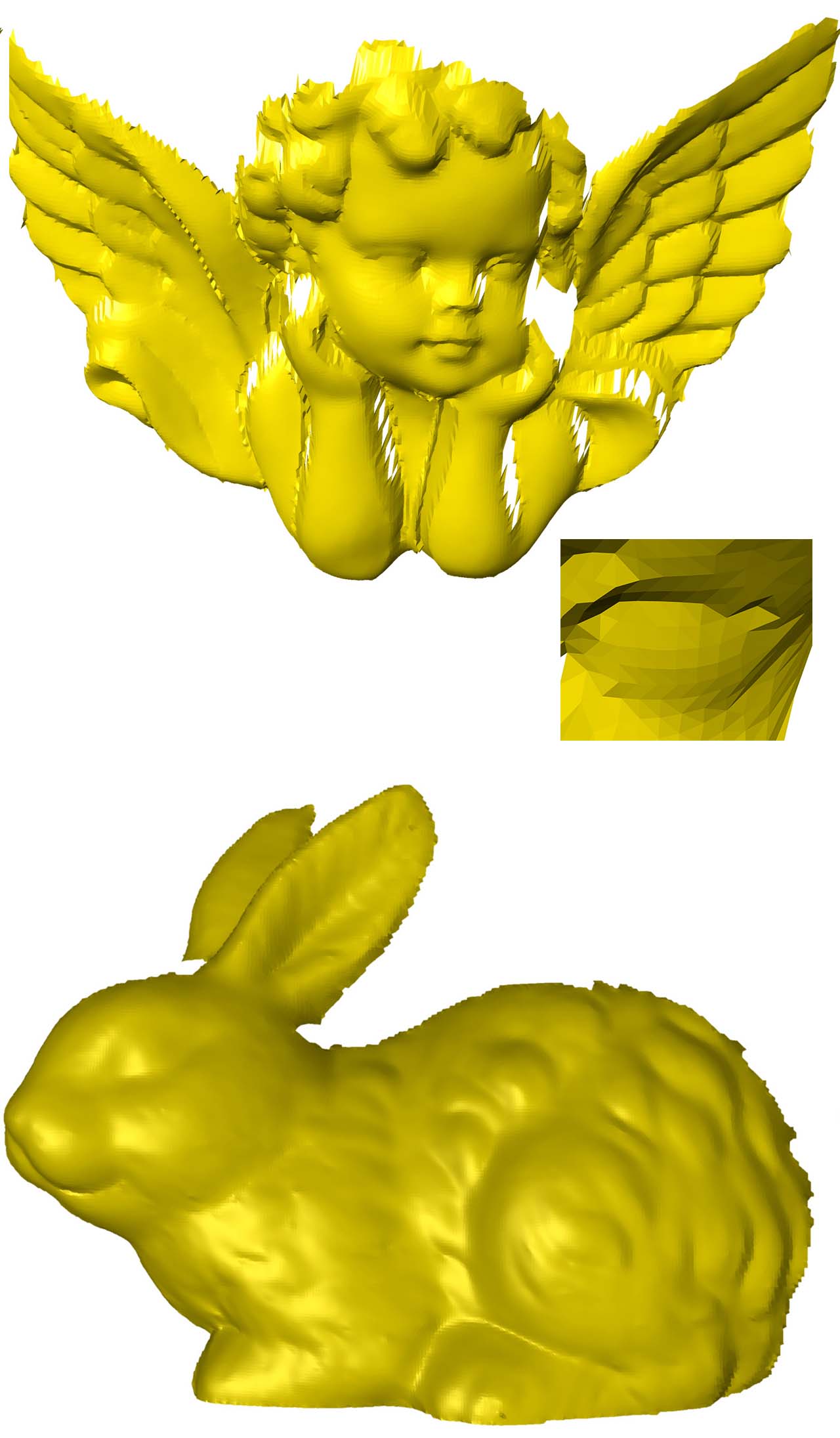} }}%
	\subfloat[\cite{BilNorm}]{{\includegraphics[width=2.4cm]{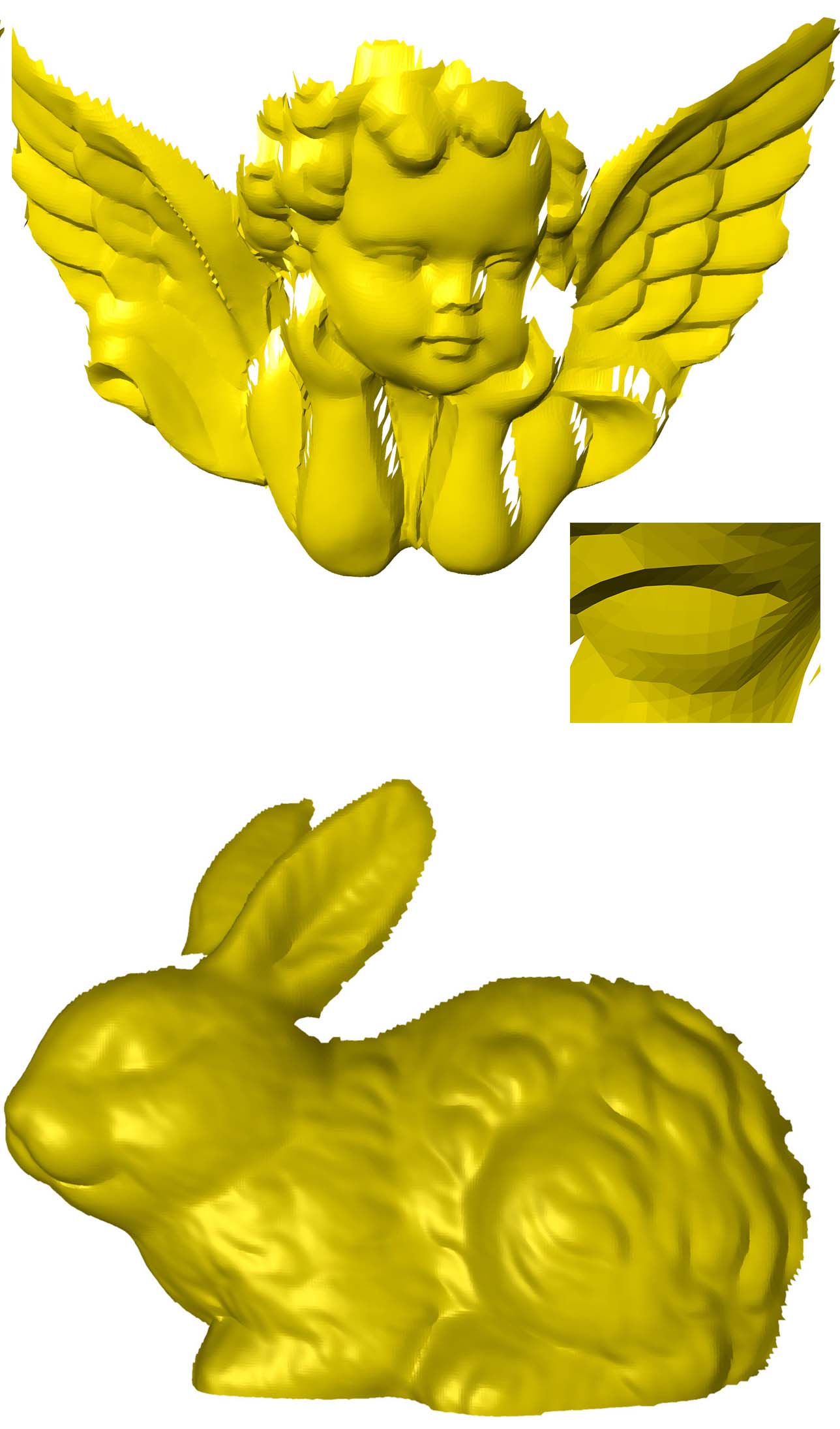} }}%
	\subfloat[\cite{L0Mesh}]{{\includegraphics[width=2.4cm]{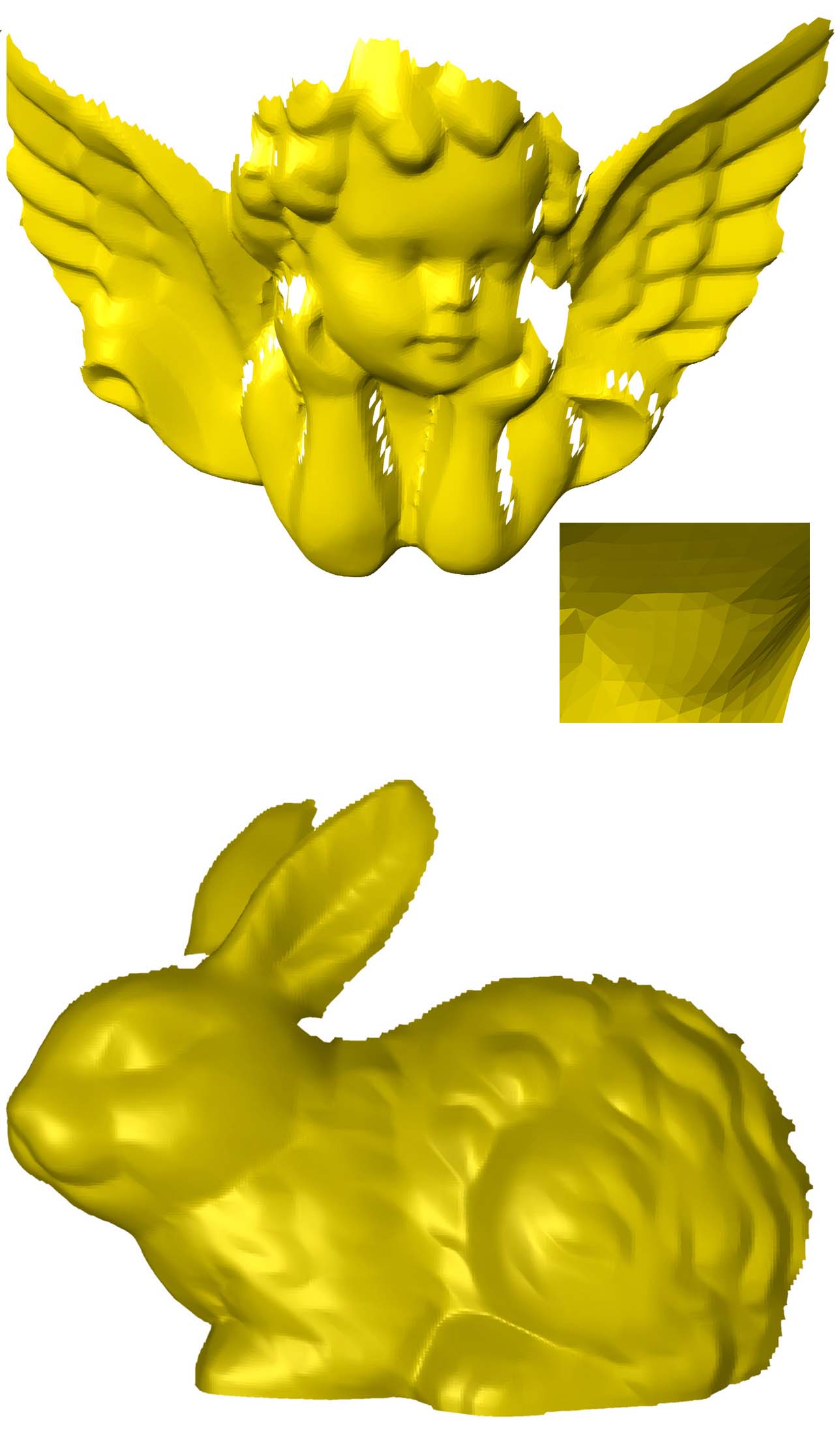} }}%
	\subfloat[\cite{Guidedmesh}]{{\includegraphics[width=2.4cm]{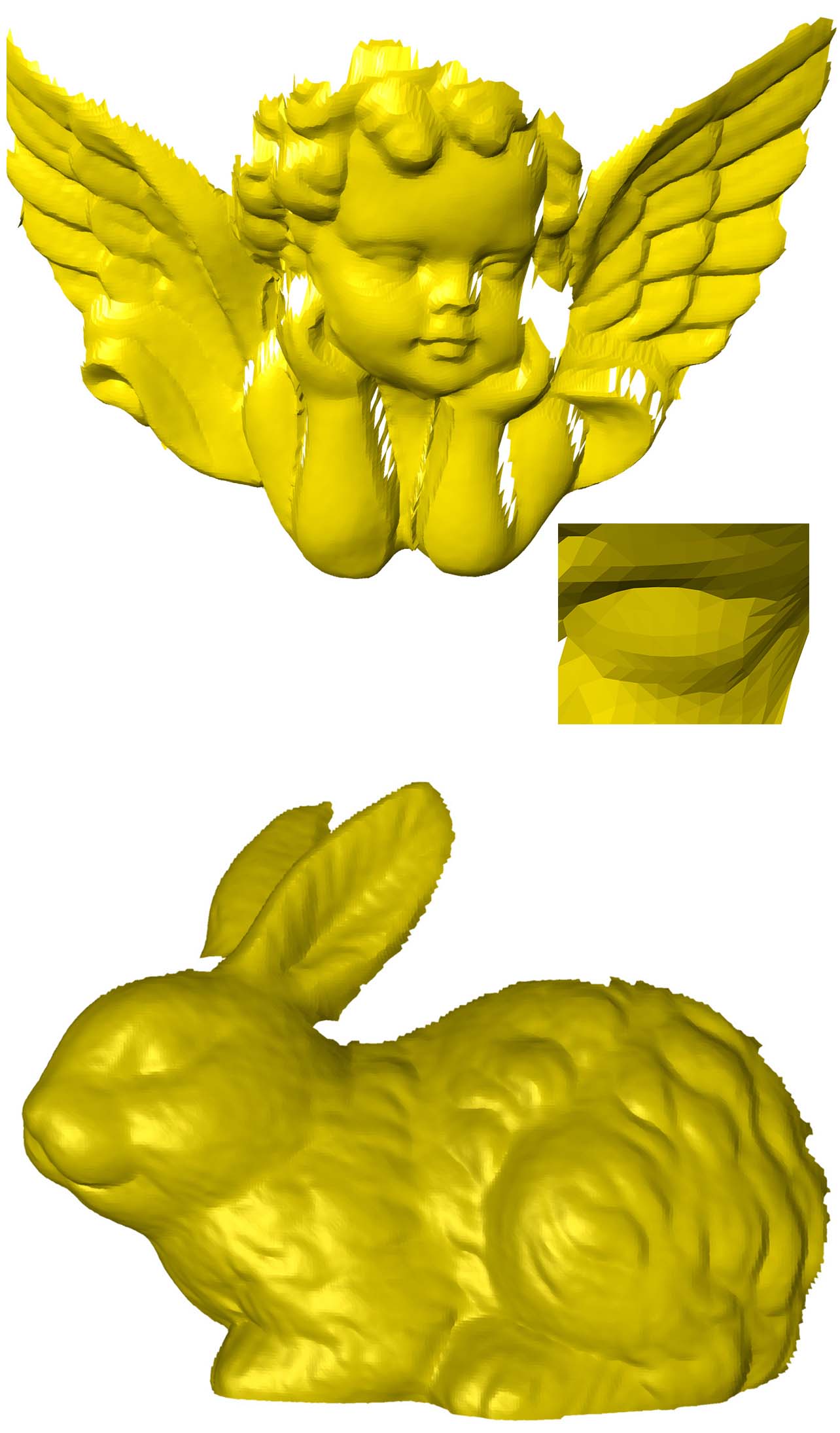} }}%
	\subfloat[Ours]{{\includegraphics[width=2.4cm]{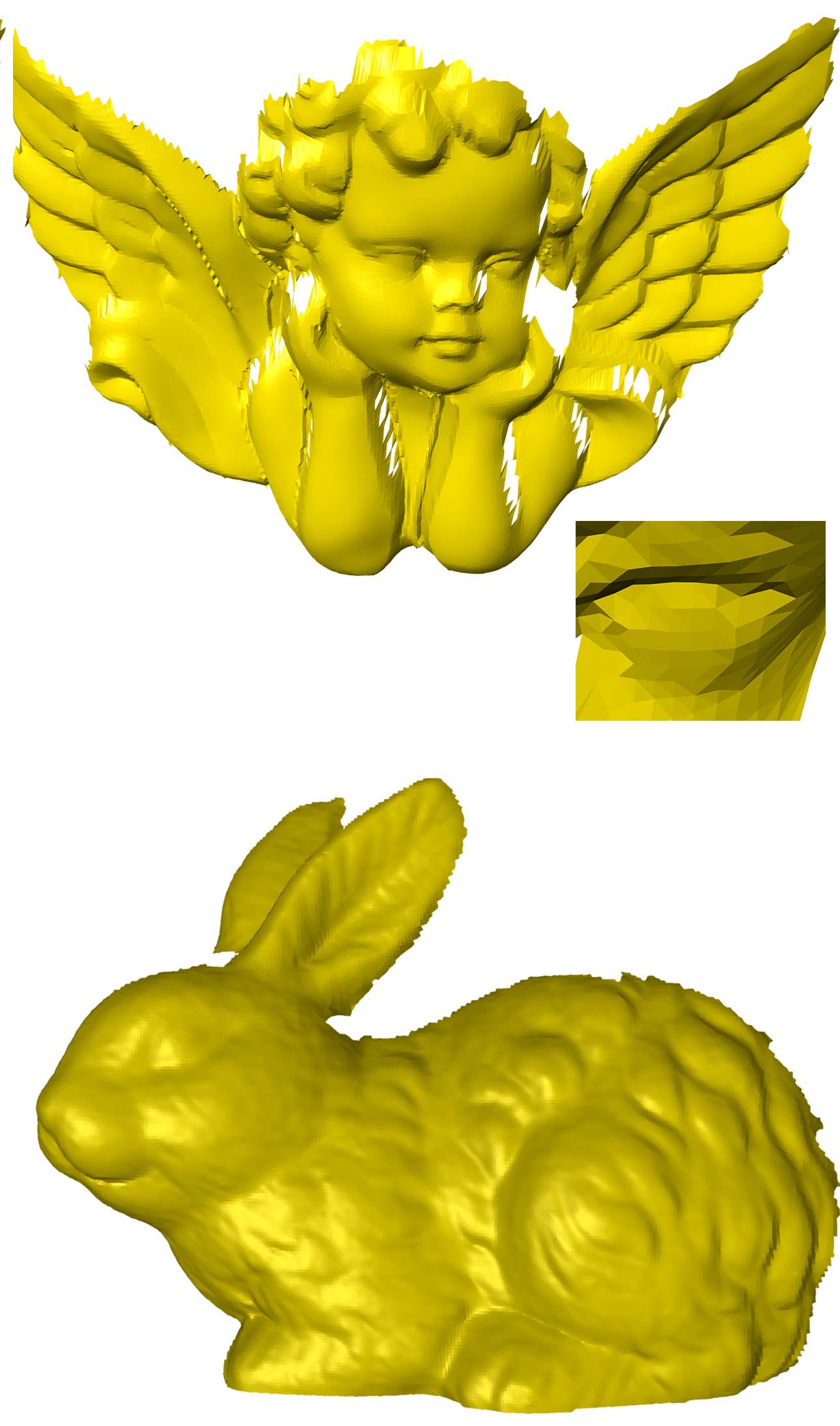} }}%
	\caption{Triangulated mesh surface ({real data}) corrupted by 3D scanner noise. Both rows show the results obtained by state-of-the-art methods and the proposed method for the Angel and the Rabbit models. }%
	\label{fig:realdata}%
\end{figure*}
\begin{figure*}[h]%
	\centering
	\subfloat[The gorgoyle model]{{\includegraphics[width=6.2cm]{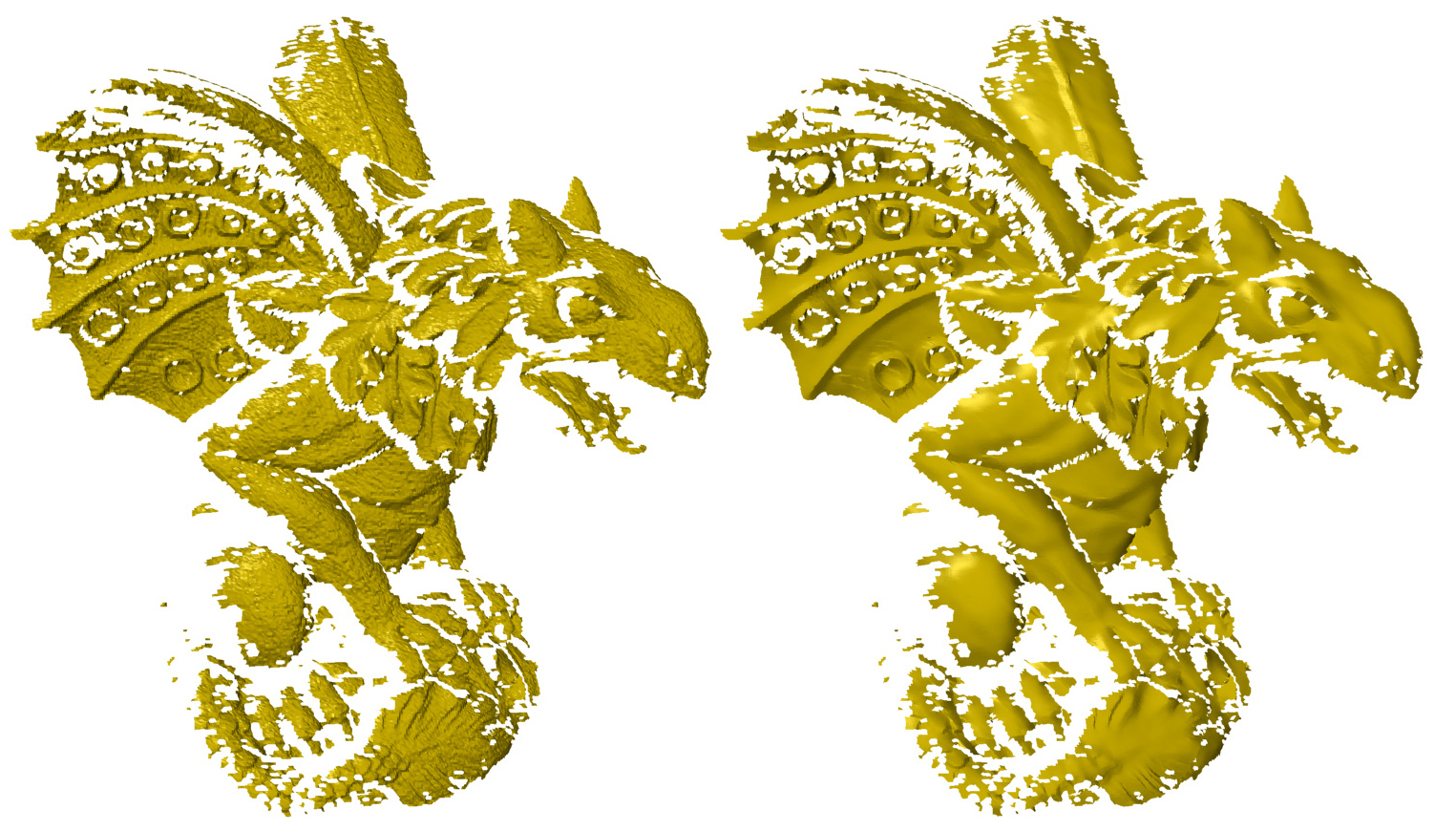} }}%
	\subfloat[The balljoint model]{{\includegraphics[width=5.5cm]{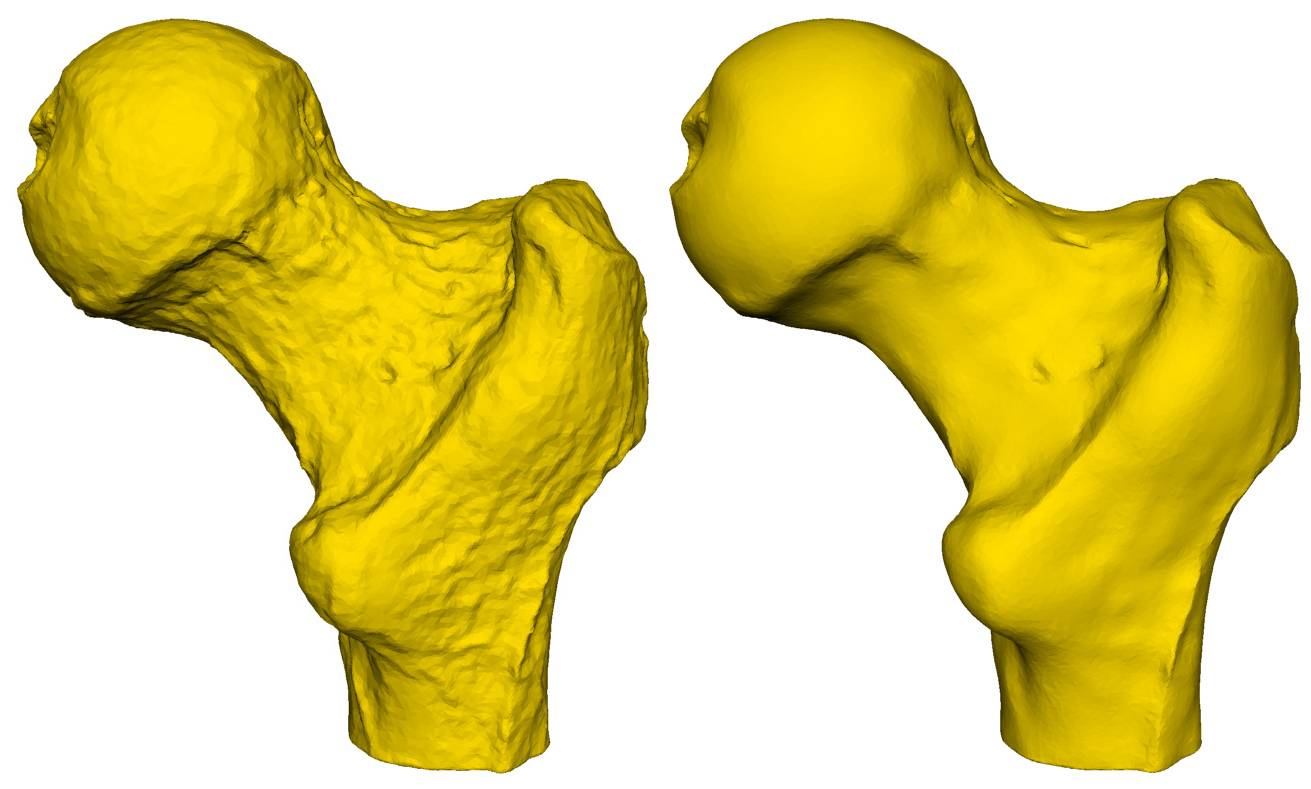} }}%
	\subfloat[The eagle model]{{\includegraphics[width=4.8cm]{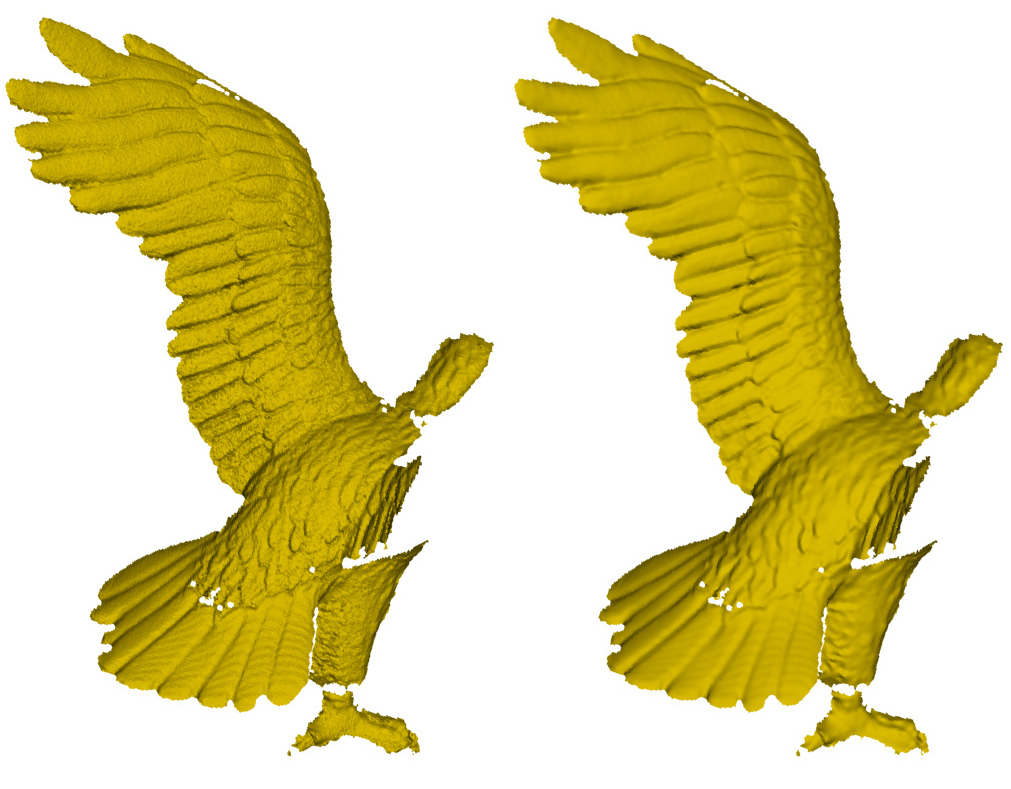} }}%
	\caption{Results obtained by our method against different kinds of real data captured by the laser scanner. The figure (a) and (c) show real life scans with a lot of holes and our method manages to produce good results.}
	\label{fig:realdiff}
\end{figure*}

\begin{figure*}[h]%
	\centering
	\subfloat[Random edge flip]{{\includegraphics[width=2.8cm]{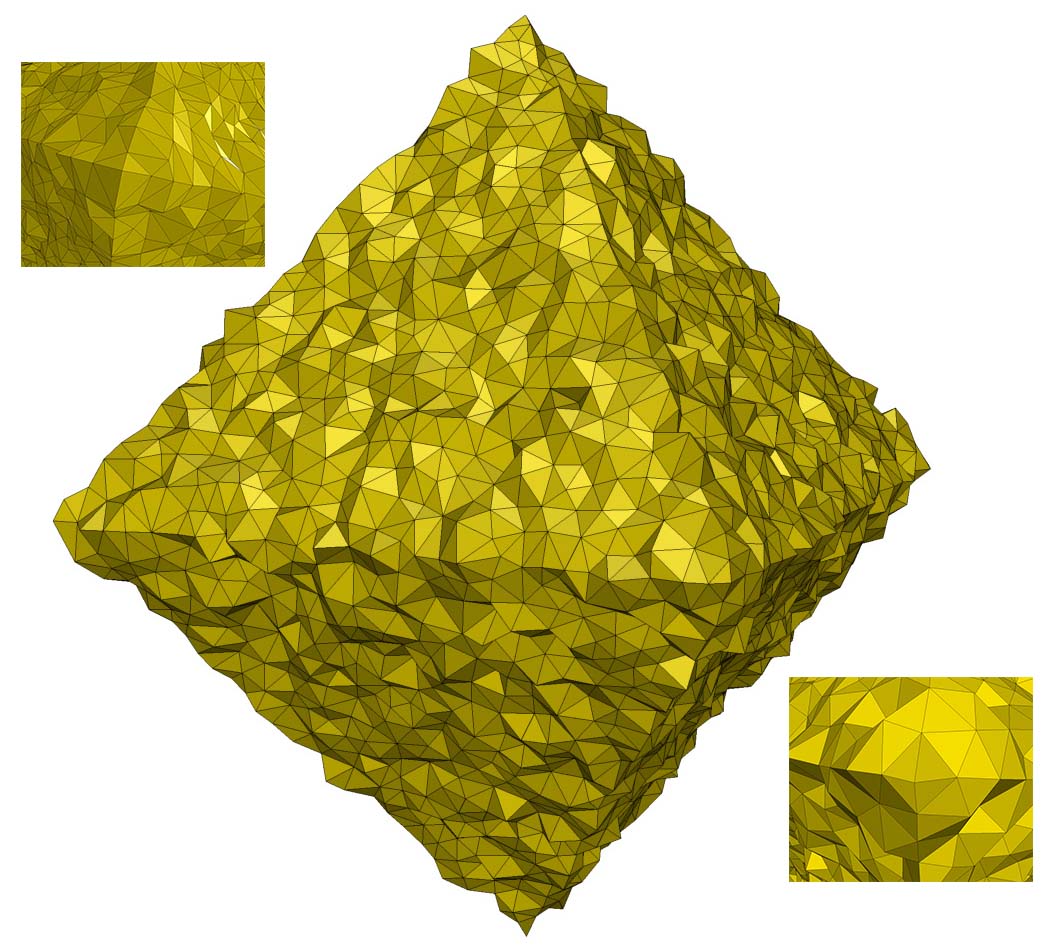} }}%
	\subfloat[\cite{BilFleish}]{{\includegraphics[width=2.8cm]{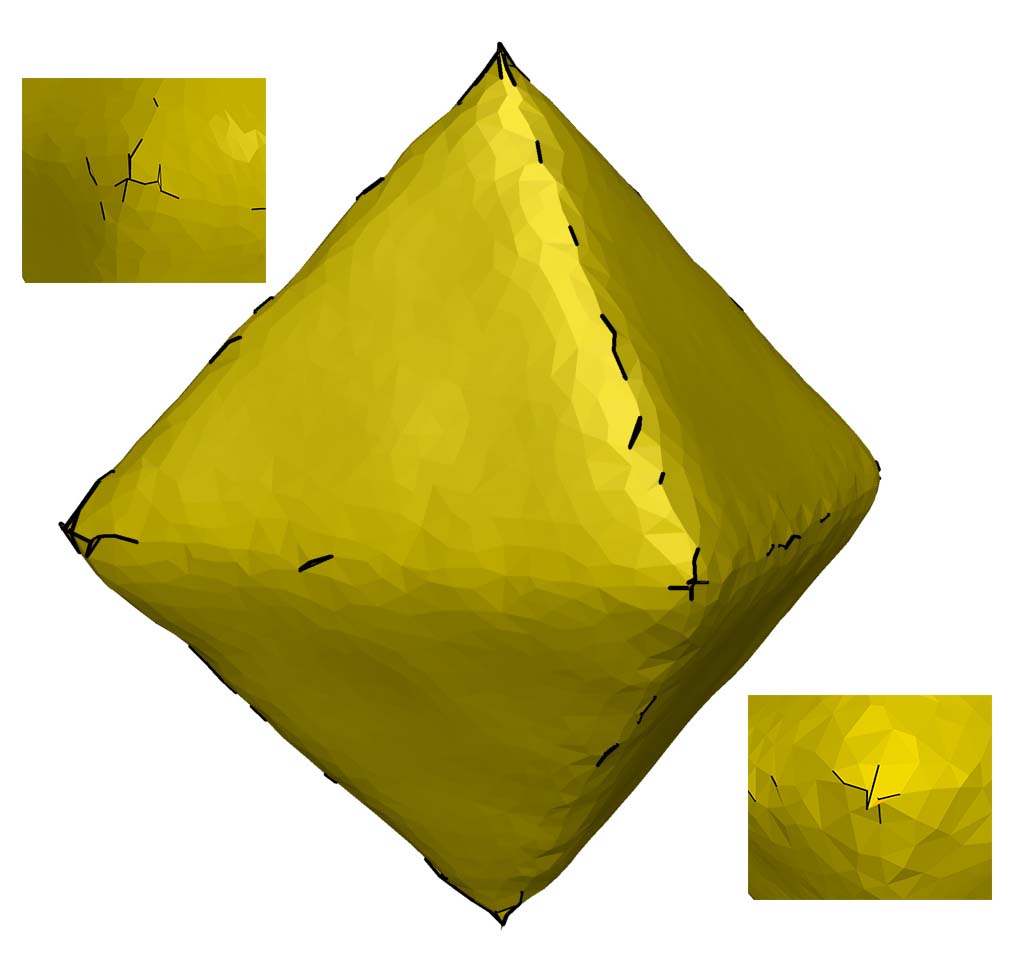} }}%
	\subfloat[\cite{aniso}]{{\includegraphics[width=2.8cm]{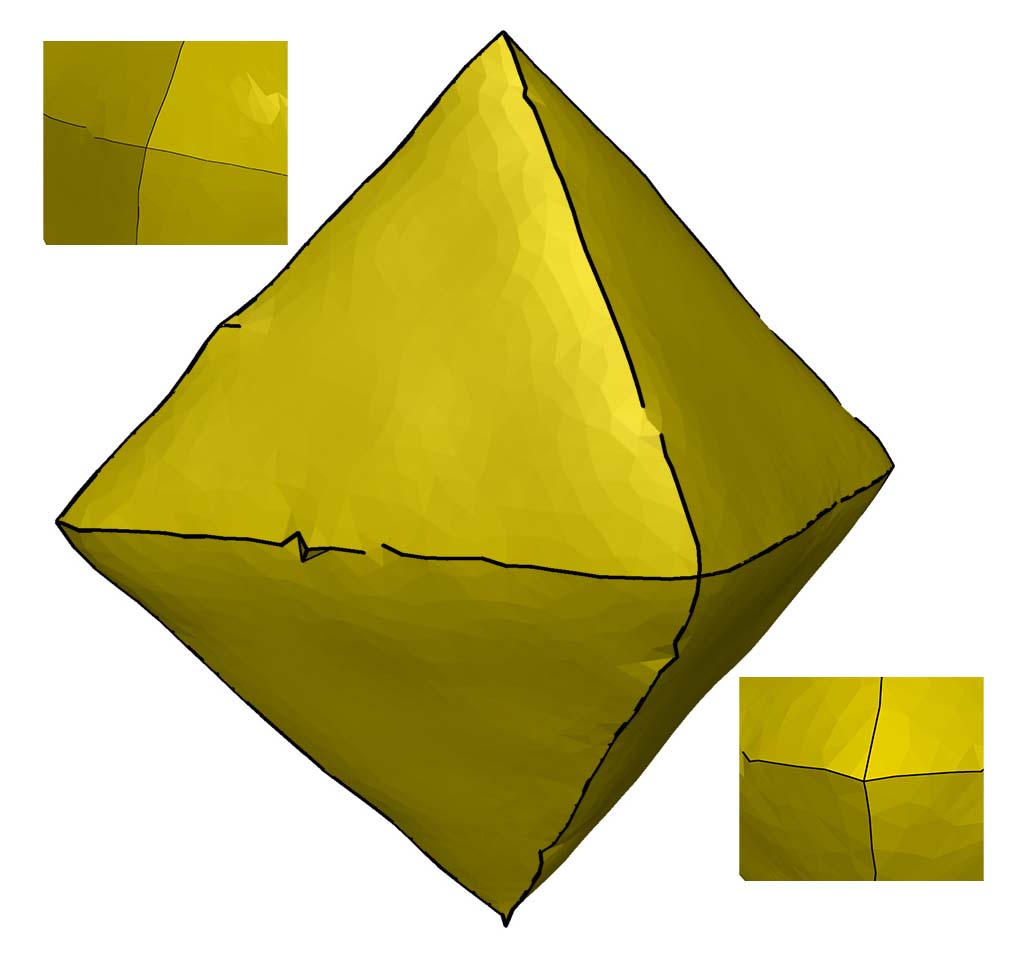} }}%
	\subfloat[\cite{BilNorm}]{{\includegraphics[width=2.8cm]{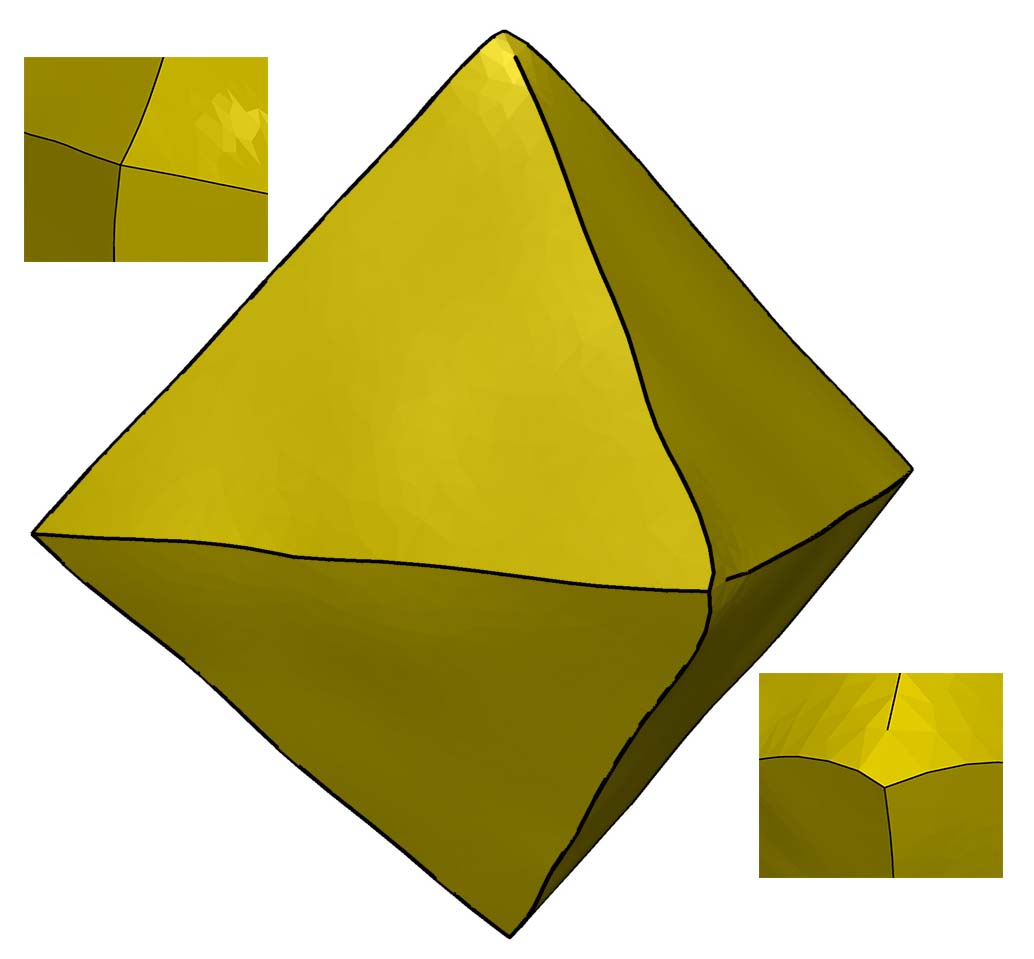} }}%
	\subfloat[\cite{L0Mesh}]{{\includegraphics[width=2.8cm]{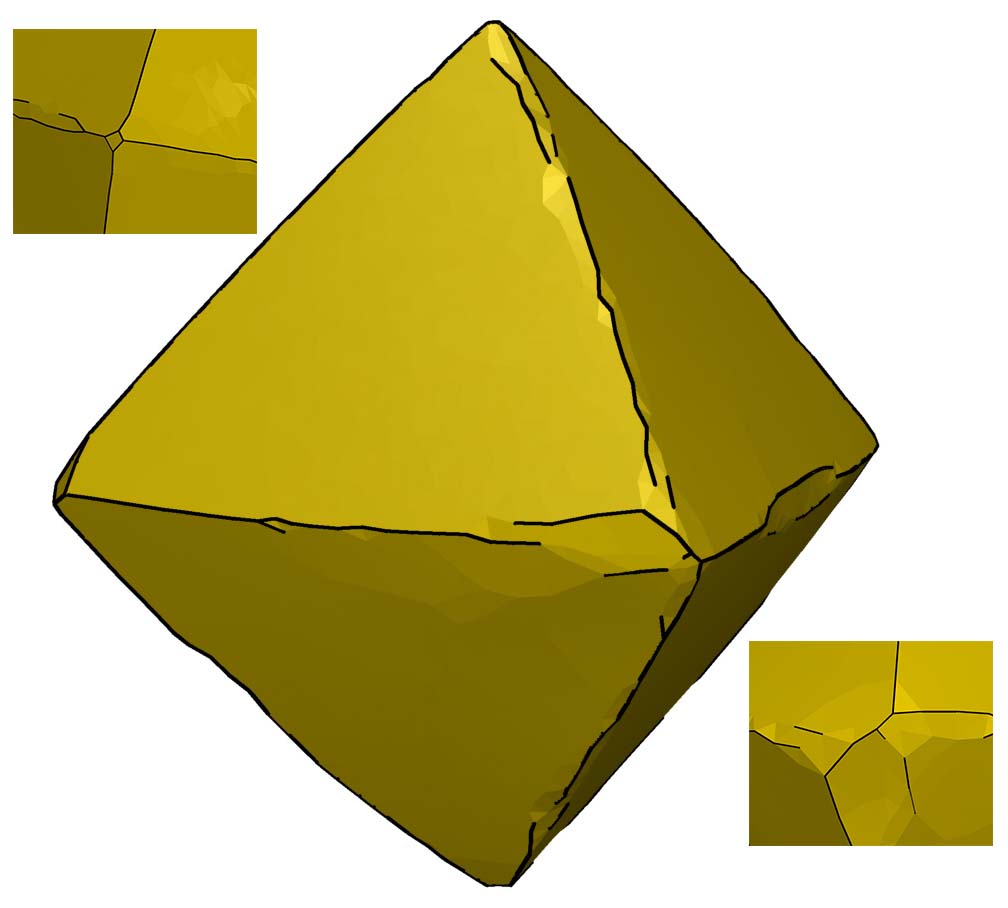} }}%
	\subfloat[Ours]{{\includegraphics[width=2.8cm]{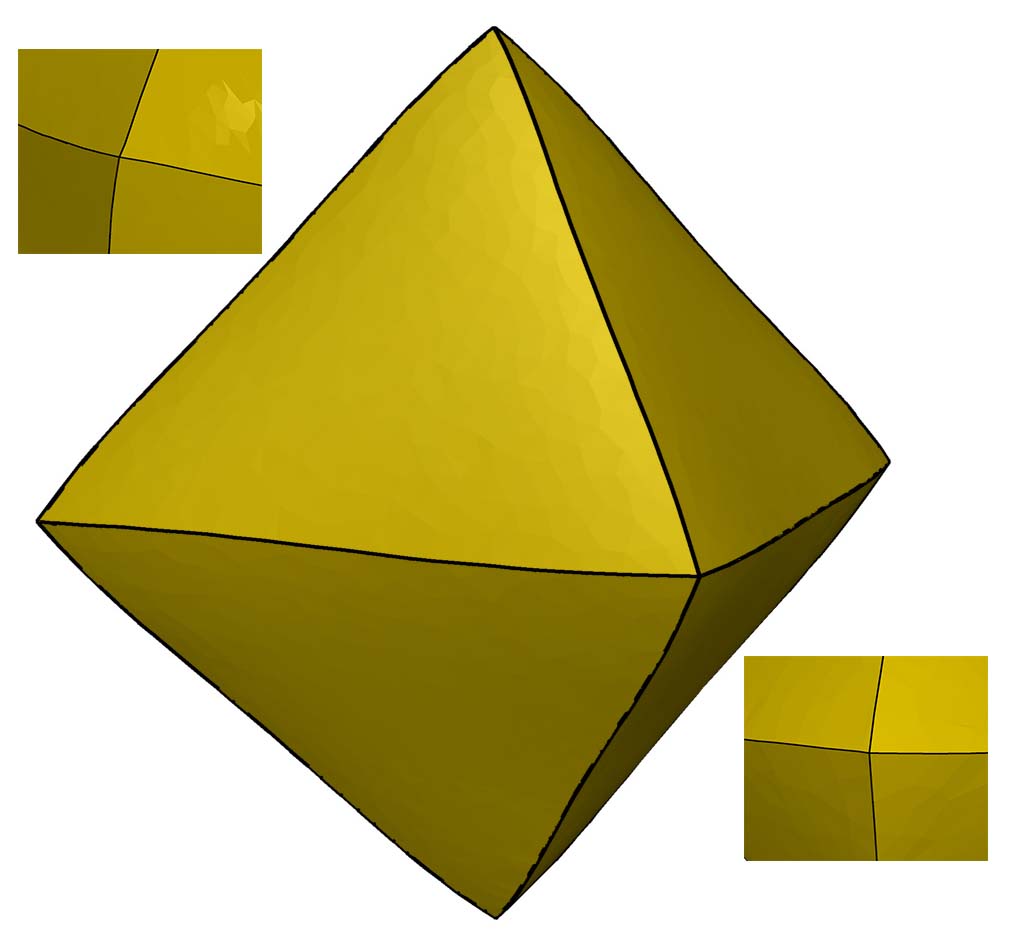} }}%
	\caption{Robustness against random edge flips after adding noise. (a) Noisy model corrupted with $\sigma_n=0.3l_e$ and  random edge flips, (b) to (f) show that the proposed method preserves better features (magnified corners) compared to state-of-the-art methods.}%
	\label{fig:edgeFlip}%
\end{figure*}

\begin{figure*}%
	\centering
	\subfloat[Noisy]{{\includegraphics[width=2.2cm]{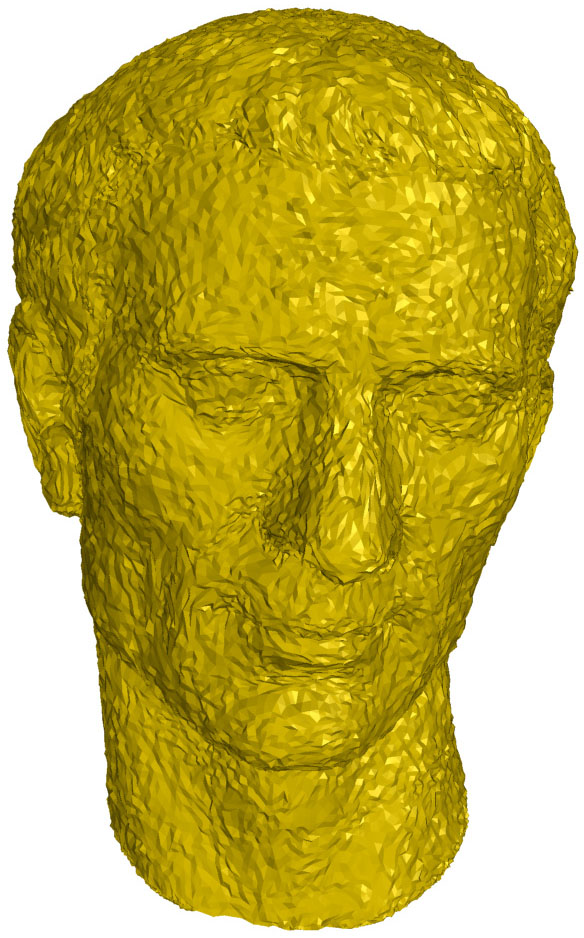} }}%
	\subfloat[\cite{BilFleish}]{{\includegraphics[width=2.2cm]{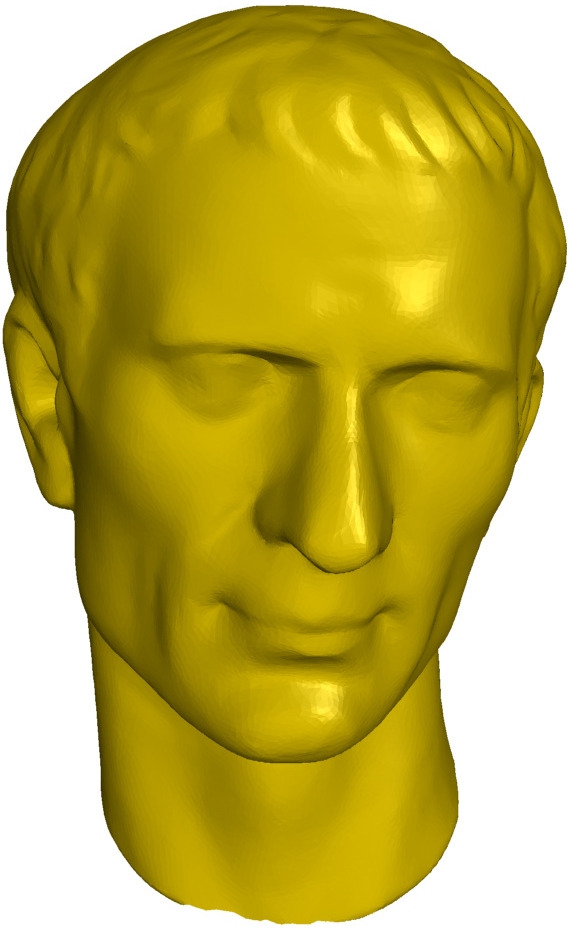} }}%
	\subfloat[\cite{aniso}]{{\includegraphics[width=2.2cm]{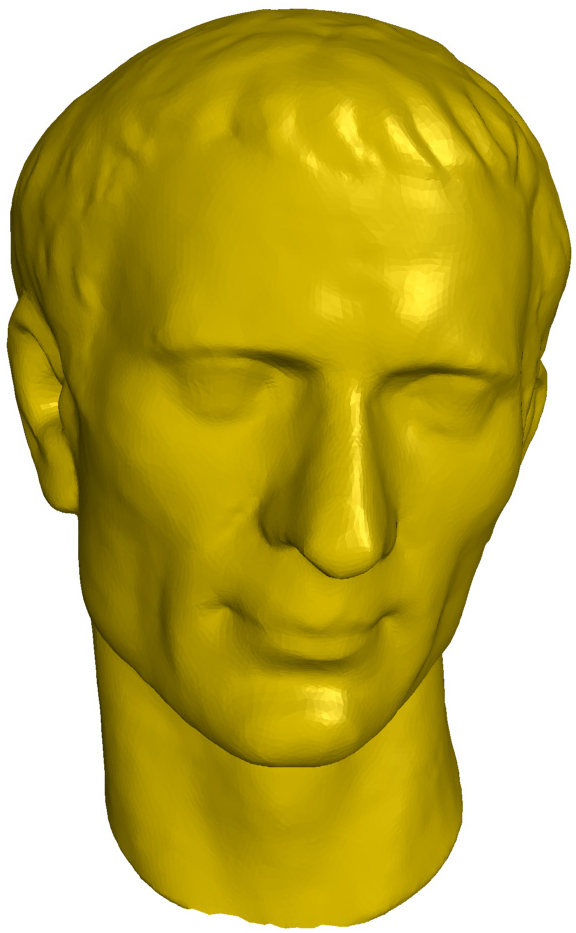} }}%
	\subfloat[\cite{BilNorm}]{{\includegraphics[width=2.2cm]{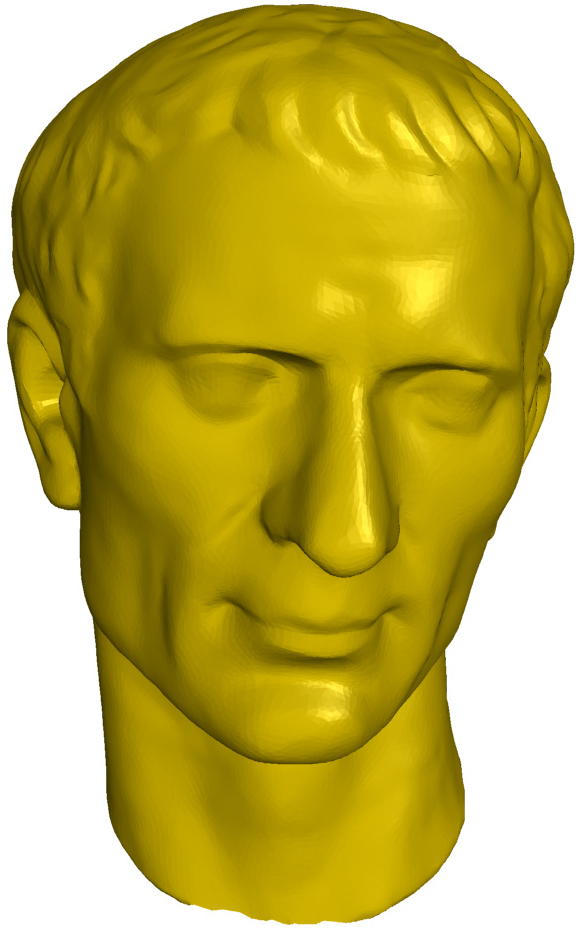} }}%
	\subfloat[\cite{L0Mesh}]{{\includegraphics[width=2.2cm]{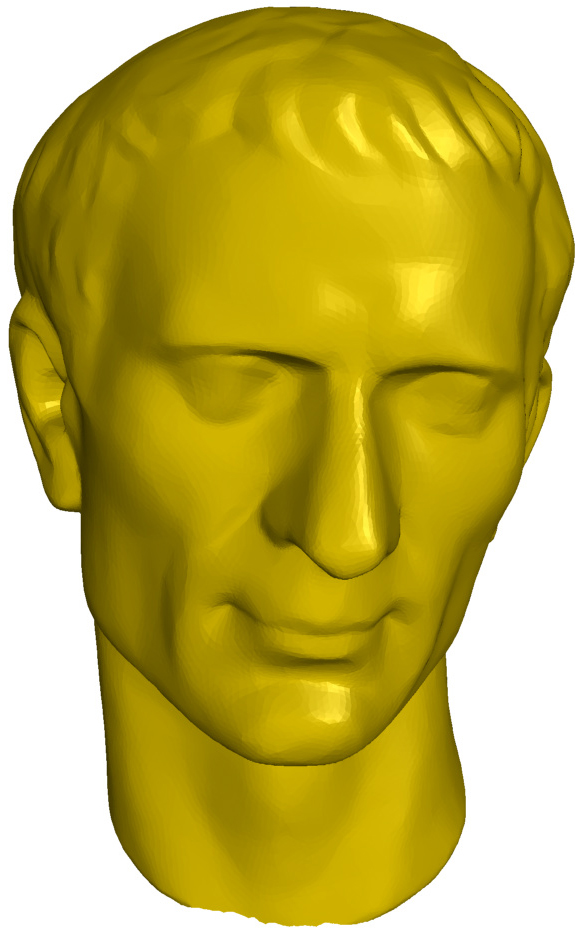} }}%
	\subfloat[\cite{Guidedmesh}]{{\includegraphics[width=2.2cm]{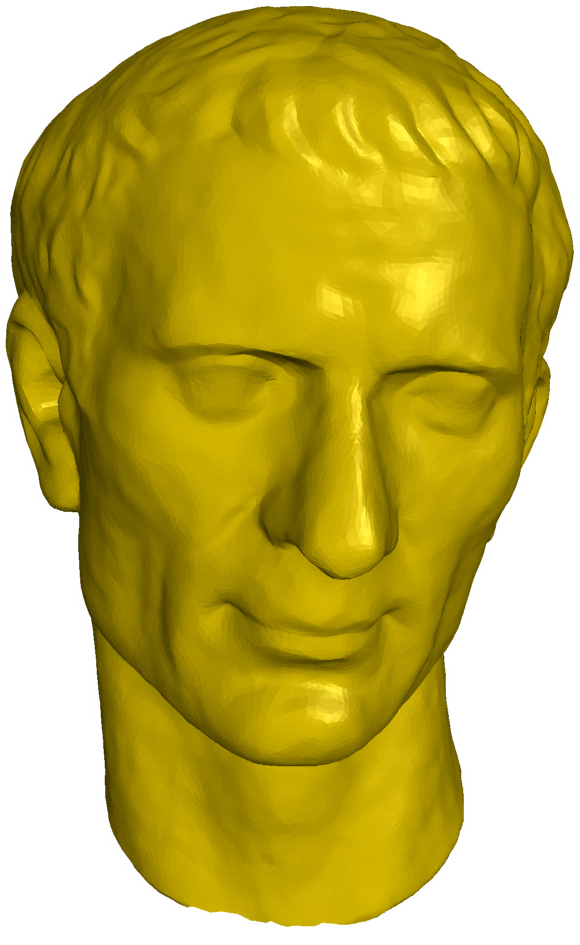} }}%
	\subfloat[\cite{robust16}]{{\includegraphics[width=2.2cm]{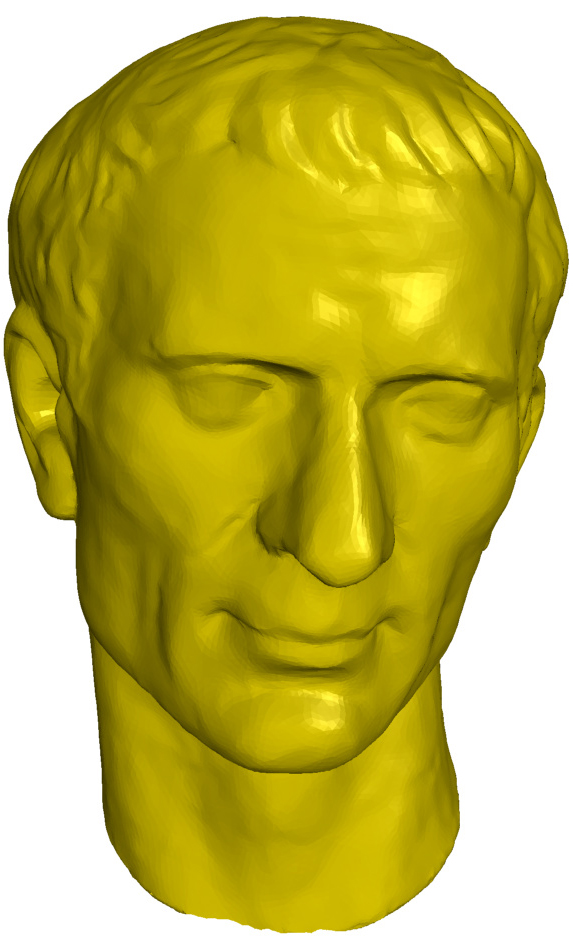} }}%
	\subfloat[Ours]{{\includegraphics[width=2.2cm]{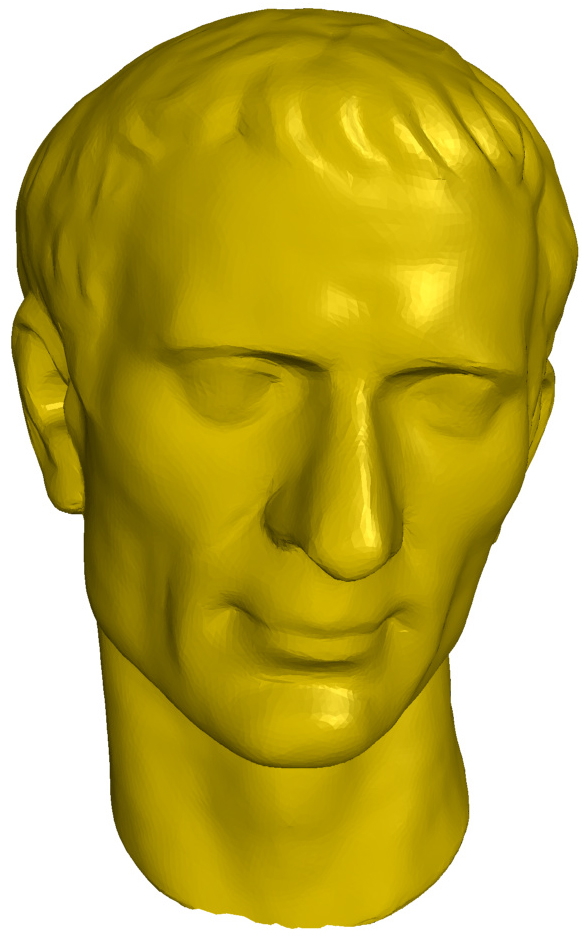} }}%
	\caption{The Julius model corrupted by Gaussian noise ($\sigma_n= 0.2l_e$) in random direction. The results are produced by state-of-the-art methods and the proposed method. Our method does not outperform state-of-the-art methods and the output is quite similar to method\cite{BilNorm}. The proposed method removes low frequency noise better than method \cite{Guidedmesh} where we can see small ripples of noise on the Julius model.  }%
	\label{fig:juli}%
\end{figure*}

\section{Experiments, Results and Discussion}
\label{exp}
\begin{figure}%
	\centering
	\subfloat[Cube (real data)]{{\includegraphics[width=3.2cm]{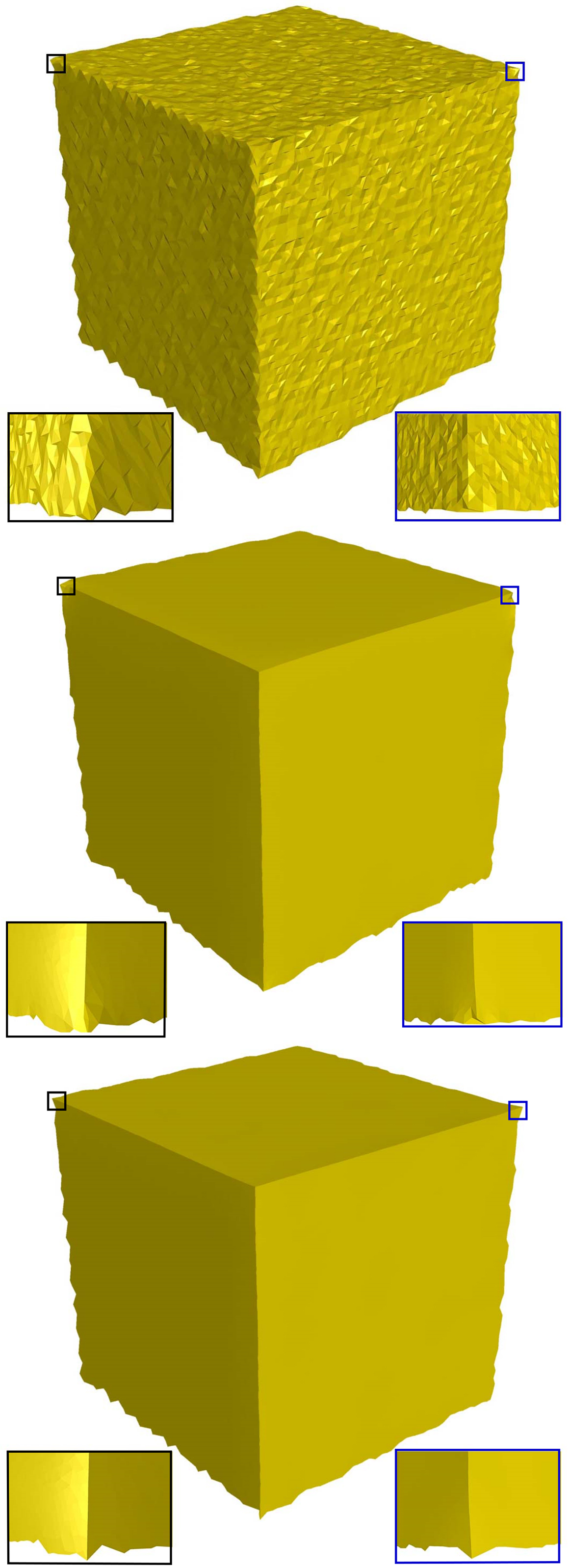} }}%
	\subfloat[Pierret model]{{\includegraphics[width=2.2cm]{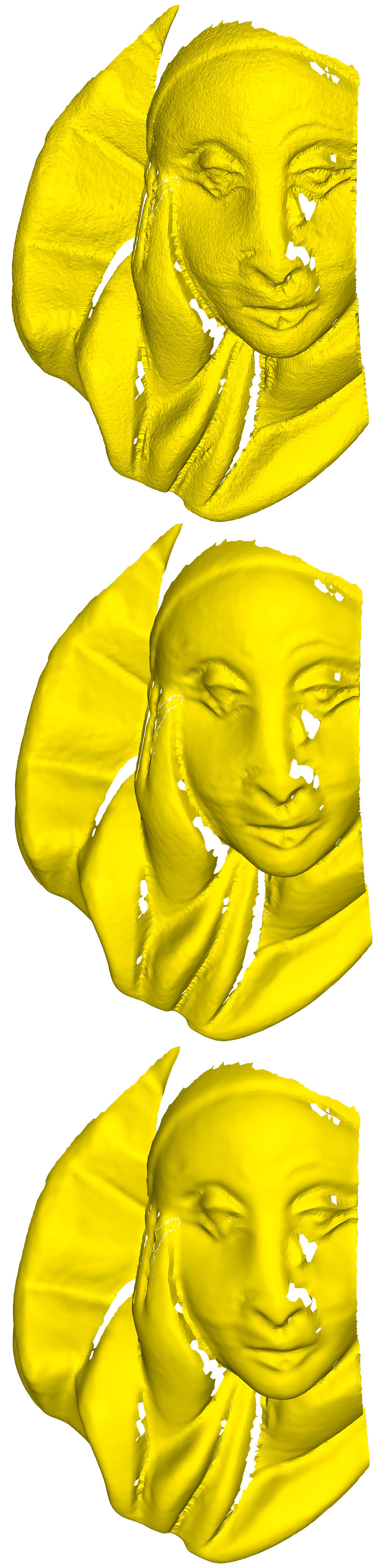} }}%
	\subfloat[Vase]{{\includegraphics[width=1.6cm]{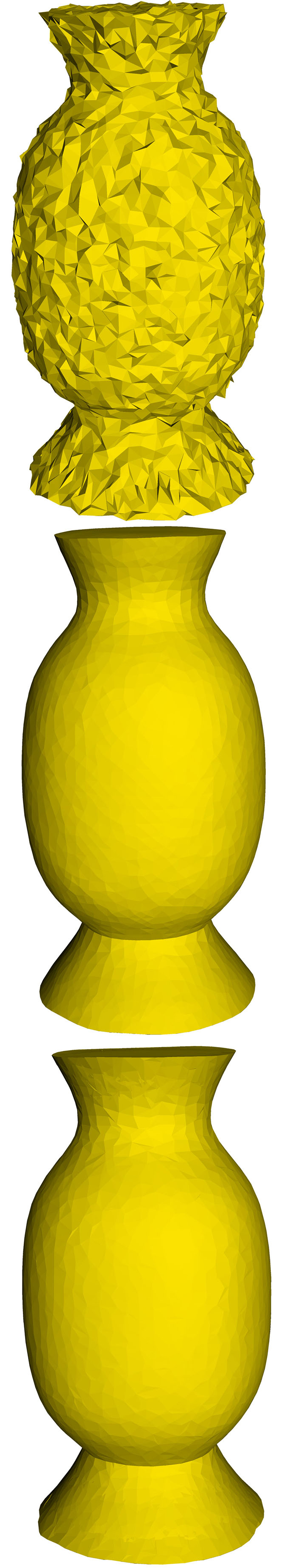} }}%
	
	\caption{Comparison with method \cite{robust16}: {Real data} (scanned Cube, Pierret model) and Vase, Julius models are corrupted by synthetic noise. Second row shows the results obtained by Lu et al.\cite{robust16} and third row presents the results obtained by the proposed method.}%
	\label{fig:compLu}%
\end{figure}

We evaluated the capacity of our algorithm on various kinds of CAD (Figure~\ref{fig:neighComp} -~\ref{fig:fanDisk}), CAGD (Figure ~\ref{fig:tauinc}, ~\ref{fig:juli}, \ref{fig:dragKd}) models corrupted with synthetic noise and real scanned data (Figure ~\ref{fig:realdata}, ~\ref{fig:realdiff}, ~\ref{fig:compLu}) models with different types of features. Noisy surfaces with non-uniform mesh corrupted with different kinds of noise (Gaussian, Impulsive, Uniform) in different random directions are also included in our experiments. We compared our method to several state-of-the-art denoising methods in which we implemented \cite{aniso}, \cite{BilNorm}, \cite{L0Mesh} and \cite{BilFleish} based on their published article and several results of \cite{AdamsKd}, \cite{Guidedmesh}, \cite{binormal} and \cite{robust16} are provided by their authors.

\textbf{Parameters:} We discussed several parameters (geometric neighbor radius $r$, dihedral angle threshold $\rho$, eigenvalue threshold $\tau$, damping factor $d$ and iteration $p$). Throughout, the whole experimentation, we fixed $\rho=0.8$, $d=3$. Effectively, there are only \textbf{3 parameters} to tune the results, in which $\tau$ is the most important as it depends on noise intensity but at the same time this parameter is not highly sensitive. We use {\boldmath$\tau \in \{0.3-0.4\}$} for synthetic data and {\boldmath$\tau \in \{0.05-0.1\}$} for real data because real data have smaller noise intensity compared to synthetic data in our experiments. The neighborhood radius $r$ depends on the number of elements within the geometric neighborhood region. We iterate several times ({\boldmath$p \in \{40-60\}$}) to obtain better result. In the quantitative comparison Table \ref{tab:quant}, the parameters are given in the following format:{\boldmath{$(\mathbb{\tau}, r, p)$}}. For the \cite{Guidedmesh} and \cite{binormal} methods, we mention \textit{Default} in the parameter column because smooth models are provided by those authors. We are following a similar pattern for other algorithms too. $(\sigma_c, \sigma_s, p)$ for \cite{BilFleish}, $(\sigma_s, p)$ for \cite{BilNorm}, $(\lambda, s, p)$ for \cite{aniso} and $(\alpha)$ for \cite{L0Mesh}, where $\sigma_s, \sigma_c$ are the standard deviation of the Gaussian function in the bilateral weighting. $s$ and $\lambda$ represent the step size and the smoothing threshold. The term $\alpha$ controls the amount of smoothing. 

\textbf{Effect of \boldmath$\tau$:} 
To see the effect of different values of $\tau$, we have experimented with two different models, the Box model (real data, with different level of features and less noise) and the Cube model (with limited features and high noise). With smaller values of $\tau \in [0.01,0.05]$, there is not much change on the cube model because of the higher noise whereas the box model manages to remove the noise and also the shallow features with increasing $\tau$. So $\tau$ is responsible for removing the noise and also for preserving features. If the feature size is smaller than the noise intensity, feature preservation is an ill-posed problem as shown in Figure~\ref{fig:tauinc}.

\textbf{Neighborhood Comparison:} Figure~\ref{fig:neighComp} shows that the geometrical neighborhood is more effective against irregular meshes compared to the topological neighborhood. The geodesic neighborhood is quite similar to the geometrical neighborhood but it is not appropriate when a model is corrupted by high intensity of noise.

\textbf{Visual Comparison:} The Block (Figure~\ref{fig:neighComp}), the Joint (Figure~\ref{fig:jointSharp}), the Cube (Figure~\ref{fig:cube}) and the Devil (Figure~\ref{fig:devRock}) have non-uniform meshes corrupted with Gaussian noise in random direction. Figure~\ref{fig:jointSharp} shows that the proposed method produces a smooth model with sharp features without creating any false features (piecewise flat areas like \cite{L0Mesh}) while method \cite{BilNorm} does not manage to remove low frequency noise (we can see smooth ripples).  Method \cite{Guidedmesh} produces good results but at the narrow cylindrical area it could not manage to retain the circular area. It also produces some false features at the non-uniform sharp corner. Method \cite{binormal} could not manage to retain the sharp features. Similarly, we can see this behavior for the Cube model in Figure~\ref{fig:cube}. The Rockerarm model (Figure~\ref{fig:rocker}) has a considerably non-uniform mesh and our method better retains sharp features (the screw part) compared to \cite{Guidedmesh},  \cite{aniso},  \cite{BilFleish} while removing the noise better compared to methods \cite{BilNorm}, \cite{L0Mesh}. Figure~\ref{fig:devRock} shows the robustness of our method against volume shrinkage. The horns of the model have the minimum shrinkage compared to state-of-the-art methods. The Fandisk model contains both cylindrical and sharp feature regions and is corrupted by high intensity Gaussian noise in random direction. Figure~\ref{fig:fanDisk} shows that the proposed method delivers both sharp features and umbilical regions without noise component and false features. Figure \ref{fig:dragKd} shows that our method effectively removes noise (around the teeth), keeps small smooth features (on the body) and creates almost null edge flips (around the claw of the dragon) compared to method \cite{AdamsKd}. In Appendix \ref{app:mcc}, the surfaces are colored by absolute value of the mean curvature to compare the proposed method with several state-of-the-art methods, in terms of suppressing noise and keeping sharp features. 

 For real data, we can not see considerable differences between the proposed method and state-of-the-art denoising methods because the noise intensity is quite low and state-of-the-art methods also produce good results. \mbox{Figure \ref{fig:realdata}} shows that our method better retains features in the right eye of the angel model, but for the Rabbit model, our result is quite similar to \cite{Guidedmesh} and better compared to other methods. Figure~\ref{fig:compLu} shows the comparison of our method to method \cite{robust16} with four different models (real and synthetic data). For the Pierret and the Cube models, our method produces smoother (non-feature region of the Pierret model) results while preserving all necessary features (corners at the cube) compared to method \cite{robust16}. For the Julius model, our method produces a smoother result, but at the cost of some fine features (around the eyes) compared to method \cite{robust16}. For the Vase model, method \cite{robust16} produces quite similar results to ours. However, the shrinkage effect is bigger in method \cite{robust16} as shown in Table \ref{tab:quant}. Figure~\ref{fig:realdiff} shows the robustness of the proposed algorithm against irregular meshes and holes in the real data. The Gorgoyle and the Eagle model have several holes and spikes, but our smoothing algorithm manages to produce a smooth surface with proper features. 

\textbf{Robustness against noise:} Our method is invariant against different kinds of noise as shown in Figure~\ref{fig:diffNoise} where the vase model is corrupted by impulsive and uniform noise. The proposed method does not produce appropriate results above a certain level of noise as shown in Figure~\ref{fig:noiseLevel}. Figure~\ref{fig:edgeFlip} shows that  our method is robust against random edge flips.

\textbf{Quantitative Comparison:}     
In this section, we give a quantitative comparison of our method with state-of-the-art methods. We are using two different parameters: $E_v$ ($L^2$ vertex-based error) and MSAE (the mean square angular error). The positional error from the original ground truth model is represented by $E_v$ and defined as \cite{vertexUpdate}:
\begin{equation*}
E_v = \sqrt{\frac{1}{3\sum_{k\in F}^{} A_k} \sum_{i\in V}^{} \sum_{j\in F_v(i)}^{} A_j {\mathrm{dist}(\tilde{v}_i, T)}^2},
\end{equation*}
where $F$ is the triangular element set and $V$ represents the set of vertices. $A_k$ and $A_j$ are the corresponding element areas and $F_v(i)$ is the number of elements in the $i^{th}$ vertex-ring. $\mathrm{dist}(\tilde{v}_i, T)$ is the closest $L^2$-distance between the newly computed vertex $\tilde{v}_i$ and the triangle $T$ of the reference model.
The MSAE computes the orientation error between the original model and the smooth model and is defined as:
\begin{equation*}
MSAE = E[\angle ({\mathbf{\tilde{n}, n}})],
\end{equation*}
where $\mathbf{\tilde{n}}$ is the newly computed face normal and $\mathbf{n}$ represents the face normal of the reference model. $E$ stands for the expectation value. The quantitative comparison Table \ref{tab:quant} shows that our method performs better for most of the models e.g. Cube, Devil, Joint etc. For some model like Fandisk, our method produces quite similar numeric errors as state-of-the-art methods. 
\begin{figure}[h]\centering
	\includegraphics[width=7cm]{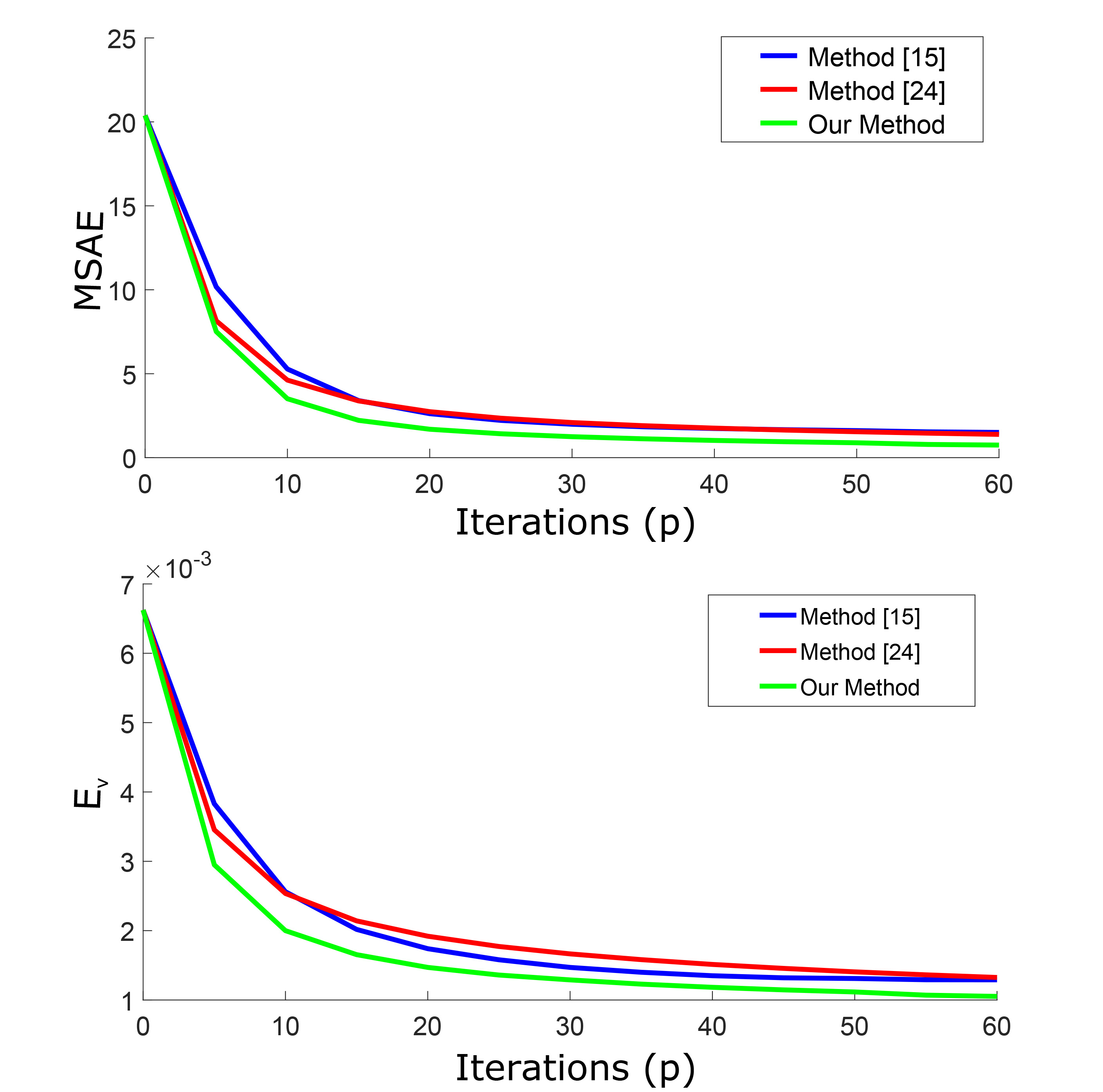}
	\caption{Convergence plot comparison between the methods \cite{aniso}, \cite{BilNorm} and our method. The error metrics MSAE and $E_v$ are computed for the Cube model and $p$ is the number of iterations. Figure shows that the proposed method has better convergence rate compared to the methods \cite{aniso} and \cite{BilNorm}. }
	\label{fig:errGraph}
\end{figure}
\textbf{Convergence:}
Our smoothing algorithm has a stable and fast convergence (as shown in Figure \ref{fig:errGraph}) because of the eigenvalue binary optimization. Modification of the eigenvalues of the ENVT will not affect the orientation of the corresponding face normal when noise is removed because the difference between two eigenvalues will be zero and also the less dominant eigenvalue will be zero. There will be no more modification on noisy surfaces after some iteration when we meet the explained scenario as shown in Figure \ref{fig:iteration}, where there is no significant change (visually) after 60 iterations. Figure~\ref{fig:errGraph} shows that the proposed method converges with minimum error compared to the methods \cite{aniso} and \cite{BilNorm}. We can see that after 40 iterations, our method is almost stable and does not produce significant changes. The eigenvalue binary optimization not only helps in preserving features, but also improves the convergence rate of the algorithm.

\textbf{Running Time Complexity:} Running time complexity of the proposed method is similar to most of the two-step denoising methods \cite{Ohtake}, \cite{Taubin01linearanisotropic}, \cite{alphaTrimming}, \cite{BilNorm}. The neighborhood computation is done by the growing disk method to compute the ENVT. The ENVT computation has the complexity of $O(c \cdot n_f \cdot p )$, where $c$ is the number of elements within the neighborhood, $n_f$ and $p$ are the numbers of elements and iterations respectively. The tensor multiplication procedure has the running complexity of $O(n_f)$. Similarly, the vertex update procedure has the complexity of $O(c \cdot n_v \cdot p )$, where $n_v$ is the number of vertices in the geometry. In general, $n_f>n_v$, so the overall complexity of the algorithm will be $O(c \cdot n_f \cdot p )$. The number of elements in the geometric neighborhood $c$ plays an important role in the running time of the algorithm as shown in Table \ref{tab:runn}. For example, the Devil model has smaller number of elements and larger running time compared to the Joint model because of the different geometric neighborhood radius. The Bilateral normal method \cite{BilNorm} uses a fix number of neighborhood elements (depending on the valence of vertex) for face normal smoothing and is a bit faster compared to the proposed method. However, the other recent two step denosing methods \cite{Guidedmesh}, \cite{binormal} are slower compared to our method because of their additional denoising steps. 

\begin{center}
	\captionof{table}{Running Time (in seconds)}
	\scalebox{0.85}{	
		
		\label{tab:runn}
		\begin{tabular}{ |c|c|c|c|c|c|c|c|l| }
			
			\hline
			Models  & Cube & Devil & Joint & Fandisk&Rockerarm&Vase \\ \hline
			Time (s) & 1.2 & 51.9 & 38.2 & 5.06 &27.5&1.1 \\ 
			\hline
		\end{tabular}
	}	
\end{center}

\section{Conclusion and Future work} 
In this paper, we presented a simple and effective tensor multiplication algorithm for feature-preserving mesh denoising. The concept of the element-based normal voting tensor (ENVT) has been introduced and eigenanalysis of this tensor leads to decoupling of features from noise. We have shown that the proposed method does not need any additional Laplacian-based smoothing technique to remove the noise, like multistage state-of-the-art methods \cite{anisobil}, \cite{NVTsmooth}, \cite{Guidedmesh}, \cite{binormal}. Our method removes noise by multiplying the ENVT to the corresponding face normal, and this reduces the complexity of the algorithm. We have introduced the concept of eigenvalue binary optimization that not only enhances sharp features but also improves the convergence rate of the proposed algorithm. The local binary neighborhood selection helps to select similar elements in the neighborhood to compute the element based normal voting tensor which avoids feature blurring during the denoising process. We provide a stochastic analysis of the noise effect on the geometry depending on the average edge length of the triangulated mesh. On the basis of this analysis, we can provide an upper bound on the noise standard deviation depending on the minimum edge length to reconstruct the smooth surface from the noisy surface. The experimental results (visual and quantitative) show the capability of the proposed algorithm. Our method produces good results not only in terms of visual but also quantitatively with all kind of data including CAD, CAGD and real data. We have also shown the robustness of the algorithm against different kinds and levels of noise. We also discussed the wrong orientation of triangles in presence of strong noise. In future work, we would like to solve problem of edge flips and extend our algorithm to point set smoothing.


%



\ifCLASSOPTIONcaptionsoff
  \newpage
\fi



%
\bibliographystyle{IEEEtran}
\bibliography{extrinsic}{}

\appendices

\section{Neighborhood definitions}
\label{app:Neigh}
\theoremstyle{definition}
\begin{definition}{Combinatorial neighborhood}
 is defined as the set of all elements connected with the vertices of the corresponding face:
 	\begin{equation*}
 	 \Omega_i = \{f_j|{v_i}_1\in f_j \lor {v_i}_2\in f_j \lor {v_i}_3\in f_j\},
 	 \end{equation*}
 	  where the neighborhood region is presented by $\Omega$ and the vertices ${v_i}_1$, ${v_i}_2$ and ${v_i}_3$ belong to the face $f_i$.
\end{definition}

 \begin{definition}{Geometrical neighborhood}
 	 is defined as the set of all elements belonging to the disk area of the desired radius and centered at the corresponding element:
 	 	\begin{equation*}
 	  \Omega_i = \{ {f_j}| \quad {\lvert c_j-c_i\rvert \leq r} \},
 	   \end{equation*}
 	   where $c_i$ and $c_j$ are the centroids of the central and neighbor element and $r$ is the radius of the neighbor disk for the geometrical neighborhood.
\end{definition}	
\begin{definition}{Geodesic neighborhood} is defined as the set of all elements within the shortest distance defined by the radius r:
	\begin{equation*}
	\Omega_i = \{f_j | \quad \mathcal{D}(f_i,f_j) \leq r \}, 
	\end{equation*} where $f_i$ is the source point and $\mathcal{D}(f_i,f_j):\mathcal{M} \to \mathbb{R}$ is a geodesic distance function on a manifold surface $\mathcal{M}$.
\end{definition}	

\begin{figure*}[h]%
	\centering
	\subfloat[Original]{{\includegraphics[width=2.1cm]{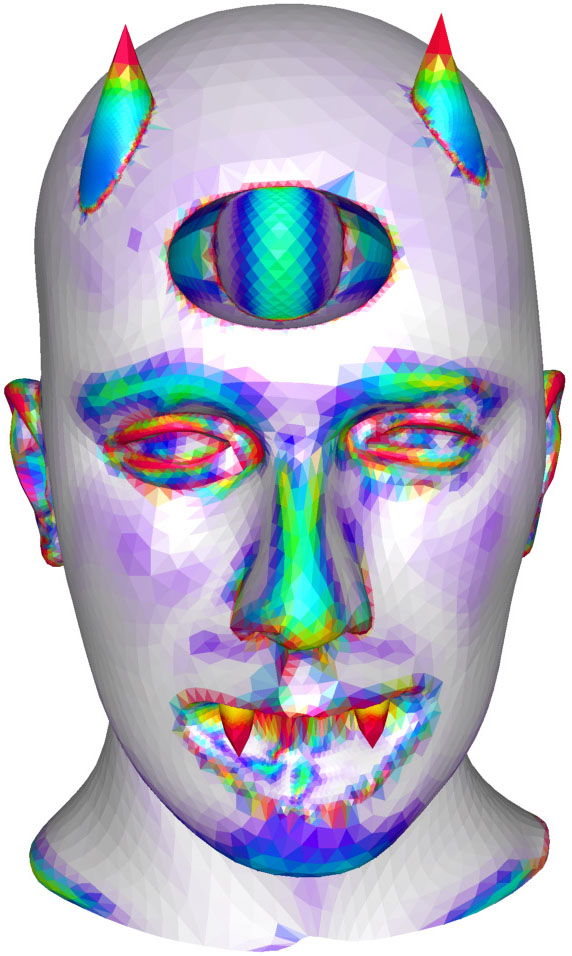} }}%
	\subfloat[Noisy]{{\includegraphics[width=2.1cm]{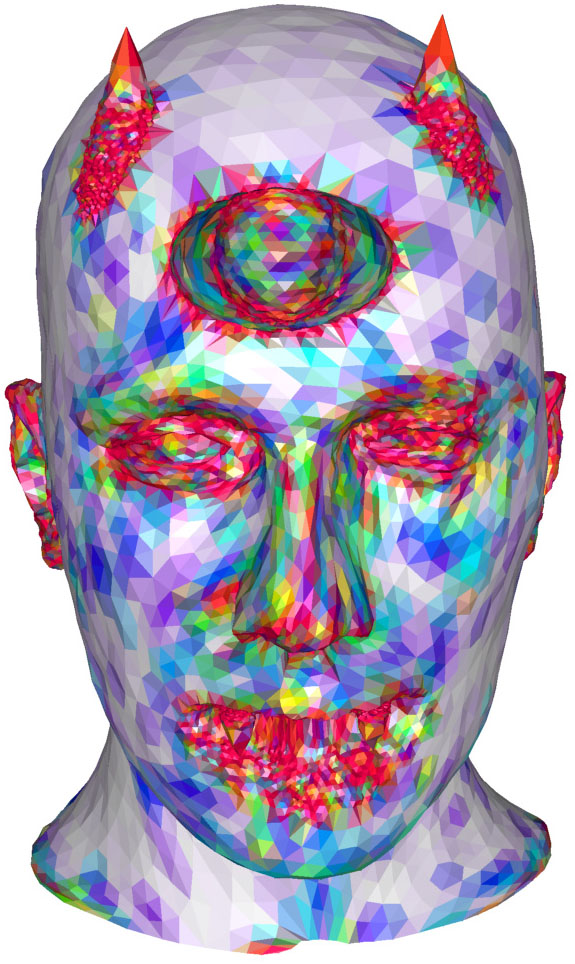} }}%
	\subfloat[\cite{aniso}]{{\includegraphics[width=2.1cm]{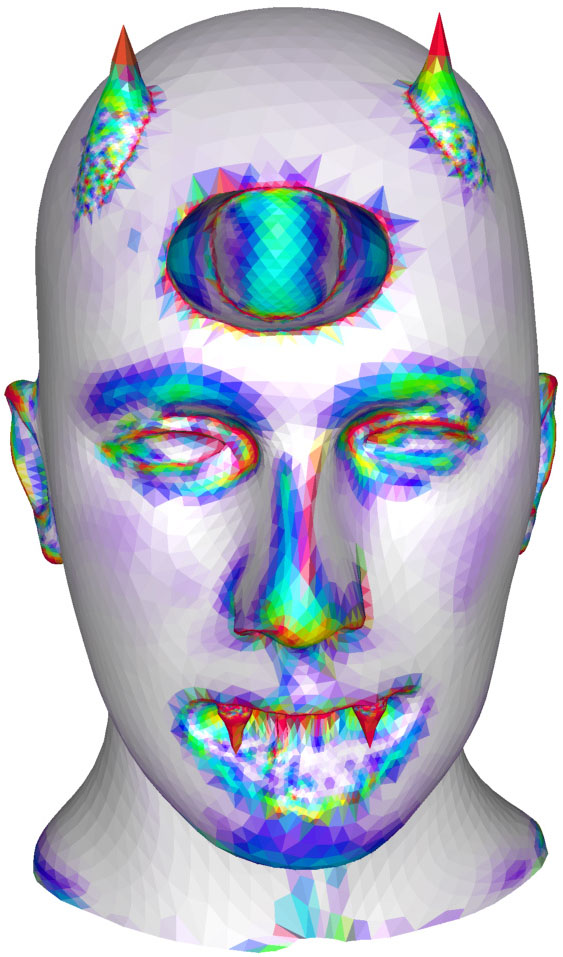} }}%
	\subfloat[\cite{BilNorm}]{{\includegraphics[width=2.1cm]{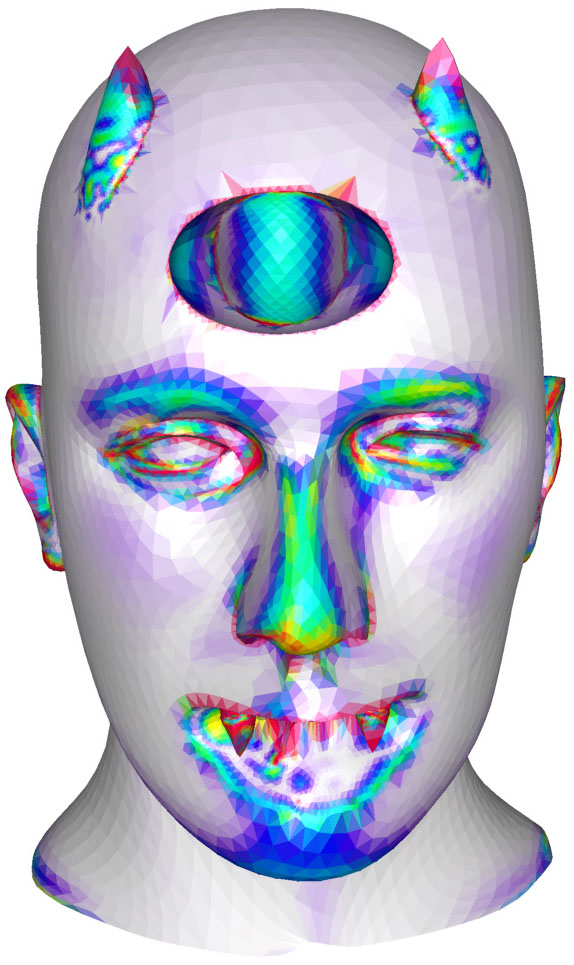} }}%
	\subfloat[\cite{L0Mesh}]{{\includegraphics[width=2.1cm]{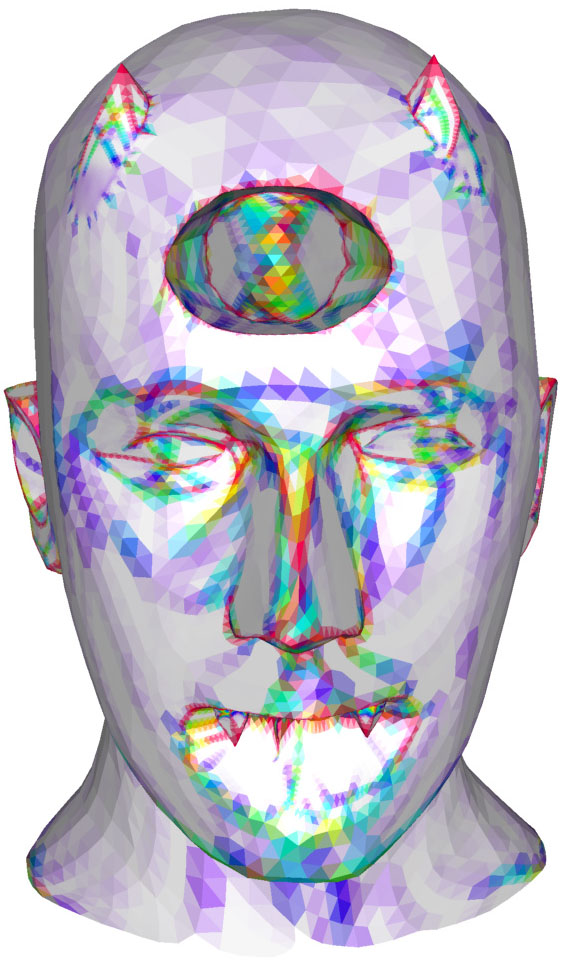} }}%
	\subfloat[\cite{Guidedmesh}]{{\includegraphics[width=2.1cm]{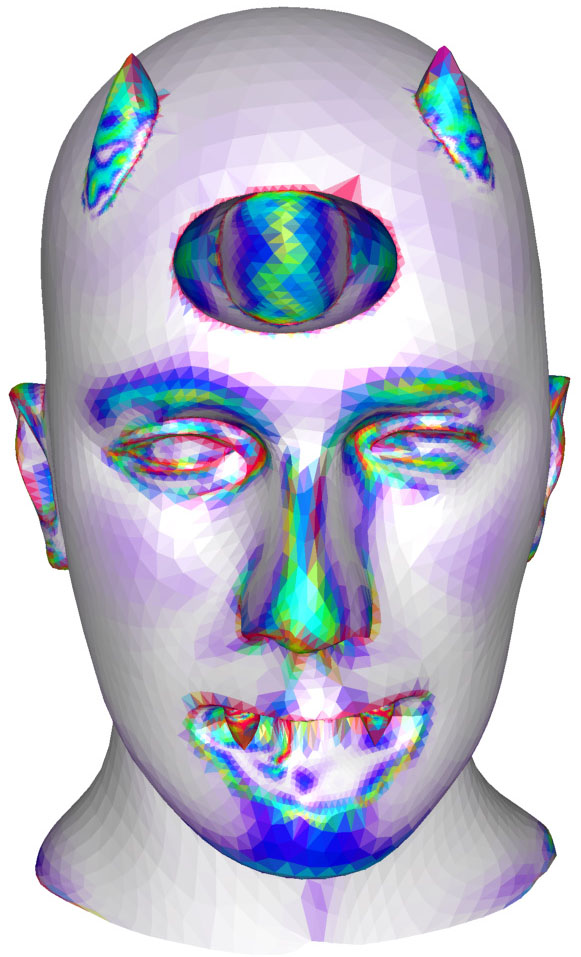} }}%
	\subfloat[\cite{binormal}]{{\includegraphics[width=2.1cm]{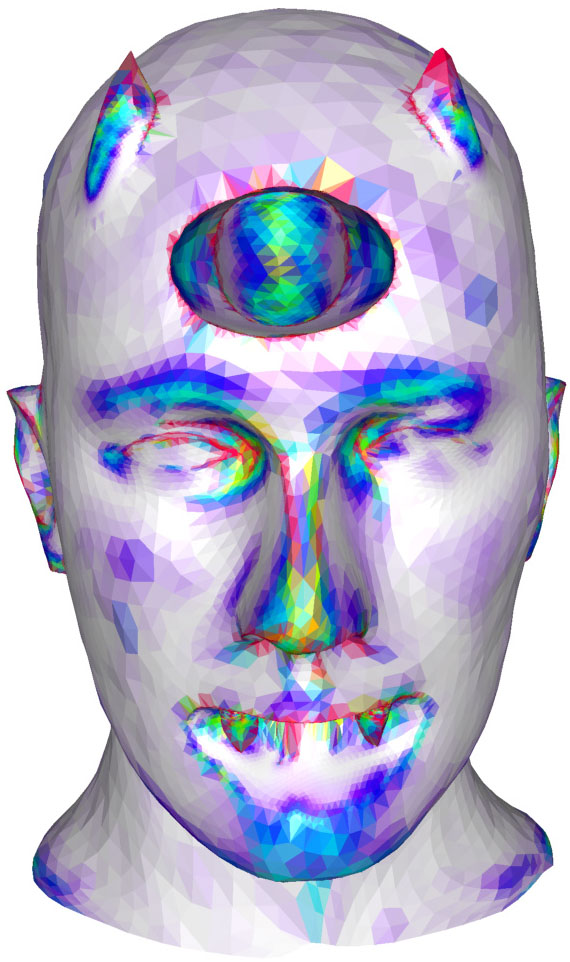} }}%
	\subfloat[Ours]{{\includegraphics[width=2.5cm]{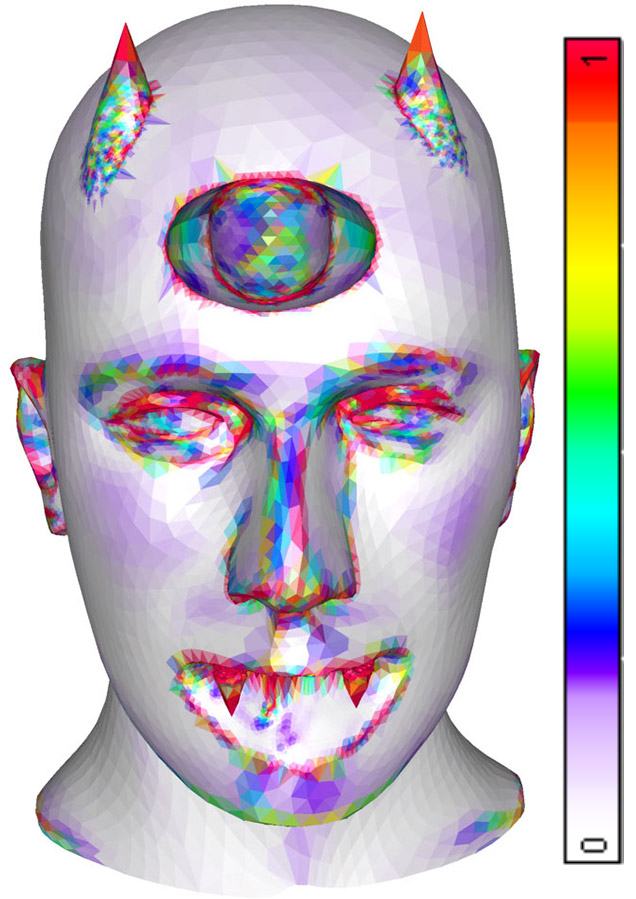} }}%
	\caption{ The Devil model consists of non-uniform triangles corrupted by Gaussian noise with standard deviation $\sigma_n= 0.15l_e$. The results produced by state-of-the-art methods and the proposed method are colored by the absolute value of the area weighted cotangent mean curvature.}%
	\label{fig:devRockmc}%
\end{figure*}

\begin{figure*}[h]%
	\centering
	\subfloat[Noisy]{{\includegraphics[width=2.1cm]{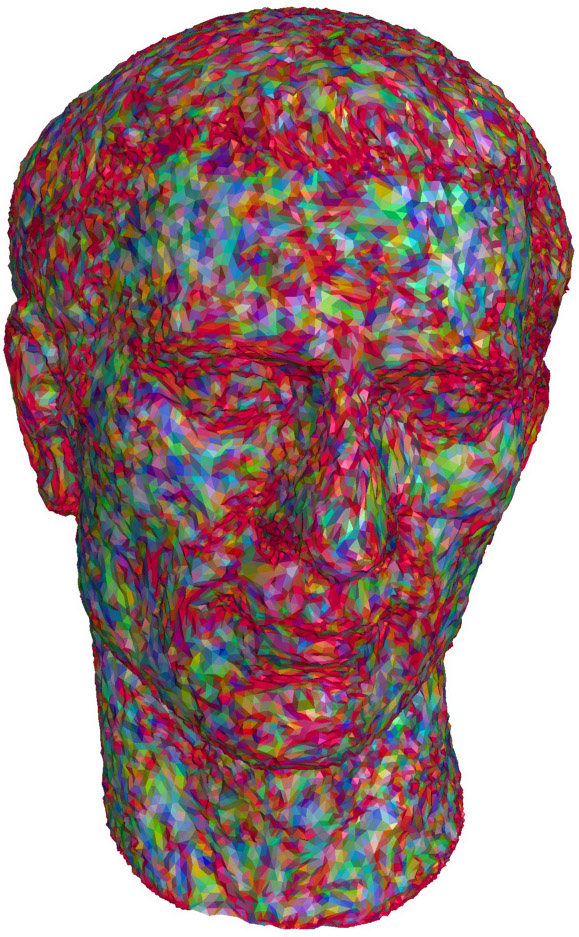} }}%
	\subfloat[\cite{BilFleish}]{{\includegraphics[width=2.1cm]{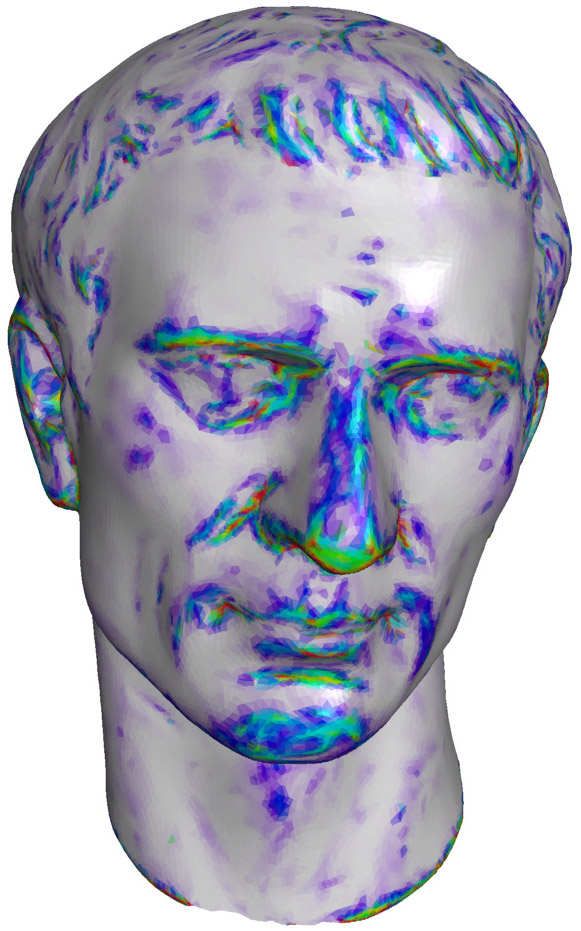} }}%
	\subfloat[\cite{aniso}]{{\includegraphics[width=2.1cm]{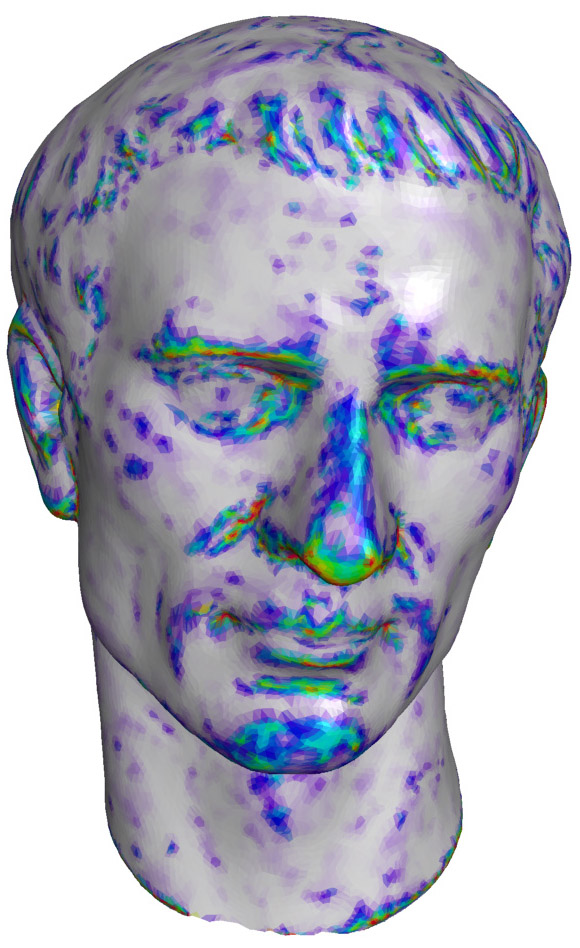} }}%
	\subfloat[\cite{BilNorm}]{{\includegraphics[width=2.1cm]{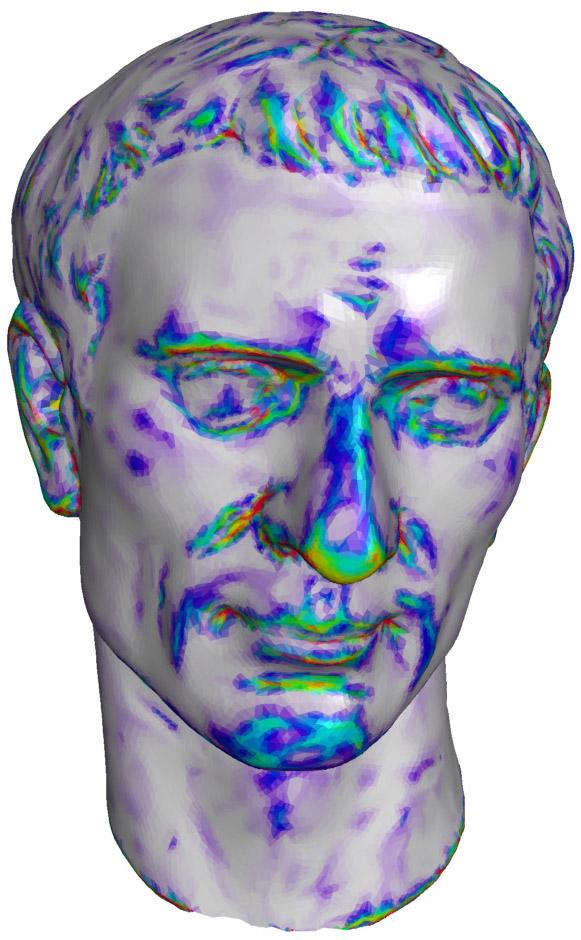} }}%
	\subfloat[\cite{L0Mesh}]{{\includegraphics[width=2.1cm]{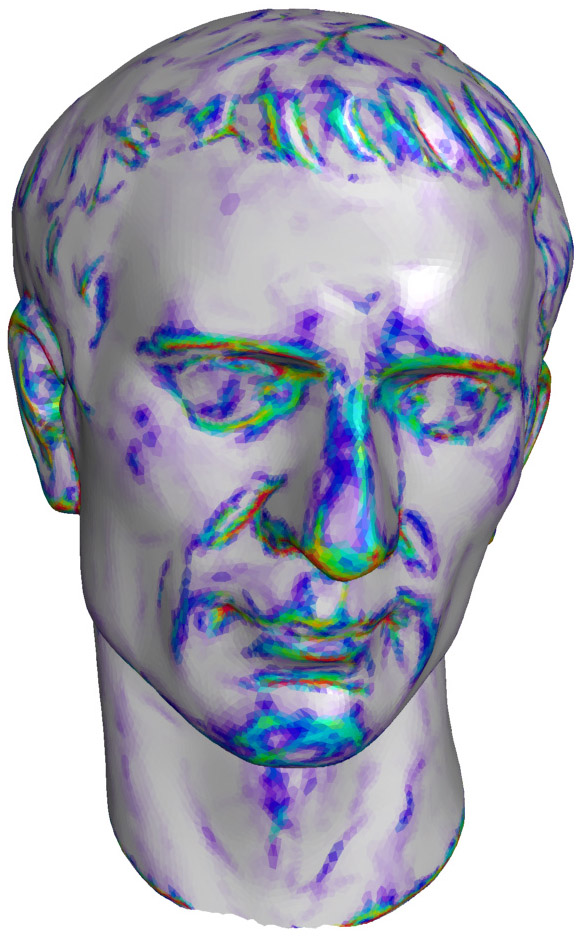} }}%
	\subfloat[\cite{Guidedmesh}]{{\includegraphics[width=2.1cm]{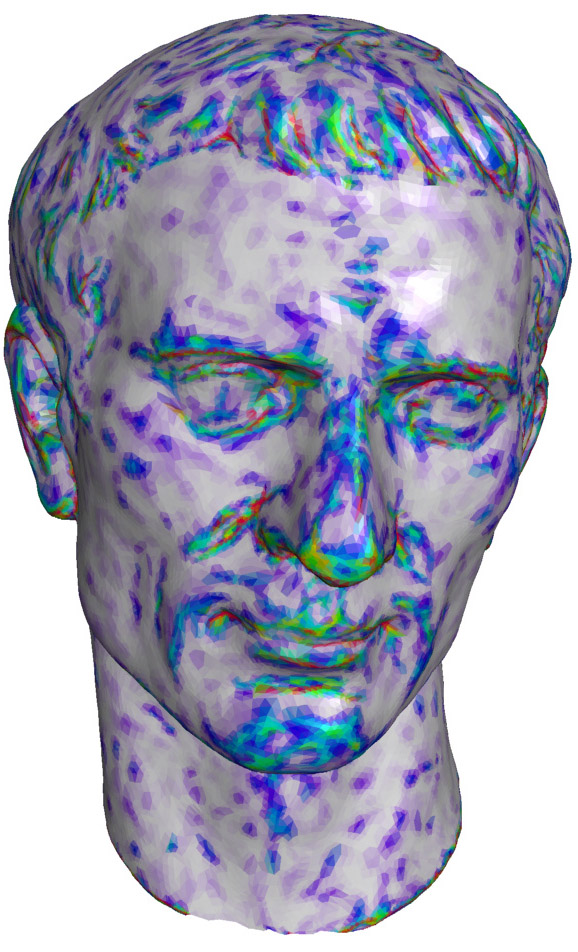} }}%
	\subfloat[\cite{robust16}]{{\includegraphics[width=2.12cm]{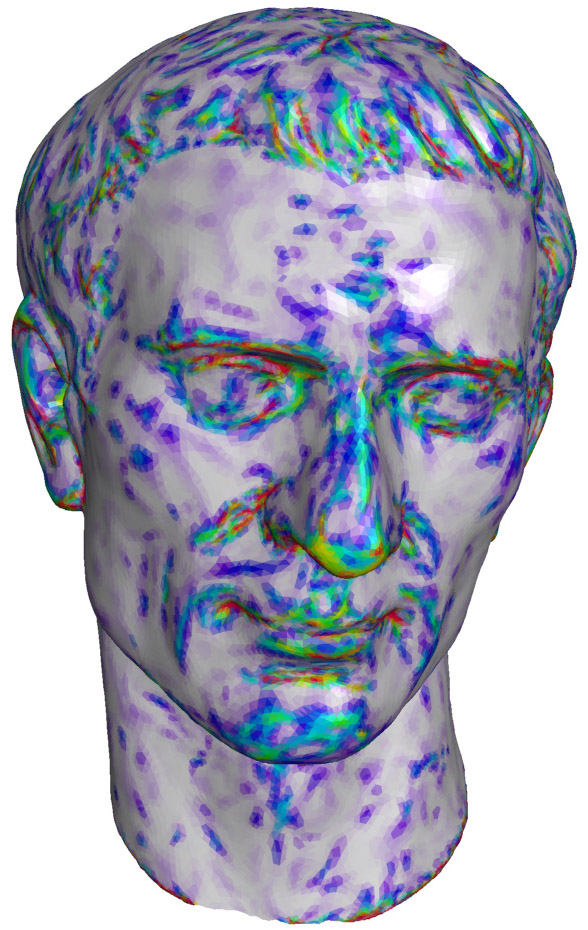} }}%
	\subfloat[Ours]{{\includegraphics[width=2.4cm]{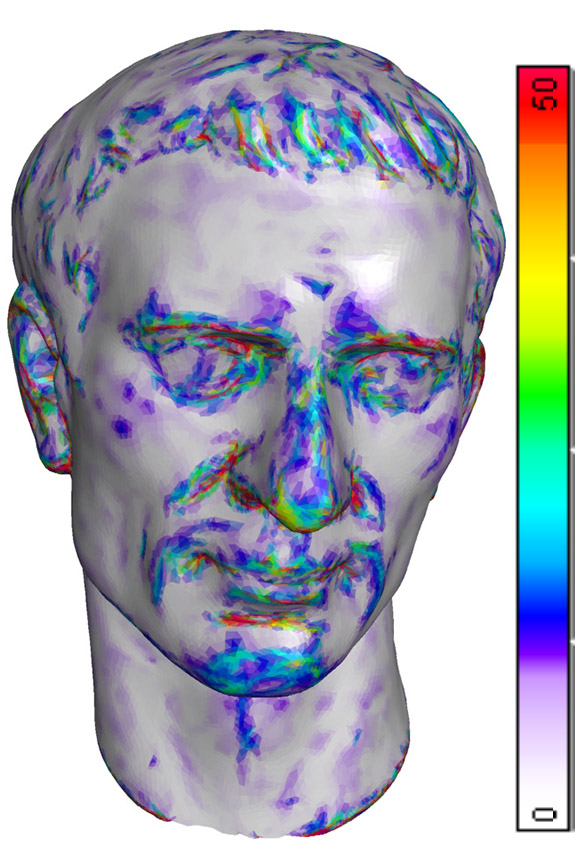} }}%
	\caption{The Julius model corrupted by Gaussian noise ($\sigma_n= 0.2l_e$) in random direction. The results which are produced by state-of-the-art methods and the proposed method show that our method is quite similar to method \cite{BilNorm} and better compared to methods \cite{Guidedmesh} and \cite{robust16}.}%
	\label{fig:julimc}%
\end{figure*}

\begin{figure*}[h]%
	\centering
	\subfloat[Original]{{\includegraphics[width=2.1cm]{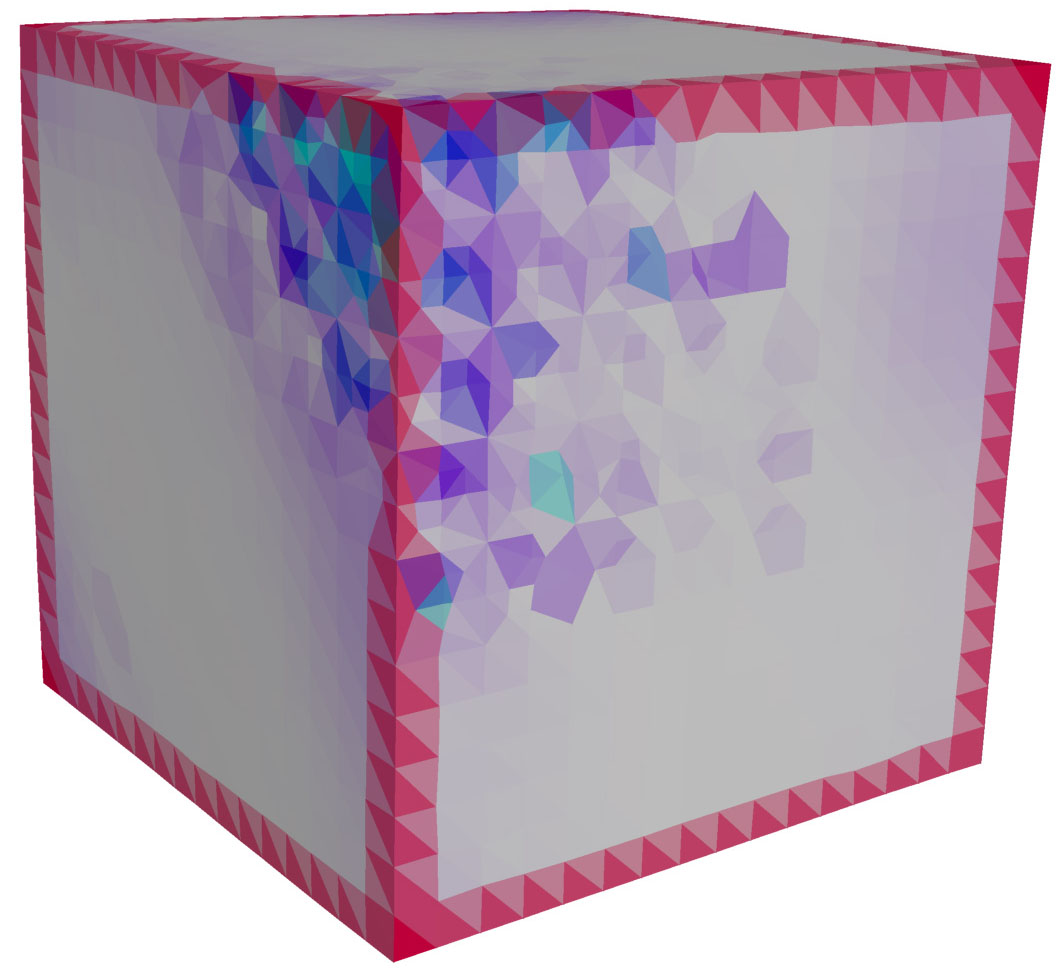} }}%
	\subfloat[Noisy]{{\includegraphics[width=2.1cm]{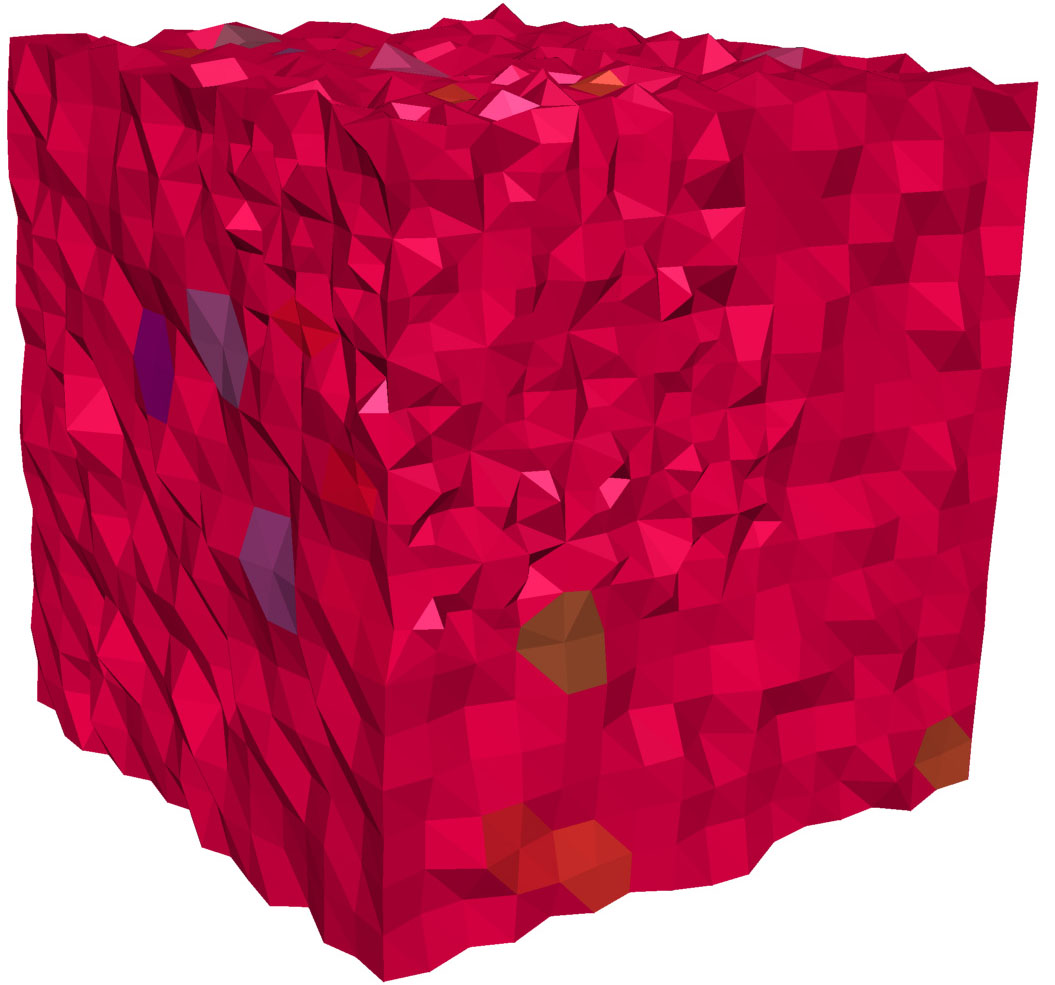} }}%
	\subfloat[\cite{aniso}]{{\includegraphics[width=2.1cm]{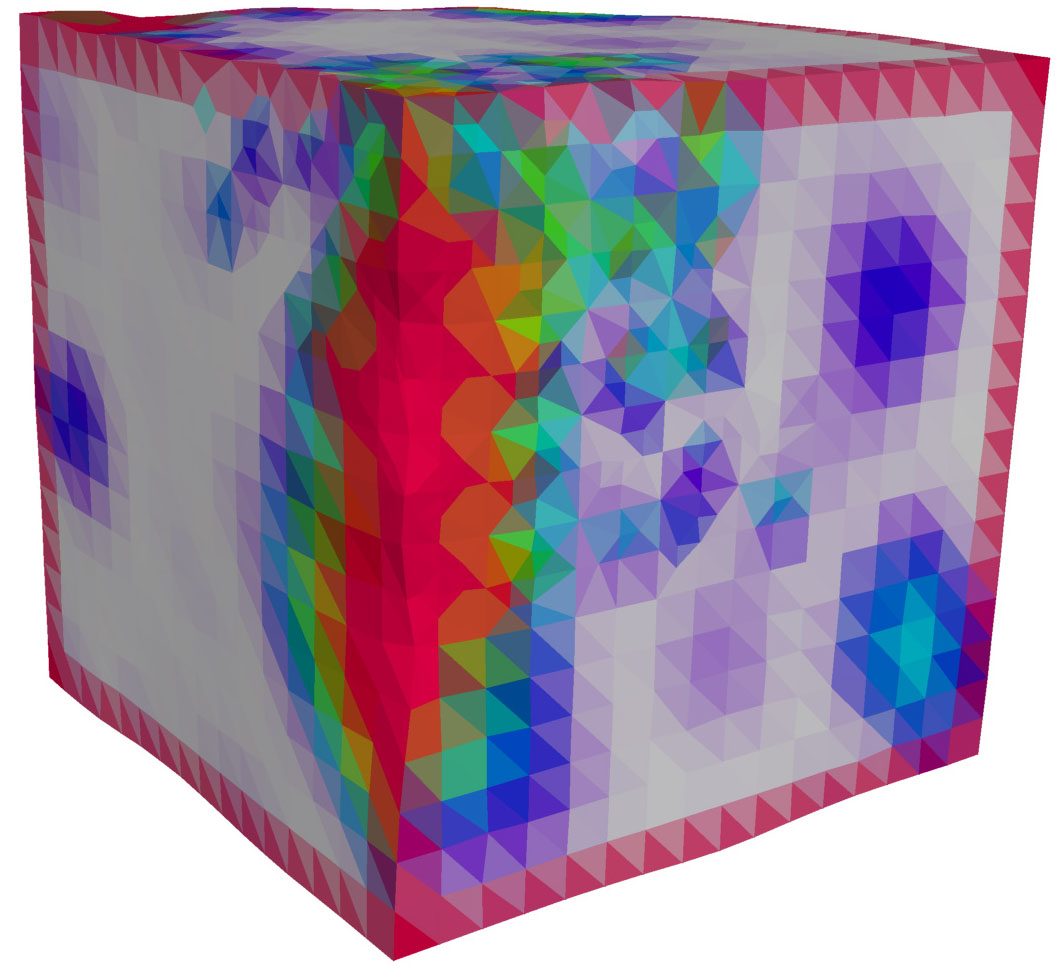} }}%
	\subfloat[\cite{BilNorm}]{{\includegraphics[width=2.1cm]{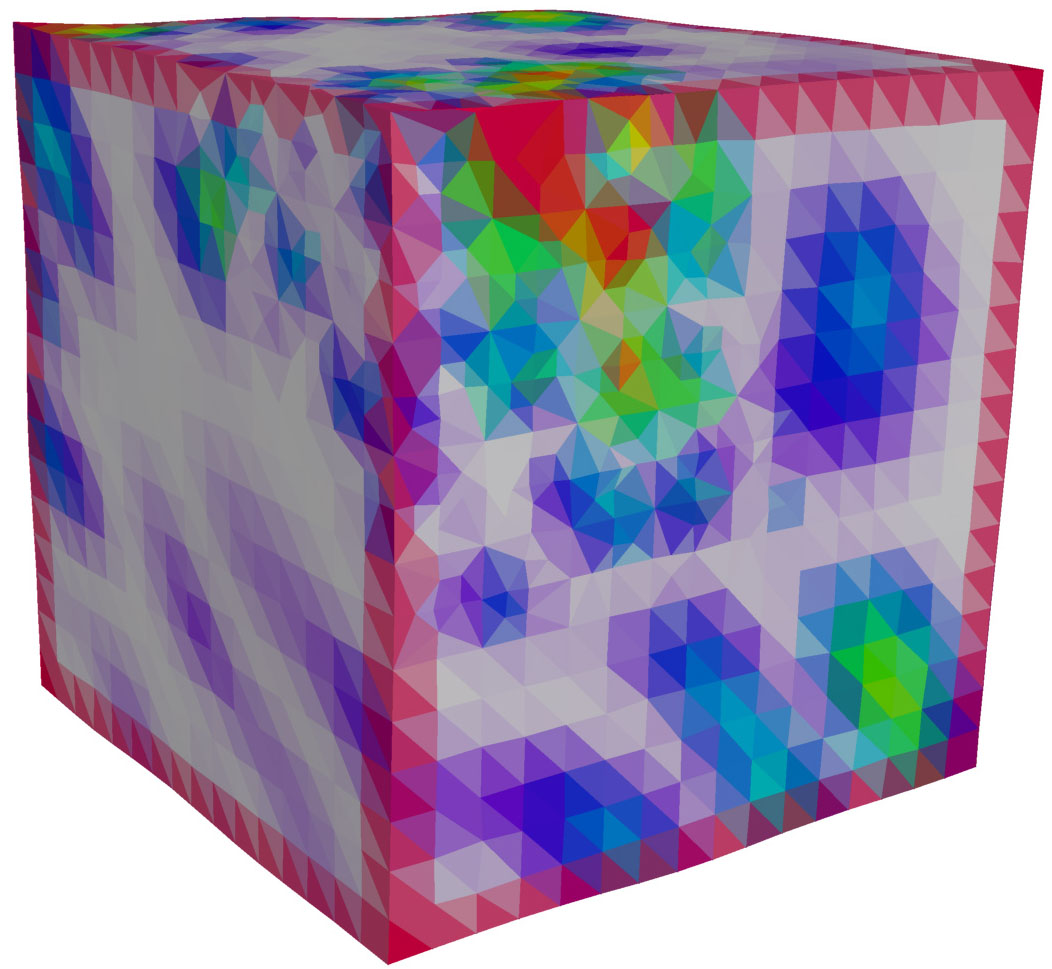} }}%
	\subfloat[\cite{L0Mesh}]{{\includegraphics[width=2.1cm]{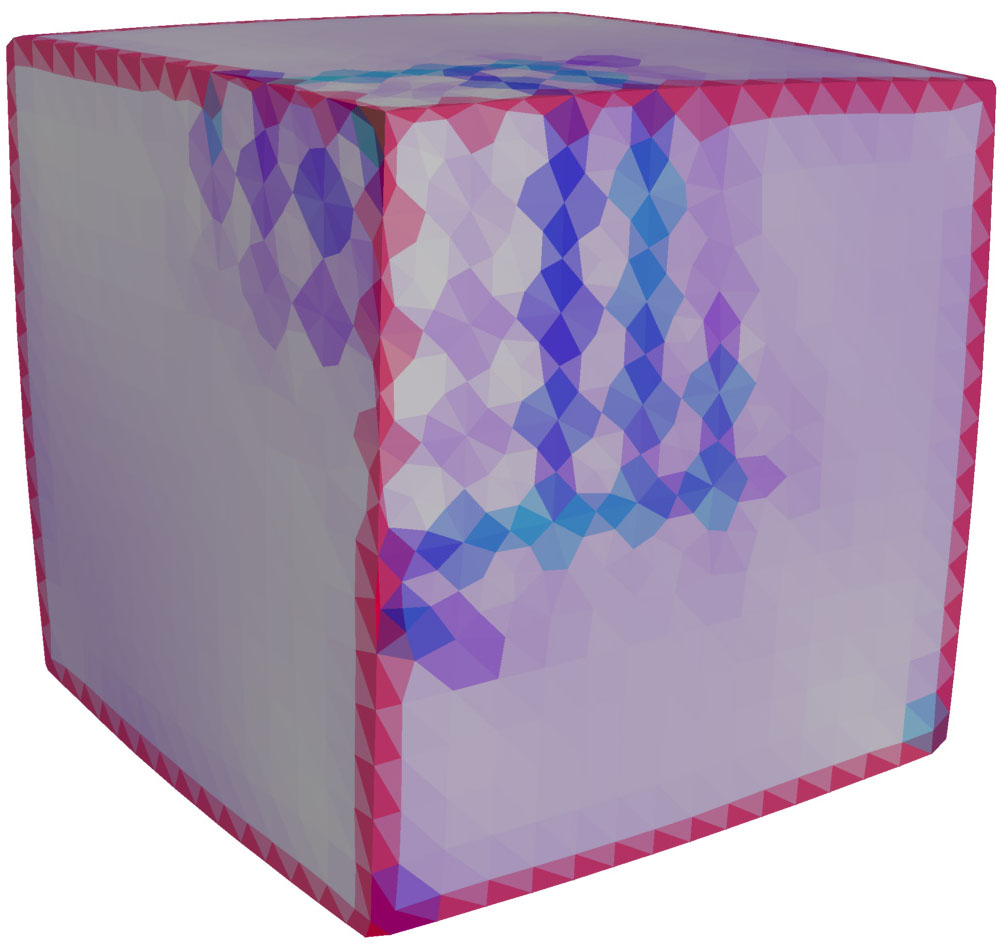} }}%
	\subfloat[\cite{Guidedmesh}]{{\includegraphics[width=2.1cm]{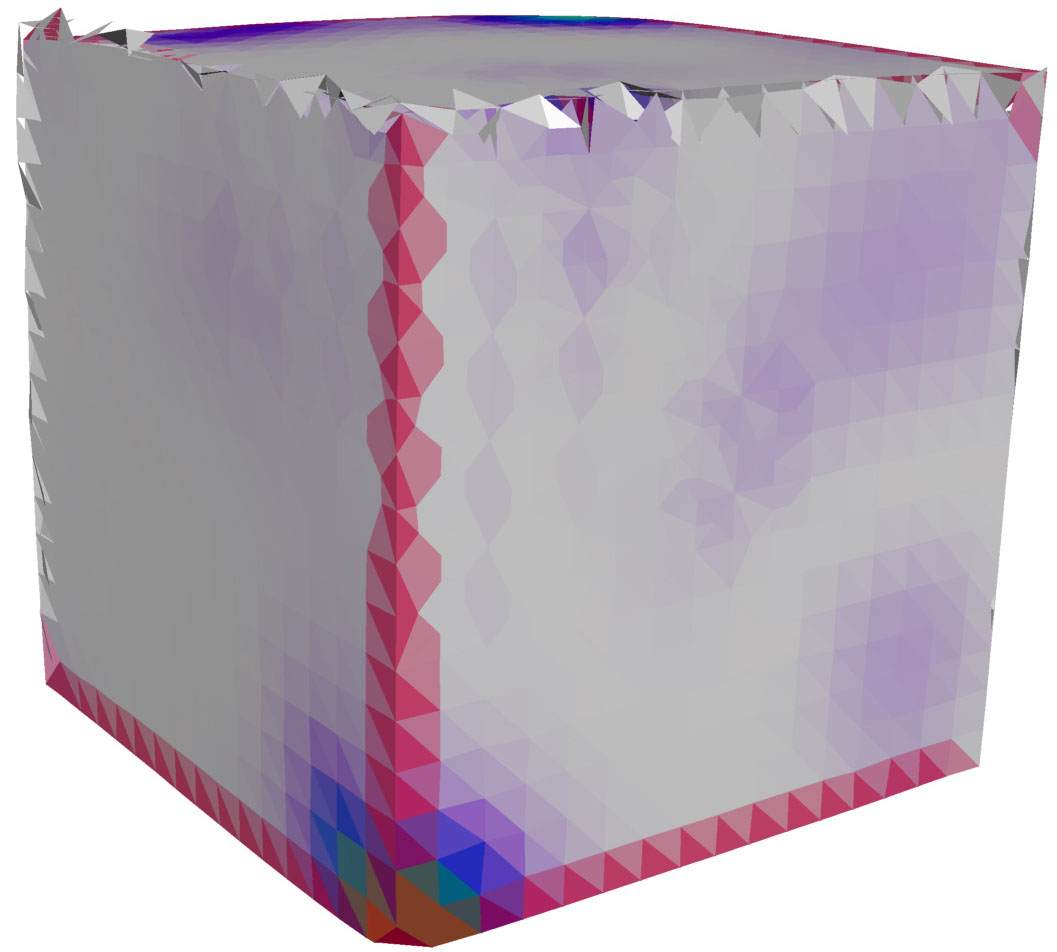} }}%
	\subfloat[\cite{binormal}]{{\includegraphics[width=2.1cm]{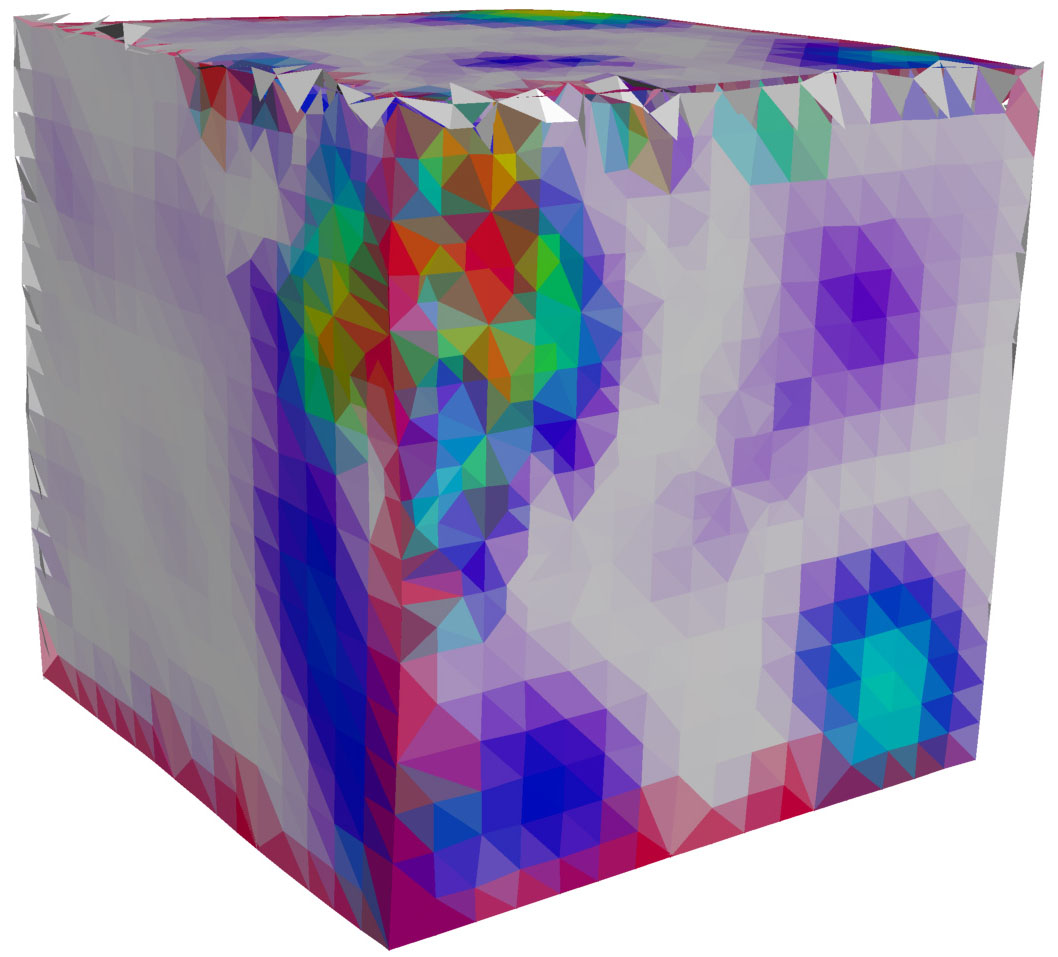} }}%
	\subfloat[Ours]{{\includegraphics[width=2.2cm]{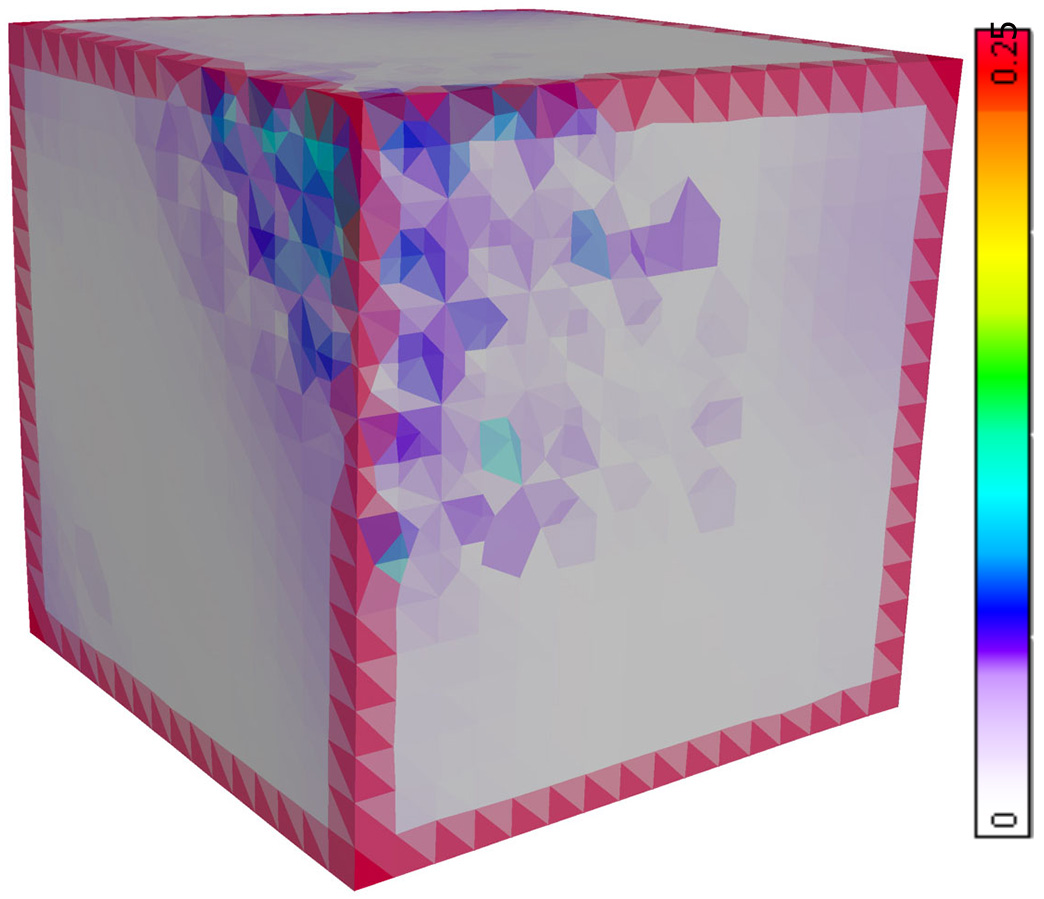} }}%
	\caption{The Cube model consists of non-uniform triangles corrupted by Gaussian noise ($\sigma_n= 0.3l_e$) in normal direction. The mean curvature coloring shows that the proposed method outperforms state-of-the-art methods.}%
	\label{fig:cubemc}%
\end{figure*}
\begin{figure}[h]\centering
	\includegraphics[width=\linewidth]{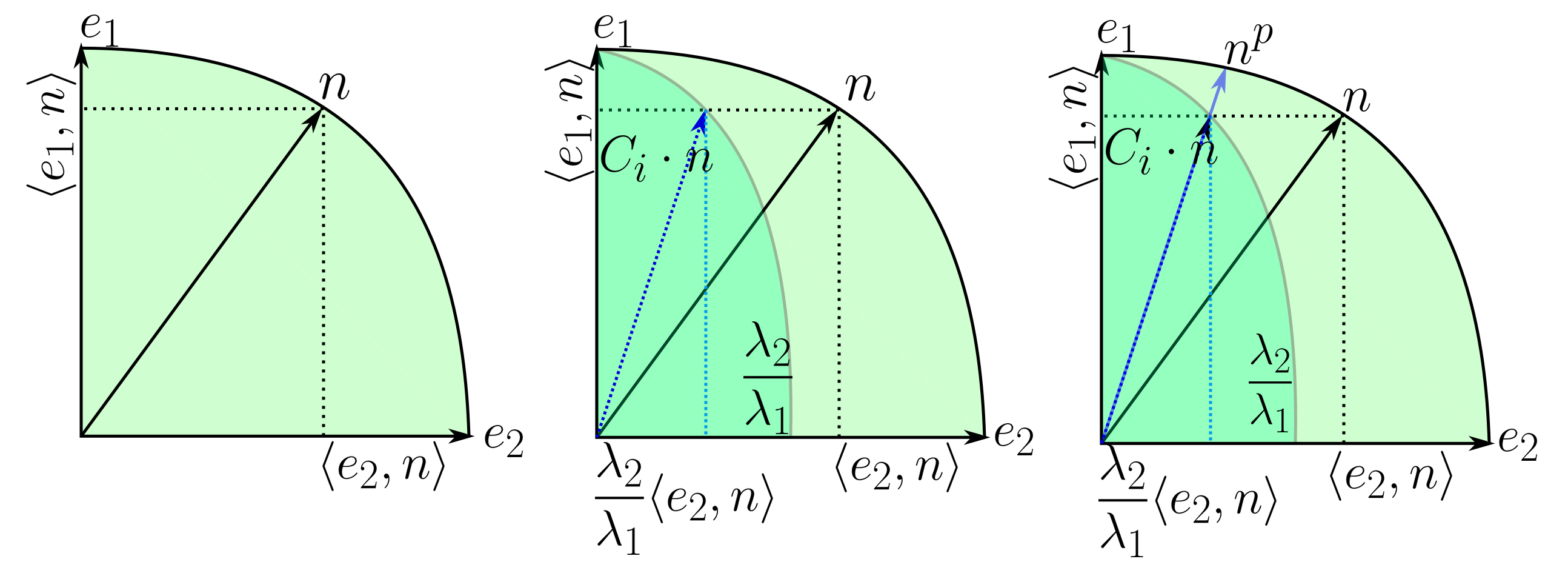}
	\caption{Shows the basic idea behind the proposed method to remove noise in $\mathbb{R}^2$ where {\boldmath$e_i$} and $\lambda_i$ represent the eigenvectors and eigenvalues of the proposed element based voting tensor and {\boldmath$n$} shows the noisy normal. We rotate the noisy normal {\boldmath$n$} towards the dominant eigendirection {\boldmath$e_1$} by a corresponding tensor multiplication. }
	\label{fig:idea}
\end{figure}
\begin{figure*}[h]%
	\centering
	\subfloat[Noisy]{{\includegraphics[width=2.4cm]{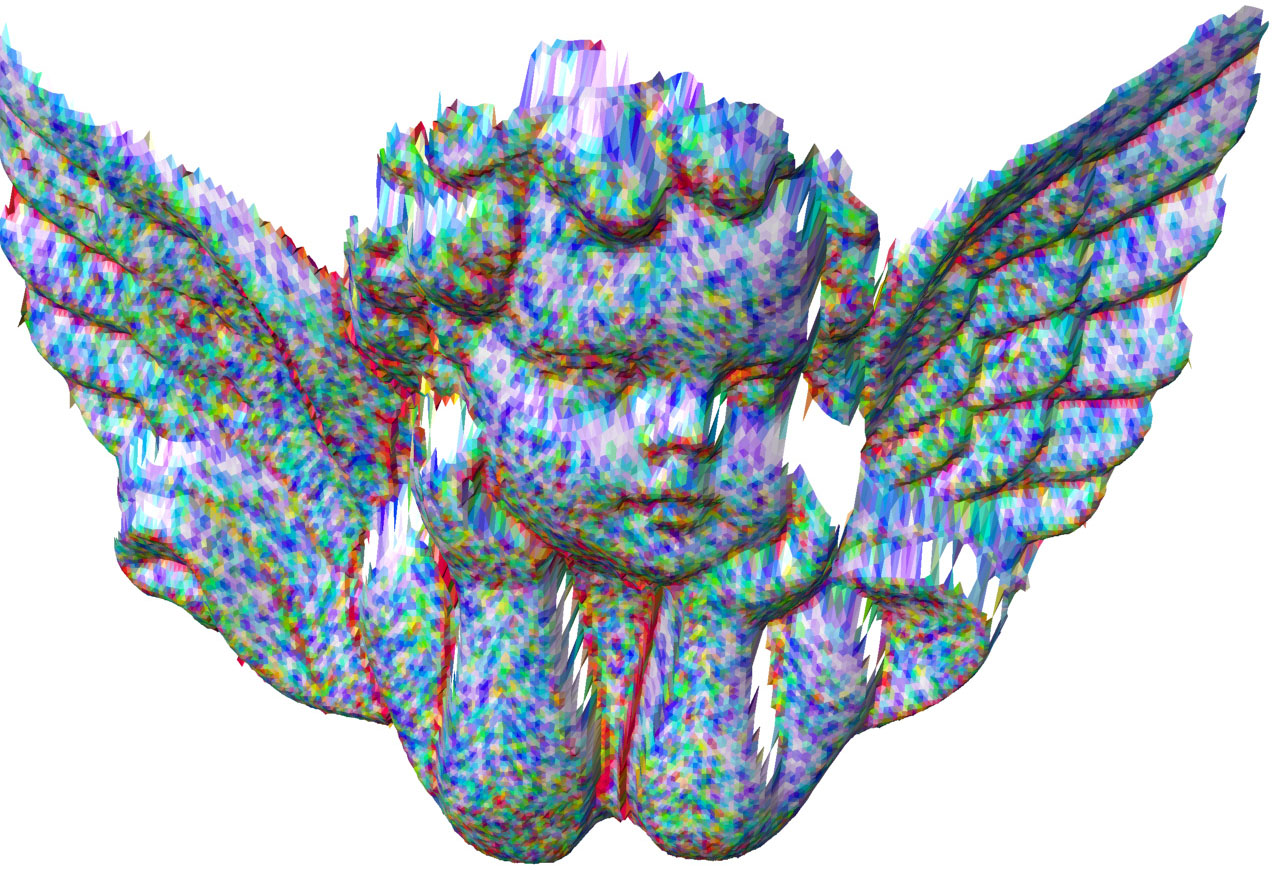} }}%
	\subfloat[\cite{BilFleish}]{{\includegraphics[width=2.4cm]{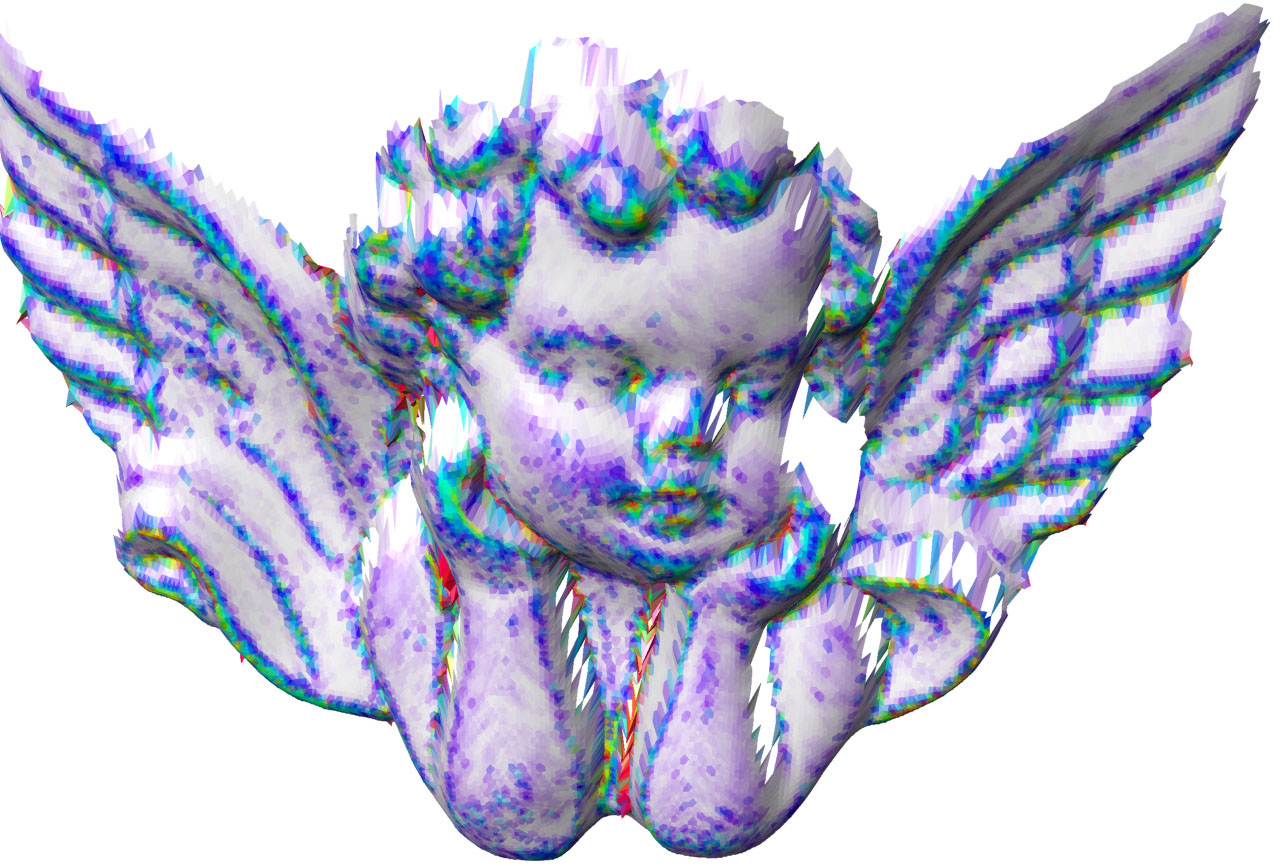} }}%
	\subfloat[\cite{aniso}]{{\includegraphics[width=2.4cm]{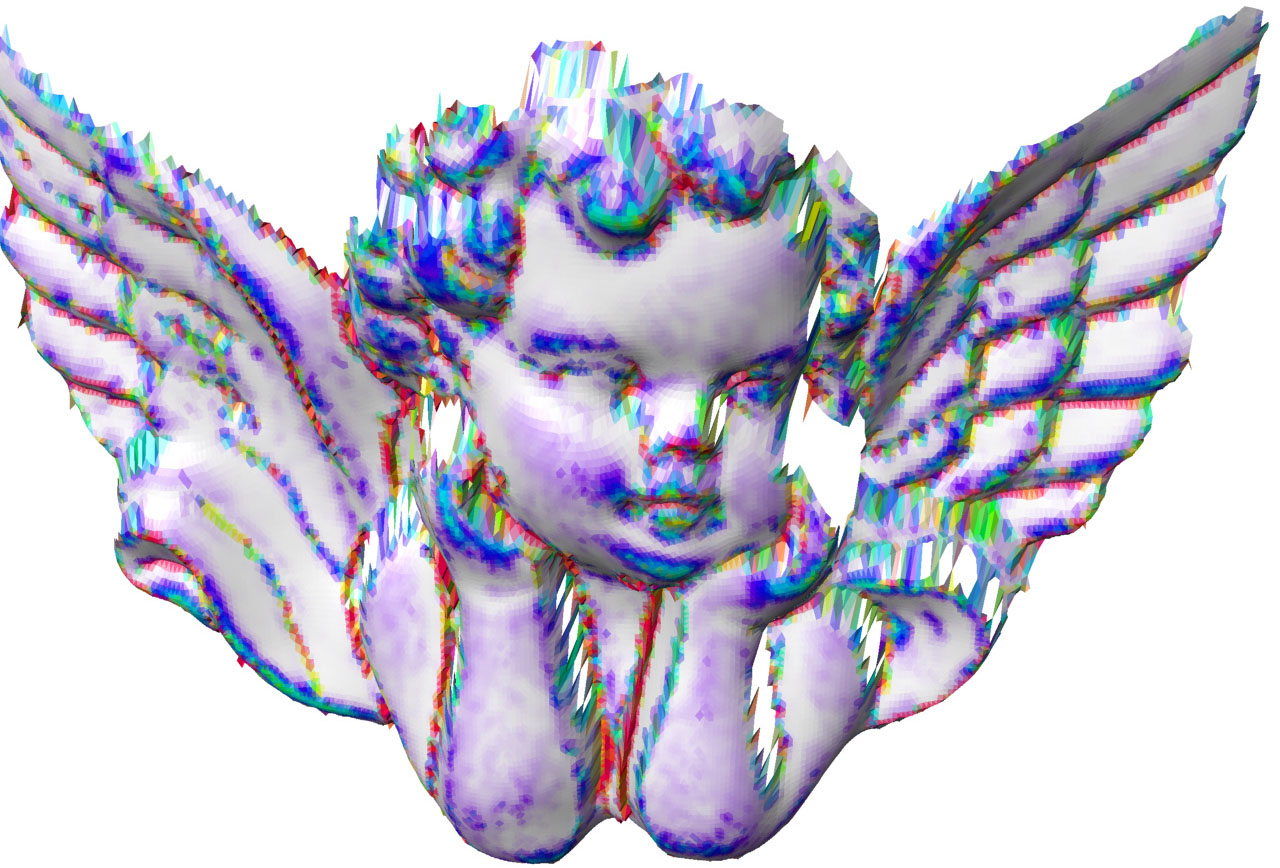} }}%
	\subfloat[\cite{BilNorm}]{{\includegraphics[width=2.4cm]{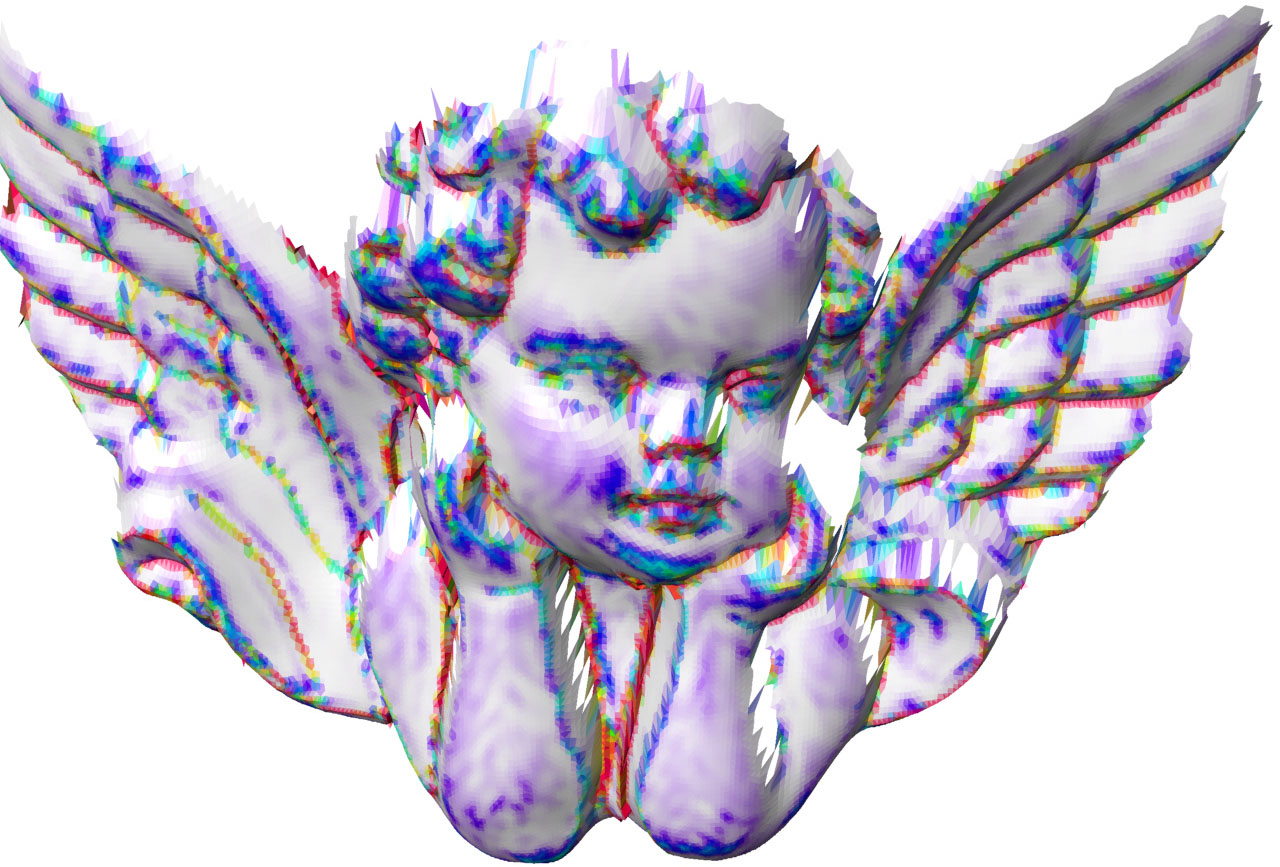} }}%
	\subfloat[\cite{L0Mesh}]{{\includegraphics[width=2.4cm]{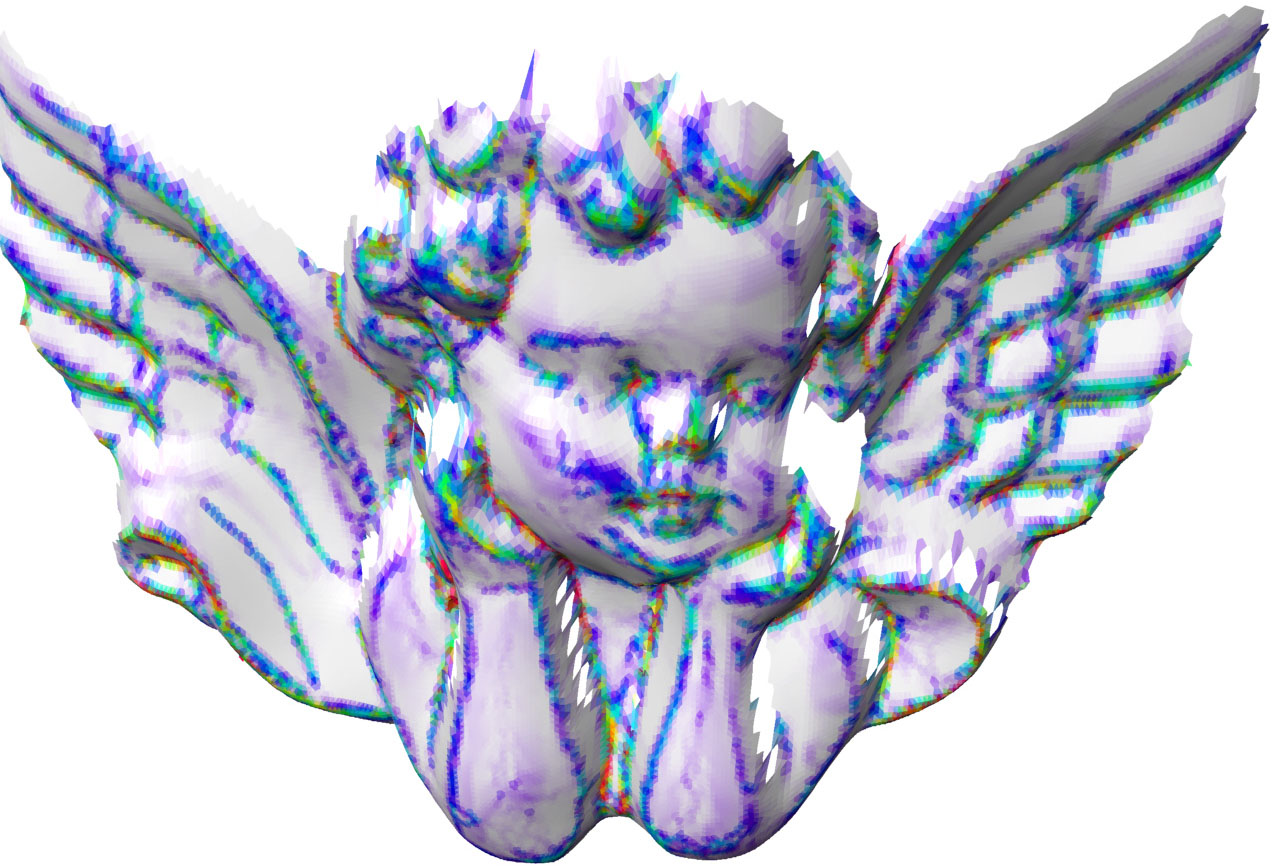} }}%
	\subfloat[\cite{Guidedmesh}]{{\includegraphics[width=2.4cm]{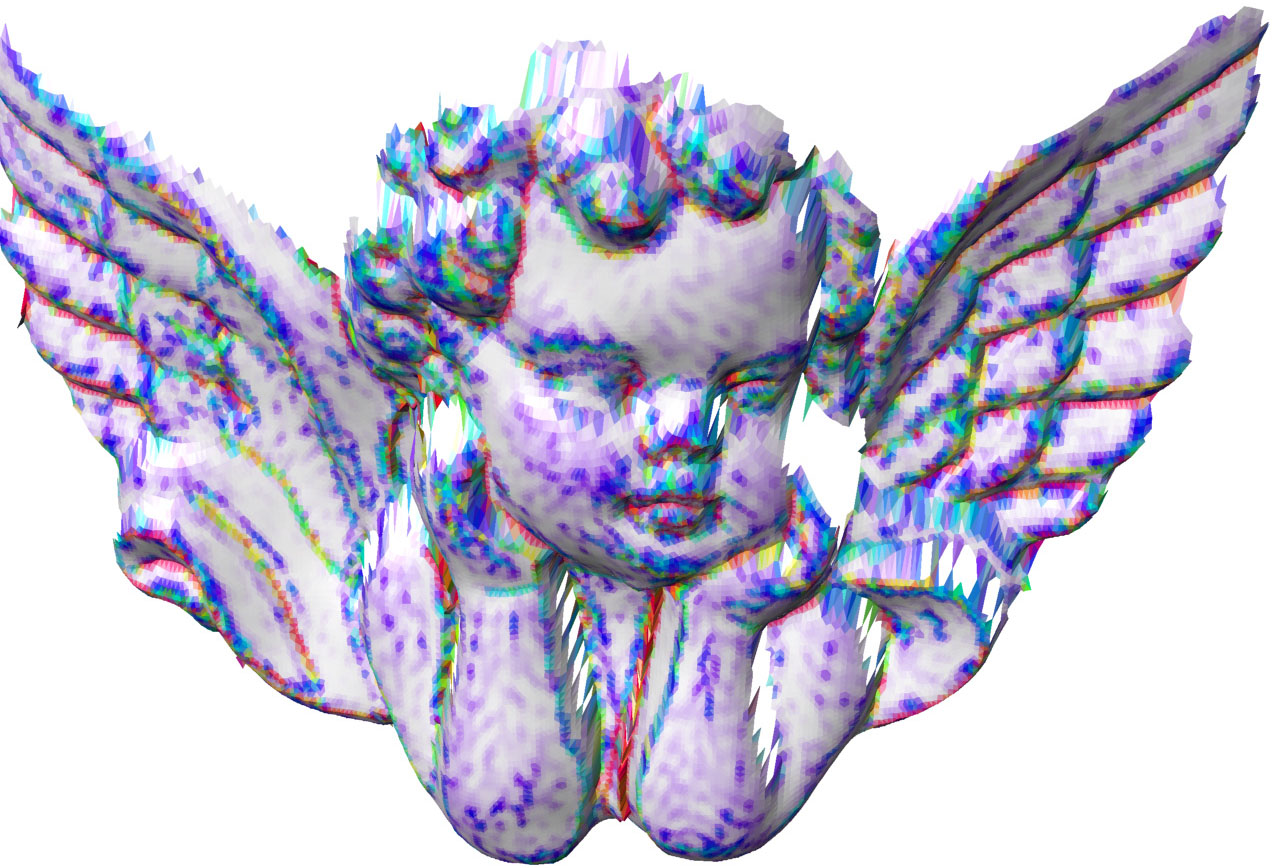} }}%
	\subfloat[Ours]{{\includegraphics[width=2.4cm]{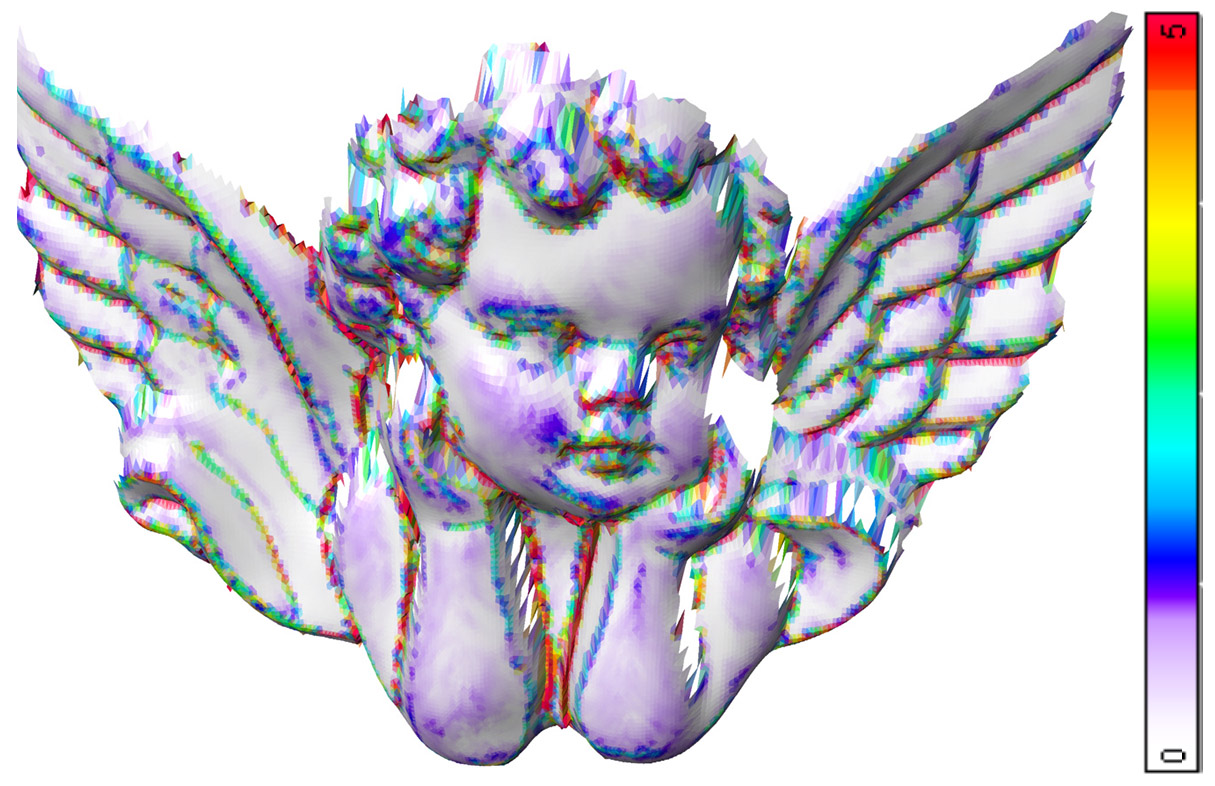} }}%
	\caption{Triangulated mesh surface ({real data}) corrupted by 3D scanner noise. The results obtained by state-of-the-art methods and the proposed method for Angel model are colored with absolute value of the area weighted mean curvature. }%
	\label{fig:realdatamc}%
\end{figure*}

\section{Relation between the Shape Operator and the ENVT}
\label{app:B}
Let us consider a smooth manifold surface $\mathcal{M}$ embedded in $\mathbb{R}^3$. We assume that the surface is orientable and it has well defined normal field $N:\mathcal{M} \to S^2$ on $\mathcal{M}$. Then we can define our proposed tensor $C_\Omega(p): \mathbb{R}^3 \to \mathbb{R}^{3\times 3}$ around a point $p$:
\begin{equation*}
C_\Omega(p) = \int_{\Omega}^{} (n\cdot n^T)d\Omega,
\end{equation*}
where $n\in N$ and $\Omega$ is the local neighborhood area of point p.

To compute the shape operator on a surface, the surface must be $C^2$ and properly oriented. The shape operator is defined on the tangent plane $\mathcal{S}:T_p\mathcal{M} \times T_p\mathcal{M} \to \mathbb{R}$. $T_p\mathcal{M}$ represents the tangent plane, spanned by the basis vectors $\xi_1$ and $\xi_2$. The eigenvalues of the shape operator $\mathcal{S}$ are the principle curvatures $\kappa_1$ and $\kappa_2$ and eigenvectors will be a basis of the tangent plane ($\xi_1, \xi_2$). The eigenvectors are called the principle curvature directions.

Represented in the orthonormal basis of the normal and the principle curvature directions, the surface normal in a local neighborhood of point $p$ can be approximated by:
\begin{equation*}
n(\xi_1,\xi_2)=[1, \kappa_1\xi_1, \kappa_2\xi_2].
\end{equation*}  
Now, we can compute the covariance matrix using the above normal vector:
\begin{equation}
n\cdot n^T = \begin{bmatrix}
1 & \kappa_1\xi_1 & \kappa_2\xi_2 \\
\kappa_1\xi_1 & \kappa_1^2\xi_1^2 & \kappa_1\xi_1\kappa_2\xi_2 \\
\kappa_2\xi_2 & \kappa_1\xi_1\kappa_2\xi_2 &\kappa_2^2\xi_2^2
\end{bmatrix}.
\end{equation}    
For a small symmetric $\Omega$, the integral for the off-diagonal components will be zero:
\begin{equation}
\int_{\Omega}^{} n\cdot n^T d\Omega = \begin{bmatrix}
1 & 0 & 0 \\
0 & \kappa_1^2\xi_1^2 & 0 \\
0 & 0 &\kappa_2^2\xi_2^2
\end{bmatrix}.
\end{equation} 
So by this approximation, the shape operator is contained in the covariance matrix as the lower right $2\times2$ sub-matrix. Therefore the second and third eigenvalue of the NVT approximate the squares of the principal curvatures $\kappa_1$ and $\kappa_2$.

\section{A Visual Representation of the mesh denoising}
\label{denoise}
Traditionally, face normal smoothing is done by rotating the concerned face normal along a geodesics on the unit sphere whereas our method aligns the face normals by projection. We project noisy face normals towards the smooth normals by multiplication of the ENVT to the corresponding face normal.

A demonstration of the ENVT multiplication to the corresponding noisy face normal in $\mathbb{R}^2$ is shown in Figure~\ref{fig:idea}. If $\lambda_2 > \lambda_1$, then it will strengthen the face normal in the desired direction $\mathbf{e}_1$ and suppresses noise in the $\mathbf{e}_2$ direction. The whole procedure consists of the following steps:
\begin{itemize}
	\item A noisy face normal $\mathbf{n}$ is decomposed according to the eigenbasis ($\mathbf{e}_1$ and $\mathbf{e}_2$) of the element based normal voting tensor, Figure~\ref{fig:idea} (left).
	\item Then, the modified eigenvalues ($\lambda_1$ and $ \lambda_2$) get multiplied in the corresponding eigendirections to suppress the weak eigenvalue in weak eigendirection, Figure~\ref{fig:idea} (middle).
	\item Finally, the new element normal $\mathbf{n}^p$ is obtained by normalizing $C_i\cdot \mathbf{n}$, in Figure~\ref{fig:idea} (right).   
\end{itemize} 

\begin{figure}[h]%
	\centering %
	\subfloat[Noisy]{{\includegraphics[width=2.55cm]{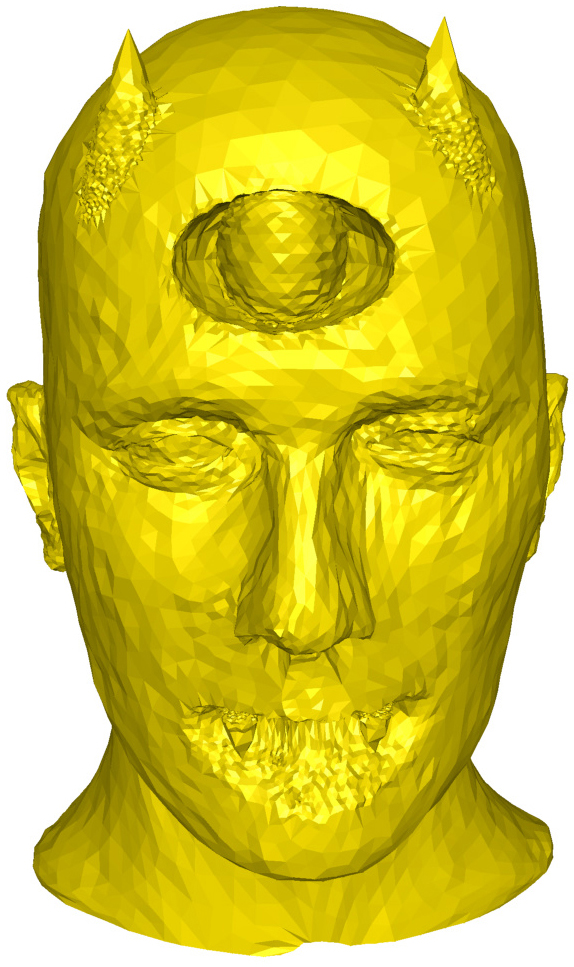} }}%
	\subfloat[RIMLS \cite{rimls}]{{\includegraphics[width=2.7cm]{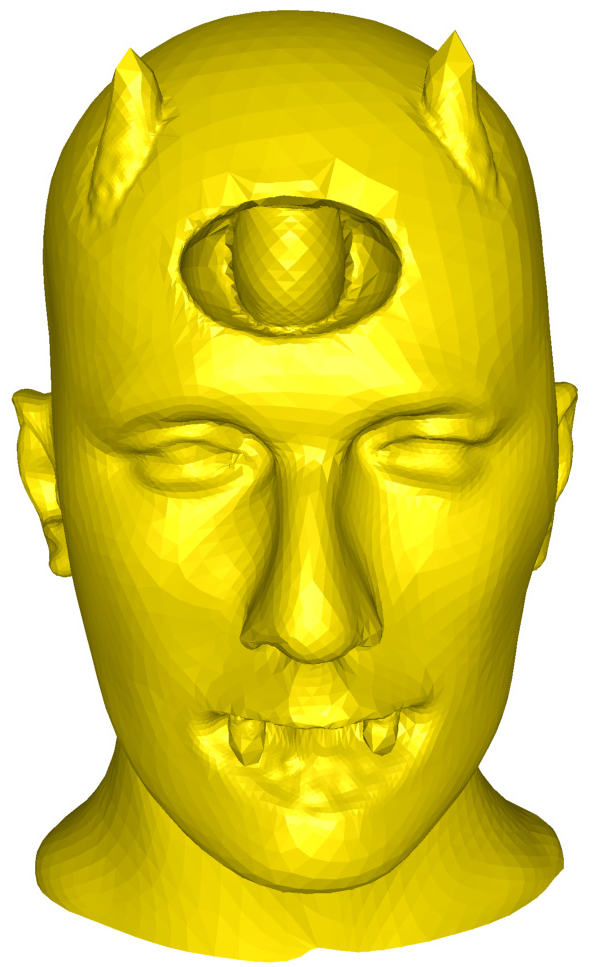} }}%
	\subfloat[Ours]{{\includegraphics[width=2.6cm]{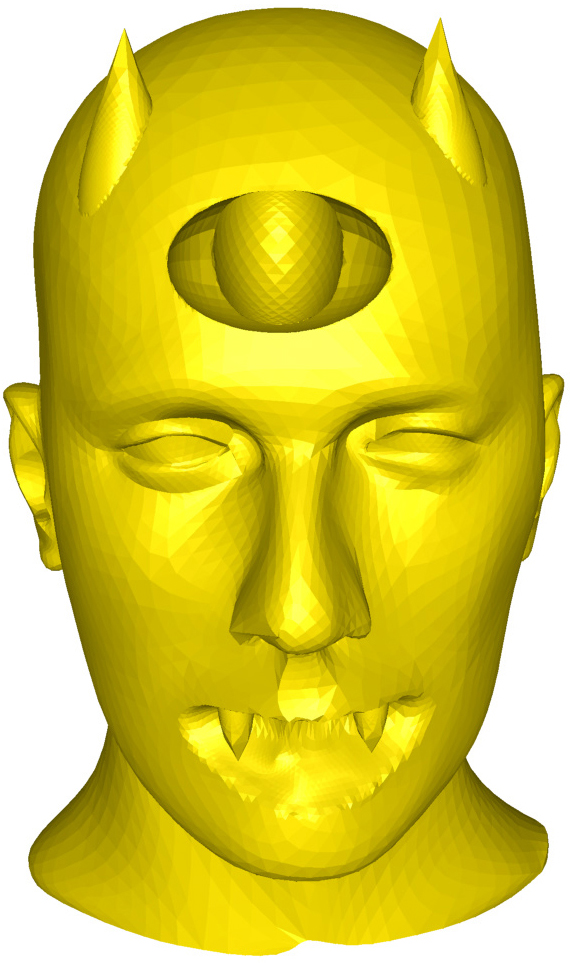} }}%
	\caption{(a) The Devil model corrupted with Gaussian noise($\sigma_n = 0.15l_e$) in the normal direction. (b) Smooth model obtained by the RIMLS method \cite{rimls}. (c) The result obtained by the proposed method.}%
	\label{fig:rimls}
\end{figure}
\section{Mean curvature coloring}
\label{app:mcc}
Mean curvature coloring can be used as a tool to show the capability of a denoising algorithm. In our experiments, we use the cotangent mean curvature operator that is computed at each vertex of the geometry \cite{aniso}. The coloring is done by using the absolute value of the mean curvature vector. Figure \ref{fig:devRockmc} shows that our method is able to remove noise quite similar to methods \cite{BilNorm} and \cite{Guidedmesh} while it produces less shrinkage and better retains the sharper features compared to the other methods. Similarly, for the Julius model (Figure \ref{fig:julimc}), the proposed method produces a smoother surface compared to methods \cite{Guidedmesh} and \cite{robust16} and retains sharper features similar to the Bilateral normal method \cite{BilNorm}, as shown in Figure \ref{fig:julimc}. The Cube model has non-uniform meshes and mean curvature coloring indicates that our method is better compared to other state-of-the-art methods in terms of noise removal and retaining sharp features. Figure \ref{fig:realdatamc} shows that the proposed method is also effective for real data (data obtained from 3D scanners).    

\section{Comparison with Robust Implicit Moving Least Squares (RIMLS)}
The RIMLS algorithm is a features preserving point set surface reconstruction method \cite{rimls}. It follows the moving least squares (MLS) method along with robust statistics to produce a high fidelity surface in presence of noise. Figure \ref{fig:rimls} shows the comparison of the proposed method with the RIMLS method.
\end{document}